\documentclass[11pt]{article}
\usepackage{xcolor}
\usepackage[margin=1in]{geometry}
\usepackage{authblk}

\title{Barron Optimal Transport I: Generative Modeling}
\author{Evan Dogariu and Joan Bruna}
\affil{Courant Institute School of Mathematics, Computing, and Data Science\\ New York University}

\usepackage{resmes}
\usepackage{fancyhdr}
\usepackage{extramarks}
\usepackage{amsmath}
\usepackage{amsthm}
\usepackage{amsfonts}
\usepackage{tikz}
\usepackage{lmodern}
\usepackage{bbm}
\usepackage[ruled,vlined]{algorithm2e}
\usepackage{xcolor}
\usepackage{hyperref}
\usepackage{changepage}
\usepackage{amssymb}
\usepackage{graphicx}\usepackage{soul}
\usepackage{etoc}
\usepackage{amsmath}

\DeclareFontFamily{OMX}{yhex}{}
\DeclareFontShape{OMX}{yhex}{m}{n}{<->yhcmex10}{}

\DeclareSymbolFont{yhlargesymbols}{OMX}{yhex}{m}{n}

\DeclareMathAccent{\widehat}{\mathord}{yhlargesymbols}{"62}

\newtheorem{theorem}{Theorem}
\newtheorem{lemma}{Lemma}
\newtheorem{question}{Question}
\newtheorem{assumption}{Assumption}
\newtheorem{remark}{Remark}
\newtheorem{proposition}{Proposition}
\newtheorem{corollary}{Corollary}

\newtheorem{definition}{Definition}
\newtheorem{example}{Example}

\newtheorem{assumptionalt}{Assumption}[assumption]

\newenvironment{assumptionp}[1]{
  \renewcommand\theassumptionalt{#1}
  \assumptionalt
}{\endassumptionalt}

\def\XXint#1#2#3{{\setbox0=\hbox{$#1{#2#3}{\int}$ }
\vcenter{\hbox{$#2#3$ }}\kern-.6\wd0}}

\DeclareMathOperator*{\argmin}{arg\,min}
 
\renewcommand{\r}{{\mathbb{R}}}

\newcommand{\n}{{\mathbb{N}}}

\newcommand{\1}{{\mathbbm{1}}}
\newcommand{\h}{\mathcal{H}}
\renewcommand{\b}{\mathcal{B}}
\newcommand{\la}{\left\langle}
\newcommand{\ra}{\right\rangle}

\renewcommand{\subset}{\subseteq}

\renewcommand{\epsilon}{\varepsilon}

\newcommand{\p}{\mathcal{P}}
\DeclareMathOperator*{\esssup}{ess\,sup}

\begin{document}
\newpage
\maketitle
\begin{abstract}
Motivated by recent applications in generative modeling and sampling, we introduce a framework for optimal measure transport where cost captures the notion of neural network complexity. In transport-based  generative models, samples from a reference distribution (e.g. Gaussian) are mapped to samples of a target distribution along ordinary or stochastic differential equations. These are implemented as deep residual networks when discretized in time, where each hidden layer approximates the associated instantaneous velocity. Thus, given a pair of target and reference measures, a natural question is to search for the most efficient neural representation that implements this transport. 

Our starting point is the kinetic formulation of OT, due to Benamou and Brenier. We replace the average kinetic $L^2$ energy by the \emph{Barron} energy \cite{bach2017breaking, weinan}, a natural norm which measures the complexity of representing a given vector field with a neural hidden layer, and which captures the adaptive properties of feature learning. This defines a metric on the space of probability measures, complementing existing Wasserstein and Stein geometries. 

In this work we examine the properties of this metric in the context of generative modeling. 
As a first application, we quantify the suboptimality of diffusion generative modeling in the Barron geometry by establishing super-polynomial score approximation lower bounds  for data generated by neural network pushforwards of the Gaussian. We then investigate the benefit of adaptivity as a way to study alternative generative models.
In a companion paper \cite{companionpaper} we leverage the Barron transport geometry for sampling applications, extending the scope of Stein variational gradient methods via feature adaptation. 

\end{abstract}

\etocdepthtag.toc{main}

\etocsettagdepth{main}{subsection}
\etocsettagdepth{appendix}{section}
\tableofcontents

\section{Introduction}
Generative modeling and sampling have become central problems in modern machine learning and scientific computing. Generative modeling -- the task of using a finite dataset to learn to generate new samples from the same underlying data distribution -- describes methods ranging from generative adversarial networks and diffusion/flow-matching models to autoregressive large language models (LLMs) \cite{goodfellow2014generative, sohl2015deep, brown2020language}. Sampling -- the task of using structural information about the target distribution (such as the unnormalized density or its score) to generate samples -- is equally important in the computational sciences: for example, it is used to explore protein conformations and molecular free-energy landscapes \cite{doi:10.1126/science.1187409, noe2019boltzmann}, to compute equilibrium properties of many-body systems \cite{metropolis1953equation}, and to simulate lattice field theories \cite{duane1987hybrid}. \par
Currently, the prevalent approaches in both tasks are rooted in a dynamical perspective, in which one starts with samples drawn from a simple base distribution $\mu$ (such as the standard Gaussian for continuous data or the empty string for language) and iteratively transports them during the sampling process toward a target distribution $\pi$; this is the case for flow-matching models \cite{lipman2023flow, liu2022flow, albergo2023stochasticinterpolantsunifyingframework} and diffusion models \cite{ho2020denoising, song2020score}, LLMs, and most sampling algorithms. If we restrict to continuous-time, then heuristically this dynamical transport from the base distribution to the target distribution is governed by the continuity equation\footnote{The dynamical transport perspective can also capture stochastic transports such as diffusion models, where (\ref{eqn:introcontinuityequation}) is replaced by a Fokker-Planck equation and (\ref{eqn:introODEformulation}) is replaced by a stochastic differential equation (SDE). Under enough regularity, this can also be achieved by adding $\nabla \log \rho_t$ to $v_t$. Our results in this paper are in the ODE setting, but the high-level ideas and many specific results transfer to the SDE setting.}
\begin{equation}
    \label{eqn:introcontinuityequation}
    \partial_t \rho_t + \nabla \cdot (\rho_t v_t) = 0,
\end{equation}
where $\rho_0 = \mu$ is the base distribution, $\rho_1 = \pi$ is the target distribution, $(\rho_t)_{t \in (0, 1)}$ describe the intermediate distributions along the sampling dynamics, and $v_t: \r^d \to \r^d$ is the instantaneous velocity field.
Under appropriate regularity, these dynamics may be implemented by the differential equation
\begin{equation}
\label{eqn:introODEformulation}
    \dot{X}_t = v_t(X_t), \qquad \rho_t = \operatorname{Law}(X_t), \qquad X_0 \sim \mu.
\end{equation}
From this dynamical perspective, both generative modeling and sampling can be viewed as the task of designing an appropriate transport, parametrized by $(v_t)_{t \in [0, 1]}$, using available samples from $\pi$ in the former and structural information on its density in the latter.

A key question is not merely whether one can transport $\mu$ to $\pi$ dynamically, but whether one can do so using vector fields that are simple to approximate in a chosen model class. For example, one may wish to restrict $v_t$ to neural networks of bounded size or to control its norm in a fixed reproducing kernel Hilbert space (RKHS), in a smoothness/Sobolev class, or in some other structured family of functions. In order to quantify this, we may wish to penalize $v_t$ using some prescribed Banach norm $\|v_t\|_{\b}$. Minimizing the accumulated cost 
\begin{equation}
\label{eqn:introaccumulatedcost}
    \int_0^1 \|v_t\|_{\b} dt
\end{equation}
over all transports from $\mu$ to $\pi$ which satisfy (\ref{eqn:introcontinuityequation}) then measures the smallest complexity required to generate $\pi$ within this model class, which informally defines an ``optimal transport'' relative to the complexity measure $\b$. 

In modern generative modeling, it has become standard to represent the drifts $v_t$ using deep residual neural networks, where each layer can be seen as an Euler discretization of an underlying neural ODE \cite{chen2018neural,chizat2026hiddenwidthdeepresnets}. 
We can thus view the forward pass of each drift network $v_t$ as a sequence of updates to the sample $X_t$, in which case the individual layers act as drifts in their own right. Phrased differently, a discretization of the dynamics (\ref{eqn:introODEformulation}) where each $v_t$ is represented by a deep residual network yields nothing more than a single very deep residual network pushing $\mu$ to $\pi$. This, in turn, is nothing more than a discretization of dynamics where each drift is a single layer of a neural network. Hence, it is without loss of generality to consider dynamical transports like (\ref{eqn:introcontinuityequation}) and (\ref{eqn:introODEformulation}) for which each $v_t$ is a shallow neural network. Such neural transports operate on two timescales: samples evolve over longer time horizons according to the target dynamics, while microscopically they evolve according to the layers of learned drift networks. In order to study the neural network size required to transport $\mu$ to $\pi$ dynamically, it is therefore natural to consider (\ref{eqn:introaccumulatedcost}) with a cost $\b$ which quantifies the size of a shallow neural network representation of $v_t$.

For that purpose we instantiate $\b$ as a \emph{Barron} space $\b_\phi$ \cite{barron0, bach2017breaking, jnwbarronOTmaps}, roughly defined as the convex hull of ridge functions of the form $x \mapsto \phi( \langle x, \theta \rangle)$ for an activation $\phi$ (see Section \ref{subsec:barronspacesdefinition}). This space models the feature learning of neural networks by representing functions as adaptively-chosen mixtures of these ridge functions (i.e. shallow neural networks), and the $\b_\phi$ norm of a function describes the size of its smallest shallow neural network representation. 
These properties make $\b = \b_\phi$ a natural choice to study what happens when transporting measures by deep networks, or equivalently by dynamics with shallow drifts.

Minimizing the accumulated cost (\ref{eqn:introaccumulatedcost}) over all transports from $\mu$ to $\pi$ satisfying the continuity equation (\ref{eqn:introcontinuityequation}) yields a metric $D_{\b_\phi}(\mu, \pi)$, which we call the {\em Barron transport metric}. In words, $D_{\b_\phi}(\mu, \pi)$ captures the size of the most efficient neural network transport from $\mu$ to $\pi$, i.e. the smallest generative model for $\pi$.
The formulation of $D_{\b_\phi}$ as a dynamical optimal transport problem allows us to deploy tools from optimal transport and control theory for this analysis. This will play only a minor role in the results of this paper, but in the companion paper \cite{companionpaper} we will study the optimal transport geometry of the metric space $(\p(\r^d), D_{\b_\phi})$ in detail, which will help us construct and analyze new sampling algorithms (see Section \ref{subsec:samplingteaser} for a preview).

\subsection{Main results}

Using the language of the Barron transport metric we can now state our main technical results, which are grouped into two categories. We first study the commonly-used denoising generative model, which defines a path $(\rho_t^\pi)_{t \in [0, 1]}$ from $\gamma_d$ to $\pi$ as the marginal law of $X_t$, the time-reversal of the Ornstein-Uhlenbeck process with initial law $\pi$. Our results show that depth is needed to model the associated drifts using neural networks -- in other words, when we choose the denoising path as the macroscopic behavior of the transport we require alternative dynamics on a nontrivial microscopic timescale. Afterward, we take a first step toward understanding such alternative behaviors by studying the role of feature learning in the broader class of dynamical generative models.

\paragraph{Denoising transport suboptimality.}
Our first technical result asks whether the denoising transport  from $\gamma_d$ to a target distribution $\pi$ is efficient in terms of shallow neural network representation. More precisely, we ask whether
the denoising\footnote{These denoising drifts are also the time-dependent scores of the marginals of the forward OU noising process (up to affine transformations), see Section \ref{subsec:prelimgenerativepaths} and equation (\ref{eqn:prelimreverseOUvelocity}).} drifts $(v_t^\pi)_{t \in [0, 1]}$ can be efficiently represented or approximated by single-layer neural networks, and whether
there exist other generative models for $\pi$ whose drifts admit smaller two-layer representations. This is equivalent to asking whether $D_{\b_\phi}(\gamma_d, \pi)$ is much smaller than $\int_0^1 \|\hat{v}_t\|_{\b_\phi}dt$, where $\hat{v}_t$ is a suitable approximation of ${v}_t^\pi$. We show that the denoising transport can be very suboptimal in this regard:

\begin{theorem}[Denoising suboptimality, informal]
\label{theorem:mainpaperreverseOUsummary}
    If $\pi$ is the pushforward of $\gamma_d$ by a ResNet of depth $\Theta(d)$ and width $O(1)$ with bounded weights and two-layer blocks using activation $\phi$, then $D_{\b_\phi}(\gamma_d, \pi) \lesssim d$. However, the following hold:
    \begin{itemize}
        \item \textbf{(Theorem \ref{theorem:mainpaperrepresentaitonlowerbound}).} There exists such a ResNet (with some $\phi$) for which  $\int_0^1 \|v_t^\pi\|_{\b_\phi}dt \gtrsim 2^d$, even though we have $\|v_t^\pi\|_{\b_\phi} < \infty$ for all $t$.
        \item \textbf{(Corollary \ref{corollary:mainpaperintegratedlowerbound}).} There exists such a ResNet (with $\phi = \operatorname{ReLU}$)\footnote{Our construction provides similar results for other activations. For example, if $\phi = \tanh$ then we get $d^{\Omega(\log(1/\epsilon)^2)}$ lower bounds, see Remark \ref{remark:mainpaperapproximationtanhscaling}.} such that for \textbf{any} approximate drifts $\hat{v}_t$ there is the implication
        \begin{equation}
            \int_0^1 \|\hat{v}_t(x) - v_t^\pi(x)\|^2 d\rho_t^\pi(x) dt\leq \epsilon^2 \implies \int_0^1 \|\hat{v}_t\|_{\b_\phi}dt \gtrsim d^{\Omega(\epsilon^{-4/5})}~,
        \end{equation}
        and moreover if each $\hat{v}_t$ is a two-layer neural network then the largest width is at least $d^{\Omega(\varepsilon^{-4/5})}$ in such cases. 
       Here, $\rho_t^\pi$ are the marginals along the ground truth denoising path.
    \end{itemize}
    In words, exact shallow representation of the denoising drifts can require exponential width, and shallow $L^2$-approximation to accuracy $\epsilon$ can require $d^{\operatorname{poly}(1/\epsilon)}$ width.\footnote{Early-stopping does not fix this. Indeed, we show in Lemma \ref{lemma:appendixearlystoppingforresnets} that if one $\epsilon$-approximates scores up to an OU time whose marginal has $\operatorname{poly}(\epsilon)$ error in either $\operatorname{TV}$, $W_1$, or $\operatorname{KL}$, then our $d^{\operatorname{poly}(1/\epsilon)}$ width lower bound still applies.}
\end{theorem}

Despite the fact that the teacher model has width $O(1)$, we find that achieving score-matching squared error $\leq \epsilon^2$ using two-layer drift models requires width of $d^{\operatorname{poly}(1/\epsilon)}$. Unlike prior works showing hardness of score-matching \cite{song2024cryptographic,vu2026computational}, the difficulty does not come from cryptographic lower-bounds or spatial regularity issues, but instead from a direct width lower bound using the Hermite rank method of \cite{bruna2026multiscale}. It therefore demonstrates a separation between using shallow and deep neural networks for the drifts/scores in denoising generative models, rather than general hardness. 

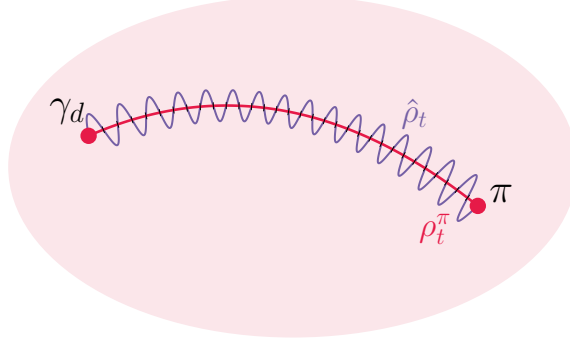
\begin{figure}[tb]
  \centering
  \resizebox{0.5\linewidth}{!}{\par
\definecolor{ambient}{HTML}{FBE6EA}
\definecolor{trajectory}{HTML}{E61A48}
\definecolor{perturbed}{HTML}{7560A5}

\def\wavecount{16}
\def\waveamplitude{2.5} \def\crossinghalflength{0.09} \pgfmathdeclarefunction{basex}{1}{\pgfmathparse{2.26+10.94*#1}}
\pgfmathdeclarefunction{basey}{1}{\pgfmathparse{5.67+4.80*#1-6.77*#1^2}}
\pgfmathdeclarefunction{slopey}{1}{\pgfmathparse{4.80-13.54*#1}}
\pgfmathdeclarefunction{speed}{1}{\pgfmathparse{sqrt(10.94^2+slopey(#1)^2)}}
\pgfmathdeclarefunction{wave}{1}{\pgfmathparse{\waveamplitude*(0.23-0.08*sin(180*#1)+0.045*#1)*sin(360*\wavecount*(0.82*#1+0.18*#1^2))}}

\begin{tikzpicture}[
  x=1cm,y=1cm,
  line cap=round,line join=round,
  smooth trajectory/.style={draw=trajectory,line width=2.1pt},
  perturbed trajectory/.style={draw=perturbed,line width=1.6pt},
  crossing bar/.style={draw=black,line width=0.65pt},
  endpoint label/.style={font=\fontsize{32}{36}\selectfont,inner sep=0pt},
  curve label/.style={font=\fontsize{27}{31}\selectfont,inner sep=0pt}
]
\fill[ambient] (8,4.80) ellipse[x radius=8cm,y radius=4.80cm];

  \coordinate (gamma) at ({basex(0)},{basey(0)});
  \coordinate (pi) at ({basex(1)},{basey(1)});

\draw[perturbed trajectory]
    plot[domain=0:1,samples=1001,variable=\t]
      ({basex(\t)-wave(\t)*slopey(\t)/speed(\t)},
       {basey(\t)+wave(\t)*10.94/speed(\t)});
  \draw[smooth trajectory]
    plot[domain=0:1,samples=101,variable=\t]
      ({basex(\t)},{basey(\t)});

\pgfmathtruncatemacro{\lastcrossing}{2*\wavecount-1}
  \foreach \k in {1,...,\lastcrossing} {
    \pgfmathsetmacro{\crossingt}{(sqrt(0.82^2+0.72*\k/(2*\wavecount))-0.82)/0.36}
    \pgfmathsetmacro{\normalx}{-slopey(\crossingt)/speed(\crossingt)}
    \pgfmathsetmacro{\normaly}{10.94/speed(\crossingt)}
    \draw[crossing bar]
      ({basex(\crossingt)-\crossinghalflength*\normalx},
       {basey(\crossingt)-\crossinghalflength*\normaly}) --
      ({basex(\crossingt)+\crossinghalflength*\normalx},
       {basey(\crossingt)+\crossinghalflength*\normaly});
  }

\fill[trajectory] (gamma) circle[radius=0.23cm];
  \fill[trajectory] (pi) circle[radius=0.23cm];
  \node[endpoint label,anchor=south east]
    at ([xshift=-0.10cm,yshift=0.34cm]gamma) {$\gamma_d$};
  \node[endpoint label,anchor=south west]
    at ([xshift=0.32cm,yshift=0.10cm]pi) {$\pi$};

\node[curve label,text=perturbed] at (11.42,6.3) {$\hat{\rho}_t$};
  \node[curve label,text=trajectory] at (12.00,3.05) {$\rho_t^\pi$};
\end{tikzpicture}}
  \caption{Diagram of the trajectories $\rho_t^\pi$ and $\hat{\rho}_t$
    connecting $\gamma_d$ to $\pi$. The ideal reverse OU path $\rho_t^\pi$ is depicted in red, while the effective path obtained by regressing the target scores with deep ResNet models is depicted in purple. The paths are matched at regular intervals, while the ResNet layers are free to implement arbitrary dynamics in between. Theorem \ref{theorem:mainpaperreverseOUsummary} suggests that this freedom is useful for parameter efficiency.} 
  \label{fig:ou-diagram}
\end{figure}

One way to interpret our result is as follows. 
We can view the entire generative process as a single very deep ResNet, for which the score-matching loss provides supervision at intermediate layers (see Section \ref{subsec:theroleofdepth}). From this perspective it is up to the practitioner how to allocate their computational budget, either by implementing many shallow drift models or a few deeper ones. In the extremes these correspond to maximal supervision (a flow driven by shallow denoising drifts) and minimal supervision (a deep single-shot denoising model), respectively. Theorem \ref{theorem:mainpaperreverseOUsummary} indicates that the former strategy can be suboptimal for the class of ResNet pushforward measures: some amount of non-denoising depth is required for efficient generative modeling of these distributions. From the two-timescale perspective, this suggests that the number of neural network layers needs to be nonnegligible with respect to the number of noise scales; see Figure \ref{fig:ou-diagram}. 

More generally, this construction is a toy model for a natural type of generative process in which features are revealed not by varying the noise level, but instead through nonlinear interactions of multi-index type. This motivates two directions for further work: (1) studying tradeoffs between network depth and denoising supervision and (2) designing algorithms which can learn more diverse types of generative models. 
We discuss future directions of research in Section \ref{sec:outlook}.

\paragraph{Feature learning.} 
In light of Theorem \ref{theorem:mainpaperreverseOUsummary}'s result that neural network-based denoising transports can require non-denoising dynamics between noise scales, it is natural to ask about the properties of such alternative dynamics.
To this end, our second technical result aims to understand the role of feature adaptation in generic dynamical generative models. We do so by comparing the settings where each drift is a two-layer network whose first layer is either fixed or adaptive, i.e. by comparing the kernel and feature learning regimes.

For a fixed kernel $k: \r^d \times \r^d \to \r^{d \times d}$ with RKHS $\h_k$, we denote by $D_{\h_k}$ the {\em Stein metric} on $\p(\r^d)$ which is obtained by the length minimization (\ref{eqn:introaccumulatedcost}) with the choice $\b = \h_k$. This was studied in \cite{duncan2023geometrysteinvariationalgradient} as the Riemannian geometry on $\p(\r^d)$ whose gradient flow structure generates the Stein variational gradient descent sampling algorithm (see Sections \ref{subsec:relatedwork} and \ref{subsec:prelimstein}), but for our purposes $D_{\h_k}$ simply measures the cost of the most efficient transport from $\mu$ to $\pi$ in terms of the RKHS norm of the drifts. The RKHS norm $\|v_t\|_{\h_k}$ captures the hardness of representing $v_t$ using frozen features (e.g. if $k$ is a random feature kernel) in the same way that the Barron norm $\|v_t\|_{\b_\phi}$ captures the hardness of representing $v_t$ using two-layer neural networks with adaptive features. Hence, the comparison between $D_{\h_k}(\mu, \pi)$ and $D_{\b_\phi}(\mu, \pi)$ provides a tool to study the role of feature learning in the task of dynamical transport from $\mu$ to $\pi$. 

In analogy to the study of Gaussian single-index models in regression \cite{multiindexsurvey}, we consider single-index flows with Gaussian source. 
We perform this comparison in both the perturbative and nonperturbative regimes:

\begin{theorem}[Feature learning comparison, informal]
\label{theorem:mainpaperfeaturelearningsummary}
    Let $\gamma_d \in \p(\r^d)$ be the standard Gaussian and let $k: \r^d \times \r^d \to \r^{d \times d}$ be rotation-equivariant (see Definition \ref{def:rotationequivariantkernel}). Define $\pi_\epsilon$ as the time-$\epsilon$ marginal of a flow initialized from $\gamma_d$ with single-index drifts with activation $\phi: \r \to \r$.\footnote{In other words $\pi_\epsilon = \operatorname{Law}(X_\epsilon)$, where $\dot{X}_t = v(X_t)$ with $X_0 \sim \gamma_d$ and $v: \r^d \to \r^d$ is a single-index vector field $v(x) = \sigma u \phi(\la x, \theta\ra)$ for some $u, \theta \in \mathbb{S}^{d-1}$ and $\sigma > 0$.} 
    \begin{itemize}
        \item \textbf{(Examples \ref{example:nonlinearshearrkhs} and \ref{example:nonlinearshearrkhssmoothed}).} For each activation $\phi$, denote
        \begin{equation}
            Q_\phi(d) := \inf_{\substack{\text{$k$ rotation-equivariant}}} \left\{\sqrt{\int_{\r^d} \operatorname{tr}k(x, x)d\gamma_d(x)}\cdot \liminf_{\epsilon \downarrow 0} \frac{D_{\h_k}(\gamma_d, \pi_\epsilon)}{D_{\b_\phi}(\gamma_d, \pi_\epsilon)}\right\}.
        \end{equation}
        Then, we have the scalings $Q_{\operatorname{ReLU}}(d) = +\infty$, $Q_{\tanh}(d) \asymp d^{d/4}$, $Q_{\sin}(d) \asymp e^{\sqrt{d}}$ as $d \to \infty$.
        \item \textbf{(Example \ref{example:mainpapernonperturbativefeaturelearning}).}  There exists a rotation-equivariant $k_\star$ s.t. for each fixed $\epsilon > 0$
        \begin{equation}
            \left(\sup_{t \in [0, 1]}\sqrt{\int_{\r^d} \operatorname{tr}k_\star(x, x)d\rho_t^\star(x)}\right)\cdot  \frac{\int_0^1 \|v_t^\star\|_{\h_{k_\star}}dt}{D_{\b_\phi}(\gamma_d, \pi_\epsilon)} \lesssim \operatorname{poly}(d)
        \end{equation}
        as $d \to \infty$, where $(v_t^\star, \rho_t^\star)_{t \in [0, 1]}$ satisfies (\ref{eqn:introcontinuityequation}) with $\rho_0^\star = \gamma_d$ and $\rho_1^\star = \pi_\epsilon$.
    \end{itemize}
    In words, the separation between $D_{\b_\phi}$ and rotation-equivariant $D_{\h_k}$ is made superpolynomially-large by perturbative single-index flows, but not by nonperturbative ones.
\end{theorem}
The first part is essentially the vector field version of the classical feature learning separation statement that single-index functions are difficult to represent in rotationally-invariant RKHSs \cite{bach2017breaking,ghorbani2021when}, and the proof is the same (via spherical harmonics/Hermite decomposition). 
However, the second part implies that this separation disappears once depth is allowed. Phrased differently, the second part says that a single-index function can be efficiently represented by a deep ResNet whose blocks use the same fixed feature basis. This observation raises the question of whether depth can compensate for a lack of adaptive feature selection (particularly in dynamical measure transport), which we leave to future work.
Note that these quantities $D_{\h_k}, D_{\b_\phi}$ consider an infimum over all transports from $\gamma_d$ to $\pi$, which may be qualitatively different than the transports that are learned in practice (e.g. Wasserstein optimal transports, denoising transports, etc.).

\subsection{Related work}
\label{subsec:relatedwork}

\paragraph{Dynamical optimal transport and associated metrics.} The Benamou-Brenier formulation of optimal transport \cite{benamou2000computational} measures transport cost by minimizing an action subject to the continuity equation, and identifies the squared 2-Wasserstein distance between two probability measures as the minimal possible action. The textbook \cite{ags} thoroughly develops this perspective in the Wasserstein case. The survey \cite{dynamicaltransport} covers some more general dynamical transport metrics coming from other choices of the action, see also \cite{elamvazhuthi2025benamou} and \cite{einav2026equivalence} for control-theoretic instantiations of this idea. 

The choice of a non-$L^p$ dynamical transport cost has direct application to the sampling problem. In \cite{duncan2023geometrysteinvariationalgradient}, a dynamical optimal transport metric is defined on $\p(\r^d)$ using transport costs coming from RKHS norms; this is called the {\em Stein geometry}, since the resulting Riemannian geometry on $\p(\r^d)$ is used to analyze the mean-field behavior of the Stein variational gradient descent algorithm \cite{liu2017steinvariationalgradientdescent} by casting it as a formal Riemannian gradient flow. Similarly, in \cite{rrw} a dynamical optimal transport metric is defined using the Radon transform, and the resulting Riemannian gradient flow structure yields a sampling algorithm with better per-step computational cost. 

\paragraph{Dynamical optimal transport for generative modeling.}
Dynamical generative modeling requires choosing or learning a generative path from data. Some recent works propose to do so using Wasserstein optimal transport. \cite{finlay2020train} and \cite{onken2021otflow} use optimal transport machinery to regularize neural ODEs and normalizing flow models, respectively.   \cite{tong2024improving} uses optimal couplings computed over minibatches in order to learn generative models. Lastly, \cite{liu2022flow} proposes a procedure in which a generative model is iteratively ``rectified" to reduce transport costs and straighten trajectories. 

Closer to our work, several recent papers propose alternative criteria for dynamical optimal transport based on velocity regularity with the goal of better generative modeling. The paper \cite{tsimpos2025optimal} investigates rescheduling of the Wasserstein dynamical transport (and more general straight-line interpolations), deriving a reparameterization of time which minimizes the largest Lipschitz constant of the drift fields. The papers \cite{pandey2025diffeomorphic} and \cite{maurais2025learning} propose to optimize dynamical transports using RKHS penalties on their velocity fields: the former derives some sample and model complexity bounds in this setting, while the latter highlights the optimal control perspective and practical algorithmic considerations. While researchers are beginning to design and study non-Wasserstein dynamical optimal transports for generative modeling, it remains unclear how the geometry of a target distribution relates to the notion of drift regularity we impose.

\paragraph{Approximation theory for deep ResNets and neural ODEs.} There is a growing body of work which studies the approximation power of deep ResNets and their continuous-time/infinite-depth limits, neural ODEs (see \cite{chizat2026hiddenwidthdeepresnets} for a quantitative study of this convergence). 

One approach is based on defining function spaces with suitable norms or metrics. The paper \cite{weinan} notes that classical Barron spaces characterize shallow representations and do not explicitly encode depth or compositional structure; the authors subsequently design flow-induced function spaces (via neural ODE representations with sufficient regularity across depth) and prove direct and inverse approximation theorems for residual networks along with statistical complexity bounds. Related approaches include the ResNet norm of \cite{jacot2026deeplearningconvexparadigm} and its connection to circuit complexity as well as the study of minimum-norm representation paths in \cite{jacot2025hamiltonian}. These works focus on understanding which functions can be represented and approximated using ResNets and neural ODEs, largely by emphasizing the role of depth and the ODE structure.

Complementary work examines which maps neural ODE architectures can represent; see \cite[Section 2]{chemnitz2025dynamical} and \cite{kuehn2026analysis}. These results explain the role of augmentation and geometric restrictions, including the nonsingular Jacobian of finite-time flows generated by smooth vector fields. Such representability results do not by themselves give bounds on the norm or network size needed to approximate a particular measure transport.
Relatedly, the work \cite{ruizbalet2023neural} investigates the universal approximation property for neural ODEs through control-theoretic arguments, and \cite{li2026universal} shows that ODEs with two-layer neural network drifts can approximate a wide class of dynamics or endpoint mappings. 

There is also a recent line of work which specifically investigates the use of neural ODEs in the representation and approximation of probability distributions via dynamical transport. In \cite{marzouk2024distribution} they establish statistical convergence guarantees for likelihood-trained neural ODEs, including nearly minimax-optimal density-estimation rates under smoothness assumptions. Their companion paper \cite{marzouk2025distribution} constructs straight-line flow representations for suitable classes of smooth transport maps and derives explicit neural ODE approximation bounds in Wasserstein distance and KL divergence, with dependence on the dimension and regularity of the source and target densities.

To summarize, the aforementioned works develop approximation theories for deep ResNets and neural ODEs, both for function approximation and recently in the context of probability measure approximation. These methods can yield approximation rates in the spirit of the shallow neural network theory, but it is still unclear which geometric properties of a data distribution allow neural ODE transports with few parameters.

\paragraph{Optimality of reverse Ornstein-Uhlenbeck transport.}
It is known that the reverse OU probability flow (which is one of the most common approaches to generative modeling, see Section \ref{subsec:prelimgenerativepaths}) does not always minimize the Benamou-Brenier action associated with Wasserstein optimal transport. In \cite{tanana2021comparison} it is shown that the endpoint map of the related reverse heat flow transport of Kim-Milman \cite{kim2012generalization} can differ from the Brenier map, see also \cite{LAVENANT2022108225} for quadratic transport cost suboptimality. However, from \cite{mikulincer2022lipschitzpropertiestransportationheat} it is also known that these reverse heat flow transports have endpoint maps with good spatial regularity.
These results leave open the question of how reverse OU transport compares under costs that measure the approximation/learning complexity of its velocity field. 

There are some notable recent works demonstrating some hardness of learning reverse OU transports. The paper
\cite{song2024cryptographic} leverages cryptographic hardness to show the hardness of learning $L^2$-approximate scores; in that construction the hardness comes from the learning problem, and the author notes that the true scores are easy to approximate with neural networks.
The recent work of \cite{vu2026computational} constructs distributions for which (1) there exists an $L^2$ approximate drift which does not sample well and (2) no poly-time drift estimator with controlled Lipschitz constant can sample well. Our approximation lower bounds are more structural as they do not rely on regularity or computational hardness: the Hermite rank argument directly lower bounds network size irrespective of the estimator's regularity or computational complexity.

\paragraph{Feature learning in RKHS and Barron spaces.}
 For a common dictionary of neurons, an RKHS fixes a feature distribution, while a Barron norm optimizes over representations and can therefore reflect features adapted to the target. In \cite{bach2017breaking},  this distinction is studied through variation-norm regularization in order to prove adaptation to unknown low-dimensional structure, i.e. {\em feature learning}. The separation between kernel methods and neural networks is further investigated in \cite{ghorbani2021when} and \cite{wu2022spectral} using tools from spherical harmonic analysis. These works analyze feature learning separations in the shallow regression setting.

 Deep and compositional kernels have also been studied \cite{shankar2020neural,liang2022mehler}, with particular focus on the effect of depth on the spectrum and complexity of their associated RKHSs. These works demonstrate improved modeling properties arising from the compositional structure, but it is still unclear to what extent depth can compensate for the lack of adaptive feature selection in a fixed-feature RKHS. This question remains poorly understood not only in the setting of regression using deep models composed of RKHS layers, but also in the context of dynamical measure transport with RKHS drifts.

\subsection{Statement of AI Usage}
Several mathematical ideas and proof arguments in this paper were suggested by ChatGPT 5.6 and Astra and subsequently developed and refined by the authors. These include the sharp scaling of RKHS tangent norms in Lemma \ref{lemma:appendixupperandlowerboundsforsingleindexgaussianflows} and the $\tanh$ deconvolution argument in Lemma \ref{lemma:appendixtiltdenominatorremoval}. For the tail control of the averaging operator in Lemma \ref{lemma:appendixaveragingoperator}, the initial ideas originated with the authors, and subsequent exchanges with Astra helped develop and refine the argument. The authors take full responsibility for all results and arguments in this paper.

\section{Preliminaries}

For some notation, we write $\p(\r^d)$ for the space of Borel probability measures on $\r^d$ (equipped with the narrow topology, see \cite[Chapter 5.1]{ags}), $\p_2(\r^d)$ for those with finite second moment, and $\gamma_d := \mathcal N(0,\1_d)$ for the standard Gaussian. If $T:\r^d\to\r^m$ is measurable and $\mu \in \p(\r^d)$, then $T_\#\mu \in \p(\r^m)$ denotes the pushforward of $\mu$ by $T$. We denote by $\|\cdot\|_\infty$ the sup norm and by $\|\cdot\|_{L^\infty}$ the essential supremum norm w.r.t. Lebesgue. All logarithms are natural.

Throughout this paper, dynamical transport between probability measures is described by the continuity equation. A narrowly continuous curve $(\rho_t)_{t\in[0,1]}\subset\p(\r^d)$ and a Borel vector field $v_t:\r^d\to\r^d$ are said to satisfy the continuity equation if 
\begin{equation}
\label{eqn:prelimcontinuity}
    \partial_t\rho_t+\nabla\cdot(\rho_t v_t)=0
\end{equation}
in the weak sense, i.e. if for every $\varphi\in C_c^\infty((0,1)\times\r^d)$ we have
\begin{equation}
\label{eqn:prelimweakcontinuity}
    \int_0^1\int_{\r^d}
    \left(
        \partial_t\varphi(t,x)
        +
        \langle v_t(x),\nabla\varphi(t,x)\rangle
    \right)
    d\rho_t(x)dt
    =0,
\end{equation} 
see (8.1.3) of \cite{ags}.
We always assume enough integrability for this expression to be well-defined; for example,
\begin{equation}
    \int_0^1\|v_t\|_{L^1(\rho_t; \r^d)}dt<\infty.
\end{equation}
If $\rho_0=\mu$ and $\rho_1=\nu$, we say that $(v_t, \rho_t)_{t\in[0,1]}$ transports $\mu$ to $\nu$.
When $v_t$ is sufficiently regular, the same transport can be described by the ODE
\begin{equation}
\label{eqn:prelimcharacteristics}
    \dot X_t=v_t(X_t),
    \qquad
    X_0\sim\mu,
    \qquad
    \rho_t=\operatorname{Law}(X_t).
\end{equation}
We use the continuity equation formulation because it does not require pointwise existence or uniqueness of the flow.

\subsection{Wasserstein metric}

\label{subsec:prelimwasserstein}

For $\mu,\nu\in\p(\r^d)$, the quadratic Wasserstein distance is defined by
\begin{equation}
\label{eqn:prelimW2}
    W_2^2(\mu,\nu)
    :=
    \inf_{\Gamma\in\Pi(\mu,\nu)}
    \int_{\r^d\times\r^d}|x-y|^2d\Gamma(x,y),
\end{equation}
where $\Pi(\mu,\nu)$ is the set of couplings of $\mu$ and $\nu$. This is finite whenever $\mu, \nu \in \p_2(\r^d)$.
There are many interesting properties of the Wasserstein space $(\p_2(\r^d), W_2)$ which we will not require, see the second half of the textbook \cite{ags} for more details. The main fact which will be relevant for us is the formula of Benamou-Brenier \cite{benamou2000computational}, which gives an equivalent dynamical description: 
\begin{equation}
\label{eqn:prelimBBenergy}
    W_2(\mu,\nu)^2
    =
    \inf_{\mu\to\nu}
    \int_0^1
    \|v_t\|_{L^2(\rho_t; \r^d)}^2dt,
\end{equation}
where the infimum is over $(v_t, \rho_t)_{t\in[0,1]}$ which transport $\mu$ to $\nu$ (i.e. which satisfy the continuity equation (\ref{eqn:prelimcontinuity}) and $\rho_0 = \mu$, $\rho_1 = \nu$).
Equivalently,
\begin{equation}
\label{eqn:prelimBBlength}
    W_2(\mu,\nu)
    =
    \inf_{\mu\to\nu}
    \int_0^1
    \|v_t\|_{L^2(\rho_t; \r^d)}dt.
\end{equation}
The equivalence of \eqref{eqn:prelimBBenergy} and \eqref{eqn:prelimBBlength} follows by reparameterizing a path to have constant speed. 

In particular, the instantaneous Wasserstein cost of a velocity field is its $L^2(\rho_t; \r^d)$ norm, which measures the average speed of particles under the current distribution. Otto \cite{ottocalculus} used this to recognize the metric geometry $(\p_2(\r^d), W_2)$ as a formal Riemannian geometry, in which the tangent vectors are identified as vector fields and the tangent norm at $\rho \in \p_2(\r^d)$ is $\|\cdot\|_{L^2(\rho; \r^d)}$. This allows us to draw inspiration from finite-dimensional Riemannian geometry to reason about the dynamical evolution of probability measures. As some examples:
\begin{itemize}
    \item (\textbf{Gradient flows}). For a sufficiently regular functional $\mathcal{F}: \p_2(\r^d) \to \r$ and an initial $\rho_0 \in \p_2(\r^d)$, one may define a curve of measures $(\rho_t)_{t \geq 0} \subset \p_2(\r^d)$ as the $W_2$ gradient flow of $\mathcal{F}$. It can be shown that $(v_t, \rho_t)_{t \geq 0}$ satisfies the continuity equation (\ref{eqn:prelimcontinuity}) with \begin{equation}
        \label{eqn:prelimsW2gradientflow}v_t(x) = -(\nabla \delta \mathcal{F}(\rho_t))(x),
    \end{equation} where $\delta \mathcal{F}(\rho): \r^d \to \r$ denotes the first variation of $\mathcal{F}$ at $\rho$. An important special case is with $\mathcal{F}(\cdot) := \operatorname{KL}(\cdot \| \pi)$ for a target $\pi \in \p_2(\r^d)$, in which case the resulting curve $(\rho_t)_{t \geq 0}$ describes the evolution of marginal distributions along the Langevin dynamics. See Chapter 11 of \cite{ags} for full details.
    \item (\textbf{Geodesics}). For two measures $\mu, \nu \in \p_2(\r^d)$, one can describe the optimal transport $(v_t, \rho_t)_{t \in [0, 1]}$ which minimizes (\ref{eqn:prelimBBenergy}) via the geodesic PDE
    \begin{equation}
    \label{eqn:prelimW2geodesics}
        v_t = \nabla \varphi_t, \quad \partial_t \varphi_t + \frac{1}{2} |\nabla \varphi_t|^2 = 0 \text{ a.e.},
    \end{equation}
    where $\varphi_t: \r^d \to \r$ are the generating potentials (see e.g. \cite[Theorem 6.2.10]{ags}).
\end{itemize}

An important property of the Wasserstein geometry is its locality: the velocity of a gradient flow (resp. geodesic) at a point $x \in \r^d$ depends only on the Euclidean gradient of the generating potential $\nabla \delta \mathcal{F}(\rho_t)$ (resp. $\varphi_t$) at $x$. This lack of cooperation between particles is part of the reason why the $W_2$ gradient flow of $\operatorname{KL}(\cdot \| \pi)$ can be simulated by the Langevin SDE, which is implemented with noninteracting particles.

We emphasize that the $L^2(\rho_t; \r^d)$ norm does not measure how complicated the function $x\mapsto v_t(x)$ is in terms of smoothness, approximation difficulty, etc. A transport $(v_t, \rho_t)_{t\in[0,1]}$ which is optimal w.r.t. $W_2$ connects two measures in a way which moves particles minimally, but by doing so it can sacrifice the spatial regularity of the induced drift fields $v_t$.

\subsection{Stein metric}
\label{subsec:prelimstein}

Let
$k:\r^d\times\r^d\to\r^{d\times d}$
be a symmetric positive-semidefinite matrix-valued kernel, and let $\h_k$ be its vector-valued reproducing kernel Hilbert space (RKHS). In particular, for every $x, a \in \r^d$ we have $k(\cdot, x)a \in \h_k$ and
\begin{equation}
\label{eqn:prelimreproducing}
    \langle v,k(\cdot,x)a\rangle_{\h_k}
    =
    \langle v(x),a\rangle \quad (\forall v \in \h_k).
\end{equation}
\begin{assumptionp}{$k$}
\label{assumption:kernel}
    Throughout the rest of this paper, we assume the kernel $k: \r^d \times \r^d \to \r^{d \times d}$ is jointly continuous (w.r.t. the operator norm). 
\end{assumptionp}
If $(v_t,\rho_t)_{t \in [0, 1]}$ is a path with $v_t\in\h_k$ we may consider its RKHS energy $\int_0^1 \|v_t\|_{\h_k}^2dt$ in the spirit of (\ref{eqn:prelimBBenergy}), in which case the corresponding transport distance is
\begin{equation}
\label{eqn:prelimsteindistance}
    D_{\h_k}(\mu,\nu)^2
    :=
    \inf_{\mu\to\nu}
    \int_0^1\|v_t\|^2_{\h_k}dt,
\end{equation}
with the infimum over solutions of the continuity equation (\ref{eqn:prelimcontinuity}) with $v_t\in\h_k$ for almost every $t$. 
Equivalently, by constant-speed reparameterization,
\begin{equation}
\label{eqn:prelimsteinlength}
    D_{\h_k}(\mu,\nu)
    =
    \inf_{\mu\to\nu}
    \int_0^1\|v_t\|_{\h_k}dt.
\end{equation}
Note that if $\|\cdot\|_{L^2(\rho; \r^d)} \lesssim \|\cdot\|_{\h_k}$ (which holds for e.g. bounded kernels) then $W_2(\mu, \nu) \lesssim D_{\h_k}(\mu, \nu)$, see \cite[Lemma 17]{duncan2023geometrysteinvariationalgradient}.

Under suitable regularity assumptions, this construction gives an extended metric
(i.e. a metric which may take the value $+\infty$) which is referred to as the {\em Stein metric} in the literature; see
\cite[Definition 15, Lemma 17]{duncan2023geometrysteinvariationalgradient}
for one precise formulation.  Importantly, the space $(\p(\r^d), D_{\h_k})$ can be equipped with a formal Riemannian geometry (since the tangent spaces $\h_k$ are Hilbert) to enable analytical methods in analogy with the Wasserstein setting. This geometry has been investigated recently in \cite{duncan2023geometrysteinvariationalgradient} as the metric on $\p(\r^d)$ whose gradient flow structure induces the Stein variational gradient descent (SVGD) algorithm \cite{liu2017steinvariationalgradientdescent}. In particular, \cite{duncan2023geometrysteinvariationalgradient} developed the Riemannian perspective and showed the following:
\begin{itemize}
    \item (\textbf{Gradient flows}). In contrast with the Wasserstein case (\ref{eqn:prelimsW2gradientflow}), the $D_{\h_k}$ gradient flow of $\mathcal{F}$ is implemented by $(v_t, \rho_t)_{t \geq 0}$ satisfying the continuity equation (\ref{eqn:prelimcontinuity}) with
    \begin{equation}
    \label{eqn:prelimssteingradientflow}
        v_t(x) = -(\mathcal{T}_{k, \rho_t} \nabla \delta \mathcal{F}(\rho_t))(x),
    \end{equation}
    where $\mathcal{T}_{k, \rho_t}: L^2(\rho_t; \r^d) \to L^2(\rho_t; \r^d)$ is the integral operator with kernel $k$ and measure $\rho_t$.\footnote{In other words, $(\mathcal{T}_{k, \rho}v)(x) = \int_{\r^d} k(x, y) v(y) d\rho(y)$.} An important special case is with $\mathcal{F}(\cdot) := \operatorname{KL}(\cdot \| \pi)$ for a target $\pi \in \p(\r^d)$, in which case the resulting curve $(\rho_t)_{t \geq 0}$ describes the evolution of marginal distributions along the mean-field limit of continuous-time SVGD. See Lemma 11 and Corollary 13 of \cite{duncan2023geometrysteinvariationalgradient} for these statements. 
    \item (\textbf{Geodesics}). For two measures $\mu, \nu \in \p(\r^d)$, one can describe a transport $(v_t, \rho_t)_{t \in [0, 1]}$ which is stationary for the energy (\ref{eqn:prelimsteindistance}) by the PDE
    \begin{equation}
    \label{eqn:prelimssteingeodesics}
        v_t = \mathcal{T}_{k, \rho_t} \nabla \varphi_t, \quad \partial_t \varphi_t + \la \nabla \varphi_t, \mathcal{T}_{k, \rho_t} \nabla \varphi_t\ra = 0 \text{ a.e.},
    \end{equation}
    where $\varphi_t: \r^d \to \r$ are the generating potentials, see \cite[Proposition 20]{duncan2023geometrysteinvariationalgradient}. This can be contrasted with the Wasserstein case (\ref{eqn:prelimW2geodesics}).
\end{itemize}

The difference between the Wasserstein and Stein geometries is easily understood through the tangent norms:
\begin{equation}
\label{eqn:prelimW2vsStein}
    \text{\textbf{Wasserstein}: }
    \|v_t\|_{L^2(\rho_t; \r^d)},
    \qquad
    \text{\textbf{Stein}: }
    \|v_t\|_{\h_k}.
\end{equation}
The first measures how far/fast particles move, while the second measures how large the velocity field is in a prescribed RKHS. Crucially, the introduction of a kernel $k$ forces particles in different locations to interact in order to decrease the RKHS norm; this can be also seen through the gradient flow and geodesic identities (\ref{eqn:prelimssteingradientflow}) and (\ref{eqn:prelimssteingeodesics}), in which the kernel integral operator $\mathcal{T}_{k, \rho}$ couples the velocities experienced by different particles. This interaction (and its implementable form using Stein integration by parts) yields the advantages of the SVGD sampling algorithm.

Any choice of kernel $k: \r^d \times \r^d \to \r^{d \times d}$ induces its own Stein geometry through the metric $D_{\h_k}$, which can be understood in two equivalent ways. One perspective is that the kernel defines a notion of vector field complexity via its RKHS $\h_k$, which yields a Benamou-Brenier-type energy minimization via (\ref{eqn:prelimsteindistance}). The other perspective is that the kernel couples spatial positions and the velocities experienced by particles at those positions via the operator $\mathcal{T}_{k, \rho}$. Of course, this interaction and the resulting quantities/algorithms depend heavily on the choice of kernel. 

Some common choices are bilinear kernels $k(x, y) = (1 + \la x, y\ra)\1_d$, Gaussian/RBF kernels $k(x, y) = e^{-\|x-y\|^2}\1_d$, and the like \cite{liu2017steinvariationalgradientdescent}; these introduce isotropic interactions between particles. As is the case in kernel-based learning methods, much can be achieved if one were able to select the ``perfect kernel" for the given data. However, in practice one must choose a kernel in advance, in which 
case the kernel captures a fixed set of features and ought not to prefer any particular direction. We will focus on the following class:

\begin{definition}[Rotation-equivariant kernel]
\label{def:rotationequivariantkernel}
A matrix-valued kernel $k$ is {\em rotation-equivariant} if
\begin{equation}
\label{eqn:prelimrotationequivariant}
    k(Qx,Qy)
    =
    Qk(x,y)Q^\top \quad (\forall x, y \in \r^d)
\end{equation}
for every orthogonal matrix $Q\in O(d)$. We denote the class of such kernels by $\mathcal K_{\operatorname{rot}}$.
\end{definition}

We emphasize that in practice (sampling with SVGD, learning dynamical transports via kernel methods or neural networks in the kernel regime, etc.) the kernels involved tend to be rotation-equivariant \cite{liu2017steinvariationalgradientdescent}. Informally, $\mathcal{K}_{\operatorname{rot}}$ can be viewed as the class of kernels which are ``data-agnostic" since they represent a fixed basis-independent feature set and cannot know any features of the data.

One of the main developments of this work will be the construction of a geometry on $\p(\r^d)$ using Barron norms instead of RKHS norms, in which case the induced geometry can be understood as ``adaptively choosing features based on the data". This perspective will yield dynamical transports and algorithms which leverage feature learning, and one contribution of this work will be the comparison between fixed features in the rotation-equivariant Stein case and learned features in the Barron case. We defer this construction until Section \ref{section:barron}.

\subsection{Generative models as dynamical transports}
\label{subsec:prelimgenerativepaths}

So far, we have discussed the geometry induced by minimization over dynamical transports from $\mu$ to $\nu$ for an arbitrary pair $\mu, \nu \in \p(\r^d)$. While these optimal transport constructions are mathematically meaningful on their own, we are interested in these geometries as analytical tools to study the task of generative modeling.\footnote{In the companion paper \cite{companionpaper} we will also leverage the same constructions to study the related task of sampling.} In this subsection we frame generative modeling as the learning of a dynamical transport $(v_t, \rho_t)_{t \in [0, 1]}$ in order to draw practical conclusions from these geometric quantities.

Let $\mu$ be a base distribution which is easy to sample from (such as the standard Gaussian $\gamma_d$),\footnote{Generally $\mu$ will be rotationally-invariant, since it must also be data-agnostic and ought not to depend on the choice of basis.} and let $\pi$ be the target data distribution. For our purposes, a (continuous-time) generative model consists of a velocity field $(v_t)_{t\in[0,1]}$ such that the ODE
\begin{equation}
\label{eqn:prelimgenerativeODE}
    \dot X_t=v_t(X_t),
    \qquad
    X_0\sim\mu,
\end{equation}
satisfies $\operatorname{Law}(X_1) = \pi$. 
This is the standard setup of a {\em flow model} \cite{chen2018neural,grathwohl2019ffjord}, and the continuity equation induced by (\ref{eqn:prelimgenerativeODE}) can also describe the marginals of a stochastic interpolant or SDE-based generative model via its probability flow ODE. Thus, the goal of the task of generative modeling is to learn a suitable $(v_t)_{t \in [0, 1]}$ using samples from the target $\pi$.

The endpoint measures do not determine the path: different generative models may connect the same $\mu$ and $\pi$ using very different intermediate distributions and velocity fields. For example, Wasserstein optimal transport chooses a path minimizing \eqref{eqn:prelimBBlength}, while flow-matching and diffusion methods usually prescribe a convenient interpolation and learn the velocity field associated with (the reversal of) that interpolation. Hence, the learning problem is not well-specified unless one chooses a particular transport between $\mu$ and $\pi$ to attempt to learn. We summarize the main choices below:

\paragraph{Reverse Ornstein-Uhlenbeck transport.}

The main prescribed path studied in this paper is the denoising path to a target $\pi \in \p(\r^{d})$ used by modern diffusion models. The Ornstein-Uhlenbeck (OU) semigroup $(P_\tau)_{\tau \geq 0}$ acts on functions according to
\begin{equation}
\label{eqn:prelimOU}
    P_\tau f(x)
    :=
    \mathbb E_{Z \sim \gamma_d}
    \left[
        f\left(
            e^{-\tau}x
            +
            \sqrt{1-e^{-2\tau}}Z
        \right)
    \right],
\end{equation}
and the adjoint operators $(P_\tau^*)_{\tau \geq 0}$ act on probability measures and describe the corresponding flow along the OU path.
Indeed, if $X_0\sim\gamma_d$ and $X_1\sim\pi$ are independent, then for $t\in[0,1]$ the interpolant
\begin{equation}
\label{eqn:prelimVPinterpolation}
    X_t
    :=
    tX_1+\sqrt{1-t^2}X_0
\end{equation}
has law given by
\begin{equation}
\label{eqn:prelimVPlaw}
    \rho_t := \operatorname{Law}(X_t) 
    =
    P_{-\log t}^* \pi,
    \quad t\in(0,1]
\end{equation}
and satisfies $\rho_0=\gamma_d$ and $\rho_1=\pi$. Here, we use the time change $\tau = - \log t$ to convert from transport time $t \in (0, 1]$ to OU time $\tau \in [0, \infty)$. Note that $\rho_t$ is absolutely continuous with smooth density, and if $\pi$ is also absolutely continuous then the relative Gaussian densities satisfy $\frac{\rho_t}{\gamma_d} = P_{-\log t}\left(\frac{\pi}{\gamma_d}\right)$.

Thus, \eqref{eqn:prelimVPlaw} gives a path from Gaussian noise to the target. Its deterministic probability flow velocity (see e.g. \cite[Theorem 6]{albergo2025stochastic}) is given by
\begin{equation}
\label{eqn:prelimreverseOUvelocity}
    v_t(x)
    =
    \frac{1}{t}
   \left( \nabla\log \frac{\rho_t}{\gamma_d}\right)(x),
    \quad t\in(0,1].
\end{equation}
Indeed, one may check
\begin{equation}
    \partial_t\rho_t
    +
    \nabla\cdot(\rho_t v_t)
    =
    0.
\end{equation}
The same velocity can be written in denoising form: Tweedie's formula (see equation (5.12) of \cite{albergo2023stochasticinterpolantsunifyingframework}) gives
\begin{equation}
\label{eqn:prelimtweedie}
    v_t(x)
    =
    \frac{
        \mathbb E[X_1\mid X_t=x]-tx
    }{1-t^2}.
\end{equation}
Hence, the reverse OU probability flow moves each noised sample toward its posterior estimate of the clean sample. These marginals $(\rho_t)_{t \in [0, 1]}$ are the same as those of the Anderson reverse SDE of a time-change of the OU process, see Section 5.1 of \cite{albergo2025stochastic}.

Given target samples $X_1\sim\pi$, we can sample the interpolation
(\ref{eqn:prelimVPinterpolation}) directly. Moreover, if
\begin{equation}
    U_t := \frac{X_1-tX_t}{1-t^2}
\end{equation}
then (\ref{eqn:prelimtweedie}) is equivalently
\begin{equation}
    v_t(x)=\mathbb E[U_t\mid X_t=x].
\end{equation}
Hence $v_t$ is the population minimizer of the square-loss regression problem
\begin{equation}
\label{eqn:prelimsscorematcingloss}
    \inf_v \mathbb E\left[\|v(t,X_t)-U_t\|^2\right]
\end{equation}
which can be learned using the finite dataset of target samples.
Equivalently one may regress the Gaussian noise or score. This gives the usual
{\em score-matching objective} used in (denoising) diffusion models, see
\cite{sohl2015deep,ho2020denoising,song2020score,albergo2025stochastic}.

Importantly, the usual score, noise-prediction, denoising, reverse-SDE, and probability flow
parameterizations of this Gaussian denoising path are equivalent up to explicit
affine transformations and time-dependent scalar factors. For concreteness, we work
throughout with the probability flow velocity in
(\ref{eqn:prelimreverseOUvelocity}), and we refer to the induced transport $(v_t, \rho_t)_{t \in [0, 1]}$ with $\rho_0 = \gamma_d$ and $\rho_1 = \pi$ as the {\em reverse OU transport to $\pi$}.
It is known that the reverse OU transport is not Wasserstein-optimal \cite{tanana2021comparison,LAVENANT2022108225}, and in this work we will ask whether this prescribed
transport is efficient in a neural network function space.

\paragraph{Flow-matching transport.}
More generally, one may prescribe an interpolation between the base and
target distributions for which the samplewise velocity is known (or estimated via finite-differences). If $X_t$ denotes
the interpolation and $\dot X_t$ its samplewise velocity, then the marginal velocity
field
\begin{equation}
    v_t(x)=\mathbb E[\dot X_t\mid X_t=x]
\end{equation}
satisfies the continuity equation for $\rho_t=\operatorname{Law}(X_t)$.
The field $v_t$ can again be learned by a square-loss regression using samples of
$(X_t,\dot X_t)$. This is the basic {\em flow-matching} construction
\cite{lipman2023flow,albergo2023building,albergo2025stochastic}.

\paragraph{Wasserstein optimal transport.}
One can instead consider the transport $(v_t, \rho_t)_{t \in [0, 1]}$ from $\mu$ to $\pi$ given by the induced $W_2$ optimal transport, i.e. the minimizer of (\ref{eqn:prelimBBenergy}). While this transport has some useful properties (particles move in straight lines, which can be helpful for discretization of the generative process), it can be hard to directly learn from target samples since the exact Wasserstein coupling is not
directly available. Practical flow-matching
methods therefore approximate it (for example using minibatch optimal transport
\cite{tong2024improving}), after which coupled points may be connected with linear interpolations and the drifts may be learned via flow matching. 

A common alternate dynamical approach, dubbed {\em rectified flow} \cite{liu2022flow}, is to first learn a generative model from samples using something like the score-matching objective (as described above). One then applies a rectification procedure which produces increasingly straight trajectories without increasing the Wasserstein transport cost. In practice, this can provide useful approximations to $W_2$ optimal transports.

\section{Barron Transport Metric}
\label{section:barron}
In the previous section, we framed the task of generative modeling as learning a collection of drifts $(v_t)_{t \in [0, 1]}$ which transport a known base measure $\mu$ (often the standard Gaussian $\gamma_d$) to the target data distribution $\pi$ via the continuity equation (\ref{eqn:prelimcontinuity}). We surveyed two different notions of dynamical optimal transport (Wasserstein and Stein) which describe the ``least complex"  transport from $\mu$ to $\pi$, where complexity of $v_t$ is measured in $L^2(\rho_t)$ in the Wasserstein case and in $\h_k$ in the Stein case. 

The Wasserstein construction is natural when the goal of an efficient transport is to minimize the movement/speed of particles, and the Stein construction is natural to describe coherent transports which are efficient with respect to a fixed set of features encoded in the kernel. However, since we will in practice be representing and approximating the drifts $v_t$ via neural networks, it makes sense to measure a transport's complexity using a function-space norm adapted to neural-network approximation.

This will be the subject of this section: in short, we define a metric on $\p(\r^d)$ which asks ``what is the minimal neural network size required to transport one measure to another"? To do so, we will leverage the notion of Barron spaces \cite{weinan}. 

\subsection{Abstract definition}
Before doing so, we first show that the dynamical length minimizations of (\ref{eqn:prelimBBlength}) and (\ref{eqn:prelimsteinlength}) can be generalized using any Banach norm (satisfying a mild regularity condition, see Assumption \ref{assumption:main}) to yield an extended metric on $\p(\r^d)$:

\begin{definition}
\label{def:abstractmetric}
    Let $\b$ be a Banach space of functions from $\r^d \to \r^d$ which satisfies Assumption \ref{assumption:main}. For every $\mu, \nu \in \p(\r^d)$, define the quantity
    \begin{equation}
        D_\b(\mu, \nu) := \inf_{\mu \to \nu} \int_0^1 \|v_t\|_\b dt,
    \end{equation}
    where the infimum is over $(v_t, \rho_t)_{t \in [0, 1]}$ transporting $\mu$ to $\nu$ via the continuity equation (\ref{eqn:prelimcontinuity}).
\end{definition}

In this work we will only prove the simplest properties of these metrics:

\begin{proposition}
\label{prop:abstractmetricsmainpaper}
     $(\p(\r^d), D_\b)$ is an extended metric space (i.e. the metric may take the value $+\infty$), and convergence in $D_\b$ implies narrow convergence of measures.
\end{proposition}
\begin{proof}
    See Section \ref{subsec:appendixproofofabstractmetrics}.
\end{proof}

 In the companion paper \cite{companionpaper} we study a more general setting where the tangent norm $\b$ is allowed to depend on $\rho_t$, in which case we give a thorough treatment of absolutely continuous curves, geodesics, and metric gradient flows using the methods of Ambrosio-Gigli-Savar\'e \cite{ags}.

\subsection{Barron spaces and their associated transport metrics}
\label{subsec:barronspacesdefinition}

Let $ \b_\phi$ be a representational Barron space, i.e. 
    \begin{equation}
    \label{eqn:abstractbarronspace}
        \|w\|_{\b_\phi} = \inf\left\{\|\mu\|_{\operatorname{TV}}:\; w(x) = \int_\Theta \phi_\theta(x) d\mu(\theta) \quad \forall x \in \r^d\right\}\end{equation}
    for some parameterization $\phi_\theta: \r^d \to \r$ and $\r^d$-valued signed measures $\mu$ on a parameter space $\Theta$. This is a Banach space of functions from $\r^d \to \r^d$, and if $\{\phi_\theta\}_{\theta \in \Theta}$ is uniformly bounded on compact sets then it satisfies Assumption \ref{assumption:main}. We define $D_{\b_\phi}$ according to Definition \ref{def:abstractmetric}, i.e.
    \begin{equation}
        \label{eqn:defofDbphi}
        D_{\b_\phi}(\mu, \nu) := \inf_{\mu \to \nu} \int_0^1 \|v_t\|_{\b_\phi}dt \quad (\forall \mu, \nu \in \p(\r^d)),
    \end{equation}
    and we refer to this as the {\em Barron transport metric}.

        \begin{remark}
        \label{remark:barronasinfoverRKHS}
    Note that since $\|\mu\|_{\operatorname{TV}} = \inf_{\substack{\eta \in \p(\Theta):\; \mu \ll \eta}} \|\frac{d\mu}{d\eta}\|_{L^2(\eta)}$, if for each $\eta \in \p(\Theta)$ we define the random feature kernel $k^{(\eta)}(x, y) := \int_\Theta \phi_\theta(x) \phi_\theta(y) d\eta(\theta) \cdot \1_d$, then for all $w$ we have
    \begin{equation}\|w\|_{\b_\phi} = \inf_{\substack{\eta \in \p(\Theta)}} \|w\|_{\h_{k^{(\eta)}}},\end{equation}
    see \cite[Proposition 3]{weinan}.
    In this way, the Barron norm is an infimum over random feature RKHS norms, and the optimal distribution over features is allowed to depend on $w$. Hence,
    \begin{equation}D_{\b_\phi}(\mu, \nu) \leq \inf_{\eta \in \p(\Theta)} D_{\h_{k^{(\eta)}}}(\mu, \nu) \quad (\forall \mu, \nu \in \p(\r^d)).\end{equation}
    This need not be an equality, since the choice of the optimal $\eta$ need not be the same for all drifts along an optimal flow.
\end{remark}
    
We will be focused on parameterizations of the form $\phi_\theta(x) = \phi(\la a, x\ra + b)$ for a continuous $\phi: \r \to \r$ and parameters $\theta = (a, b)$ which describe two-layer neural networks.\footnote{Technically, to represent the concrete Barron spaces of Definitions \ref{def:barronspaces} and \ref{def:barronspaceshomogeneous} in the form of (\ref{eqn:abstractbarronspace}) we would take $\phi_\theta(x) = \phi(\la a, x\ra + b) / (1 + \|a\| + |b|)$ when $\phi$ is bounded and $\phi_\theta(x) = \phi(\la a, x\ra + b) / ( \|a\| + |b|)^s$ when $\phi$ is positively $s$-homogeneous.}

\begin{assumptionp}{$\phi$}
\label{assumption:activationfunciton}
    Throughout the rest of this paper, we assume the activation $\phi: \r \to \r$ is continuous and either bounded or positively $s$-homogeneous for some $s > 0$ (i.e. $\phi(\lambda z) = \lambda^s\phi(z)$ for all $\lambda \geq 0$). 
\end{assumptionp}

  We will use the following two formulations of Barron space, depending on whether $\phi$ is bounded (e.g. $\tanh$, $\sin$, etc.) or positively $s$-homogeneous (e.g. a monomial, a power of the ReLU or leaky ReLU, etc.). In the bounded case we may need to additionally constrain the size of the neurons $\|a\| + |b| \leq R$, so we allow our definition to accommodate this; unless otherwise stated we always take $R = \infty$.
    
    \begin{definition}[Barron space, bounded]
    \label{def:barronspaces}
    Let $\phi: \r \to \r$ be continuous and bounded, and let $R \in (0, \infty]$.
        We denote by $\b_\phi^R$ the Banach space of vector fields $v: \r^d \to \r^d$ with norm
        \begin{equation}\|v\|_{\b_\phi^R} = \inf_\mu \left\{\int_{\r^d \times \r} (1 + \|a\| + |b|) d|\mu|(a, b):\; v(\cdot) = \int_{\r^d  \times \r} \phi(\la a, \cdot\ra + b)d\mu(a, b)\right\}\end{equation}
        where $\mu$ ranges over signed vector-valued measures on $ \r^d \times \r$ s.t. $\|a\| + |b| \leq R$ for $\mu$-a.e. $(a, b)$. If $R = \infty$ we denote this space by $\b_\phi$ for simplicity.
        
        In particular, the finite two-layer neural network with $\mu_j, a_j \in \r^d$ and $b_j \in \r$ given by
        \begin{equation}v(\cdot) = \sum_{j=1}^m \mu_j \phi(\la a_j, \cdot\ra + b_j)\end{equation}
        has Barron norm $\|v\|_{\b_\phi} \leq \sum_j \|\mu_j\| (1 + \|a_j\| + |b_j|)$.
    \end{definition}
    \begin{definition}[Barron space, homogeneous]
    \label{def:barronspaceshomogeneous}
    Let $\phi: \r \to \r$ be continuous and positively $s$-homogeneous for some $s > 0$. We denote by $\b_\phi$ the Banach space of vector fields $v: \r^d \to \r^d$ with norm
        \begin{equation}\|v\|_{\b_\phi} = \inf_\mu \left\{\int_{\r^d \times \r} (\|a\| + |b|)^s d|\mu|(a, b):\; v(\cdot) = \int_{\r^d  \times \r} \phi(\la a, \cdot\ra + b)d\mu(a, b)\right\}\end{equation}
        where $\mu$ ranges over signed vector-valued measures on $ \r^d \times \r$.
        
        In particular, the finite two-layer neural network with $\mu_j, a_j \in \r^d$ and $b_j \in \r$ given by
        \begin{equation}v(\cdot) = \sum_{j=1}^m \mu_j \phi(\la a_j, \cdot\ra + b_j)\end{equation}
        has Barron norm $\|v\|_{\b_\phi} \leq \sum_j \|\mu_j\| ( \|a_j\| + |b_j|)^s$.
        
    Note that if $\phi(z) = z$ is the identity then every $v \in \b_\phi$ must be affine $v(x) = Mx + c$, in which case $\|v\|_{\b_\phi} = \|M\|_* + \|c\|$ with $\|\cdot\|_*$ the nuclear norm.
    \end{definition}
    \begin{remark}
        We will abuse notation and use $\b_\phi$ to denote both of these Banach spaces, where we implicitly use the correct definition depending on whether $\phi$ is bounded or positively $s$-homogeneous. We will only ever refer to $\b_\phi^R$ in the bounded case, and unless otherwise stated we always use $\b_\phi$ (i.e. the $R = \infty$ case).
    \end{remark}

The Barron norm is an atomic variation norm whose atoms are normalized individual neurons; finite two-layer networks correspond to finite atomic decompositions, while the Barron space permits arbitrary mixtures. Unlike an RKHS with a fixed feature distribution, the Barron norm allows the relevant features to be chosen adaptively.
 As we will see, this $D_{\b_\phi}$ metric can (1) help us understand the dynamical transport of measures using two-layer mean-field neural network drifts and (2) produce interesting adaptive sampling algorithms. The first point is the main subject of this paper, while the second point will be the subject of the companion paper \cite{companionpaper}.

  In general $\b_\phi$ is a Banach rather than Hilbert space, so unlike the Stein geometry there is no canonical Riemannian structure which we use here. Despite this, the spaces $(\p(\r^d), D_{\b_\phi})$ enjoy many dynamical properties akin to those of $W_2$ and $D_{\h_k}$; the companion paper develops the metric geometry of these spaces in greater detail.

\subsection{Interpretation as a ResNet norm}

The Barron transport metric has a direct interpretation in terms of deep residual networks (ResNets). Recall that a ResNet with $L$ residual blocks can be written in the form
\begin{equation}
    h^{\ell+1}(x)
    =
    h^\ell(x)
    +
    \frac{1}{L}v_\ell(h^\ell(x)),
    \quad
   ( \ell\in \{0,\ldots,L-1\}),
\end{equation}
where each $v_\ell:\r^d\to\r^d$ is represented by a neural network. As $L\to\infty$, under some regularity this yields the Euler discretization of the {\em neural ODE}
\begin{equation}
    \dot X_t=v_t(X_t), \quad (t \in [0,1])
\end{equation}
under the relabeling $t \leftrightarrow \ell / L$. This continuous-depth interpretation of ResNets as neural ODEs is standard, see e.g. \cite{weinan} and the quantitative $L \to \infty$ convergence theory of Chizat \cite{chizat2026hiddenwidthdeepresnets}.

Suppose that the residual blocks $v_\ell$ are two-layer networks with activation $\phi$. A natural intrinsic complexity measure for the ResNet is
\begin{equation}
    \frac{1}{L}\sum_{\ell=0}^{L-1}\|v_\ell\|_{\b_\phi}.
\end{equation}
In the continuous-depth limit this becomes
\begin{equation}
\label{eqn:continuousresnetpathnorm}
    \int_0^1\|v_t\|_{\b_\phi}dt.
\end{equation}
Thus, the quantity appearing in the definition of $D_{\b_\phi}$ is exactly the continuous-depth Barron ResNet norm of the corresponding neural ODE.

This also has a direct interpretation in terms of network parameters. For example, in the case of a bounded activation, if
\begin{equation}
    v_\ell(x)
    =
    \sum_{j=1}^{m_\ell}
    \mu_{j,\ell}\phi(\la a_{j,\ell},x\ra+b_{j,\ell}),
\end{equation}
is a finite-width two-layer network then Definition \ref{def:barronspaces} gives
\begin{equation}
    \|v_\ell\|_{\b_\phi}
    \leq
    \sum_{j=1}^{m_\ell}
    \|\mu_{j,\ell}\|
    \left(1+\|a_{j,\ell}\|+|b_{j,\ell}|\right).
\end{equation}
Hence the ResNet norm is bounded by the layer-averaged parameter norm of the residual blocks. The analogous statement holds for positively $s$-homogeneous activations using Definition \ref{def:barronspaceshomogeneous}.
Consequently, the quantity
\begin{equation}
\label{eqn:barronresnetinterpretation}
    D_{\b_\phi}(\mu,\nu)
    =
    \inf_{\mu\to\nu}
    \int_0^1\|v_t\|_{\b_\phi}dt
\end{equation}
can be interpreted as the minimum continuous-depth ResNet norm required to transport $\mu$ to $\nu$ using a neural ODE with two-layer blocks.

This is the main reason we use the Barron transport geometry to study generative modeling: $D_{\b_\phi}(\mu,\nu)$ measures the complexity of the simplest neural dynamics connecting the two distributions, rather than the amount by which particles move or the complexity of the dynamics in a fixed feature space. In particular, the resulting $D_{\b_\phi}$ balls around $\mu$ consist of those measures which admit a neural network generative model of bounded size in the following sense:
\begin{itemize}
    \item Lower bounds on $D_{\b_\phi}(\mu, \nu)$ correspond to lower bounds on the average Barron norm of each layer of any deep ResNet/neural ODE which transports $\mu$ to $\nu$. 
    \item Conversely, upper bounds on $D_{\b_\phi}(\mu, \nu)$ prove the existence of a deep ResNet/neural ODE transporting $\mu$ to $\nu$ whose layer-averaged Barron norm is upper bounded. 
\end{itemize}
Thus, bounds on $D_{\b_\phi}(\mu, \nu)$ provide a quantitative understanding of the approximation complexity for dynamical transport from $\mu$ to $\nu$ using neural ODEs. 

\begin{question}
   One could also define a finite-depth distance via $D_{\b_\phi}^{(L)}(\mu, \nu) := \inf \frac{1}{L}\sum_{\ell = 1}^L \|v_\ell\|_{\b_\phi}$, where the infimum is over blocks $(v_\ell)_{\ell = 1}^L$ s.t. $F_\# \mu = \nu$ with $F := (\1 + \frac{1}{L}v_L) \circ \ldots \circ (\1 + \frac{1}{L}v_1)$. Quantifying the convergence $D_{\b_\phi}^{(L)} \to D_{\b_\phi}$ as $L \to \infty$ is useful to transfer our $D_{\b_\phi}$ results to real-world ResNets, which we leave for future work.
\end{question}

Our perspective of a continuous-depth ResNet norm via a geometry on $\p(\r^d)$ with tangent norm $\|\cdot\|_{\b_\phi}$ is complementary to the approaches of \cite{weinan, jacot2026deeplearningconvexparadigm}. These works study the approximation properties of deep ResNets by carefully designing function spaces that see the whole ResNet as a single function, which requires some way to capture the sensitivity of function outputs w.r.t. changes at intermediate layers. In \cite{weinan} this is needed to control the $L^2$ approximation error of finite-width ResNets, since error accumulates across depth in a way that leads to a path norm/multiplicative norm. In \cite{jacot2026deeplearningconvexparadigm} this is used to describe circuit complexity of a function in terms of its ResNet representations.
By contrast, the $D_{\b_\phi}$ metric leverages the energy-minimization perspective of dynamical optimal transport with fixed endpoints to encode depth, even though the cost itself is local and additive across depth. 

\begin{remark}
    The relaxation to measures may also be useful beyond generative modeling in the study of ResNet approximation theory. For example, any infinitely-deep and infinitely-wide ResNet (i.e. a neural ODE) with bounded activations which interpolates a dataset whose inputs have marginal distribution $\mu$ and whose outputs have marginal distribution $\nu$ must have norm at least $ D_{\b_\phi}(\mu, \nu)$ (which is $\gtrsim W_2(\mu, \nu)$). By using the results of \cite{chizat2026hiddenwidthdeepresnets}, one should be able to get an analogous statement for ResNets of finite depth and width which have been trained for finite time. Alternatively, by using the results of \cite{jacot2026deeplearningconvexparadigm} one can hope to understand the minimal circuit complexity needed for such a task. This offers an approach to ResNet approximation lower bounds in terms of the geometry of the data distributions themselves, which we leave for future work.\footnote{Note that such an approach discards information about any coupling between inputs and outputs. Lower bounds would therefore be weaker, as they would capture the difficulty of transforming one distribution into another (as opposed to transforming each input into its corresponding output). The power of this perspective is that by relaxing to measures, we can naturally encode model depth through continuity equation dynamics while still using simple Barron technology.}
\end{remark}

\subsection{Comparisons with other notions of distance}

As we increase the regularity of $\phi$ we also increase the regularity of functions in $\b_\phi$, which can cause the distance $D_{\b_\phi}$ to increase. A useful first question to ask is how strong the $D_{\b_\phi}$ distance is under regularity assumptions on $\phi$, since this can help contextualize any quantitative results we obtain regarding $D_{\b_\phi}$. 
We note here a particularly clean lower bound on the Barron transport metric $D_{\b_\phi}$ in terms of either Wasserstein distance or KL divergence: the latter is interesting since it forgets all geometric information about a target distribution.

\begin{proposition}[Wasserstein and KL lower bounds]
\label{prop:wassersteinkllowerbound}
    Let $\phi: \r \to \r$ satisfy Assumption \ref{assumption:activationfunciton} and suppose that $\|\phi\|_\infty\leq M < \infty$. Then, there is the Wasserstein lower bound
    \begin{equation}
        D_{\b_\phi}(\mu, \nu) \geq \frac{W_\infty(\mu, \nu)}{M} \quad (\forall \mu, \nu \in \p(\r^d)).
    \end{equation}
    If we assume further that $\phi$ is $M$-Lipschitz then there is the KL lower bound
\begin{equation}
        D_{\b_\phi}(\gamma_d, \pi) \geq \frac{1}{4M} \cdot \min\left\{\frac{\operatorname{KL}(\pi \| \gamma_d)}{\sqrt{d}},\; \sqrt{\operatorname{KL}(\pi \| \gamma_d)}\right\} \quad (\forall \pi \in \p(\r^d)).
    \end{equation}
\end{proposition}
\begin{proof}
    See Section \ref{subsec:proofofWassersteinKLlowrbounds}. Similar lower bounds hold for any $D_\b$ metric with $\b \hookrightarrow L^\infty$ and $\b \hookrightarrow W^{1,\infty}$, respectively, as well as with $\gamma_d$ replaced by any log-smooth measure with sufficient tail control, see Lemma \ref{lemma:appendixwassersteinandKLcontrol}.
\end{proof}

This result helps us understand the foliation structure of $(\p(\r^d), D_{\b_\phi})$ in the case of bounded $\phi$, since any measures $\mu, \nu$ which have infinite $W_\infty$ distance (or KL under Lipschitzness and sufficient regularity of either $\mu$ or $\nu$) also have infinite $D_{\b_\phi}$ distance. For example, $D_{\b_\phi}(\mu, \nu) = +\infty$ if $\mu$ has compact support but $\nu$ does not, if $\mu$ and $\nu$ are mutually singular, if $\mu$ and $\nu$ have different tail scales at infinity, etc.
    
\subsection{Optimal transport distances between Gaussians}
In some of our main results, we will be comparing the Stein $D_{\h_k}$ and Barron $D_{\b_\phi}$ metrics to gain some intuition about the complexity of generative modeling.
In order to perform an initial comparison between these distances, we will inspect the case where both endpoints are Gaussians. If we restrict to bilinear kernels $k$ and linear activations $\phi$, the spaces $\h_k$ and $\b_\phi$ consist of affine drifts and so a Gaussian initial measure remains Gaussian along every finite-length admissible flow. This reduces the infinite-dimensional minimization in (\ref{eqn:defofDbphi}) to a tractable finite-dimensional matrix control problem, for which we can get matching upper and lower bounds. For further simplicity we restrict one of the endpoints to be the standard Gaussian $\gamma_d$, leading to:

\begin{proposition}
\label{prop:optimalgaussiantogaussian}
    Let $\pi := \mathcal{N}(m, \Sigma)$ with $\Sigma \succ 0$. Then, the following hold up to universal constants:
    \begin{itemize}
        \item \textbf{(Wasserstein). }$W_2(\gamma_d, \pi) = \sqrt{\|m\|^2 + \|\Sigma^{1/2} - \1_d\|_F^2}$.
        \item \textbf{(Stein).} $D_{\h_k}(\gamma_d, \pi) \asymp \log(1 + \|m\|) + \|\log \Sigma\|_F$ for $k(x, y) = (1 + \la x, y\ra)\1_d$.
        \item \textbf{(Barron).} $D_{\b_\phi}(\gamma_d, \pi) \asymp \log(1 + \|m\|) + \|\log \Sigma\|_*$ for $\phi(z) = z$.
    \end{itemize}
    Here, $\|\cdot\|_F$ is the Frobenius norm and $\|\cdot\|_*$ is the nuclear norm.
\end{proposition}
\begin{proof}
This follows from Lemmas \ref{lemma:W2distancebetweengaussians}, \ref{lemma:steingaussiantogaussian}, and \ref{lemma:barrongaussiantogaussian}. Interestingly, the transports realizing the upper bounds for linear Stein and linear Barron are almost the same. They move the mean and covariance in separate motions: the mean movement uses a spatial preconditioning by $mm^\top$ to get the logarithmic dependence, while the log-covariance linearly interpolates from $0$ to $\log \Sigma$. See Remark \ref{remark:steinbarronlineargaussiantogaussianmeanmovement} for more details.
\end{proof}

Compared to the Wasserstein distance, the linear Stein and Barron distances have improved dependence on $m$ (and $\Sigma$ if $\Sigma \succeq \1_d$). On the other hand, if there is a covariance eigenvalue $\sigma^2 \ll 1$ then linear Stein and Barron blow up like $\log \frac{1}{\sigma^2}$ while Wasserstein stays bounded. Furthermore, we have $D_{\h_k} \lesssim D_{\b_\phi}$ with qualitatively similar behaviors.
\begin{remark}
    Note that by Proposition \ref{prop:wassersteinkllowerbound}, if $\|\phi\|_\infty < \infty$ then $D_{\b_\phi}(\gamma_d, \pi) = +\infty$ whenever $\Sigma \neq \1_d$, and when the covariances are equal we have $D_{\b_\phi}(\gamma_d, \pi)  \gtrsim \|m\|$. By contrast, Proposition \ref{prop:optimalgaussiantogaussian} shows that an activation with linear growth allows a logarithmic cost of translation and can connect two Gaussians of different covariance. Hence, using an activation with linear growth instead of a bounded one makes an affine shift exponentially easier to implement via a neural network flow (at least in the Gaussian case).
\end{remark}

\section{Approximation Theory for Generative Paths}

Having introduced the Barron transport metric $D_{\b_\phi}$, we now use it to study two approximation questions for generative modeling. 
 In practice, we choose $\mu$ to be easy to sample from (such as the standard Gaussian $\gamma_d$) and $\nu$ a potentially complicated data distribution (which we will denote $\pi$).

First, we compare the reverse OU transport to the Barron optimal transport in order to study the role of depth in denoising models.
In particular, in Section \ref{subsec:reverseOU} we ask when the reverse OU transport is optimal (up to constants) w.r.t. the Barron energy. We answer in the affirmative for centered Gaussians and single-index tilts, but in the negative for noncentered Gaussians and single-index flows (Proposition \ref{prop:reverseOUgaussiantogaussian} and Example \ref{example:reverseOUsingleindexflowsandtilts}). To clarify the situation, we construct a measure $\pi$ (a pushforward of $\gamma_d$ by a deep narrow ResNet) for which the denoising path has exponential $\b_\phi$ path norm, but which a simple time-varying single-index flow can easily generate and so $D_{\b_\phi}(\gamma_d, \pi)$ is small (Theorem \ref{theorem:mainpaperrepresentaitonlowerbound}). 
This suggests that the reverse OU transport is suboptimal for modeling neural network pushforwards of the Gaussian, but the obstruction we prove is in terms of exact representation and only for small OU times. 
To finish, we strengthen this result by constructing another neural network pushforward $\pi$ for which any shallow neural network that $\epsilon$-approximates the reverse OU drifts in $L^2$ must have width $d^{\operatorname{poly}(1/\epsilon)}$ (Theorem \ref{theorem:mainpaperapproximationlowerbound}). Hence, any shallow score networks which achieve small score-matching loss on this data must have large width, irrespective of how they are learned. This implies that deeper drift networks are needed to enable efficient generative modeling for these data distributions.

 In Section \ref{subsec:featurelearning} we instead aim to quantify the benefit of feature learning by comparing the Barron $D_{\b_\phi}$ distance to the Stein $D_{\h_k}$ distance, where $\h_k$ is the RKHS of a rotation-equivariant kernel $k$. We already saw that for Gaussian-to-Gaussian transports with linear $\phi$ and bilinear $k$ they behave similarly (Proposition \ref{prop:optimalgaussiantogaussian}). However, there are also the classical feature learning separation results of e.g. \cite{bach2017breaking,ghorbani2021when} which exhibit functions that are easy to represent in Barron spaces (i.e. using shallow neural networks) but are difficult to represent in rotation-invariant RKHS (i.e. using random feature models). In the generative modeling setting, we show that perturbatively-small single-index pushforwards/flows can yield measures in high dimension for which the ratio ${D_{\h_k}}/{D_{\b_\phi}}$ becomes very large (Examples \ref{example:nonlinearshearrkhs} and \ref{example:nonlinearshearrkhssmoothed}), but that this separation breaks down in the nonperturbative regime since deep RKHS transports can model single-index flows (Example \ref{example:mainpapernonperturbativefeaturelearning}). In other words, the classical shallow feature learning separation holds locally but fails once depth is introduced.
These preliminary results suggest that RKHS-based generative models can be fairly expressive, and that a separation with neural network generative models (if it exists) might instead come from the learning procedure rather than approximation capabilities. 
For now the comparison between RKHS-based and neural network-based generative modeling remains poorly understood, and we consider it important to compare these paradigms from both approximation and computational perspectives.

To summarize, our results point toward two themes in the approximation theory of generative modeling/dynamical measure transport: 
\begin{itemize}
    \item The reverse OU/denoising transport (which is common in practice) cannot always be efficiently modeled using shallow neural networks, even when there exist other transports which can be. In such settings, non-denoising depth is required.
    \item The Barron optimal transport is an efficient alternative when data has preferred directions/low intrinsic dimension, and an RKHS optimal transport which leverages depth may enjoy similar benefit. 
\end{itemize}
Taken together, these results demonstrate that the Barron/RKHS dynamical optimal transports have desirable properties for generative modeling that the current denoising approach can fail to capture.
Our study of the approximation question therefore motivates a generative modeling approach based on algorithms which (approximately) solve the $D_{\b_\phi}$ or $D_{\h_k}$ optimal transport problems; in Section \ref{subsec:algorithms} we give some heuristic algorithmic ideas, but the computational question is wide open and left for future work.

\subsection{Reverse OU (sub)optimality}
\label{subsec:reverseOU}
As mentioned in Section \ref{subsec:prelimgenerativepaths}, the reverse OU transport from $\gamma_d$ to $\pi$ is a common design choice for generative models. 
A natural question to ask is whether this transport is efficient to implement with shallow neural networks, or whether there can exist more economical alternative transports using smaller shallow networks. 
Another way to phrase this question is to ask whether the reverse OU transport is optimal in terms of the $D_{\b_\phi}$ energy (\ref{eqn:defofDbphi}), perhaps up to multiplicative constants. To help us approach this question, we define the following quantity:

\begin{lemma}[Condition for reverse OU (sub)optimality]
\label{lemma:reverseOUsuboptimality}
Let $\pi \in \p(\r^d)$ and $\b$ a Banach space of functions from $\r^d \to \r^d$, and define
\begin{equation}
    \mathcal{E}_{\b}(\pi) := \int_0^\infty \left\|\nabla \log \frac{P_\tau^* \pi}{\gamma_d}\right\|_{\b}d\tau,
\end{equation}
where $(P_\tau)_{\tau > 0}$ is the OU semigroup. If $(v_t)_{t \in [0, 1]}$ is any time-reparameterization\footnote{By this we mean that $v_t =  \frac{\dot\alpha(t)}{\alpha(t)} \nabla\log \frac{\rho_{\alpha(t)}}{\gamma_d}$, where $\rho_s = P_{-\log s}^* \pi$ and $\alpha: [0, 1] \to [0, 1]$ is any absolutely continuous, strictly increasing function with $\alpha(0) = 0$, $\alpha(1) = 1$.} of the reverse OU transport from $\gamma_d$ to $\pi$, then 
\begin{equation}
    \mathcal{E}_{\b}(\pi) = \int_0^1 \|v_t\|_{\b}dt \geq D_{\b}(\gamma_d, \pi).
\end{equation}

\end{lemma}
\begin{proof}
    This follows from reparameterizing from transport time $t \in [0, 1]$ to OU time $\tau \in [0, \infty)$.
\end{proof}

Naturally $D_{\b}(\gamma_d, \pi) \leq \mathcal{E}_{\b}(\pi)$, and they are comparable if and only if the reverse OU/denoising transport from $\gamma_d $ to $ \pi$ has complexity (w.r.t. $\b$) comparable to the optimal one. 
Hence, we can use this quantity $\mathcal{E}_{\b}(\pi)$ to understand the (sub)optimality of the reverse OU generative model for various choices of target measure $\pi \in \p(\r^d)$. 

As before, we will consider different target measures $\pi$ of increasing complexity. We start with the linear Gaussian-to-Gaussian case, followed by simple nonlinear single-index perturbations of the Gaussian, after which we look at nonperturbative pushforwards of the Gaussian by deep narrow ResNets. 
In each setting, we will focus on the Barron case $\b = \b_\phi$ in order to draw conclusions about the implementation of score-matching diffusion/denoising models using shallow neural network drifts.\footnote{Note that since the Barron norm is an infimum over random-feature RKHS norms (see Remark \ref{remark:barronasinfoverRKHS}), many of these results will naturally extend to the corresponding kernel settings.}

We will see that for very simple data (pushforwards of the Gaussian by a small teacher neural network) the reverse OU transport can fail to see the correct relationship between coordinates, as the relevant structure is not visible by varying the noise level. In such cases, it instead uses brute force and requires very wide networks to represent or approximate the OU scores, even though there exists a small alternative generative model (namely, the teacher network). These constructions provide a good toy model of a type of obstruction which denoising models with shallow neural network drifts can encounter on simple data. Our hope is that they can (1) help us understand the structures in real-world data which influence the role of depth in denoising models and (2) guide the design of other alternative generative modeling paradigms; we leave these investigations to future work.

\subsubsection{Gaussian targets}

We again start by inspecting the setting of linear drift function spaces and $\pi$ a Gaussian:
\begin{proposition}[Reverse OU, Gaussian-to-Gaussian]
\label{prop:reverseOUgaussiantogaussian}
    Let $\pi := \mathcal{N}(m, \Sigma)$ with $\Sigma \succ 0$. Then, the following hold up to universal constants:
    \begin{itemize}
        \item \textbf{(Wasserstein).} $\mathcal{E}_{W_2}(\pi) \asymp W_2(\gamma_d, \pi)$.\footnote{Here, we denote $\mathcal{E}_{W_2}(\pi) := \int_0^\infty \|\nabla \log \frac{P_\tau^* \pi}{\gamma_d} \|_{L^2(P_\tau^*\pi)}d\tau$ to be the unsquared Benamou-Brenier kinetic energy of the reverse OU transport.}
        \item \textbf{(Stein).} 
        If $k(x, y) = (1 + \la x, y\ra)\1_d$ then
\begin{equation}
        \mathcal{E}_{\h_k}\left(\pi\right) \gtrsim \frac{\|m\|}{\max\{1, \;\|\Sigma\|_{\operatorname{op}}\}} + \frac{1}{2}\|\log \Sigma\|_F.\end{equation}
        However, if $m = 0$ and $\Sigma = \operatorname{diag}(e^{d/\sqrt{1}}, \ldots, e^{d / \sqrt{d}})$ then $\mathcal{E}_{\h_k}\left(\pi\right) \gtrsim \sqrt{\log d} \cdot D_{\h_k}(\gamma_d, \pi)$.
        \item \textbf{(Barron).}
        If $\phi(z) = z$ then
\begin{equation}
        \mathcal{E}_{\b_\phi}\left(\pi\right) \gtrsim \frac{\|m\|}{\max\{1, \;\|\Sigma\|_{\operatorname{op}}\}} + \frac{1}{2}\|\log \Sigma\|_*.\end{equation}Furthermore, if $m = 0$ then $\mathcal{E}_{\b_\phi}\left(\pi\right) \asymp D_{\b_\phi}(\gamma_d, \pi)$.
    \end{itemize}
\end{proposition}
\begin{proof}
    This follows from Lemma \ref{lemma:reverseOUgaussiantogaussian} and Example \ref{example:reverseOUsteinGaussiansuboptimality}.
\end{proof}

Hence, for Gaussians the reverse OU transport is always optimal up to constants w.r.t. $W_2$, and for centered Gaussians it is also optimal up to constants w.r.t. linear Barron (but curiously not linear Stein). However, when moving the mean this transport can be arbitrarily suboptimal w.r.t. linear Stein/Barron (recall from Proposition \ref{prop:optimalgaussiantogaussian} that the optimal transport should have logarithmic dependence on $\|m\|$).

\subsubsection{Perturbative single-index flows and tilts}
As our simplest nonlinear examples, we consider targets $\pi$ which are perturbations of the standard Gaussian by single-index functions. We consider two types of perturbations: flowing by a single-index vector field and tilting by a single-index scalar function. 

\begin{example}[Reverse OU, perturbative single-index flows and tilts]
\label{example:reverseOUsingleindexflowsandtilts}
    Fix a globally-Lipschitz $\phi: \r \to \r$ which is bounded below, and fix any $\sigma \in (0, \infty)$ and $u, \theta \in \mathbb{S}^{d-1}$ s.t. $|\la u, \theta\ra| < 1$. For $\epsilon > 0$ define the flow perturbation
    \begin{equation}
        \label{eqn:flowsandtiltsexampleflow}\pi_\epsilon^{\operatorname{flow}} := \operatorname{Law}(X_\epsilon), \quad \dot{X}_t = \sigma u \phi(\la X_t, \theta\ra), \quad X_0 \sim \gamma_d
    \end{equation}
    and for $\epsilon > 0$ small enough that $1 + \epsilon h_\phi \geq 0$ define the tilt perturbation
    \begin{equation}
        \pi_\epsilon^{\operatorname{tilt}} := (1 + \epsilon h_\phi) \gamma_d, \quad h_\phi(x) := \phi(\la x, \theta\ra) - \mathbb{E}_{\xi \sim \mathcal{N}(0, 1)} [\phi(\xi)].
    \end{equation}
    If $\phi \in \{\operatorname{ReLU},\tanh, \sin\}$ then it holds that
    \begin{equation}
        \liminf_{\epsilon \downarrow 0} \frac{\mathcal{E}_{\b_\phi}\left({\pi_\epsilon^{\operatorname{flow}}}\right)}{D_{\b_\phi}(\gamma_d, \pi_\epsilon^{\operatorname{flow}})} = +\infty,\quad 
        \limsup_{\epsilon \downarrow 0} \frac{\mathcal{E}_{\b_\phi}\left({\pi_\epsilon^{\operatorname{tilt}}}\right)}{D_{\b_\phi}(\gamma_d, \pi_\epsilon^{\operatorname{tilt}})} \leq 23.
    \end{equation}
\end{example}
\begin{proof}
    See Section \ref{subsec:proofofreverseOUsingleindexflowsandtilts}. Morally, the measure $\pi_\epsilon^{\operatorname{flow}}$ from (\ref{eqn:flowsandtiltsexampleflow}) has a reverse OU transport whose drifts have a term like $\theta \la \cdot, u\ra \phi'(\la \cdot, \theta\ra)$; this is a product of two neurons, and it is not representable by two-layer networks for the activations we consider. This obstruction does not occur in the case of tilts, in which the reverse OU transport has a term like $\theta \phi'(\la \cdot, \theta\ra)$.
\end{proof}

Hence, the reverse OU transport is Barron-suboptimal for small single-index flows/pushforwards of $\gamma_d$, whereas it is Barron-optimal (up to constants) for small single-index tilts.
This example hints that in nonlinear settings the reverse OU transport can sometimes be suboptimal w.r.t. the $L^1([0, 1] \to \b_\phi)$ energy, but the mechanism identified here is perturbative and somewhat opaque.

\subsubsection{Intuition: triangular shear measures}
In the following subsections, we aim to clarify the situation with nonperturbative constructions of target distributions $\pi \in \p(\r^d)$ for which the reverse OU drifts (i.e. the time-dependent scores along the OU process, see Section \ref{subsec:prelimgenerativepaths}) are difficult to model with shallow neural networks despite the fact that $D_{\b_\phi}(\gamma_d, \pi)$ is small (i.e. there are alternative generative models for $\pi$ with simple shallow drifts). The high-level intuition is that there can exist compositional/hierarchical structure in the data which is not visible by varying the noise level, in which case the reverse OU transport fails to leverage it. We will show that this arises naturally for data which is drawn from a pushforward of the Gaussian by a deep neural network.

A good toy example to understand this mechanism is a simple construction in which $\pi \in \p(\r^d)$ is a pushforward of the Gaussian $\gamma_d$ by a triangular shear map $T: \r^d \to \r^d$ of the form
\begin{equation}
\label{eqn:mainpapertriangularshearmap}
    T(z, u) := (z, u + \Psi(z)), \quad (z \in \r^{d-1}, u \in \r)
\end{equation}
for some function $\Psi: \r^{d-1} \to \r$. Note that $\pi^\Psi := T_\# \gamma_d \in \p(\r^d)$ has Gaussian density \begin{equation}\frac{\pi^\Psi}{\gamma_d}(z, u) = \exp(u\Psi(z) - \Psi(z)^2/2),\end{equation} and so the last coordinate of the Gaussian score $\nabla \log \frac{\pi^\Psi}{\gamma_d}$ is equal to $\Psi$. 
Hence, the reverse OU drifts must model $\Psi$ (or an OU-smoothed version of it) and so they inherit $\Psi$'s complexity. 

Heuristically, the benefit of dynamical measure transport is that it allows us to avoid learning an entire endpoint map by instead learning drifts and using composition to increase the modeling capacity; this is analogous to how deep neural networks often have improved efficiency over shallow ones \cite{eldan2016power,telgarsky2016benefits}. The beauty of the reverse OU/denoising generative model is that it reduces the task from learning the entire transport to regression of the drifts \cite{chen2023sampling}, which in our analogy may be viewed as a form of layerwise supervision of a deep model. However, the observation of the previous paragraph implies that for pushforwards by triangular shears like (\ref{eqn:mainpapertriangularshearmap}), the scores are as complex as the  endpoint map itself. In our analogy, this means that the reverse OU supervision destroys the natural modeling benefit of depth which dynamical transport usually provides.

Of course, this intuition we are describing need not hold at positive OU times $\tau > 0$, since the actual drifts involve the scores of the OU-smoothed target $P_\tau^* \pi^\Psi$ and may be simpler. Thus, the technical part of our task is to show that the complexity of $\Psi$ persists for a non-negligible amount of OU time. In Section \ref{subsec:appendixtriangularshearmeasures} we identify two different mechanisms for this:
\begin{itemize}
    \item \textbf{(Smoothness).} By some elementary OU stability estimates, we show in Lemma \ref{lemma:stupidnonsenserevised} that large derivatives of $\Psi$ yield large derivatives of the drifts for positive OU times. Although Gaussian smoothing removes oscillations quickly, we find that the size of the derivative outweighs the rate of smoothing for derivatives of order at least 3. This third derivative obstruction becomes a Barron norm obstruction for activations with bounded third derivative, in which case $\mathcal{E}_{\b_\phi}(\pi^\Psi)$ can become large (see Corollary \ref{corollary:representationlowerbound}). 
    
    We emphasize that this lower bound is only encountered when exactly representing the full OU path and need not show up in practice. For example, if $\Psi$ has oscillations of frequency $\exp(d)$ then the third derivative of the score is of size $\exp(3d)$ for OU times $\tau \lesssim \exp(-2d)$; with any early-stopping which is $\gtrsim \exp(-2d)$ this obstruction is avoided. Moreover, by Girsanov one need only learn $L^2$-approximate drifts for accurate sampling \cite{chen2023sampling}, which means that an approximate generative model does not need to match the third derivatives of the score. Hence, while this mechanism suggests reverse OU suboptimality for some triangular shear measures, it leaves a lot to be desired.

    \item \textbf{(Hermite rank).} In \cite[Theorem 4.18]{bruna2026multiscale}, it was proved that $L^2$ approximation of a function using two-layer neural networks encounters a quantitative width lower bound in terms of the spectral tails of the function's Hermite coefficient tensors; this is because shallow networks of small width have low-rank Hermite coefficients. Importantly, since the Wiener chaos decomposition diagonalizes the OU operator it is easy to track the evolution along the OU flow. Thus, for $\Psi$ which have heavy Hermite spectral tails across Hermite degrees, we can show that a low score matching loss with shallow network drifts implies large width (see Proposition \ref{prop:appendixdaapproximationlowerbound}). Unlike the smoothness obstruction, this lower bound has more practical significance since it (1) describes the hardness of $L^2$-approximate score modeling rather than exact representation, (2) directly targets the width, and (3) persists longer under OU flow.
\end{itemize}
Both smoothness and Hermite rank can be viewed as notions of depth complexity of $\Psi$, since they obstruct modeling by shallow networks but not deep ones (depth can exponentially increase both spatial irregularity and rank, but width cannot). By showing that the depth complexity of $\Psi$ (which is fully present in the scores at $\tau = 0$) propagates to the time-dependent scores along the OU flow, we make precise the intuition that ``modeling the reverse OU drifts is as hard as modeling the full endpoint map for these pushforward distributions". 
Hence, if $\Psi$ has large depth complexity then the reverse OU transport fails to decompose it into simpler pieces. 

It is easy to construct examples for which an alternative transport succeeds in this goal. The obvious choice is to take $\Psi$ itself to be a deep neural network, in which case the triangular shear map $T$ from (\ref{eqn:mainpapertriangularshearmap}) is a deep ResNet and an alternative transport with shallow drifts may be built from the layers of the network itself.
In the next two subsections, we will design such ResNet pushforwards to exhibit large third derivatives (by using deep networks to compose many nonlinearities) and high Hermite rank (by using wide three-layer networks and activations with heavy Hermite tails), which together yield Theorem \ref{theorem:mainpaperreverseOUsummary}.

\subsubsection{ResNet pushforwards: an exponential representation lower bound}
We first instantiate a triangular shear measure $\pi^\Psi$ whose generating map $\Psi$ has large third derivatives but admits a compositional decomposition into shallow pieces (and therefore a Barron-efficient alternative transport). This is accomplished by chaining together many nonlinearities across the layers of a deep network, which can amplify the nonlinearity to create derivatives that are exponential in the depth. Hence, we can construct a pushforward of $\gamma_d$ by a ResNet of depth $\Theta(d)$ with two-layer blocks of width $O(1)$ for which the following holds:

\begin{theorem}[Shallow representation lower bound, ResNet pushforward]
\label{theorem:mainpaperrepresentaitonlowerbound}
    Fix $d \geq 4$ and let $\phi(t) := \max\{0, t\}^3$ be the $\operatorname{ReLU}^3$ activation. Then, there exists a ResNet $T_\theta: \r^d \to \r^d$ of depth $\Theta(d)$ and width $O(1)$ (which uses a smoothed $\operatorname{ReLU}^3$ activation) s.t. if $\pi^\theta := (T_\theta)_\# \gamma_d \in \p(\r^d)$ then 
    \begin{equation}
    \label{eqn:mainpaperrepresentationseparation}
        \mathcal{E}_{\b_\phi}(\pi^\theta) \gtrsim 2^d, \quad D_{\b_\phi}(\gamma_d, \pi^\theta) \lesssim d.
    \end{equation}
    Furthermore, the OU scores $s_\tau^\theta := \nabla \log P_\tau^* \pi^\theta$ and $b_\tau^\theta := \nabla \log \frac{P_\tau^* \pi^\theta}{\gamma_d}$ are in $\b_\phi$ for every $\tau \in [0, \infty)$.
\end{theorem}
\begin{proof}
The precise statement is given in Proposition \ref{prop:appendixrepresentaitonupperlowerbounds}.
The hard instance constructs a triangular shear $T_\theta(z, u) = (z, \psi(u) + \Psi_\theta(z))$, where $\Psi_\theta: \r^{d-1} \to \r$ is a deep narrow network of the form
\begin{equation}
\label{eqn:mainpaperformofrepresentationlowerbound}
   \Psi_\theta(z) = \omega\left(z_{d-1} + \Lambda_\omega \omega\left(z_{d-2} + \Lambda_\omega \omega\left( \cdots + \Lambda_\omega \omega\left(z_1\right)\right)\right)\right)
\end{equation}
for a scalar function $\omega \in C^5_b(\r)$ with $\Lambda_\omega$ a constant depending only on $\omega$. Alternatively, this can be viewed as a scalar readout (from the $d^{th}$ coordinate) of a ResNet where each hidden layer $\ell$ has width 1, input feature $e_\ell$, output feature $e_{\ell + 1}$, and activation $\Lambda_\omega \omega$.  

The lower bound shows the third derivative of $\Psi_\theta$ is large and uses some OU stability estimates to imply the same for $s_\tau^\theta, b_\tau^\theta$ for small $\tau$ (see Corollary \ref{corollary:representationlowerbound}). 
The upper bound simply computes (\ref{eqn:mainpaperformofrepresentationlowerbound}) using an ODE with single-index drifts (namely, the layers of $\Psi_\theta$), writes the result to the $u$ coordinate, and then reverses the computation to return $z$ to its initial state. The function $\psi: \r \to \r$ is designed in order to have the right polynomial tails, which is needed to ensure the OU scores are in $\b_{\operatorname{ReLU}^3}$. 
The teacher activation implementing $T_\theta$ is a smoothed $\operatorname{ReLU}^3$ function, which is chosen so that both $\omega$ and $\psi$ can be represented using $O(1)$ width. 
\end{proof}

\begin{remark}
    The exponential separation result of (\ref{eqn:mainpaperrepresentationseparation}) is fairly general: $\omega$ can be essentially any nondegenerate $C^5_b$ function, $\psi$ can have any tail behavior, $\phi$ can be any activation with bounded third derivatives which can represent $\omega$ and $\psi_\theta$, the features need not be an orthonormal set (though they must be chained so that the output feature of each layer is the input feature of the next layer), and the lower bound is exponential in the depth (see Lemma \ref{lemma:appendixrepresentationlowerboundresnets} for a general statement). 
    
    However, our proof that the scores are Barron relies heavily on the $\operatorname{ReLU}^3$ structure (since we use a weighted-Sobolev-to-Barron embedding, see Lemma \ref{lemma:appendixsobolevtobarron}). This requires the right tails for $\psi$, which in turn requires choosing a teacher activation that can implement both $\omega$ and $\psi$.
\end{remark}

In words, the distribution $\pi^\theta$ encodes a sequence of very simple coordinatewise movements, and performing them in the correct order yields a generative model for $\pi^\theta$ which can be implemented with small shallow neural network drifts. However, the reverse OU transport is unaware of this structure and attempts to adjust the complicated coordinate directly using $ \Psi_\theta$, whose Barron norm increases exponentially with depth (the composition of Barron functions can cause an exponential increase in Barron norm, see \cite{eldan2016power,telgarsky2016benefits,weinan}). 
Since the scores are Barron for all OU times, this exponential obstruction is quantitative rather than a statement that the scores escape the Barron space near the data side.

\begin{figure}[!htb]
    \centering
    \includegraphics[width=0.95\linewidth]{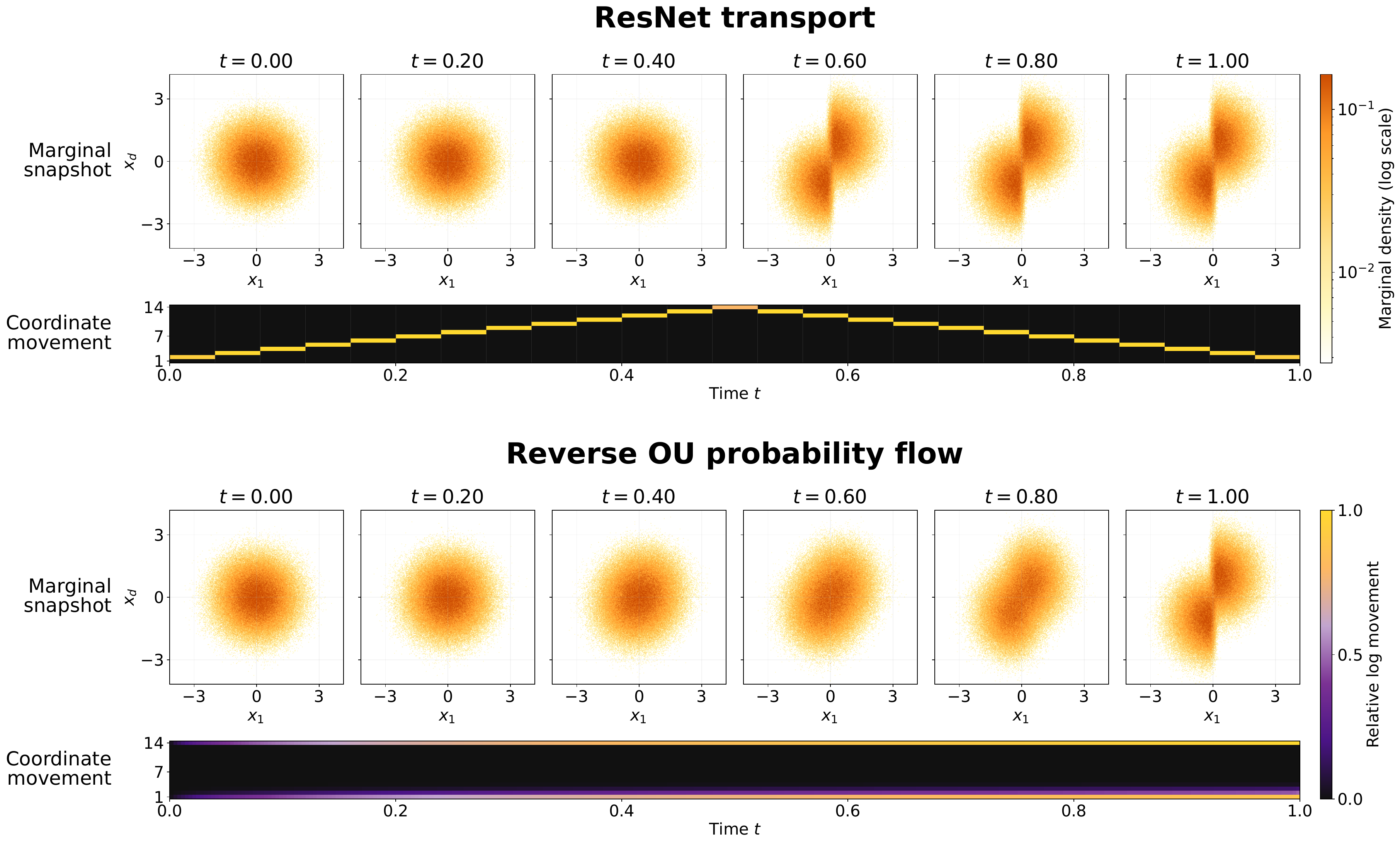}
    \caption{
    Marginal densities of 2-dimensional slices (first and last coordinates) of $\rho_t$, along with histograms showing which coordinates are in motion, for the ResNet and reverse OU transports from $\gamma_d$ to the lower bound instance $\pi^\theta$ constructed in Theorem \ref{theorem:mainpaperrepresentaitonlowerbound} (with $\omega = \tanh$, $\psi = \operatorname{Id}$, and $d=14$ for simplicity). 
    Starting from $t=1$ and looking backward, we see that the OU process adjusts the $x_d$ coordinate immediately, whereas the single-index ODE waits until $t \approx 0.5$ to do so (since it must first adjust all other coordinates).}
    \label{fig:reverseOUcascademarginals}
\end{figure}

In Figure \ref{fig:reverseOUcascademarginals}  we visually compare the reverse OU transport for $\pi^\theta$ to the more efficient generative model (which is obtained by computing $\Psi_\theta$ using single-index drifts, writing the output to the final coordinate, and reversing the computation to return the first $d-1$ coordinates to their initial state). Our earlier intuition is displayed nicely: the ResNet transport greedily and efficiently moves one coordinate at a time and only touches the final coordinate once the computation is done, whereas the reverse OU transport attempts to move the hard coordinate instantly.

\begin{remark}
    The cascade-across-coordinates structure in (\ref{eqn:mainpaperformofrepresentationlowerbound}) is akin to the Gibbs scan, which is known in sampling theory as a way to sample complicated measures via simple iterated conditioning. These are the types of measures from which an algorithm like chained Langevin \cite{cheng2025hardnesssamplingmixturedistributions} is designed to sample. Our result shows that denoising models can fail to take advantage of this structure.
\end{remark}

\subsubsection{ResNet pushforwards: an approximation lower bound}
The prior subsection showed that for a certain deep ResNet pushforward $\pi^\theta$, exact representation of the reverse OU/denoising generative model using shallow neural network drifts with polynomially-bounded weights requires exponential average width. That being said, we emphasize that the exponential obstruction we identify only occurs at exponentially-small OU times and relies on large third derivatives: a practical pipeline using early-stopping and $L^2$-accurate score estimates can yield a generative model which approximates $\pi^\theta$ without encountering the lower bound of Theorem \ref{theorem:mainpaperrepresentaitonlowerbound}. 
In this subsection we will remove both of these caveats with a construction for which any fixed-accuracy shallow drift approximation to the reverse OU generative model requires large width in high dimension. 

More precisely, we instantiate a triangular shear measure $\pi^\Psi$ whose generating map $\Psi$ has high Hermite rank and therefore cannot be efficiently approximated by shallow models. This is achieved by combining many independent features in a nonlinear way; the independent features create high rank, and a heavy Hermite tail of the nonlinear readout amplifies this effect by placing them in high Hermite degrees. We do so using a pushforward of $\gamma_d$ by a ResNet of depth $\Theta(d)$ with two-layer blocks of width $O(1)$, or alternatively by a three-layer network of width $\Theta(d)$.

For notation, if $\pi^\theta = (T_\theta)_\# \gamma_d \in \p(\r^d)$ is the Gaussian pushforward by a ResNet $T_\theta: \r^d \to \r^d$ with parameters $\theta$ then we denote by $\rho_\tau^\theta := P_\tau^* \pi^\theta$ the OU flow for $\tau \in [0, \infty)$. We write
$s_\tau^\theta := \nabla \log  \rho_\tau^\theta$ and $b_\tau^\theta := \nabla \log \frac{\rho_\tau^\theta}{\gamma_d}$ for the OU scores w.r.t. Lebesgue and the Gaussian, respectively.

\begin{theorem}[Shallow approximation lower bound, ResNet pushforward] 
\label{theorem:mainpaperapproximationlowerbound}
Fix $\epsilon > 0$ smaller than a universal constant. Then, there exists $d_0 = d_0(\epsilon)$ s.t. for each $d \geq d_0$ there exists a ResNet $T_\theta: \r^d \to \r^d$ of depth $\Theta(d)$ and width $O(1)$ with ReLU activations with the following property for a universal $c > 0$:

If $0 \leq \tau \lesssim \epsilon^{4/3}$ and $\hat{s}_\tau: \r^d \to \r^d$ is any two-layer neural network of width $m$ (i.e. a sum of $m$ ridge functions) for which
\begin{equation}
    \|s_\tau^\theta - \hat{s}_\tau\|_{L^2(\rho_\tau^\theta; \r^d)} \leq \epsilon,
\end{equation}
then it holds that $m \gtrsim d^{c\epsilon^{-4/3}}$ and $\|\hat{s}_\tau\|_{\b_\phi} \gtrsim \epsilon d^{\frac{c}{2}\epsilon^{-4/3}}$ for all $\phi: \r \to \r$ satisfying Assumption \ref{assumption:activationfunciton}.
The same statements hold with $s_\tau^\theta$ replaced by $b_\tau^\theta$, and the implied constants are independent of $\epsilon, d, \tau, \hat{s}_\tau,m$. 
\end{theorem}
\begin{remark}
    The ResNet $T_\theta: \r^d \to \r^d$ also does not depend on $\epsilon$. In other words, for each dimension $d \in \n$ there is a fixed ResNet pushforward s.t. for any $\epsilon > 0$ the reverse OU scores are hard to $\epsilon$-approximate by shallow networks in dimension $\geq d_0(\epsilon)$. Moreover, $D_{\b_{\operatorname{ReLU}}}(\gamma_d, \pi^\theta) \lesssim \sqrt{d}$ with a universal constant.
\end{remark}
\begin{proof}
    The hard instance constructs a triangular shear $T_\theta(z, u) = (z, u + \Psi_\theta(z))$, where $\Psi_\theta: \r^{d-1} \to \r$ is a wide three-layer network of the form
    \begin{equation}
        \Psi_\theta(z) = \chi\left(z_{d-1} + \frac{1}{\sqrt{d-2}}\sum_{\ell = 1}^{d-2} g(z_\ell)\right)
    \end{equation}
    for scalar functions $\chi, g: \r \to \r$. Note that $T_\theta$ can be also expressed as a deep narrow ResNet, see Section \ref{subsec:appendixresnetapproximation}. In this ResNet each layer is a shear (i.e. the input and output features are orthogonal) and so it can be implemented by a ReLU flow; since the weights are $\frac{1}{\sqrt{d-2}}$, this means that  $D_{\b_{\operatorname{ReLU}}}(\gamma_d, \pi^\theta) \lesssim \sqrt{d}$.

    By Lemma \ref{lemma:appendixhermitedecayforapproximation}, with $\chi(t) = \max\{0, t\} - \max\{0, t-1\}$ and any centered non-affine Lipschitz $g$ we get $\mathcal{V}_{\chi, g}(K, 2K) \gtrsim K^{-3/2}$ with constants depending only on $g$, and hence $K(\epsilon) \gtrsim \epsilon^{-4/3}$.
    We choose $g(t) = \max\{0, t\} - \frac{1}{\sqrt{2\pi}}$ for simplicity.
    The claim follows by applying Corollary \ref{corollary:approximationlowerbound} with $L = d-2$. 
\end{proof}

In words, Theorem \ref{theorem:mainpaperapproximationlowerbound} says that for a ReLU ResNet pushforward $\pi^\theta$ of this sort and any fixed $\epsilon > 0$, any shallow network which $\epsilon$-approximates the true score at any OU time $\tau \lesssim \epsilon^{4/3}$ must have width $\gtrsim d^{c\epsilon^{-4/3}}$.   Hence, a practical denoising sampling setup (using early-stopping and $L^2$-accurate scores) which aims to achieve small error must model scores in this window and will therefore encounter our lower bound. More precisely, standard sampling guarantees for practical denoising models (e.g. \cite{chen2023sampling, chen2026highaccuracy}) decompose the end-to-end sampling error along the lines of \begin{equation}
\label{eqn:mainerrordecomposition}
        \text{sampling error } \lesssim \text{ early-stopping error} + \text{$L^2$ score-estimation error} + \text{discretization error}.
    \end{equation}
    In Lemma \ref{lemma:appendixearlystoppingforresnets}, we show that early-stopping at OU time $\tau \asymp \epsilon^{4/3}$ incurs $\sqrt{\operatorname{KL}}$ error $\gtrsim \epsilon$ and both $W_1$ and $\operatorname{TV}$ error $\gtrsim \epsilon^{4/3}$ for the hard instance of Theorem \ref{theorem:mainpaperapproximationlowerbound}. Our lower bound therefore implies that a denoising setup using shallow network drifts which gets the first two terms of (\ref{eqn:mainerrordecomposition}) to $\lesssim \delta$ requires width $\gtrsim d^{\operatorname{poly}(1/\delta)}$. 
This is a more substantial hardness result than that of Theorem \ref{theorem:mainpaperrepresentaitonlowerbound}, since it implies that the practical denoising score-matching procedure will fail on these measures when using shallow drift networks of insufficient width.

Theorem \ref{theorem:mainpaperapproximationlowerbound} is also different from the prior score-matching hardness works of \cite{song2024cryptographic,vu2026computational}, since the obstruction does not come from computational/cryptographic difficulty nor spatial regularity but rather from the approximation capabilities of shallow neural networks. Instead of revealing an unavoidable issue that score-matching encounters, this construction highlights an approximation-theoretic separation between modeling scores with shallow and deep drift networks. In Section \ref{subsec:theroleofdepth}, we will discuss some conclusions and intuitions regarding the design of generative models which can be drawn from this result.

\begin{remark}
\label{remark:mainpaperapproximationtanhscaling}
The same construction with $\chi = g = \tanh$ yields a similar result:
for $0 \leq \tau \lesssim \epsilon$,  we get that $m \gtrsim d^{c \log(1/\epsilon)^2}$ and $\|\hat{s}_\tau\|_{\b_\phi} \gtrsim \epsilon d^{\frac{c}{2}\log(1/\epsilon)^2}$. To show this, we apply the other result from Lemma \ref{lemma:appendixhermitedecayforapproximation} to see that $K(\epsilon) \gtrsim \log(1/\epsilon)^2$, after which we use Corollary \ref{corollary:approximationlowerbound}. In this case, getting the first two terms of (\ref{eqn:mainerrordecomposition}) to $\lesssim \delta$ using shallow score networks requires width $\gtrsim d^{\Omega(\log(1/\delta)^2)}$.
\end{remark}

We finish up by showing a simpler integrated form of the lower bound, which is the one quoted in the informal statement of Theorem \ref{theorem:mainpaperreverseOUsummary}.

    \begin{corollary}[Integrated shallow approximation lower bound]
    \label{corollary:mainpaperintegratedlowerbound}
        Fix $T > 0$ and $0 < \epsilon < \epsilon_0(T)$. Then, for each $d \geq d_0(\epsilon)$ there exists a ResNet $T_\theta: \r^d \to \r^d$ of depth $\Theta(d)$ and width $O(1)$ with ReLU activations with the following property for a universal $c > 0$:

        If $(\hat{s}_\tau)_{\tau \in [0, T]}$ is a measurable family of two-layer neural networks of width $m(\tau)$ for which
        \begin{equation}
        \label{eqn:mainpaperintegratederrorassumption}
            \int_0^T \|s_\tau^\theta - \hat{s}_\tau\|^2_{L^2(\rho_\tau^\theta; \r^d)}d\tau \leq \epsilon^2,
        \end{equation}
        then $\esssup_{\tau \in [0, T]}m(\tau) \gtrsim d^{c\epsilon^{-4/5}}$ and for all $\phi: \r \to \r$ satisfying Assumption \ref{assumption:activationfunciton} we have
        \begin{equation}
        \label{eqn:mainpaperintegratederrorclaim}
            \int_0^T \|\hat{s}_\tau\|_{\b_\phi}d\tau \gtrsim \epsilon^{7/5} d^{\frac{c}{2}\epsilon^{-4/5}} \quad \text{and}\quad \int_0^T \|\hat{s}_\tau\|_{\b_\phi}^2d\tau \gtrsim \epsilon^{2} d^{{c}\epsilon^{-4/5}}.
        \end{equation}
        The same statements hold with $s_\tau^\theta$ replaced by $b_\tau^\theta$, and the implied constants are independent of $T, \epsilon, d, (\hat{s}_\tau)_{\tau \in [0, T]}$.
    \end{corollary}
    \begin{proof}
        We let $a > 0$ be the constant s.t. Theorem \ref{theorem:mainpaperapproximationlowerbound} holds for $\tau \in [0, a \delta^{4/3}]$ with pointwise accuracy parameter $\delta$. We take $\delta = (2\epsilon^2/a)^{3/10}$ and $\tau_\delta = a\delta^{4/3}$. This implies that $\tau_\delta \delta^2 = 2\epsilon^2$ while $\delta \asymp \epsilon^{3/5}$ and $\tau_\delta \asymp \epsilon^{4/5}$. For sufficiently small $\epsilon$, Theorem \ref{theorem:mainpaperapproximationlowerbound} applies at accuracy $\delta$ and time $\tau_\delta \leq T$. If we set $e_\tau := \|s_\tau^\theta - \hat{s}_\tau\|_{L^2(\rho_\tau^\theta; \r^d)}$ and $G := \{\tau \in [0, \tau_\delta]:\; e_\tau \leq \delta\}$ then by (\ref{eqn:mainpaperintegratederrorassumption}) we have 
        \begin{equation}
            |[0, \tau_\delta] \setminus G| \leq \frac{1}{\delta^2}\int_0^{\tau_\delta}e_\tau^2 d\tau \leq \frac{\epsilon^2}{\delta^2} = \frac{\tau_\delta}{2}.
        \end{equation}
        Hence, $|G| \geq \tau_\delta / 2$. For every $\tau \in G$ we have $m(\tau) \gtrsim d^{c\delta^{-4/3}}$ and $\|\hat{s}_\tau\|_{\b_\phi} \gtrsim \delta d^{\frac{c}{2}\delta^{-4/3}}$, from which the claim (\ref{eqn:mainpaperintegratederrorclaim}) follows by integration over $G$.
    \end{proof}

\subsection{Feature learning separation}
\label{subsec:featurelearning}

We now aim to study dimensional separations between the $D_{\b_\phi}$ and rotation-equivariant $D_{\h_k}$ distances as a way to understand the role of feature learning vs. learning drifts in a rotation-equivariant kernel space. We do this first in a perturbatively-small regime where matching upper and lower bounds are possible, after which we will present an initial result in the nonperturbative setting. These results together yield Theorem \ref{theorem:mainpaperfeaturelearningsummary}.

We start in a simple setting where $\pi$ is a small pushforward of a rotation-invariant measure by a single-index ReLU map:

\begin{example}[Perturbative feature learning, single-index ReLU pushforward]
    \label{example:nonlinearshearrkhs}
Let $d \geq 4$ and $\mu \in \p_2(\r^d)$ be rotationally-invariant with $\int_{\r^d} |x|^2 d\mu(x) > 0$. Let $\phi(z) = \max\{0, z\}$ be the ReLU activation, fix $u, \theta \in \mathbb{S}^{d-1}$ with $\la u, \theta\ra \neq 0$, and for $\epsilon > 0$ define
\begin{equation}w(x) := u \phi(\la x, \theta\ra), \quad \nu_\epsilon := (\1 + \epsilon w)_\# \mu \in \p_2(\r^d).\end{equation}
Then, for any rotation-equivariant $k \in \mathcal{K}_{\operatorname{rot}}$ which satisfies Assumption \ref{assumption:kernel} we have
\begin{equation}\lim_{\epsilon \downarrow 0} \frac{D_{\h_k}(\mu, \nu_\epsilon)}{D_{\b_\phi}(\mu, \nu_\epsilon)} = +\infty .\end{equation}
\end{example}
\begin{proof}
See Section \ref{subsection:generativemodelingproofs}.
\end{proof}

This is essentially a restatement of the folklore fact that ``a single-index ReLU function does not lie in any rotation-invariant RKHS for $d \geq 4$" (see \cite[Appendix D]{bach2017breaking}) in the setting of vector-valued functions and rotation-equivariant RKHS. For completeness we provide a proof by combining a first-order dual lower bound with a standard argument involving spherical harmonics; in the ReLU case the sum over harmonics diverges for $d \geq 4$.\footnote{Since $x \mapsto \la x, u\ra_+$ is in $H^{s}(\mathbb{S}^{d-1})$ for every $s < \frac{3}{2}$, it turns out that for $d \in \{2,3\}$ the ReLU neuron lies in a rotation-invariant RKHS for $d < 4$.} 

The proof method via spherical harmonics (or Hermite decompositions when the source is Gaussian) is general, and can be used to show matching upper and lower bounds for perturbatively-small single-index flows (see Lemma \ref{lemma:appendixupperandlowerboundsforsingleindexgaussianflows} for a general statement). We apply it here to exhibit some quantitative separations for other activations:

\begin{example}[Perturbative feature learning, autonomous single-index flows]
\label{example:nonlinearshearrkhssmoothed}
    Let $d\geq3$ and $\gamma_d$ be the standard Gaussian. Fix $u, \theta \in \mathbb{S}^{d-1}$, and for $\epsilon > 0$ define $\nu_\epsilon := \operatorname{Law}(X_\epsilon)$ using the flow
    \begin{equation}
    \label{eqn:mainpapersingleindexflow}
    \dot{X}_t =  u\phi(\la X_t, \theta\ra), \quad X_0 \sim \gamma_d.\end{equation}
    For an activation $\phi: \r \to \r$ satisfying Assumption \ref{assumption:activationfunciton}, define
    \begin{equation}
        Q_{\phi, d} :=  \inf_{\substack{k \in \mathcal{K}_{\operatorname{rot}} \\ k \text{ satisfies Assumption \ref{assumption:kernel}}}}\left\{\sqrt{\int_{\r^d} \operatorname{tr}k(x, x) d\gamma_d(x)} \cdot \liminf_{\epsilon \downarrow 0} \frac{D_{\h_k}(\gamma_d, \nu_\epsilon)}{D_{\b_{\phi}}(\gamma_d, \nu_\epsilon)}\right\} .
    \end{equation}
    Then, the following scalings hold as $d \to \infty$:
    \begin{equation}
        Q_{\tanh, d} \asymp d^{\frac{d}{4} \cdot (1+o(1))}, \qquad Q_{\sin, d} \asymp e^{\sqrt{d} \cdot (1+o(1))}.
    \end{equation}    
\end{example}
\begin{proof}
See Section \ref{subsection:generativemodelingproofs}.
    In Lemma \ref{lemma:appendixupperandlowerboundsforsingleindexgaussianflows} we show matching upper and lower bounds (up to constants) for Gaussian single-index flows in terms of the Hermite coefficients of the activation. In Lemmas \ref{lemma:appendixtanhhermitescaling} and \ref{lemma:appendixtsinhermitescaling} we compute the asymptotics uniformly in $u, \theta$ for $\tanh$ and $\sin$, respectively.
\end{proof}

\begin{figure}[!htb]
    \centering
    \includegraphics[width=1.0\linewidth]{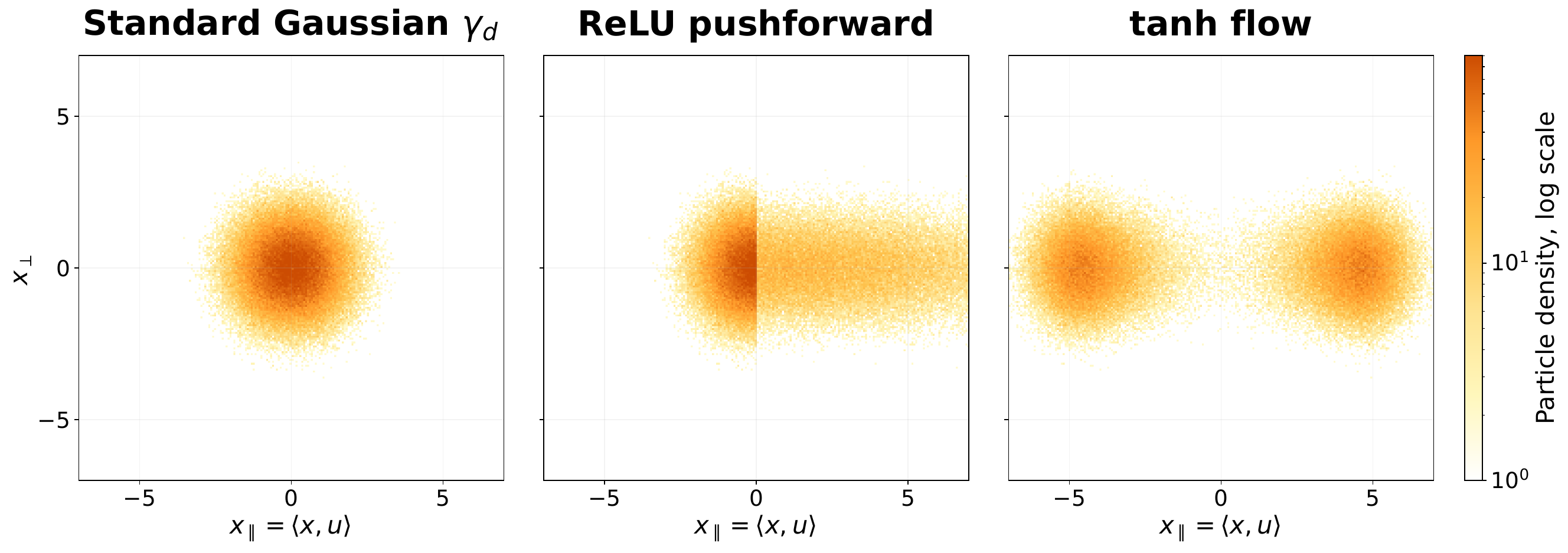}
    \caption{Densities of 2-dimensional slices (the $u$-coordinate and one perpendicular to $u$) for the standard Gaussian $\gamma_d$, the measure $\nu_\epsilon$ constructed in Example \ref{example:nonlinearshearrkhs} with a ReLU pushforward, and the measure $\nu_\epsilon$ constructed in Example \ref{example:nonlinearshearrkhssmoothed} with a smooth $\tanh$ flow; we take $d=6$ and $\epsilon = 4$ for visualization purposes. The ReLU pushforward has a clear nonsmooth transition, while the $\tanh$ flow almost looks like a Gaussian mixture.}
    \label{fig:featurelearningplots}
\end{figure}

Examples \ref{example:nonlinearshearrkhs} and \ref{example:nonlinearshearrkhssmoothed} exhibit feature learning separations in a high dimensional perturbative regime for a nonsmooth ReLU transport (where the ratio can grow unbounded in fixed $d$) and for smooth $\tanh$ or $\sin$ transports (where the ratio is superpolynomial in $d$), respectively. In Figure \ref{fig:featurelearningplots} we plot two-dimensional slices of some of these measures to help visualize their differences and the nature of the separation.

\begin{remark}
The use of spherical harmonics to lower bound RKHS norms of single-index functions is well-known in the field; our only contribution to this story is the application of this technique to the local geometry of the $D_\b$ metrics/the measure transport problem.
\end{remark}

Based on this initial perturbative result, it is tempting to conclude that ``for small $\epsilon$, there is the classical superpolynomial-in-$d$ separation between rotation-equivariant $D_{\h_k}$ and $D_{\b_\phi}$ when transporting to the endpoints of $\epsilon$-length single-index flows". Examples \ref{example:nonlinearshearrkhs} and \ref{example:nonlinearshearrkhssmoothed} indicate that this is the case when $\epsilon \downarrow 0$ with $d$ fixed, but it is entirely possible that this separation only kicks in when $\epsilon$ is very tiny, say $\epsilon \lesssim e^{-d}$. After all, by taking $\epsilon \downarrow 0$ we are effectively constraining a transport to use time-independent drifts, which does not allow an RKHS-based transport to leverage depth.\footnote{For a more precise geometric perspective, by taking $\epsilon \downarrow 0$ we are measuring the norm of the original single-index drift in the tangent spaces of the manifolds $(\p(\r^d), D_{\h_k})$ and $(\p(\r^d), D_{\b_\phi})$. By doing so, we reason about the RKHS norm of the prescribed transport instead of the infimum over all transports.}

It turns out that this is exactly the case: in the macroscopic/nonperturbative setting, there exist shortcuts which can be taken to lower the total RKHS energy to $\operatorname{poly}(d)$. More precisely, we construct an efficient alternate flow which matches the endpoint map of a time-invariant single-index flow by paying a one-time cost which is $\operatorname{poly}(d)$ and independent of $\epsilon$. We demonstrate this mechanism in the simple setting where $\la u, \theta\ra = 0$ and $\phi$ is smooth and bounded, but we expect it to apply more generally:

\begin{example}[Nonperturbative feature learning, autonomous single-index flows] 
\label{example:mainpapernonperturbativefeaturelearning}
Let $d \geq 3$, fix $\phi \in C^\infty_b(\r)$ and $u, \theta \in \mathbb{S}^{d-1}$ with $\la u, \theta \ra = 0$, and define $\nu_T := \operatorname{Law}(X_T)$ for the flow (\ref{eqn:mainpapersingleindexflow}). 
Then, there exists a smooth rotation-equivariant $k_\star \in \mathcal{K}_{\operatorname{rot}}$ (depending only on $d, \phi, T$) and a transport $(v_t, \rho_t)_{t \in [0, 1]}$ from $\gamma_d$ to $\nu_T$ using $\operatorname{poly}(d)$-Lipschitz drifts which satisfies
\begin{equation}
    \left(\sup_{t \in [0, 1]} \sqrt{\int_{\r^d} \operatorname{tr}k_\star(x, x) d\rho_t(x)}\right) \cdot \int_0^1 \|v_t\|_{\h_{k_\star}}dt \lesssim \operatorname{poly}(d).
\end{equation}
    
\end{example}
\begin{proof}
The precise nonasymptotic statement is given in Proposition \ref{prop:appendixnonperturbativefeaturelearning}, and it holds with $\gamma_d$ replaced by any initial $\rho_0 \in \p(\r^d)$ for which $(\int_{\r^d}|x|^d d\rho_0(x))^{1/d} \lesssim \operatorname{poly}(d)$. The method is similar to how the linear Stein metric achieves logarithmic dependence on the mean in Proposition \ref{prop:optimalgaussiantogaussian}: there is an initial compression/preconditioning phase in the direction $u$, followed by an exponentially-shrunk version of the original single-index transport, followed by a decompression phase. Since the compression/decompression phases have costs logarithmic in the compression factor, we may efficiently shrink the single-index phase to achieve polynomial total cost.
\end{proof}

Thus, endpoint maps of (time-invariant) single-index flows can be efficiently represented using drifts which lie in a rotation-equivariant RKHS.
Taken literally, this result says that for $\epsilon \gtrsim \frac{1}{\operatorname{poly}(d)}$ we no longer have the superpolynomial separations between rotation-equivariant $D_{\h_k}$ and $D_{\b_\phi}$ that we saw in the $\epsilon \downarrow 0$ regime in Examples \ref{example:nonlinearshearrkhs} and \ref{example:nonlinearshearrkhssmoothed}. More philosophically, it indicates depth can allow RKHS methods to model some functions which require feature learning in the shallow case. In the setting of generative modeling, this result lends some credibility to the idea of learning dynamical transports by representing drifts in RKHS, see e.g. \cite{fisher2021measure, pandey2025diffeomorphic, maurais2025learning}. 

Our investigation only scratches the surface of the comparison between the $D_{\b_\phi}$ and rotation-equivariant $D_{\h_k}$ metrics: all we have shown is that the classical shallow separations persist locally in the dynamical measure transport problem, but they break down once depth and time-dependent drifts are allowed. We consider it an important direction for future work to investigate the efficiency of modeling more complex data using drifts which lie in rotation-equivariant RKHS, particularly for targets which are flows/pushforwards by neural networks.

\section{Outlook}
\label{sec:outlook}

In this paper, we introduced the Barron transport metric $D_{\b_\phi}$ as a technical tool to help us study approximation theory for generative modeling. In this section, we highlight some open questions and directions for future research as well as a potential application beyond generative modeling.

\subsection{Algorithmic and computational ideas}
\label{subsec:algorithms}

Our results can be summarized as follows: representing drifts with neural networks (which utilize feature learning) can provably allow dynamical transports with more efficient representations, but the standard denoising score-matching algorithm can fail to find such transports even in very natural cases. 
Similarly, it is folklore that algorithms which attempt to learn the Wasserstein optimal transport with neural networks can suffer from lack of drift regularity.\footnote{See the counterexample of Caffarelli \cite{cafarellicounterexample}, which suffers from a discontinuous optimal transport map due to the target's nonconvex support. Any ODE with e.g. Lipschitz drifts cannot represent such a map, and so neither can any flow model with Barron drifts using a Lipschitz activation.} Hence, the current paradigms for learning dynamical generative models encounter neural network representation bottlenecks, even on easy data for which simple generative models with shallow drifts exist. 

This opens up the obvious next question: do there exist efficient algorithms which can (approximately) solve dynamical optimal transport problems like $D_{\h_k}$ or $D_{\b_\phi}$ given access to uncoupled samples from source and target measures? Such an algorithm would yield an approach to dynamical generative modeling which is built to seek efficient representation via kernel methods or neural networks; this may prove useful on real-world data in high dimension. By contrast, the nonexistence of such an algorithm would yield some information about the fundamental difficulty of learning generative models from samples. Furthermore, given the expressivity of transports with RKHS drifts (see Example \ref{example:mainpapernonperturbativefeaturelearning}) it is also valuable to study the algorithmic differences between the $D_{\h_k}$ and $D_{\b_\phi}$ problems.
The algorithmic/computational aspects of this question are poorly understood and constitute a valuable direction for future work.

\begin{remark}[Time regularity of geodesics]
    An efficient algorithm to learn dynamical transports must somehow discretize $[0, 1]$, which relies on the regularity-in-time of the optimal transport $(v_t, \rho_t)_{t \in [0, 1]}$. At the current moment we do not have precise control on this (since the $D_{\h_k}$ and $D_{\b_\phi}$ energies do not directly penalize irregularity of $t \mapsto v_t$). 
    
    A promising direction involves a priori estimates on the geodesic PDE. As an example, in the Stein case with a Matérn/Sobolev kernel $\h_k \cong H^s$, the PDE (\ref{eqn:prelimssteingeodesics}) and Sobolev multiplication estimates ensure that any constant-speed $D_{\h_k}$ geodesic has $\sup_{t \in [0, 1]}\|\partial_tv_t\|_{H^{s-1}} \lesssim D_{\h_k}(\rho_0, \rho_1)$. This approach will be formalized and developed for the general case in \cite{companionpaper}.
\end{remark}

In light of the deep ResNet pushforward counterexamples from Theorems \ref{theorem:mainpaperrepresentaitonlowerbound} and \ref{theorem:mainpaperapproximationlowerbound}, we can formulate the following slightly more precise ResNet version of this question: 

\begin{question}[Efficient generative modeling of ResNet pushforwards]
\label{question:resnetpushforwards}
    Given a teacher ResNet $F: \r^d \to \r^d$ and samples from its Gaussian pushforward $\pi = F_\# \gamma_d$, is there an efficient algorithm which learns a drift field $(v_t)_{t \in [0, 1]}$ transporting $\gamma_d$ to $\pi$? In other words, can we efficiently learn a deep ResNet/neural ODE which maps $\gamma_d$ to $\pi$?
    What properties of $F$ allow such an algorithm? 
Ideally, we want the learned transport to have path norm/Barron cost comparable to that of $F$.
\end{question}
\begin{remark}[Shifted transport metric]
    To make this question more practical, it suffices to transport $\gamma_d$ to a measure which approximates $\pi$ in some metric or divergence. Allowing the target to be shifted within a small ball can substantially facilitate transport; this tradeoff between controlled metric displacement and dynamical transport is closely related to the shifted composition methodology of \cite{altschuler2026shifted}. In our setting, this would correspond to the quantity 
    \begin{equation}
        \widetilde{D}_{\mathcal{B},\varepsilon}( \mu, \pi) = \inf \left\{ \int_0^1 \| \hat{v}_t \|_{\mathcal{B}} dt: \; \int_0^1 \| \hat{v}_t - v_t \|_{L^2(\rho_t; \r^d)}^2 dt \leq \varepsilon^2\right\},
    \end{equation}
    where the infimum is over transports $(v_t, \rho_t)_{t \in [0, 1]}$ satisfying $\partial_t \rho_t + \nabla \cdot( v_t \rho_t)=0$ and $ \rho_0 = \mu, \rho_1 = \pi $.
    Under mild regularity of $\hat{v}_t$ (e.g. Lipschitz regularity, see \cite[Prop. 3]{albergo2023building}), the drift approximation can be translated into Wasserstein control of the transport endpoint $W_2(\hat{\pi}, \pi) \lesssim \varepsilon$.  
\end{remark}

There are some strong prior works due to Chen et al. regarding the computational hardness of learning  pushforwards of the Gaussian by shallow neural networks: in \cite{sitanlearningpolynomialtransformations} they show that a Gaussian pushed forward by a two-layer neural network with polynomial activations can be efficiently learned under either quadratic or low-rank structure, while in \cite{chen2022learning} they show by a moment-matching argument that a Gaussian pushed forward by a two-layer ReLU neural network with $\log(d)$ hidden neurons cannot always be efficiently learned in the statistical query model.  

These results mirror the regression setting, where small Barron norm of a target function does not necessarily imply efficient learnability \cite[Theorem 4]{csqlowerboundrelunets} but a suitable sparse\footnote{In fact, in the regression setting the sparsity threshold for $\operatorname{poly}(d)$ complexity is also $\log(d)$ hidden neurons without further assumptions, which aligns with the lower bound of \cite{chen2022learning}.} neural network representation or special activation structure does (this is the subject of recent work on the multi-index model, see e.g. the survey \cite{multiindexsurvey}). Based on this, we can pose a question simpler than Question \ref{question:resnetpushforwards} as a concrete next step toward understanding the computational landscape of generative modeling:

\begin{question}[Efficient generative modeling of multi-index pushforwards]
\label{question:multiindexpushforwards}
    Given a teacher multi-index vector field $F: \r^d \to \r^d$ of the form $F(x) = \sum_{j=1}^m \sigma_j u_j \phi(\la x, \theta_j\ra)$  and samples from its Gaussian pushforward $\pi = (\operatorname{Id} + F)_\# \gamma_d$, is there an efficient algorithm which learns a generative model for $\pi$? Which properties of $F$ allow such an algorithm (e.g. $m = O(1)$)?
\end{question}

When $m = 1$ this problem is closely related to non-Gaussian component analysis (NGCA), which is known to exhibit statistical-computational gaps; see \cite{diakonikolas2022nongaussian} and references therein. The result of Example \ref{example:reverseOUsingleindexflowsandtilts} shows that the standard denoising algorithm will have difficulty representing the reverse OU drifts of such a target using two-layer neural networks,\footnote{Note that a flow by a single-index ReLU vector field can be represented by a single-index ReLU pushforward due to homogeneity, and vice versa.}
but the overall computational hardness of this problem is not completely understood. A key difference from the multi-index regression setting is that 
a learner can only observe moments from the $\pi$ marginal as opposed to cross-moments of the induced coupling of $(\gamma_d, \pi)$, which makes it easier to hide information by matching moments. We leave the investigation into Question \ref{question:multiindexpushforwards} for future work.

\subsection{The role of depth and layerwise supervision}
\label{subsec:theroleofdepth}

In a practical generative model, we are constrained by the amount of inference-time computation required to generate a sample. Consider a dynamical generative model (a flow/SDE model performing e.g. denoising or optimal transport), and suppose that the drifts are modeled using ResNets. The layers of each ResNet can be viewed as a sequence of separate drifts in their own right: from the perspective of a particle, any update of the form $x \mapsto x + \eta F(x)$ is a ``drift". 
The natural quantity measuring sequential computation is the number of drift evaluations times the number of layers in each drift network, since this captures how many basic operations each particle must undergo during generation. 
The major difference lies in how the transport is learned: each drift ResNet is trained to perform a specific operation (e.g. in denoising models they attempt to match the reverse OU drifts), while an individual layer of such a drift ResNet may implement any subroutine to help. 

Put simply, we may view the entire generative process as a single very deep ResNet which we train by supervised regression on various intermediate layers of the practitioner's choosing. 
Indeed, once the total depth is fixed by computational constraints, it is up to the designer whether to allocate more computation toward deeper drift models or more drift evaluations per generation; this is equivalent to less or more layerwise supervision, respectively. This raises a natural algorithmic question, which we pose for denoising models specifically:

\begin{question}[Allocation of depth/layerwise supervision]
    For a fixed total depth of the generative process of a denoising model, what is the ``best" allocation between the number of neural network score evaluations and the depth of each score neural network? Here, ``best" should be measured in terms of both training complexity and sample quality.
\end{question}

The two extremes are (1) minimal supervision, in which case we train one deep ResNet pushing the Gaussian to the target distribution and (2) maximal supervision, in which case there are many shallow  networks trained via score-matching. The first approach is reminiscent of generative adversarial networks (GANs) \cite{goodfellow2014generative}; although these have been famously outperformed by iterative dynamical methods such as diffusion models \cite{gansucks}, recently there has been much progress in minimizing the number of denoising steps required during generation \cite{salimans2022progressive,song2023consistency,kim2024consistency}. 
In Theorems \ref{theorem:mainpaperrepresentaitonlowerbound} and \ref{theorem:mainpaperapproximationlowerbound} we showed the potential suboptimality of the second approach, since shallow score networks can cause inefficiencies due to the nature of the reverse OU supervision. 
The optimal allocation is likely data-dependent and somewhere in the middle of these two extremes, and it is a useful direction of future work to understand the underlying tradeoffs in both simple and real-world settings.

\subsection{Application to sampling}
\label{subsec:samplingteaser}
We defined the Barron transport metric $D_{\b_\phi}$ as a dynamical optimal transport problem with tangent norm $\|\cdot\|_{\b_\phi}$. It turns out that any choice of tangent norm will yield a (Finsler) manifold geometry on $\p(\r^d)$, with an associated theory of absolutely continuous curves, geodesics, and gradient flows in terms of the continuity equation.
Importantly, the gradient flow of $\operatorname{KL}(\cdot \| \pi)$ w.r.t. such a geometry will always yield an algorithm for sampling from $\pi$ given query access to $\nabla \log \pi$. This is the perspective underlying modern analysis of the SVGD sampling algorithm as a $D_{\h_k}$ gradient flow (see \cite{duncan2023geometrysteinvariationalgradient}), as well as the recently-introduced Radon-Wasserstein geometry and its induced sampling algorithm \cite{rrw}.

In the companion paper \cite{companionpaper}, we leverage this observation to design other sampling algorithms through careful selection of the tangent norms. Doing so will require a rigorous study of the optimal transport geometry of these generalized dynamical metrics in the spirit of Ambrosio-Gigli-Savar\'e \cite{ags}. 
In this subsection we will ignore mathematical rigor and the full generality, and instead we give a taste of the sampling application in the case of the Barron transport metric $D_{\b_\phi}$ in order to further motivate the study of the $(\p(\r^d), D_{\b_\phi})$ geometry.

Indeed, by taking the gradient flow of $\operatorname{KL}(\cdot \| \pi)$ in the metric space $(\p(\r^d), D_{\b_\phi})$ (in the sense of \cite[Chapter 1]{ags}) and discretizing it, we arrive at the following algorithm for sampling from $\pi\propto e^{-V}$ with potential $V: \r^d \to \r$ via a cloud of interacting particles:

\begin{algorithm}[!htb]
\caption{Barron transport sampler}
\label{alg:barrontransportsampler}
\KwIn{Initial particles $X_1,\ldots,X_N\in\r^d$, activation
$\phi:\r\to\r$, step size $\eta>0$.}

\While{not converged}{
    $v_n \gets \nabla V(X_n)$ for $n\in[N]$\;

    $(\sigma^\star,u^\star,\theta^\star)
    \gets
    \displaystyle
    \argmin_{\substack{\sigma>0\\u,\theta\in\mathbb S^{d-1}}}
    \left\{
    \frac{\sigma}{N}\sum_{n=1}^N
    \left[
    \phi(\la X_n,\theta\ra)\la u,v_n\ra
    -
    \la u,\theta\ra\phi'(\la X_n,\theta\ra)
    \right]
    +\frac{\sigma^2}{2}
    \right\}$\;

    $X_n \gets X_n+\eta \cdot \sigma^\star u^\star
    \phi(\la X_n,\theta^\star\ra)$ for $n\in[N]$\;
}
\end{algorithm}

This interacting particle sampler regresses a single-index vector field at each step, along which it coherently flows all the particles. This is very similar in spirit to the SVGD algorithm, except that the velocity field is selected in a way which utilizes feature learning instead of kernel regression.
Algorithm \ref{alg:barrontransportsampler} has favorable behavior on target distributions which have hidden low-dimensional structure. As some illustrative examples which are strongly log-concave:

\begin{enumerate}
    \item[(1)] \textbf{(Hidden subspace).} If our target is $\pi^{(1)} := \mathcal{N}\left(0, (\1_d - \frac{1}{\log d}P)^{-1}\right)$ for any orthogonal projection $P = P^\top = P^2$ with $\operatorname{rank}(P) = \Theta(\log^2 d)$, then the following hold when initializing from $\gamma_d$:
    \begin{itemize}
        \item Langevin Monte Carlo (LMC) with any deterministic, nonnegative step size schedule requires $\Omega(\sqrt{d})$ steps to decrease KL by a constant factor.
        \item Mean-field SVGD (i.e. the infinite-particle population version) with a Gaussian or Matérn kernel and any  bounded stable\footnote{By this, we mean that a descent lemma is satisfied for the KL, i.e. $\operatorname{KL}(\rho_{n+1} \| \pi) \leq \operatorname{KL}(\rho_n \| \pi) - c \eta_n \|v_n\|_{\h_k}^2$ with $c>0$ independent of dimension.} step size schedule requires $\Omega(d)$ steps to decrease KL by a constant factor.
        \item The mean-field Barron transport sampler (with linear activation $\phi(z) = z$ and step size $\eta = 1/2$) achieves KL $\leq \epsilon$ in $O\left((\log d)^2 (1 + \log \log (1/\epsilon))\right)$ steps.  
    \end{itemize}
    \item[(2)] \textbf{(Single-index tilt).} If our target is $\pi^{(2)} \propto e^{-a \psi (\la x, \theta\ra) - \|x\|^2/2}$ for some hidden $a > 0$, $\theta \in \mathbb{S}^{d-1}$, and an even $C^2$ $\psi: \r \to \r$ for which $\pi^{(2)}$ is strongly log-concave, then the following hold when initializing from $\gamma_d$ assuming positive finite initial KL:
    \begin{itemize}
        \item LMC with any deterministic, nonnegative step size schedule requires $\Omega(\sqrt{d})$ steps to decrease KL by a constant factor.
        \item Mean-field SVGD with a Gaussian or Matérn kernel and any bounded stable step size schedule requires $\Omega(d)$ steps to decrease KL by a constant factor.
        \item The mean-field Barron transport sampler (with activation\footnote{Or any activation $\phi$ with $\|\psi'\|_{\b_\phi} < \infty$.} $\phi = \psi'$ and a dimension-independent step size) achieves KL $\leq \epsilon$ in $O(\log (1/\epsilon))$ steps.
    \end{itemize}
\end{enumerate}

\begin{figure}[!htb]
    \centering
    \includegraphics[width=1.0\linewidth]{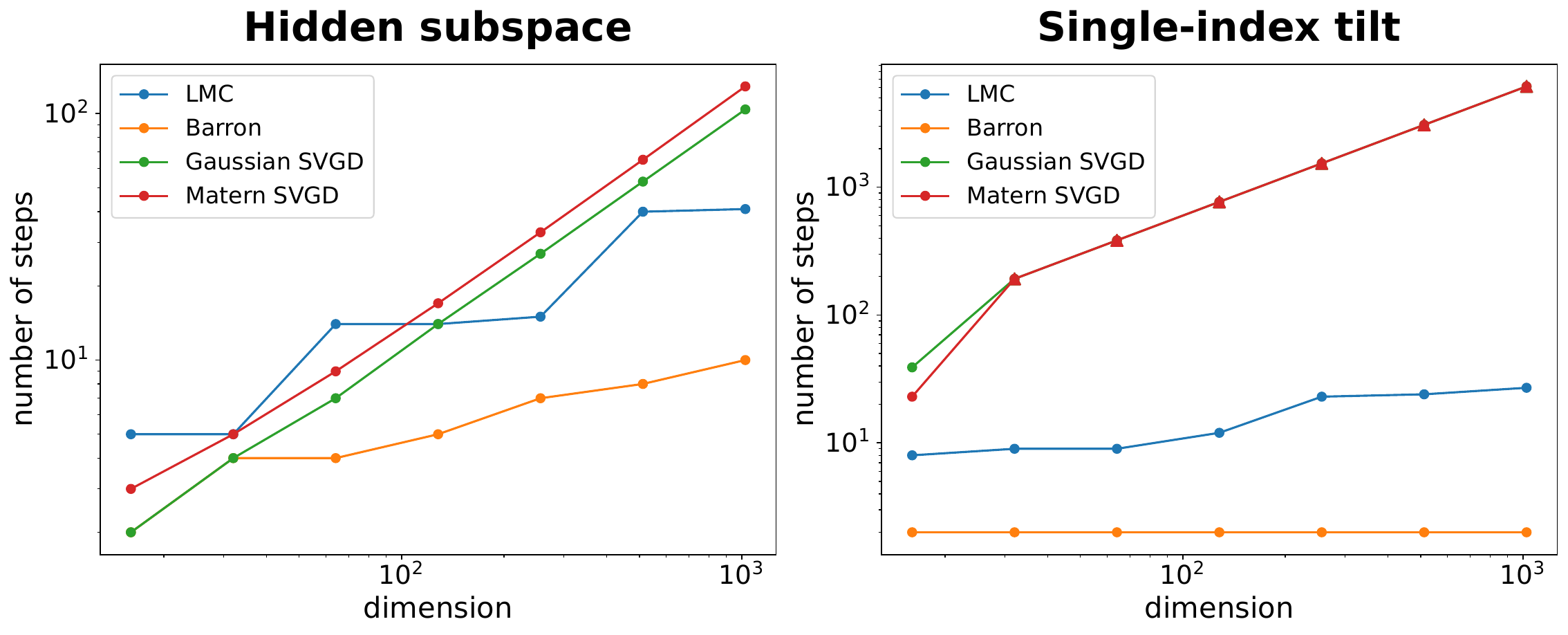}
    \caption{
    Plots of the number of steps required so that $\operatorname{KL}(\rho_n \| \pi) \leq \frac{1}{4}  \operatorname{KL}(\rho_0 \| \pi)$ as a function of $d$ for the constructions $\pi \in \{\pi^{(1)}, \pi^{(2)}\}$ listed above. For each fixed dimension, we sweep over fixed learning rates (and bandwidth/smoothness in the SVGD cases) and plot the best performance; triangles indicate a failure to reach the threshold KL in an allotted number of steps. The hidden subspace is of dimension $\lfloor \log^2(d) / 4\rfloor$, and the single-index tilt is constructed using $a = 0.1$ and $\psi(z) = 0.256 \cos(z) + 1.32 \cos(2z)$. }
    \label{fig:samplingfigure}
\end{figure}

Here $a, \psi$, and the step size bounds are independent of dimension, and Gaussian and Matérn kernels are normalized to $\operatorname{tr}k(x, x) = d$.
 This behavior is numerically verified in Figure \ref{fig:samplingfigure}.

In words, the Barron transport sampler is provably useful for discovering hidden directions and choosing adaptive step sizes in those directions, whereas other samplers can suffer from trying to apply similar step sizes in all directions or in a fixed feature basis. This phenomenon persists for more complex target measures sharing these properties (multi-index tilts, products of these examples, etc.). Indeed, one can show dimension-free convergence rates for the discretized $D_{\b_\phi}$ sampler on log-smooth targets satisfying a log-Sobolev inequality for which the relative scores along the sampling path have small Barron norm.
 We leave the derivation of this update rule, convergence rates, and the math/intuition behind the above examples to the companion paper \cite{companionpaper}.

\newpage
\bibliographystyle{plain} \bibliography{refs} \par
\newpage
\appendix

\etocdepthtag.toc{appendix}
\begingroup
\renewcommand{\contentsname}{Appendix Contents}

\etocsettagdepth{main}{none}
\etocsettagdepth{appendix}{subsection}

\tableofcontents
\endgroup

\section{Proofs}
This section contains the proofs of various claims from the main paper.

\subsection{Construction of the $D_\b$ metrics}
\label{subsec:appendixproofofabstractmetrics}
Motivated by dynamical formulations of optimal transport, we define general metrics on the space $\p(\r^d)$ of Borel probability measures on $\r^d$. Informally, if $\mu, \nu \in \p(\r^d)$ then we say that $(v_t, \rho_t)_{t \in [0, 1]}$ {transports $\mu$ to $\nu$} if we have the continuity equation
\begin{equation}
\label{eqn:informalcontinuityequation}
    \partial_t \rho_t + \nabla \cdot (\rho_t v_t) = 0, \quad \rho_0 = \mu, \quad \rho_1 = \nu.
\end{equation}
Let $\b$ be any Banach space of functions from $\r^d \to \r^d$. We assume the following mild regularity of $\b$ to ensure that we can test against smooth compactly-supported functions:

\begin{assumption}
\label{assumption:main}
Throughout this paper we assume that for every $\varphi \in C^\infty_c(\r^d)$,
\begin{equation}
    \sup_{\|v\|_\b \leq 1} \sup_{x \in \r^d} \left|\la v(x), \nabla \varphi(x)\ra\right| < \infty.
\end{equation}
We will always identify vectors $v \in \b$ with their representative functions $v(x)$.
\end{assumption}
We can now define a notion of distance between the measures $\mu, \nu$ by minimizing an accumulated cost $\int_0^1 \|v_t\|_\b^p dt$ over all transports from $\mu$ to $\nu$. A rigorous construction is given below, where we emphasize that the choice of $p$ does not matter due to time-reparameterization:

\begin{proposition}
\label{prop:independentofp}
     For any $\mu, \nu \in \p(\r^d)$, let $\mathcal{S}_{\mu \to \nu}$ denote the class of pairs $(v_t, \rho_t)_{t \in [0, 1]}$ s.t.:
    \begin{enumerate}
        \item[(i)] $t \mapsto \rho_t$ is narrowly continuous,
        \item[(ii)] $v_t \in \b$ for a.e. $t$ and the maps $t \mapsto \|v_t\|_{\b} $ and $ (t, x) \mapsto v(t, x)$ are Borel measurable,
        \item[(iii)] 
        $\int_0^1 \int_{\r^d} |v_t(x)| d\rho_t(x) dt < \infty,$
        \item[(iv)] the continuity equation (\ref{eqn:informalcontinuityequation}) holds in the weak sense, see (8.1.3) of \cite{ags}.
    \end{enumerate}
    For every $p \in [1, \infty)$, define the quantity
    \begin{equation}D_{\b, p}(\mu, \nu) := \inf_{(v_t, \rho_t)_{t \in [0, 1]} \in \mathcal{S}_{\mu \to \nu}} \left(\int_0^1 \|v_t\|_{\b}^p dt\right)^{1/p},\end{equation}
    with a value of $+\infty$ if no admissible transport of finite energy exists.
    Then, it holds that 
    \begin{equation}
    \label{eqn:independentofp}
    D_{\b, p}(\mu, \nu) = D_{\b, p'}(\mu, \nu) \quad (\forall p, p' \in [1, \infty)). \end{equation}
\end{proposition}

\begin{proof}
    It suffices to prove that $D_{\b, p}(\mu, \nu) = D_{\b, 1}(\mu, \nu)$ for every $p \in [1, \infty)$. Naturally, for any admissible pair Hölder's inequality implies
    \begin{equation}\int_0^1 \|v_t\|_{\b} dt \leq \left(\int_0^1 \|v_t\|_{\b}^p dt\right)^{1/p}\end{equation}
    and so $D_{\b, 1}(\mu, \nu) \leq D_{\b, p}(\mu, \nu)$. If $D_{\b, 1}(\mu, \nu) = +\infty$ then the reverse inequality is trivial, and otherwise let $(v_t, \rho_t)_{t \in [0, 1]} \in \mathcal{S}_{\mu \to \nu}$ be admissible with finite length
    \begin{equation}L := \int_0^1 \|v_t\|_{\b} dt < \infty.\end{equation}
    Fix $\delta > 0$ and define
    \begin{equation}a_\delta(t) := \frac{\int_0^t \|v_\tau\|_{\b}d\tau + \delta t}{L + \delta}.\end{equation}
    Then, $a_\delta: [0, 1] \to [0, 1]$ is strictly increasing and absolutely continuous with $a'_\delta(t) = \frac{\|v_t\|_{\b} + \delta}{L+\delta} > 0$ for a.e. $t$, and so its inverse $\theta_\delta := a_\delta^{-1}$ is $\frac{L+\delta}{\delta}$-Lipschitz and therefore absolutely continuous. Define
    \begin{equation}\tilde{\rho}_\tau := \rho_{\theta_\delta(\tau)}, \quad \tilde{v}_\tau := \theta'_\delta(\tau) v_{\theta_\delta(\tau)}.\end{equation}
    For any $\varphi \in C^\infty_c((0, 1) \times \r^d)$, if we define $\psi(t, x) := \varphi(a_\delta(t), x)$ then
    testing $\psi$\footnote{Here, we may test $\psi$ by approximating $a_\delta$ in $W^{1,1}$ by smooth functions and using dominated convergence due to (iii).} against
    the original continuity equation (\ref{eqn:informalcontinuityequation}) gives
    \begin{align}0 &= \int_0^1 \int_{\r^d} \left(a_\delta'(\tau) \partial_t\varphi(a_\delta(\tau), x) + \la v_\tau(x), \nabla_x \varphi(a_\delta(\tau), x) \ra\right) d\rho_\tau(x) d\tau\\
     &= \int_0^1 \int_{\r^d} \left(\partial_t\varphi(t, x) + \la  \theta_\delta'(t){v}_{\theta_\delta(t)}(x), \nabla_x \varphi(t, x) \ra\right) d\tilde{\rho}_t(x) dt \\
    &= \int_0^1 \int_{\r^d} \left(\partial_t\varphi(t, x) + \la \tilde{v}_t(x), \nabla_x \varphi(t, x) \ra\right) d\tilde{\rho}_t(x) dt,
    \end{align}
    where the second line uses the absolutely continuous time change $t =a_\delta(\tau) $ along with the chain rule. Similarly, change of variables implies
    \begin{equation}\int_0^1 \int_{\r^d} |\tilde{v}_t| d\tilde{\rho}_t dt = \int_0^1 \int_{\r^d} \theta_\delta'(t) |v_{\theta_\delta(t)}| d\rho_{\theta_\delta(t)} dt = \int_0^1 \int_{\r^d} |v_\tau| d\rho_\tau d\tau < \infty\end{equation}
    and so $(\tilde{v}_t, \tilde{\rho}_t)_{t \in [0, 1]} \in \mathcal{S}_{\mu \to \nu}$. Furthermore, for a.e. $t \in [0, 1]$ we know
    \begin{equation}\|\tilde{v}_t\|_{\b} = \theta'_\delta(t) \|v_{\theta_\delta(t)}\|_{\b} = \frac{L+\delta}{\|v_{\theta_\delta(t)}\|_{\b} + \delta} \cdot \|v_{\theta_\delta(t)}\|_{\b} \leq L + \delta\end{equation}
    and therefore
    \begin{equation}D_{\b, p}(\mu, \nu) \leq \left(\int_0^1 \|\tilde{v}_t\|_{\b}^pdt\right)^{1/p} \leq L + \delta.\end{equation}
    Taking $\delta \downarrow 0$ and taking the infimum over all finite-length admissible pairs proves the claim.
\end{proof}

\begin{definition}[$D_\b$ metric]
\label{def:dbpmetrics}
For $\mu, \nu \in \p(\r^d)$ we denote the class $\mathcal{S}_{\mu \to \nu}$ as in Proposition \ref{prop:independentofp} and we denote the shared quantity (\ref{eqn:independentofp}) by
    \begin{equation}
    \label{eqn:defofDb}
    D_{\b}(\mu, \nu) :=   \inf_{(v_t, \rho_t)_{t \in [0, 1]} \in \mathcal{S}_{\mu \to \nu}} \int_0^1 \|v_t\|_{\b} dt,\end{equation}
    with a value of $+\infty$ if no admissible transport of finite length exists. 
\end{definition}

It is elementary to check the following properties, which justify us calling $D_\b$ a metric. Due to Assumption \ref{assumption:main}, this metric produces a topology stronger than the narrow topology:

\begin{proposition}
\label{prop:quasimetric}
$D_{\b}$ is an extended metric (i.e. a metric which may take the value $+\infty$). In particular, the following hold for all $\mu, \nu \in \p(\r^d)$:
\begin{enumerate}
    \item[(i)] $D_{\b}(\mu, \nu) = D_{\b}(\nu, \mu)$.
    \item[(ii)] $D_{\b}(\mu, \nu) = \inf_{\rho \in \p(\r^d)}\left\{ D_{\b}(\mu, \rho) + D_{\b}(\rho, \nu)\right\}$.
    \item[(iii)] $D_{\b}(\mu, \nu) = 0 \iff \mu = \nu$.
\end{enumerate}
Furthermore, if $(\mu_n)_{n \in \n} \subset \p(\r^d)$ and $\mu \in \p(\r^d)$ are s.t. $\lim_{n \to \infty}D_{\b}(\mu_n, \mu) = 0$ then we have the narrow convergence $\mu_n \rightharpoonup \mu$.
\end{proposition}

\begin{proof}
    (i) follows from the observation that if $(v_t, \rho_t) \in \mathcal{S}_{\mu \to \nu}$ then $(-v_{1-t}, \rho_{1-t}) \in \mathcal{S}_{\nu \to \mu}$ and $\|-v_{1-t}\|_{\b} = \|v_{1-t}\|_{\b}$. 
    
    Our proof of (ii) is taken from \href{https://mathoverflow.net/questions/166307/pde-based-triangle-inequality-for-optimal-transportation}{this MathOverflow post}. Suppose first that
    \begin{equation}d_1 := D_{\b,1}(\mu, \rho), \quad d_2 := D_{\b,1}(\rho, \nu)\end{equation}
    are both finite (if not, the inequality is trivial). Fix $\epsilon > 0$ and let $(v_t, \rho_t) \in \mathcal{S}_{\mu \to \rho}$ and $(w_t, \eta_t) \in \mathcal{S}_{\rho \to \nu}$ be s.t.
    \begin{equation}\int_0^1 \|v_t\|_{\b} dt \leq d_1 + \epsilon, \quad\int_0^1 \|w_t\|_{\b} dt \leq d_2 + \epsilon.\end{equation}
Note that if $\tau \in (0, 1)$ and we define
    \begin{equation}\tilde{v}_t := \begin{cases}
        v_{t/\tau}/\tau & t \leq \tau \\
        w_{(t-\tau)/(1-\tau)}/(1-\tau) & t > \tau
    \end{cases}, \quad \tilde{\rho}_t := \begin{cases}
        \rho_{t/\tau} & t \leq \tau \\
        \eta_{(t - \tau)/(1-\tau)} & t > \tau
    \end{cases},\end{equation}
    then $\tilde{\rho}_{\tau} = \rho$ and $(\tilde{v}, \tilde{\rho}) \in \mathcal{S}_{\mu \to \nu}$. Furthermore,
    \begin{align}\int_0^1 \|\tilde{v}_t\|_{\b}dt &={\tau^{-1}}\int_0^{\tau} \|{v}_{t/\tau}\|_{\b} dt+ {(1-\tau)^{-1}}\int_{\tau}^{1} \|{w}_{(t-\tau)/(1-\tau)}\|_{\b} dt  \\&\leq (d_1+\epsilon) + (d_2+\epsilon), \end{align}
    and therefore taking the infimum over transports implies
    \begin{equation}D_{\b,1}(\mu, \nu) \leq d_1 + d_2 + 2\epsilon.\end{equation}
    Taking $\epsilon \downarrow 0$ shows the $\leq$ direction; for equality we may take $\rho = \mu$.

    For (iii), note that if $D_{\b,1}(\mu, \nu) = 0$ then for each $n \in \n$ there is some $(v_t^n, \rho_t^n) \in \mathcal{S}_{\mu \to \nu}$ s.t. 
    \begin{equation}\int_0^1 \|v_t^n\|_{\b} dt \leq \frac{1}{n}.\end{equation}
    For any $\varphi \in C^\infty_c(\r^d)$, by testing against the continuity equation we find
    \begin{equation}\int \varphi d\nu - \int \varphi d\mu = \int_0^1 \int \la v_t^n, \nabla \varphi\ra d\rho_t^n dt.\end{equation}
    Applying Assumption \ref{assumption:main} and denoting by $C_\varphi$ the implied constant, we find
    \begin{align}\left|\int \varphi d\nu - \int \varphi d\mu\right| &\leq C_\varphi  \cdot \left(\int_0^1 \|v_t^n\|_{\b} dt\right) \leq \frac{C_\varphi }{n}.\end{align}
    Letting $n \to \infty$ implies $\int \varphi d\nu = \int \varphi d\mu$, and since this holds for all $\varphi \in C^\infty_c(\r^d)$, we find $\mu = \nu$. If $\mu = \nu$ then the null drift is admissible, proving (iii).

    For the last statement, let $(v_t, \rho_t)_{t \in [0, 1]} \in \mathcal{S}_{\mu \to \nu}$ be admissible. Integrating the continuity equation against $\varphi$ yields
\begin{equation}\int_{\r^d} \varphi d\nu - \int_{\r^d} \varphi d\mu = \int_0^1 \int_{\r^d} \la v_t, \nabla \varphi\ra d\rho_t dt\end{equation}
and so 
\begin{align}
    \left|\int_{\r^d} \varphi d\nu - \int_{\r^d} \varphi d\mu\right| &\leq C_\varphi \int_0^1 \|v_t\|_{\b} dt,
\end{align}
Taking the infimum over admissible transports,
\begin{equation}
     \left|\int_{\r^d} \varphi d\nu - \int_{\r^d} \varphi d\mu\right|  \leq C_\varphi D_{\b}(\mu, \nu).
\end{equation}
Suppose now that $D_{\b}(\mu_n, \mu) \to 0$. The above estimate implies that
\begin{equation}\int_{\r^d} \varphi d\mu_n \to \int_{\r^d} \varphi d\mu \quad (\forall \varphi \in C^\infty_c(\r^d)),\end{equation}
from which narrow convergence follows from e.g. \cite[Remark 5.1.6]{ags}.
\end{proof}

We finish by showing that the $D_{\h_k}$ and $D_{\b_\phi}$ constructions we consider satisfy Assumption \ref{assumption:main}:

\begin{lemma}
    Assumption \ref{assumption:main} is satisfied in either of the following cases:
    \begin{itemize}
        \item if a kernel $k: \r^d \times \r^d \to \r^{d \times d}$ satisfies Assumption \ref{assumption:kernel} and $\b = \h_k$.
        \item if an activation $\phi: \r \to \r$ satisfies Assumption \ref{assumption:activationfunciton} and $\b = \b_\phi$.
    \end{itemize}
\end{lemma}
\begin{proof}
    Let $\varphi \in C^\infty_c(\r^d)$, and let $K := \operatorname{spt}(\varphi)$ be its compact support set. It suffices to show that there is some $C_K < \infty$ s.t.
    \begin{equation}
        \sup_{x \in K}\|v(x)\| \leq C_K \|v\|_\b \quad (\forall v \in \b).
    \end{equation}
    In the Stein case, we know that 
    \begin{equation}
        \|v(x)\| \leq \|v\|_{\h_k} \cdot \|k(x, x)\|_{\operatorname{op}}^{1/2},
    \end{equation}
    and so compactness of $K$ and continuity of $k$ yield that $C_K := \sup_{x \in K} \|k(x, x)\|_{\operatorname{op}}^{1/2} < \infty$. In the Barron case, if $\phi$ is bounded then we know
    \begin{equation}
        \|v(x)\| \leq \|v\|_{\b_\phi} \cdot \|\phi\|_\infty,
    \end{equation}
    while if $\phi$ is positively $s$-homogeneous then
    \begin{equation}
        \|v(x)\| \leq \|v\|_{\b_\phi} \cdot \max\{|\phi(1)|, |\phi(-1)|\} \cdot (1 + \|x\|)^s.
    \end{equation}
    \end{proof}

\subsection{Wasserstein and KL lower bounds on $D_{\b}$}
\label{subsec:proofofWassersteinKLlowrbounds}

In this subsection we prove some general lower bounds on the dynamical transport distances: namely, if $\b\hookrightarrow C_b(\r^d)$ with sup norm then $W_\infty \lesssim D_\b$ while if $\b \hookrightarrow W^{1,\infty}(\r^d)$ then $\operatorname{KL} \lesssim D_\b + D_\b^2$ for sufficiently regular endpoints. These do not play a role in the results of the main paper, but they are useful tools and offer interesting consequences for our generative modeling applications.

\begin{lemma}[Wasserstein and KL control]
\label{lemma:appendixwassersteinandKLcontrol}
    Let $\b$ be a Banach space of functions from $\r^d \to \r^d$ s.t.
    \begin{equation}
    \label{eqn:appendixboundedM0}
        \|w\|_\infty \leq M_0 \|w\|_\b \quad (\forall w \in \b)
    \end{equation}
    for some $M_0 < \infty$. Then, for every $p \in [1, \infty]$ we have 
    \begin{equation}
        W_p(\mu, \nu) \leq M_0 D_\b(\mu, \nu) \quad (\forall \mu, \nu \in \p(\r^d)).
    \end{equation}
    Suppose further that $\b \hookrightarrow W^{1, \infty}$ continuously and
    \begin{equation}
        \label{eqn:appendixboundedM1}\|\operatorname{div}w\|_{L^\infty} \leq M_1 \|w\|_\b \quad (\forall w \in \b)
    \end{equation}
    for some $M_1 < \infty$. Let $\mu \propto e^{-V}$ be s.t. $\nabla V$ is $\beta$-Lipschitz and $S_\mu := \mathbb{E}_{\mu}\|\nabla V\| < \infty$. Then,
        \begin{equation}
        \operatorname{KL}(\nu \| \mu) \leq (M_0 S_\mu + M_1) D_{\b}(\mu, \nu) + \frac{\beta M_0^2}{2}D_{\b}(\mu, \nu)^2 \quad (\forall \nu \in \p(\r^d)).
    \end{equation}
\end{lemma}
\begin{proof}
Fix an admissible $(v_t, \rho_t)_{t \in [0, 1]} \in \mathcal{S}_{\mu \to \nu}$ and write $L := \int_0^1 \|v_t\|_\b dt$. By the superposition principle for the weak continuity equation, there is a probability measure $\mathbb{Q}$ on absolutely continuous curves $\omega:[0, 1] \to \r^d$ s.t. $(e_t)_\# \mathbb{Q} = \rho_t$ and $\dot{\omega}_t = v_t(\omega_t)$ for $\mathbb{Q}$-a.e. $\omega$. Hence,
\begin{equation}
    |\omega_1 - \omega_0| \leq \int_0^1 \|v_t\|_\infty dt \leq M_0 L
\end{equation}
for $\mathbb{Q}$-a.e. $\omega$, which means $(e_0, e_1)_\# \mathbb{Q}$ is a coupling of $\mu$ and $\nu$ supported on pairs with distance at most $M_0L$. Taking an infimum over transports yields $W_\infty(\mu, \nu) \leq M_0 D_\b(\mu, \nu)$.

Now, assume that $\b \hookrightarrow W^{1,\infty}$ continuously and that we take Lipschitz representatives. This means that
\begin{equation}
    \int_0^1 (\|v_t\|_\infty + \operatorname{Lip}(v_t)) dt < \infty
\end{equation}
for any finite-action drifts $(v_t, \rho_t)_{t \in [0, 1]}$, which ensures that the corresponding ODE has a unique flow $X_t$ with $(X_t)_\# \mu = \rho_t$; this lets us upgrade from the weak continuity equation to this ODE flow for any finite-action path. Now, the change of variables and Jacobian formulas imply
\begin{equation}
\label{eqn:proofappendixklchangeofvariables}
    \operatorname{KL}(\nu \| \mu) = \int_{\r^d} \left(V(X_1(x)) - V(x) - \int_0^1 \operatorname{div}v_t(X_t(x)) dt\right) d\mu(x).
\end{equation}
Since $\nabla V$ is $\beta$-Lipschitz we know
\begin{equation}
    V(X_1(x)) - V(x) \leq \la \nabla V(x), X_1(x) - x\ra + \frac{\beta}{2}\|X_1(x) - x\|^2.
\end{equation}
We know $\|X_1(x) - x\| \leq \int_0^1 \|v_t\|_\infty dt \leq M_0 L$ as before, and also
\begin{equation}
    -\int_0^1 \operatorname{div}v_t(X_t(x))dt \leq \int_0^1 \|\operatorname{div}v_t\|_{L^\infty} dt \leq M_1 L.
\end{equation}
Plugging these into (\ref{eqn:proofappendixklchangeofvariables}) yields that
\begin{equation}
    \operatorname{KL}(\nu \| \mu) \leq M_0 L \int_{\r^d} \|\nabla V\| d\mu + \frac{\beta M_0^2}{2}L^2 + M_1L,
\end{equation}
after which the claim follows by taking an infimum over transports.
\end{proof}

In the Stein case $\b = \h_k$, the properties (\ref{eqn:appendixboundedM0}) and (\ref{eqn:appendixboundedM1}) are implied by $\sup_x \|k(x, x)\|_{\operatorname{op}}^{1/2} \leq M_0$ and $\sup_x \left(\operatorname{div}_x \operatorname{div}_y k(x, y)|_{y=x}\right)^{1/2} \leq M_1$, respectively. In the Barron case $\b = \b_\phi$, they are implied by $\|\phi\|_\infty \leq M_0$ and $\|\phi'\|_{L^\infty} \leq M_1$, respectively. This directly implies Proposition \ref{prop:wassersteinkllowerbound}:

\begin{proof}[Proof of Proposition \ref{prop:wassersteinkllowerbound}]
The $W_\infty$ bound follows from Lemma \ref{lemma:appendixwassersteinandKLcontrol}.
   For the KL bound, the assumptions of Lemma \ref{lemma:appendixwassersteinandKLcontrol} hold with $M_0, M_1 = M$, and with $\mu = \gamma_d$ we have $\beta = 1$ and $S_{\gamma_d} \leq \sqrt{d}$. Thus, with $K := \operatorname{KL}(\pi \| \gamma_d)$ and $D := D_{\b_\phi}(\gamma_d, \pi)$ we find
   \begin{equation}
       K \leq M(\sqrt{d} + 1) D + \frac{M^2}{2}D^2 \leq 2M\sqrt{d}D + \frac{M^2}{2}D^2.
   \end{equation}
   At least one term on the right must be at least $K/2$, and so either $MD \geq \frac{K}{4\sqrt{d}}$ or $MD \geq \sqrt{K}$.
\end{proof}

\subsection{Gaussian-to-Gaussian transport}
We first analyze the optimal and reverse OU transports in the setting where the source and target are both Gaussian. For $W_2$ optimal transport and reverse OU the drifts are always affine, which means that intermediate measures remain Gaussian and so we can analyze mean and covariance dynamics explicitly without relying on perturbative arguments. For the $D_{\h_k}$ and $D_{\b_\phi}$ settings we must impose this structure ourselves, i.e. restrict to 
kernels and activations which are affine. 

For simplicity, throughout this subsection we take $\gamma_d = \mathcal{N}(0, \1_d)$ the standard Gaussian and $\pi = \mathcal{N}(m, \Sigma)$ an arbitrary nondegenerate Gaussian with $\Sigma \succ 0$. The general case should follow through similar methods.

\subsubsection{Optimal transport distances}
First, we recall that the Wasserstein distance has an exact form for Gaussians:
\begin{lemma}[Example 1.19 of \cite{jnwtextbook}]
\label{lemma:W2distancebetweengaussians}
    $W_2(\gamma_d, \pi) = \sqrt{\|m\|^2 + \|\Sigma^{1/2} - \1_d\|_F^2}$.
\end{lemma}

Moving on to the Stein case, we restrict to bilinear kernels in order to ensure affine drifts. This lets us reduce to a finite-dimensional matrix control problem which can be solved up to universal constants:

\begin{lemma}[Stein distance between Gaussians]
\label{lemma:steingaussiantogaussian}
    Let $\alpha > 0$ and $A \in \r^{d \times d}$ be symmetric and positive definite, and let $k(x, y) := (\alpha + x^\top A y) \1_d$ (which implies Assumption \ref{assumption:kernel} is satisfied). 
    If we denote
    \begin{equation}D := \frac{1}{\sqrt{\|A\|_{\operatorname{op}}}} \left(\log\left(1 + \sqrt{\frac{\|A\|_{\operatorname{op}}}{\alpha}}\|m\|\right) + \frac{1}{2} \|\log \Sigma\|_F\right), \end{equation}
    for $\|\cdot\|_F$ the Frobenius norm, then with $\kappa(A)$ the condition number of $A$ it holds that
    \begin{equation}\frac{1}{2}D \leq {D_{\h_k}(\gamma_d, \pi)} \leq 2\sqrt{\kappa(A)}D.\end{equation}
    In particular, with $k(x, y) = (1 + \la x, y\ra)\1_d$ we have
    \begin{equation}D_{\h_k}(\gamma_d, \pi) \asymp \log \left(1 + \|m\|\right) + \|\log \Sigma\|_F.\end{equation}
\end{lemma}
\begin{proof}
Let us denote $a := \|A\|_{\operatorname{op}}$.
   Every $v \in \h_k$ is of the form $v(x) = Mx + c$ with the norm
   \begin{equation}\|v\|_{\h_k}^2 = \|MA^{-1/2}\|_F^2 + \frac{\|c\|^2}{\alpha}.\end{equation}
   Since these are affine, any finite-length admissible transport starting from $\gamma_d$ remains Gaussian; we write
   \begin{equation}
   \label{eqn:prooftherhoandvofchoice}
       \rho_t = \mathcal{N}(m_t, \Sigma_t), \quad v_t(x) = M_tx + c_t
   \end{equation}
   which follow the dynamics
   \begin{equation}\label{eqn:proofmatrixdynamics}\dot{m}_t = M_t m_t + c_t, \quad \dot{\Sigma}_t = M_t\Sigma_t + \Sigma_t M_t^\top, \quad m_0 = 0, \Sigma_0 = \1_d.\end{equation}
   Let us denote
   \begin{equation}
   \label{eqn:proofrkhsnorminthebilinearcase}
       \ell_t := \|v_t\|_{\h_k} = \left(\|M_tA^{-1/2}\|_F^2 + \frac{\|c_t\|^2}{\alpha}\right)^{1/2}.
   \end{equation}
   If we let $L_t$ solve the dynamics
   \begin{equation}
       \label{eqn:proofcholeskydynamics}
       \dot{L}_t = M_t L_t, \quad L_0 = \1_d
   \end{equation}
   then $\Sigma_t = L_t L_t^\top$. If $s_1(t), \ldots, s_d(t)$ are the singular values of $L_t$ then
we may take an SVD $L_t = U_t S_t V_t^\top$ with $S_t = \operatorname{diag}(s_1(t), \ldots, s_d(t))$. In other words, $L_t v_j = s_j(t) u_j$ with $u_j, v_j$ the left and right unit singular vectors, after which the standard formula yields \begin{equation}s_j'(t) = u_j^\top \dot{L}_t v_j = u_j^\top M_tL_t v_j = s_j(t) u_j^\top M_t u_j\end{equation}
whenever singular values do not cross, and so for a.e. $t$ we have
\begin{equation}\frac{d}{dt} \log s_j(t) = u_j^\top M_t u_j.\end{equation}
Since $u_j^\top M_tu_j$ are the diagonal entries of $U_t^\top M_t U_t$ we find
\begin{equation}
\label{eqn:proofell2estimatesingularvalues}
    \sum_{j=1}^d |u_j^\top M_tu_j|^2 \leq \sum_{i,j = 1}^d \left|(U_t^\top M_tU_t)_{ij}\right|^2 = \|U_t^\top M_tU_t\|_F^2 = \|M_t\|_F^2
\end{equation}
and therefore we have
   \begin{equation}\left(\sum_{j=1}^d \left|\frac{d}{dt}\log s_j(t)\right|^2\right)^{1/2} \leq \|M_t\|_F.\end{equation}
   Integrating over time and using that $\log s_j(0) = 0$,
    \begin{equation}\left(\sum_{j=1}^d |\log s_j(1)|^2 \right)^{1/2} = \|\log s(1)\|_2 \leq \int_0^1 \left\|\frac{d}{dt} \log s(t)\right\|_2 dt \leq \int_0^1 \|M_t\|_F dt.\end{equation}
    In particular, since the eigenvalues of $\Sigma$ are $s_1(1)^2, \ldots, s_d(1)^2$ we find that
    \begin{equation}\|\log \Sigma\|_F^2 = \sum_{j=1}^d \left|\log(s_j(1)^2)\right|^2 = 4 \|\log s(1)\|_2^2\end{equation}
    and hence
    \begin{equation}\frac{1}{2}\|\log \Sigma\|_F \leq \int_0^1 \|M_t\|_F dt \leq \sqrt{a} \int_0^1 \ell_t dt.\end{equation}
    For the mean, if $r_t := \|m_t\|$ then 
    \begin{equation}\dot{r}_t \leq \|M_tm_t\| + \|c_t\| \leq \sqrt{a}r_t \|M_tA^{-1/2}\|_F + \sqrt{\alpha}\frac{\|c_t\|}{\sqrt{\alpha}} \leq \sqrt{ar_t^2 + \alpha} \cdot \left(\|M_tA^{-1/2}\|_F^2 + \frac{\|c_t\|^2}{\alpha}\right)^{1/2}\end{equation}
    and so
    \begin{equation}\frac{1}{\sqrt{a}}\frac{d}{dt} \left[\sinh^{-1} \left(\sqrt{\frac{a}{\alpha}}r_t\right)\right] = \frac{\dot{r}_t}{\sqrt{ar_t^2 + \alpha}} \leq \ell_t.\end{equation}
    Integrating in time and using that $r_0 = 0$, we find
    \begin{equation}\sinh^{-1}\left(\sqrt{\frac{a}{\alpha}} \|m\|\right) \leq \sqrt{a}\int_0^1 \ell_t dt.\end{equation}
    We may now take an infimum over finite-energy transports $(v_t, \rho_t)_{t \in [0, 1]}$ and use that $\sinh^{-1}(z) \geq \log(1+z)$ to find
    \begin{equation}D_{\h_k}(\gamma_d, \pi) \geq \frac{1}{\sqrt{a}} \max\left\{\frac{1}{2}\|\log \Sigma\|_F, \; \log\left(1 + \sqrt{\frac{a}{\alpha}} \|m\|\right)\right\} \geq \frac{1}{2}D.\end{equation}
    For the upper bound, we start with the triangle inequality: with $\pi_0 := \mathcal{N}(0, \Sigma)$ the centered Gaussian with correct variance we have
    \begin{equation}D_{\h_k}(\gamma_d, \pi) \leq D_{\h_k}(\gamma_d, \pi_0) + D_{\h_k}(\pi_0, \pi)\end{equation}
    by Proposition \ref{prop:quasimetric}(ii). To bound the first term, we take $\rho_t := \mathcal{N}(0, \Sigma^t)$ and $v_t(x) := M_tx$ with $M_t := \frac{1}{2}\log \Sigma$, so that we have
    \begin{equation}\frac{d}{dt}\Sigma^t = M_t \Sigma^t + \Sigma^t M_t.\end{equation}
This is an admissible transport with time-invariant drift, and we use (\ref{eqn:proofrkhsnorminthebilinearcase}) to calculate
\begin{equation}
\label{eqn:prooftriangleinequalityterm1}
    D_{\h_k}(\gamma_d, \pi_0) \leq \int_0^1 \|v_t\|_{\h_k}dt = \frac{1}{2}\|(\log \Sigma)A^{-1/2}\|_F \leq \frac{1}{2\sqrt{\lambda_{\text{min}}(A)}} \|\log \Sigma\|_F.
\end{equation}
If $m = 0$ the desired bound trivially holds; otherwise, write $u := \frac{m}{\|m\|}$. We define the affine maps
\begin{equation}
\label{eqn:proofmeantermupperbound1}
    T_t(x) := (\1 + (s(t) - 1)uu^\top)x + h(t) u
\end{equation}
for $s: [0, 1] \to (0, \infty)$ and $h: [0, 1] \to \r$ to be determined. We take $s(0) = s(1) = 1$, $h(0) = 0$, and $h(1) = \|m\|$ to ensure that if we define $\rho_t := (T_t)_\# \pi_0$ then $\rho_0 = \pi_0$ and $\rho_1 = \pi$. The Eulerian velocity of this path is $v_t(x) = M_tx + c_t$ for
\begin{equation}
\label{eqn:proofmeantermupperbound2}
    M_t = \frac{s'(t)}{s(t)}uu^\top, \quad c_t = \left(h'(t) - \frac{s'(t)}{s(t)}h(t)\right)u
\end{equation}
and so by (\ref{eqn:proofrkhsnorminthebilinearcase}) we have
\begin{equation} \label{eqn:proofrkhsnorminthebilinearcase2}\|v_t\|_{\h_k}^2 = \beta \left(\frac{s'(t)}{s(t)}\right)^2 + \frac{1}{\alpha} \left(h'(t) - \frac{s'(t)}{s(t)}h(t)\right)^2,
\end{equation}
where we denote $\beta := u^\top A^{-1}u$.
Hence, if we change variables to
\begin{equation}y_t := \frac{1}{s(t)}, \quad z_t := \frac{h(t)}{\sqrt{\alpha \beta}s(t)}\end{equation}
then the derivatives satisfy
\begin{equation}\dot{y}_t = \frac{-s'(t)}{s(t)^2} , \quad \dot{z}_t = \frac{1}{\sqrt{\alpha \beta}} \frac{h'(t) s(t) - h(t) s'(t)}{s(t)^2}.\end{equation}
Using (\ref{eqn:proofrkhsnorminthebilinearcase2}) we find
\begin{equation}\int_0^1 \|v_t\|_{\h_k} dt = \sqrt{\beta} \cdot \int_0^1 \frac{\sqrt{\dot{y}_t^2 + \dot{z}_t^2}}{y_t}dt.\end{equation}
This is exactly $\sqrt{\beta}$ times the length of the curve $w(t) := (z_t, y_t) \in \mathbb{H}$, where $\mathbb{H} := \{(x, y):\; y > 0\}$ is the Poincaré upper half-plane with metric $dw^2 = \frac{dx^2 + dy^2}{y^2}$. Using the \href{https://en.wikipedia.org/wiki/Poincaré_half-plane_model#Distance_calculation}{formula for the distance in the metric space $\mathbb{H}$}, if $w:[0, 1] \to \mathbb{H}$ is a geodesic then
\begin{align}\int_0^1 \|v_t\|_{\h_k}dt &= \sqrt{\beta} \cdot d_{\mathbb{H}}(w(0), w(1)) \\&=\sqrt{\beta} \cdot 2 \sinh^{-1}\left(\frac{\|w(1) - w(0)\|_{\r^2}}{2\sqrt{y_0y_1}}\right)\\
&= \sqrt{\beta} \cdot 2\sinh^{-1}\left(\frac{\|m\|}{2\sqrt{\alpha \beta}}\right).
\end{align}
We take $w(t) = (z_t, y_t)$ to be any constant-speed hyperbolic geodesic from $w(0) = (0, 1)$ to $w(1) = (\|m\|/\sqrt{\alpha \beta}, 1)$ and define $s = 1/y$ and $h = \sqrt{\alpha \beta }z / y$ in order to ensure that this holds.
Since $\sinh^{-1}(z/2) \leq \log(1+z)$ and $\frac{1}{a}\leq \beta \leq \frac{1}{\lambda_{\text{min}}(A)}  $ we find
\begin{equation}D_{\h_k}(\pi_0, \pi) \leq \frac{1}{\sqrt{\lambda_{\text{min}}(A)}}\cdot 2 \log\left(1 + \sqrt{\frac{a}{\alpha}}\|m\|\right).\end{equation}
Combining this with (\ref{eqn:prooftriangleinequalityterm1}) and the triangle inequality, we find
\begin{equation}D_{\h_k}(\gamma_d, \pi) \leq \frac{1}{\sqrt{\lambda_{\text{min}}(A)}} \cdot \left(\frac{1}{2}\|\log \Sigma\|_F + 2 \log \left(1 + \sqrt{\frac{a}{\alpha}}\|m\|\right)\right) \leq 2\sqrt{\kappa(A)} D.\end{equation}
\end{proof}

The Barron setting with identity activation has more or less the same behavior, with the key difference of changing the matrix norm from Frobenius ($\ell^2$ on singular values) to nuclear ($\ell^1$ on singular values):

\begin{lemma}[Barron distance between Gaussians]
\label{lemma:barrongaussiantogaussian}
    Let $\phi(z) = z$ (which satisfies Assumption \ref{assumption:activationfunciton}). Then, with $\|\cdot\|_*$ denoting the nuclear norm it holds that
    \begin{equation}\frac{1}{2}\left(\log\left(1 + \|m\|\right) + \frac{1}{2} \|\log \Sigma\|_*\right) \leq D_{\b_\phi}(\gamma_d, \pi) \leq 3\left(\log\left(1 + \|m\|\right) + \frac{1}{2} \|\log \Sigma\|_*\right).\end{equation}
\end{lemma}
\begin{proof}
 As in the proof of Lemma \ref{lemma:steingaussiantogaussian}, since $v \in \b_\phi$ has $v(x) = Mx + c$ then any finite-length admissible transport starting from $\gamma_d$ remains Gaussian. Let $\rho_t, v_t, M_t, c_t, m_t, \Sigma_t, L_t$ be as in (\ref{eqn:prooftherhoandvofchoice}), (\ref{eqn:proofmatrixdynamics}), and (\ref{eqn:proofcholeskydynamics}), let $s_1(t), \ldots, s_d(t)$ be the singular values of $L_t$, and let us denote
    \begin{equation}\ell_t := \|v_t\|_{\b_\phi} = \|M_t\|_* + \|c_t\|.\end{equation}
    If we replace the estimate (\ref{eqn:proofell2estimatesingularvalues}) with its $\ell_1$ version
    \begin{equation}\sum_{j=1}^d |u_j^\top M_tu_j| \leq \|M_t\|_*\end{equation}
    then we instead find that
    \begin{equation}\|\log s(1)\|_1 \leq \int_0^1 \left\|\frac{d}{dt}\log s(t)\right\|_1 dt \leq \int_0^1 \|M_t\|_* dt.\end{equation}
    In particular, since the eigenvalues of $\Sigma$ are $s_1(1)^2, \ldots, s_d(1)^2$ we have
    \begin{equation}\frac{1}{2} \|\log \Sigma\|_* = \|\log s(1)\|_1 \leq \int_0^1 \ell_tdt.\end{equation}
    For the mean, with $r_t := \|m_t\|$ then
    \begin{equation}\dot{r}_t \leq \|M_tm_t\| + \|c_t\| \leq r_t \|M_t\|_{\operatorname{op}} + \|c_t\| \leq r_t \|M_t\|_* + \|c_t\| \leq (1 + r_t) \ell_t,\end{equation}
    or in other words
    \begin{equation}\frac{d}{dt} \log (1+r_t) \leq \ell_t.\end{equation}
    Integrating and applying $r_0 = 0$ yields
    \begin{equation}\log(1+\|m\|) \leq \int_0^1 \ell_t dt,\end{equation}
    and so we find
    \begin{equation}D_{\b_\phi}(\gamma_d, \pi) \geq \max\left\{\frac{1}{2}\|\log \Sigma\|_*,\; \log \left(1 + \|m\|\right)\right\}.\end{equation}
    For the upper bound, we apply the same triangle inequality
    \begin{equation}D_{\b_\phi}(\gamma_d, \pi) \leq D_{\b_\phi}(\gamma_d, \pi_0) + D_{\b_\phi}(\pi_0, \pi)\end{equation}
    with $\pi_0 := \mathcal{N}(0, \Sigma)$. For the first term, we use the interpolation $\rho_t = \mathcal{N}(0, \Sigma^t)$ and $v_t(x) = M_tx$ with $M_t = \frac{1}{2}\log \Sigma$ to see that
    \begin{equation}D_{\b_\phi}(\gamma_d, \pi_0) \leq \frac{1}{2}\|\log \Sigma\|_*.\end{equation}
    Note that in light of our lower bound, the above is equality. 
    
    For the mean term in the upper bound, fix $s: [0, 1] \to (0, \infty)$ and $h: [0, 1] \to \r$ with $s(0) = s(1) = 1$, $h(0) = 0$, and $h(1) = \|m\|$ to be determined. Define $T_t$ as in (\ref{eqn:proofmeantermupperbound1}) and $\rho_t := (T_t)_\# \pi_0$, and let $v_t(x) = M_tx + c_t$ for $M_t, c_t$ as in (\ref{eqn:proofmeantermupperbound2}). This is an admissible transport from $\pi_0$ to $\pi$ with
    \begin{equation}
    \label{eqn:proofbarronnormformeanupperbound}
        \|v_t\|_{\b_\phi} = \left|\frac{s'(t)}{s(t)}\right| + \left|h'(t) - \frac{s'(t)}{s(t)}h(t)\right|.
    \end{equation}
    If $m= 0$ then we are done, and for $m \neq 0$ we take
    \begin{equation}s(t) := \begin{cases}
        (1 + \|m\|)^{-3t} & 0 \leq t < \frac{1}{3},\\
        (1 + \|m\|)^{-1} & \frac{1}{3} \leq t < \frac{2}{3}, \\
        (1 + \|m\|)^{-3(1-t)} & \frac{2}{3} \leq t \leq 1,
    \end{cases} \qquad h(t) := \begin{cases}
       0 & 0 \leq t < \frac{1}{3},\\
       \frac{(3t-1)\|m\|}{1 + \|m\|} & \frac{1}{3} \leq t < \frac{2}{3}, \\
        \|m\|s(t) & \frac{2}{3} \leq t \leq 1.
    \end{cases} \end{equation}
    Then, we have $s(0) = s(1) = 1$, $h(0) = 0$, and $h(1) = \|m\|$. Furthermore, using (\ref{eqn:proofbarronnormformeanupperbound}) we have
    \begin{equation}\|v_t\|_{\b_\phi} \leq \begin{cases}
     \left|\frac{s'(t)}{s(t)}\right|   & 0 \leq t < \frac{1}{3},\\
       |h'(t)| & \frac{1}{3} \leq t < \frac{2}{3}, \\
        \left|\frac{s'(t)}{s(t)}\right|  & \frac{2}{3} \leq t \leq 1,
    \end{cases}\end{equation}
    where $h = 0$ on the first interval, $s' = 0$ on the second, and $h' - \frac{s'}{s}h = 0$ on the third. Hence, since $|\frac{s'}{s} | = |(\log s)'|= 3\log(1+\|m\|)$ on the first and third intervals we have
    \begin{align}
        D_{\b_\phi}(\pi_0, \pi) &\leq \int_0^1 \|v_t\|_{\b_\phi} dt \\&\leq \frac{1}{3} \cdot (3 \log(1 + \|m\|)) + \frac{1}{3} \cdot \frac{3\|m\|}{1 + \|m\|} + \frac{1}{3} \cdot (3 \log(1+\|m\|)) \\&\leq 3 \log(1+\|m\|),
    \end{align}
    where we used that $\frac{z}{1+z} \leq \log(1+z)$. Thus,
    \begin{equation}D_{\b_\phi}(\gamma_d, \pi) \leq \frac{1}{2}\|\log \Sigma\|_* + 3 \log(1 + \|m\|) .\end{equation}
\end{proof}

\begin{remark}
\label{remark:steinbarronlineargaussiantogaussianmeanmovement}
The transports realizing the upper bounds for linear Stein and linear Barron are almost the same: they both use the triangle inequality (i.e. they move the mean and covariance in separate steps) and they use the exact same transport to move the covariance. Furthermore, the transports which move the mean both use a compress-translate-expand strategy to achieve logarithmic dependence on $\|m\|$. This is exactly analogous to the structure of a hyperbolic geodesic, and it matches the fact that hyperbolic distance between horizontally-separated points is logarithmic in the Euclidean separation.
\end{remark}

\subsubsection{Reverse OU energies}

We will also take this opportunity to compute the action of the reverse OU transport $\mathcal{E}_\b({\pi})$ from Lemma \ref{lemma:reverseOUsuboptimality} for the case $\pi = \mathcal{N}(m, \Sigma)$ with $\Sigma \succ 0$. Doing so helps us understand more about the reverse OU (sub)optimality:

\begin{lemma}[Reverse OU energies for Gaussians]
\label{lemma:reverseOUgaussiantogaussian}
    If we denote $C_a := (1-a^2)\1_d + a^2 \Sigma$ then the following bounds hold:
    \begin{itemize}
        \item \textbf{(Wasserstein).} With $\mathcal{E}_{W_2}(\pi) = \int_0^\infty \|\nabla \log \frac{P_\tau^* \pi}{\gamma_d}\|_{L^2(P_\tau^* \pi)} d\tau$, we have
        \begin{equation}W_2(\gamma_d, \pi) \leq \mathcal{E}_{W_2}(\pi) \leq \frac{\pi}{2} W_2(\gamma_d, \pi).\end{equation}

        \item \textbf{(Stein).} With $\b = \h_k$ for $k(x, y) = (1 + \la x, y\ra)\1_d$, we have 
        \begin{equation}\mathcal{E}_{\h_k}(\pi) \geq \frac{1}{\sqrt{2}}\left( \int_0^1 \|C_a^{-1}m\| da + \frac{1}{2} \|\log \Sigma\|_F\right).\end{equation}
        \item \textbf{(Barron).} With $\b = \b_\phi$ for $\phi(z) = z$, we have \begin{equation}\mathcal{E}_{\b_\phi}(\pi) = \int_0^1 \|C_a^{-1}m\| da + \frac{1}{2} \|\log \Sigma\|_* .\end{equation}
    \end{itemize}
\end{lemma}

\begin{proof}
    Letting $a := e^{-\tau}$, we know that
    $P_\tau^*\pi = \mathcal{N}(am, C_a)$
    and therefore that
    \begin{equation}\left(\nabla \log \frac{P_\tau^*\pi}{\gamma_d}\right)(x) = x - C_a^{-1}(x - am) = a^2 (\Sigma - \1_d)C_a^{-1}x + aC^{-1}_am.\end{equation}
    This is an affine drift, and we will now compute the respective tangent norms.

    For the Wasserstein case, if $X_a \sim \mathcal{N}(am, C_a)$ then
    \begin{equation}\left(\nabla \log \frac{P_\tau^*\pi}{\gamma_d}\right)(X_a) = am + a^2 (\Sigma - \1_d) C_a^{-1} (X_a - am)\end{equation}
    and so we use $d\tau = -da / a$ to find
    \begin{align}\mathcal{E}_{W_2}(\pi) &= \int_0^1 \left(\|m\|^2 + a^2 \|(\Sigma - \1_d)C_a^{-1/2}\|_F^2\right)^{1/2}da \\&= \int_0^1 \left(\|m\|^2 + \sum_{j=1}^d \left|\frac{a(\sigma_j^2 - 1)}{\sqrt{1 -a^2 + a^2\sigma_j^2}}\right|^2\right)^{1/2}da,\end{align}
    where $\sigma^2_1, \ldots, \sigma_d^2$ are the eigenvalues of $\Sigma$.
    Now, for any $a \in [0, 1)$ and $\sigma > 0$ we have that
    \begin{align}
        0 &\leq (1 - a^2(\sigma + 1))^2 \\&= 1 - 2a^2(\sigma + 1) + a^4 (\sigma + 1)^2 \\&= 1 + a^2 (\sigma^2 - 1) - a^2(1-a^2)(\sigma+1)^2,
    \end{align}
    and so $1-a^2 + a^2\sigma^2  \geq a^2(1-a^2)(\sigma + 1)^2$. In other words,
    \begin{equation}\frac{a^2(\sigma + 1)^2}{{1 - a^2 + a^2\sigma^2}} \leq \frac{1}{{1 - a^2}}\end{equation}
    and so multiplying by $(\sigma - 1)^2$ implies the inequality
    \begin{align}
       \left|\frac{a(\sigma_j^2 - 1)}{\sqrt{1 -a^2 + a^2\sigma_j^2}}\right|^2 \leq \frac{(\sigma_j - 1)^2}{1 - a^2} \quad (\forall a \in [0, 1)).
    \end{align}
    By Lemma \ref{lemma:W2distancebetweengaussians} we know $W_2(\gamma_d, \pi) = (\|m\|^2 + \sum_{j=1}^d (\sigma_j - 1)^2)^{1/2}$, and so
    \begin{align}
        \mathcal{E}_{W_2}(\pi) \leq W_2(\gamma_d, \pi) \cdot \int_0^1 \frac{1}{\sqrt{1-a^2}} da = \frac{\pi}{2}W_2(\gamma_d, \pi).
    \end{align}
    The lower bound $W_2(\gamma_d, \pi) \leq  \mathcal{E}_{W_2}(\pi) $ is obvious because $W_2$ is the infimum over all transports.

    For the Stein case with $k(x, y) = (1 + \la x, y\ra)\1_d$, we know $d\tau = -da / a$ and so
    \begin{equation}
    \label{eqn:proofsteinreverseouenergygaussian}
        \mathcal{E}_{\h_k}(\pi) = \int_0^1 \left(\|C_a^{-1}m\|^2 + a^2 \|(\Sigma - \1_d)C_a^{-1}\|_F^2\right)^{1/2}da ,
    \end{equation}
    which implies that
    \begin{equation}\mathcal{E}_{\h_k}(\pi) \geq \frac{1}{\sqrt{2}}\int_0^1 \left(\|C_a^{-1}m\| + a\|(\Sigma - \1_d)C_a^{-1}\|_F \right)da.\end{equation}
    Now, since $\frac{d}{da}\log C_a = 2a(\Sigma - \1_d)C_a^{-1}$ (note that $\Sigma - \1_d$ and $C_a$ commute) and $\log C_0 = 0, \log C_1 = \log \Sigma$, we have
    \begin{equation}
    \label{eqn:proofthefrobeniustriangleinequalitystuff}
        \frac{1}{2}\|\log \Sigma\|_F = \left\|\int_0^1 a(\Sigma - \1_d)C_a^{-1}da\right\|_F \leq \int_0^1 a\|(\Sigma - \1_d)C^{-1}_a\|_F da.
    \end{equation}

    For the Barron case with $\phi(z) = z$, we have
    \begin{equation}\mathcal{E}_{\b_\phi}(\pi) = \int_0^1 \left(\|C_a^{-1}m\| + a\|(\Sigma - \1_d)C_a^{-1}\|_*\right)da,\end{equation}
after which the result follows from the equalities
\begin{align}
    \int_0^1 a\|(\Sigma - \1_d)C_a^{-1}\|_* da = \sum_{j=1}^d \int_0^1 \frac{a|\sigma_j^2 - 1|}{1 +a^2(\sigma_j^2-1)}da = \frac{1}{2}\sum_{j=1}^d |\log \sigma_j^2| = \frac{1}{2}\|\log \Sigma\|_*.
\end{align}
\end{proof}

Looking at this result, one might hope for a matching upper bound in the linear Stein case (or at least for the $\|\log \Sigma\|_F$ dependence to be sharp up to constants). We show that this cannot be the case, albeit in a very mild way:

\begin{example}
\label{example:reverseOUsteinGaussiansuboptimality}
Let $\pi = \mathcal{N}(0, \Sigma)$ for $\Sigma = \operatorname{diag}(e^{d/\sqrt{1}}, e^{d/\sqrt{2}}, \ldots, e^{d/\sqrt{d}})$. Then, with $k(x, y) = (1 + \la x, y\ra)\1_d$ we have
\begin{equation}\mathcal{E}_{\h_k}\left(\pi\right) \geq  \frac{\sqrt{\log d}}{24} \cdot \|\log \Sigma\|_F \geq \frac{\sqrt{\log d}}{12} \cdot D_{\h_k}(\gamma_d, \pi).\end{equation}
\end{example}

\begin{proof}
    In the centered case, from (\ref{eqn:proofsteinreverseouenergygaussian}) we find that if $\sigma^2_1, \ldots, \sigma_d^2 \geq 1$ are the eigenvalues of $\Sigma$ then
    \begin{equation}\mathcal{E}_{\h_k}\left(\pi\right) = \int_0^1 \left(\sum_{j=1}^d \left(\frac{a(\sigma_j^2 - 1)}{1 + a^2(\sigma_j^2 - 1)}\right)^2\right)^{1/2} da.\end{equation}
    For $a \in [\sigma_j^{-1}, \sigma_{j+1}^{-1})$ and all $i \leq j$ we have $a^2(\sigma_i^2 - 1) \geq 1 - e^{-d/\sqrt{i}} \geq 1 - e^{-\sqrt{d}} \geq \frac{1}{2}$, and so
    \begin{equation}\frac{a(\sigma_i^2 - 1)}{1 + a^2(\sigma_i^2 - 1)} = \frac{1}{a} \cdot \frac{a^2(\sigma_i^2 - 1)}{1 + a^2(\sigma_i^2 - 1)} \geq \frac{1}{3a}.\end{equation}
    Plugging this in and denoting $\sigma_{d+1}^2 = e^{d / \sqrt{d+1}}$ for simplicity, we split the integral to get
\begin{align}
    \mathcal{E}_{\h_k}\left(\pi\right) &\geq \sum_{j=1}^d\int_{\sigma_j^{-1}}^{\sigma_{j+1}^{-1}} \left(\sum_{i=1}^j \left(\frac{a(\sigma_i^2 - 1)}{1 + a^2(\sigma_i^2 - 1)}\right)^2\right)^{1/2} da
\geq \sum_{j=1}^d \int_{\sigma_j^{-1}}^{\sigma_{j+1}^{-1}} \frac{\sqrt{j}}{3a}da \\
    &= \frac{1}{3}\sum_{j=1}^d \sqrt{j} \cdot \log\left(\frac{\sigma_j}{\sigma_{j+1}}\right)
    = \frac{1}{6}\sum_{j=1}^d \sqrt{j} \cdot \left(\frac{d}{\sqrt{j}} - \frac{d}{\sqrt{j+1}}\right) \\
    &= \frac{d}{6} \sum_{j=1}^d \left(1 - \sqrt{\frac{j}{j+1}}\right) \geq \frac{d}{6} \sum_{j=1}^d \frac{1}{2(1+j)} \\
    &\geq \frac{d}{24} \sum_{j=1}^d \frac{1}{j},
\end{align}
where we used that $1 - \sqrt{\frac{z}{1+z}} \geq \frac{1}{2(1+z)}$ for $z > 0$. On the other hand, we have
\begin{equation}\|\log \Sigma\|_F = d\left(\sum_{j=1}^d \frac{1}{j}\right)^{1/2}\end{equation}
and so
\begin{equation}\frac{\mathcal{E}_{\h_k}\left(\pi\right)}{\|\log\Sigma\|_F} \geq \frac{1}{24}\left(\sum_{j=1}^d \frac{1}{j}\right)^{1/2} \geq \frac{\sqrt{\log d}}{24}.\end{equation}
The comparison with $D_{\h_k}(\gamma_d, \pi)$ comes from (\ref{eqn:prooftriangleinequalityterm1}).
\end{proof}

\subsection{Perturbative bounds for rotation-equivariant Stein distances}
We first prove a perturbative first-order lower bound on the $D_{\h_k}$ distance along flows when the kernel is jointly continuous:

\begin{lemma}[First-order $D_{\h_k}$ lower bound]
\label{lemma:epsilonfirstorder}
Let $k: \r^d \times \r^d \to \r^{d \times d}$ satisfy Assumption \ref{assumption:kernel}.
    Suppose that $\varphi \in C_c^\infty(\r^d)$, $\mu \in \p(\r^d)$, and $w : [0, 1] \times \r^d \to \r^d$ satisfy the following:
    \begin{enumerate}
\item[(i)] $L :=   
        \sup_{\|u\|_{\h_k} \leq 1} \int_{\r^d} \la u, \nabla \varphi\ra d\mu
        \in [0, \infty)$.
        \item[(ii)] $w$ is jointly measurable and each $w_t$ is globally-Lipschitz with $\int_0^1 (|w_t(0)| + \operatorname{Lip}(w_t))dt <\infty$.
    \end{enumerate}
    For each $\epsilon > 0$ define $\nu_\epsilon := \operatorname{Law}(X_1)$ according to the flow
    \begin{equation}\dot{X}_t = \epsilon w_t(X_t), \quad X_0 \sim \mu.\end{equation}
    Then, with $F_1(x) := \int_0^1 w_t(x) dt$ we have
    \begin{equation}\liminf_{\epsilon \downarrow 0}\frac{D_{\h_k}(\mu, \nu_\epsilon)}{\epsilon} \geq \frac{\left|\int_{\r^d} \la \nabla \varphi, F_1\ra d\mu\right|}{L}\end{equation}
    where if $L = 0$ then the right hand side is $\infty$ if $\left|\int_{\r^d} \la \nabla \varphi, F_1\ra d\mu\right| > 0$ and 0 otherwise.
\end{lemma}
\begin{proof}
    For $\rho \in \p(\r^d)$ we denote
    \begin{equation}
        g_\rho := \int_{\r^d} k(\cdot, x) \nabla \varphi(x) d\rho(x) \in \h_k, \quad G(\rho) := \|g_\rho\|_{\h_k}.
    \end{equation}
    By the reproducing property, we know that
    \begin{equation}
        G(\rho) = \sup_{\|u\|_{\h_k} \leq 1}\int_{\r^d} \la u, \nabla \varphi\ra d\rho
    \end{equation}
    and so $G(\mu) = L$. Moreover, the reproducing property also ensures
    \begin{equation}
        G(\rho)^2 = \iint_{\r^d \times \r^d} \la \nabla \varphi(x), k(x, y) \nabla \varphi(y)\ra d\rho(x) d\rho(y).
    \end{equation}
    This integrand is bounded and continuous since $k$ is jointly continuous and $\nabla \varphi$ is compactly supported, and so $G$ is narrowly continuous. 

Assumption (ii) implies the Carathéodory conditions, yielding a unique flow. 
    Let $T_t^\epsilon$ be the flow of $\epsilon w_t$, so that $\nu_\epsilon = (T_1^\epsilon)_\# \mu$. For each $x \in \r^d$ we know that $T_t^\epsilon(x) \to x$ as $\epsilon \downarrow 0$ uniformly in $t$ by Grönwall's lemma. By integrating the continuity equation, we have
    \begin{equation}
        \frac{\varphi(T_1^\epsilon(x)) - \varphi(x)}{\epsilon} = \int_0^1 \la \nabla \varphi(T_t^\epsilon(x)) , w_t(T_t^\epsilon(x))\ra dt.
    \end{equation}
    Denoting $K := \operatorname{spt}(\nabla \varphi)$, then by compactness of $K$ we know the integrand is bounded in absolute value by
    \begin{equation}
        \|\nabla \varphi\|_\infty \sup_{y \in K}|w_t(y)| \leq C_K(|w_t(0)| + \operatorname{Lip}(w_t)).
    \end{equation}
    This bound is integrable in $t$ by (ii) and is independent of $x, \epsilon$, which by dominated convergence and integration over $\mu$ implies
    \begin{equation}
        \label{eqn:proofdualitylowerbound}\lim_{\epsilon \downarrow 0} \frac{\int_{\r^d} \varphi d\nu_{\epsilon} - \int_{\r^d} \varphi d\mu}{\epsilon} = \int_{\r^d} \la \nabla \varphi, F_1\ra d\mu.
    \end{equation}
    If we define 
    \begin{equation}
        \omega(r) := \sup_{D_{\h_k}(\mu, \rho) \leq r} G(\rho)
    \end{equation}
    then since $D_{\h_k}$ convergence implies narrow convergence (Proposition \ref{prop:quasimetric}) and $G$ is narrowly continuous we have
    \begin{equation}
        \lim_{r \downarrow 0}\omega(r) = G(\mu) = L.
    \end{equation}
    Let us denote
    \begin{equation}
        \ell := \liminf_{\epsilon \downarrow 0} \frac{D_{\h_k}(\mu, \nu_\epsilon)}{\epsilon}.
    \end{equation}
    If $\ell = \infty$ the claim is immediate, and otherwise we take $\epsilon_n \downarrow 0$ realizing the $\liminf$ and admissible curves $(v_t^n, \rho_t^n)_{t \in [0, 1]}$ transporting $\mu$ to $\nu_{\epsilon_n} $ for which
    \begin{equation}
        A_n := \int_0^1 \|v_t^n\|_{\h_k} dt \leq D_{\h_k}(\mu, \nu_{\epsilon_n}) + \epsilon_n^2.
    \end{equation}
    Hence, $A_n \to 0$ and moreover $D_{\h_k}(\mu, \rho_t^n) \leq \int_0^t \|v_s^n\|_{\h_k}ds \leq A_n$. Therefore, $G(\rho_t^n) \leq \omega(A_n)$ and so
    \begin{equation}
        \left|\int_{\r^d} \varphi d\nu_{\epsilon_n} - \int_{\r^d}\varphi d\mu\right| \leq \int_0^1 \|v_t^n\|_{\h_k} G(\rho_t^n) dt \leq A_n \omega(A_n)
    \end{equation}
    by the continuity equation and duality. Dividing by $\epsilon_n$, we find
    \begin{equation}
        \left|\frac{\int_{\r^d} \varphi d\nu_{\epsilon_n} - \int_{\r^d} \varphi d\mu}{\epsilon_n} \right| \leq \frac{A_n}{\epsilon_n} \omega(A_n) \leq \left(\frac{D_{\h_k}(\mu, \nu_{\epsilon_n})}{\epsilon_n} + \epsilon_n\right) \omega(A_n).
    \end{equation}
    Taking the $n \to \infty$ limit and using $\omega(A_n) \to L$ along with (\ref{eqn:proofdualitylowerbound}) implies the claim.
\end{proof}

If $\mu$ is a rotation-invariant measure and $k$ is a rotation-equivariant kernel (see Definition \ref{def:rotationequivariantkernel}), then we may get a more concrete lower bound using test functions constructed via higher-order spherical harmonics; this is inspired by the approach of Bach \cite{bach2017breaking}. This will require some theory of polynomials on the sphere, for which we refer to the textbook of Atkinson-Han \cite{atkinson2012spherical}.

\begin{lemma}[Rotation-equivariant $D_{\h_k}$ perturbative lower bound]
\label{lemma:rkhslowerboundspherical}
     Let $d \geq 3$ and $\mu \in \p_2(\r^d)$ be rotationally-invariant and not $\delta_0$. Let $w: [0, 1] \times \r^d \to \r^d$ be jointly measurable with each $w_t$ globally-Lipschitz with $\int_0^1 (|w_t(0)| + \operatorname{Lip}(w_t))dt < \infty$, and for $\epsilon > 0$ define $\nu_\epsilon := \operatorname{Law}(X_1)$ according to the flow
    \begin{equation}\dot{X}_t = \epsilon w_t(X_t), \quad X_0 \sim \mu.\end{equation}
Then, for any $k \in \mathcal{K}_{\operatorname{rot}}$ satisfying Assumption \ref{assumption:kernel} with $\int_{\r^d}\operatorname{tr}k(x, x) d\mu(x) \in (0, \infty)$ we have
\begin{equation}
    \label{eqn:sphericalclaimquadratic}\sqrt{\int_{\r^d} \operatorname{tr}k(x, x) d\mu(x)} \cdot \liminf_{\epsilon \downarrow 0} \frac{D_{\h_k}(\mu, \nu_\epsilon)}{\epsilon} \geq \frac{1}{\sqrt{5}} \cdot \frac{ \sum_{\substack{n \in \n}} \sqrt{a_{n,d}}A_n}{\sqrt{\int_{\r^d} {|x|^2}d\mu(x)}} ,
\end{equation}
where we define
\begin{align}a_{n,d} &:= \frac{1}{n(n+d-2)} {n+d-2 \choose n},\\ A_n &:= \sup_{\substack{Y \in \mathbb{Y}_n^d \\ \|Y\|_{L^2(\sigma)} = 1}} \left|\int_{\r^d} \la \nabla_x \left[|x|^2 Y\left(\frac{x}{|x|}\right)\right], \int_0^1 w_t(x)dt\ra d\mu(x)\right|\end{align}
for $\mathbb{Y}_n^d$ the space of degree-$n$ spherical harmonics (see \cite[Definition 2.7]{atkinson2012spherical}) and $\sigma$ the uniform probability measure on $\mathbb{S}^{d-1}$.

\end{lemma}
\begin{remark}
    Note that for any $x \in \r^d$ we have the spherical gradient identity
\begin{equation}\nabla_x \left[|x|^2 Y\left(\frac{x}{|x|}\right)\right] = 2Y\left(\frac{x}{|x|}\right) {x} + |x| \nabla_{\mathbb{S}^{d-1}} Y\left(\frac{x}{|x|}\right).\end{equation}
Alternatively, if $F_1(x) := \int_0^1 w_t(x) dt$ is $C^1$ and $DF_1$ is sufficiently integrable we can apply integration by parts to get
\begin{equation}A_n = \sup_{\substack{Y \in \mathbb{Y}_n^d \\ \|Y\|_{L^2(\sigma)} = 1}} \left|\int_{\r^d} Y\left(\frac{x}{|x|}\right) \left[(d+1)x^\top F_1(x) + x^\top DF_1(x) x - |x|^2 \operatorname{div}F_1(x)\right] d\mu(x)\right|.\end{equation}
\end{remark}

\begin{proof}[Proof of Lemma \ref{lemma:rkhslowerboundspherical}]
In the statement and in what follows, we set $|x|^2Y(x/|x|)$, $h(R)/R$, and quantities like it to equal 0 when $x,R=0$.
Let $F_1(x) := \int_0^1 w_t(x) dt$, which we know lies in $L^2(\mu; \r^d)$ by the assumptions on $w$.
By Lemma \ref{lemma:epsilonfirstorder}, we find that
\begin{equation}
\label{eqn:proofsphericaldageneralsteinlowerbound}
    \liminf_{\epsilon \downarrow 0} \frac{D_{\h_k}(\mu, \nu_\epsilon)}{\epsilon} \geq \sup_{\varphi} \frac{|\int_{\r^d } \la \nabla \varphi, F_1\ra d\mu|}{\sqrt{\int_{\r^d\times \r^d} \la \nabla \varphi(x), k(x, y) \nabla \varphi(y)\ra d\mu(x) d\mu(y)}}
\end{equation}
where the supremum ranges over $\varphi \in C^\infty_c(\r^d)$. We will choose $\varphi$ as a weighted sum of spherical harmonics.

Let $\mathbb{Y}_n^d$ be the linear space of degree-$n$ spherical harmonics on $\mathbb{S}^{d-1}$ (see \cite[Definition 2.7]{atkinson2012spherical}) and $N_{n, d} := \operatorname{dim}\mathbb{Y}_n^d$ its dimension, which by (2.10) of \cite{atkinson2012spherical} satisfies
\begin{equation}
    \label{eqn:proofsphericalharmonicdimscaling}
    N_{n, d} = \frac{(2n+d - 2)(n+d-3)!}{n!(d-2)!} = \frac{2n+d-2}{(d-2)!} \prod_{j=1}^{d-3}(n+j) \quad (\forall n \in \n, 
    \;d \geq 3)
\end{equation}
with $N_{0, d} = 1$ and $N_{n, 2} = 2$ for all $n \in \n$. For each $n$, let $\{Y_{n,j}:\; 1 \leq j \leq N_{n,d}\}$ be an $L^2(\sigma)$-orthonormal basis of $\mathbb{Y}_n^d$.

Since $X \sim \mu$ is rotationally-invariant we may write $X = R\Theta$ for $\Theta \sim \sigma$ drawn uniformly on $\mathbb{S}^{d-1}$ independently of $R \in [0, \infty)$. Fix $h \in C^\infty_c((0, \infty))$ to be nonnegative with $\mathbb{E}[h(R)/R] > 0$ and set $h(0) = 0$.
We consider $\mathbb{Y}_n^d$ as a representation of $O(d)$, where each $Q \in O(d)$ is represented by the automorphism $\rho_n(Q)$ on $\mathbb{Y}_n^d$ sending $Y \mapsto Y \circ Q^\top$. 
For each degree $n$ we define the linear map
\begin{equation}J_n: \mathbb{Y}_n^d \to L^2(\mu; \mathbb{R}^d), \quad (J_nY)(x) := \nabla_x \left[h(|x|) Y\left(\frac{x}{|x|}\right) \right].\end{equation}
By the chain rule, for every $Q \in O(d)$ and $Y \in \mathbb{Y}_n^d$ we have
\begin{equation}J_n(Y \circ Q^\top) = Q \circ (J_n Y) \circ Q^\top.\end{equation}
Since $k$ is rotation-equivariant and $\mu$ is rotation-invariant, we know that the kernel integral operator $\mathcal{T}_{k, \mu}$ commutes with $f \mapsto Q \circ f \circ Q^\top$ and so for any $Y \in \mathbb{Y}_n^d$ and $Z \in \mathbb{Y}_m^d$ we have
\begin{equation}\la J_n(Y \circ Q^\top), \mathcal{T}_{k, \mu} J_m(Z \circ Q^\top)\ra_{L^2(\mu; \mathbb{R}^d)} = \la J_nY, \mathcal{T}_{k, \mu}J_mZ\ra_{L^2(\mu; \mathbb{R}^d)},\end{equation}
or in other words that
\begin{equation}
    (J_n^* \mathcal{T}_{k, \mu}J_m) \rho_m(Q) = \rho_n(Q) (J_n^* \mathcal{T}_{k, \mu}J_m) \quad (\forall Q \in O(d)).
\end{equation}
Hence, $J_n^* \mathcal{T}_{k, \mu}J_m$ is an $O(d)$-equivariant linear map from the representation $(\mathbb{Y}_m^d, \rho_m)$ to $(\mathbb{Y}_n^d, \rho_n)$. 
By \cite[Theorem 2.10]{atkinson2012spherical} these representations are irreducible. By (\ref{eqn:proofsphericalharmonicdimscaling}) and $d \geq 3$ we know that $N_{n,d}$ is strictly increasing in $n$, which means that $\mathbb{Y}_n^d$, $\mathbb{Y}_m^d$ are not isomorphic for $n \neq m$. Therefore, Schur's lemma from representation theory implies that
\begin{equation}J_n^* \mathcal{T}_{k, \mu}J_m = \delta_{n,m} q_n \1_{\mathbb{Y}_n^d} \quad(\forall n, m \in \n)\end{equation}
for some scalars $q_n \geq 0$. Hence,
\begin{equation}
    \label{eqn:proofsphericalschurclaim}
    \la J_nY, \mathcal{T}_{k, \mu}J_mZ\ra_{L^2(\mu; \mathbb{R}^d)} = \delta_{n,m} q_n \la Y, Z\ra_{L^2(\sigma)} \quad (\forall Y \in \mathbb{Y}_n^d, Z \in \mathbb{Y}_m^d).
\end{equation}
Furthermore, spherical differentiation yields
\begin{equation}(J_nY)(r\theta) = h'(r) Y(\theta) \theta + \frac{h(r)}{r}\nabla_{\mathbb{S}^{d-1}} Y(\theta),\end{equation}
which implies that for every $Y \in \mathbb{Y}_n^d, Z \in \mathbb{Y}_{m}^d$ we have
\begin{align}
    \la J_nY, J_mZ\ra_{L^2(\mu; \mathbb{R}^d)} &= \mathbb{E}_{R, \Theta}\la h'(R) Y(\Theta) \Theta + \frac{h(R)}{R} \nabla_{\mathbb{S}^{d-1}}Y(\Theta), \; h'(R) Z(\Theta) \Theta + \frac{h(R)}{R} \nabla_{\mathbb{S}^{d-1}}Z(\Theta)\ra \\
    &= \mathbb{E}_{R, \Theta}\left[h'(R)^2 Y(\Theta) {Z(\Theta)}\right] + \mathbb{E}_{R, \Theta}\left[\frac{h(R)^2}{R^2} \la \nabla_{\mathbb{S}^{d-1}} Y(\Theta), \nabla_{\mathbb{S}^{d-1}} Z(\Theta)\ra_{\mathbb{R}^d}\right],
\end{align}
where the cross-terms vanish in the second line because $\nabla_{\mathbb{S}^{d-1}}Y(\Theta)$ (which is tangent to the sphere at $\Theta$) and $\Theta$ (which is normal to the sphere at $\Theta$) are orthogonal, and similarly for $Z$. By independence of $R, \Theta$ we find 
\begin{align}
    \la J_nY, J_mZ\ra_{L^2(\mu; \mathbb{R}^d)} &= \mathbb{E}[h'(R)^2] \la Y, Z\ra_{L^2(\sigma)} + \mathbb{E}\left[\frac{h(R)^2}{R^2}\right] \la \nabla_{\mathbb{S}^{d-1}} Y, \nabla_{\mathbb{S}^{d-1}} Z\ra_{L^2(\sigma; \mathbb{R}^d)} \\
    &= \mathbb{E}[h'(R)^2] \la Y, Z\ra_{L^2(\sigma)} + m(m+d-2)\mathbb{E}\left[\frac{h(R)^2}{R^2}\right] \la Y, Z\ra_{L^2(\sigma)},
\end{align}
where the second line follows from equation (3.20) of \cite{atkinson2012spherical}. Combining this with the fact that $\mathbb{Y}_n^d \perp \mathbb{Y}_m^d$ for $n \neq m$ (see \cite[Corollary 2.15]{atkinson2012spherical}), for every $Y \in \mathbb{Y}_n^d, Z \in \mathbb{Y}_m^d$ we find
\begin{equation}
\label{eqn:proofsphericalinnerproductnoTkmu}
    \la J_nY, J_mZ\ra_{L^2(\mu; \mathbb{R}^d)} = \delta_{n,m} \left(\mathbb{E}[h'(R)^2] + n(n+d-2) \mathbb{E}\left[\frac{h(R)^2}{R^2}\right]\right) \la Y, Z\ra_{L^2(\sigma)} .
\end{equation}
Hence, the collection
\begin{equation}\left\{\frac{J_nY_{n,j}}{\sqrt{\mathbb{E}[h'(R)^2] + n(n+d-2)\mathbb{E}[h(R)^2/R^2]}}:\quad n \in \n,\; 1 \leq j \leq N_{n,d}\right\}\end{equation}
is an orthonormal family in $L^2(\mu; \mathbb{R}^d)$. Therefore, since $\mathcal{T}_{k, \mu}$ is positive and trace-class we know
\begin{align}\operatorname{tr}(\mathcal{T}_{k, \mu}) &\geq \sum_{n \in \n} \sum_{j=1}^{N_{n,d}} \la \frac{J_nY_{n,j}}{\sqrt{\mathbb{E}[h'(R)^2] + n(n+d-2)\mathbb{E}[h(R)^2/R^2]}}, \mathcal{T}_{k, \mu}\frac{J_nY_{n,j}}{\sqrt{\mathbb{E}[h'(R)^2] + n(n+d-2)\mathbb{E}[h(R)^2/R^2]}}\ra
\\ &= \sum_{n \in \n} \frac{N_{n,d}q_n}{\mathbb{E}[h'(R)^2] + n(n+d-2)\mathbb{E}[h(R)^2/R^2]},
\end{align}
where in the second line we applied (\ref{eqn:proofsphericalschurclaim}). In particular, using that $\operatorname{tr}(\mathcal{T}_{k, \mu}) = \int_{\r^d} \operatorname{tr}k(x, x) d\mu(x)$ implies that 
\begin{equation}
\label{eqn:proofsphericaltracelowerbound}
    \int_{\r^d}\operatorname{tr} k(x, x) d\mu(x) \geq \sum_{n \in \n} \frac{N_{n,d}q_n}{\mathbb{E}[h'(R)^2] + n(n+d-2)\mathbb{E}[h(R)^2/R^2]}.
\end{equation}
Now, for each $n$ let us write
\begin{equation}
\label{eqn:proofsphericalsupremumwithcutoffh}
A_n^{(h)} := \sup_{\substack{Y \in \mathbb{Y}_n^d \\ \|Y\|_{L^2(\sigma)} = 1}} \left|\int_{\r^d} \la J_nY, F_1\ra d\mu\right|\end{equation}
and let us take $(W_{n,\ell}^{(h)})_{\ell \in \n} \subset \mathbb{Y}_n^d$ to be approximate maximizers with $\|W_{n,\ell}^{(h)}\|_{L^2(\sigma) }=1$ and
\begin{equation}
\label{eqn:proofsphericalapproximatemaximizers}
A_{n,\ell}^{(h)} := \int_{\r^d} \la J_nW_{n,\ell}^{(h)}, F_1\ra d\mu \geq A_n^{(h)} - \frac{1}{\ell},\end{equation}
where we note that $J_n$ depends on $h$.
By equation (3.20) of \cite{atkinson2012spherical} we know that
\begin{equation}
    \label{eqn:proofsphericalsobolevnormofZ}
    \|\nabla_{\mathbb{S}^{d-1}} W_{n,\ell}^{(h)}\|_{L^2(\sigma; \r^d)}^2 = n(n+d-2).
\end{equation}
Now, for each degree $n$ and each $\ell$ we define the function $\varphi_{n,\ell}^{(h)}: \r^d \to \r$ by
\begin{equation}\varphi_{n,\ell}^{(h)}(r\theta) := h(r) W_{n,\ell}^{(h)}(\theta). \end{equation}
Since $h$ vanishes near $0$ and $W_{n,\ell}^{(h)}$ is a spherical polynomial we know $\varphi_{n,\ell}^{(h)} \in C^\infty_c(\r^d)$, and by definition we have
$\nabla \varphi_{n,\ell}^{(h)} = J_n W_{n,\ell}^{(h)}$. In particular, applying (\ref{eqn:proofsphericalschurclaim}) with $n=m$ and $Y=Z=W_{n,\ell}^{(h)}$ implies
\begin{equation}
    \label{eqn:proofsphericalRKHSnorm2}
   \la \nabla \varphi_{n,\ell}^{(h)}, \mathcal{T}_{k, \mu}\nabla \varphi_{n,\ell}^{(h)}\ra_{L^2(\mu; \r^d)} = q_n \|W_{n,\ell}^{(h)}\|_{L^2(\sigma)}^2 = q_n
\end{equation}
using that $\|W_{n,\ell}^{(h)}\|_{L^2(\sigma)} = 1$, and so the scalars $q_n$ are identified. 

We proceed by fixing a finite subset $\Lambda \subset \n$ with $q_n > 0$ for all $n \in \Lambda$ and defining
\begin{equation}\varphi^{(\ell, h)} := \sum_{n \in \Lambda} \frac{ A_{n,\ell}^{(h)}}{q_n}\varphi_{n,\ell}^{(h)} \in C^\infty_c(\r^d).\end{equation}
By definition of $A_{n,\ell}^{(h)}$ and (\ref{eqn:proofsphericalschurclaim}),(\ref{eqn:proofsphericalRKHSnorm2}) we have
\begin{equation}\int_{\r^d} \la \nabla \varphi^{(\ell, h)}, F_1\ra d\mu = \sum_{n \in \Lambda} \frac{ (A_{n,\ell}^{(h)})^2}{q_n}= \la \nabla \varphi^{(\ell, h)}, \mathcal{T}_{k, \mu}\nabla \varphi^{(\ell, h)}\ra_{L^2(\mu; \r^d)} .\end{equation}
The Stein lower bound of (\ref{eqn:proofsphericaldageneralsteinlowerbound}) therefore implies that
\begin{equation}\liminf_{\epsilon \downarrow 0} \frac{D_{\h_k}(\mu, \nu_\epsilon)}{\epsilon} \geq \left(\sum_{n \in \Lambda} \frac{(A_{n,\ell}^{(h)})^2}{q_n}\right)^{1/2},\end{equation}
which we combine with (\ref{eqn:proofsphericaltracelowerbound}) to find that
\begin{align}\sqrt{\int_{\r^d} \operatorname{tr}k(x, x) d\mu(x)} \cdot \liminf_{\epsilon \downarrow 0} \frac{D_{\h_k}(\mu, \nu_\epsilon)}{\epsilon}&\geq \left(\sum_{n \in \Lambda} \frac{(A_{n,\ell}^{(h)})^2}{q_n}\right)^{1/2}\\&\quad  \cdot \left(\sum_{n \in \Lambda} \frac{N_{n,d}q_n}{\mathbb{E}[h'(R)^2] + n(n+d-2)\mathbb{E}[h(R)^2/R^2]}\right)^{1/2}.\end{align}
Applying Cauchy-Schwarz in $\r^{|\Lambda|}$, we find that
for all finite $\Lambda \subset \n$ with $q_n > 0$ for $n \in \Lambda$ we have
\begin{equation}
\sqrt{\int_{\r^d} \operatorname{tr}k(x, x) d\mu(x)} \cdot \liminf_{\epsilon \downarrow 0} \frac{D_{\h_k}(\mu, \nu_\epsilon)}{\epsilon} \geq \sum_{n \in \Lambda} \frac{\sqrt{N_{n,d}} \left|A_{n,\ell}^{(h)}\right|}{\sqrt{\mathbb{E}[h'(R)^2] + n(n+d-2) \mathbb{E}[h(R)^2 / R^2]}}.
\end{equation}
If $q_n = 0$ but $A_{n,\ell}^{(h)}\neq 0$ then the left hand side must be infinite, and so this holds for all finite $\Lambda \subset \n$. On any such finite subset we may take $\ell \to \infty$ and apply (\ref{eqn:proofsphericalapproximatemaximizers}) to get
\begin{equation}
    \label{eqn:proofsphericalsteinlowerboundfull}
    \sqrt{\int_{\r^d} \operatorname{tr}k(x, x) d\mu(x)} \cdot \liminf_{\epsilon \downarrow 0} \frac{D_{\h_k}(\mu, \nu_\epsilon)}{\epsilon} \geq \sum_{n \in \Lambda} \frac{\sqrt{N_{n,d}} \cdot A_n^{(h)}}{\sqrt{\mathbb{E}[h'(R)^2] + n(n+d-2) \mathbb{E}[h(R)^2 / R^2]}}.
\end{equation}

We will now take $h(r)$ to be smooth compactly-supported approximations to $h_\star(r) = r^2$. Indeed, we define \begin{equation}
\label{eqn:proofsphericalsupremumwithcutoffdefinitionofJstar}
J_n^\star: \mathbb{Y}_n^d \to L^2(\mu; \mathbb{R}^d), \quad (J_n^\star Y)(x) := \nabla_x \left[h_\star(|x|) Y\left(\frac{x}{|x|}\right)\right] ,\end{equation}
take $\chi \in C_c^\infty(\r)$ with $0 \leq \chi \leq 1$ equal to $1$ on $[-1, 1]$ and equal to $0$ outside $[-2, 2]$, and set
\begin{equation}
    h_j(r) := r^2 \chi(r/j) (1-\chi(jr)).
\end{equation}
Then, $|h'_j(r)| + |h_j(r)/r| \lesssim r$ uniformly in $r$ and converge pointwise to $2r$ and $r$, respectively. By dominated convergence, $h_j'/2, h_j/r \to r$ in $\operatorname{L}^2(\operatorname{Law}(R))$. Thus, if we define $A_n^{(h_j)}$ and $J_n^{(h_j)}$ analogously to (\ref{eqn:proofsphericalsupremumwithcutoffh}) and (\ref{eqn:proofsphericalsupremumwithcutoffdefinitionofJstar}) then we have $J_n^{(h_j)} \to J_n^\star$ uniformly on the unit ball of $\mathbb{Y}_n^d$. 
Therefore,
\begin{equation}A_n^{(h_j)} \to \sup_{\substack{Y \in \mathbb{Y}_n^d \\ \|Y\|_{L^2(\sigma)} = 1}} \left|\int_{\r^d} \la J_n^\star Y, F_1\ra d\mu\right| =: A_n \quad (\forall n \in \n).\end{equation}
We also find that
\begin{equation}M_2(\mu)^2 := \mathbb{E}_{X \sim \mu}|X|^2 = \frac{1}{4}\mathbb{E}[h'_\star(R)^2] = \mathbb{E}\left[\frac{h_\star(R)^2}{R^2}\right],\end{equation}
which we plug into (\ref{eqn:proofsphericalsteinlowerboundfull}) to see
\begin{equation}
    \label{eqn:proofsphericalsteinlowerboundfull3}
    \sqrt{\int_{\r^d} \operatorname{tr}k(x, x) d\mu(x)} \cdot \liminf_{\epsilon \downarrow 0} \frac{D_{\h_k}(\mu, \nu_\epsilon)}{\epsilon} \geq \frac{1}{M_2(\mu)}\sum_{n \in \Lambda} \frac{\sqrt{N_{n,d}} \cdot A_n}{\sqrt{4+ n(n+d-2) }}.
\end{equation}
This holds for any finite subset $\Lambda \subset \n$, and we may increase it monotonically to hold for any subset $\Lambda \subset \n$. The claim (\ref{eqn:sphericalclaimquadratic}) then follows from combining (\ref{eqn:proofsphericalharmonicdimscaling}) and $2n+d-2 \geq n+d-2$ to find
\begin{equation}\frac{N_{n,d}}{4 + n(n+d-2)} \geq \frac{N_{n,d}}{5n(n+d-2)} \geq \frac{1}{5n(n+d-2)} {n+d-2 \choose d-2} \quad (\forall n \in \n, \; d \geq 3).\end{equation}
\end{proof}

In the case where $\mu = \gamma_d$ is the Gaussian and $w_t = w$ is a single-index drift, we can get matching lower and upper bounds on the infinitesimal behavior of $D_{\h_k}$ in terms of the Hermite spectrum of the activation function:

\begin{lemma}[Rotation-equivariant $D_{\h_k}$ bounds for single-index Gaussian flows]
\label{lemma:appendixupperandlowerboundsforsingleindexgaussianflows}
   Fix $d \geq 3$. Let $\phi: \r \to \r$ be $C^{d+2}$ with $\|\phi'\|_{C^{d+1}} < \infty$ and fix $\sigma > 0$ and $u, \theta \in \mathbb{S}^{d-1}$. For each $\epsilon > 0$ define $\nu_\epsilon := \operatorname{Law}(X_\epsilon)$ according to the flow 
    \begin{equation}
        \dot{X}_t = \sigma u \phi(\la X_t, \theta\ra), \quad X_0 \sim \gamma_d.
    \end{equation}
    If we denote the quantity
    \begin{equation}
        Q := \inf_{\substack{k \in \mathcal{K}_{\operatorname{rot}} \\ k \text{ satisfies Assumption \ref{assumption:kernel}} \\ \int_{\r^d} \operatorname{tr}k(x, x) d\gamma_d(x) \in (0, \infty)}} \left\{ \sqrt{\int_{\r^d} \operatorname{tr}k(x, x) d\gamma_d(x)} \cdot \liminf_{\epsilon \downarrow 0} \frac{D_{\h_k}(\gamma_d, \nu_\epsilon)}{\epsilon}\right\}
    \end{equation}
     and use the probabilist's Hermite\footnote{By this, we mean that $\{\operatorname{He}_j\}_j$ are the monic Hermite polynomials with $\la \operatorname{He}_i, \operatorname{He}_j\ra_{L^2(\gamma_1)} = \delta_{i,j} \cdot j!$. In particular, there is the decomposition $\phi = \sum_{j=0}^\infty {\hat{\phi}_j}/{j!}\operatorname{He}_j$ in $L^2(\gamma_1)$.} coefficients $\hat{\phi}_j := \mathbb{E}_{\xi \sim \mathcal{N}(0, 1)}[\phi(\xi) \operatorname{He}_j(\xi)]$ to define
    \begin{equation}
    \label{eqn:appendixdefinitionofDpphicforthegaussianbounds}
        \mathcal{D}_\phi(c) := \sum_{n = 0}^\infty N_{n,d}  \left(\sum_{\substack{\ell = 0\\n+2\ell \geq 1}}^\infty \left(\frac{m\Gamma(d/2)}{2^{m}\ell! \Gamma(n + \ell + d/2)}\right) \cdot |\hat{\phi}_{m-1}|^2 \cdot \left(c^2 + (1-c^2) \frac{n(n + d - 2)}{m^2(d-1)}\right)\right)^{1/2}
    \end{equation}
    with $m = n+2\ell$,
    then there are the lower and upper bounds
    \begin{equation}
        \sigma \mathcal{D}_\phi(\la u, \theta\ra) \leq Q \leq \sqrt{2} \sigma \mathcal{D}_\phi(\la u, \theta\ra) .
    \end{equation}
    The lower bound holds under the weaker assumption that $\phi$ is globally-Lipschitz. 
\end{lemma}
\begin{proof}
    Fix $w : \r^d \to \r^d$ globally-Lipschitz. 
    For every $m \in \n$, the $m^{th}$ scalar Hermite eigenspace $\h_m$ splits orthogonally as
    \begin{equation}
        \h_m = \bigoplus_{\substack{n, \ell \in \n \cup \{0\} \\ m = n + 2\ell}} \h_{n,\ell}
    \end{equation}
    where each $\h_{n,\ell}$ is isomorphic to $\mathbb{Y}_n^d$ (here, $n$ is the angular momentum/spherical harmonic degree and $\ell$ is the radial index). Choose the orthonormal basis $\{P_{n,\ell, i}\}_{i=1}^{N_{n,d}}$ of $\h_{n,\ell}$ w.r.t. $L^2(\gamma_d)$
    given by 
    \begin{equation}
    \label{eqn:proofappendixdefinitionofthe}
        P_{n,\ell,i}(r\omega) := (-1)^\ell\sqrt{\frac{\ell! \Gamma(d/2)}{2^n \Gamma(n+\ell+d/2)}} r^n L_\ell^{(n+d/2 - 1)} (r^2/2) Y_{n,i}(\omega)
    \end{equation}
    for $r \geq 0$ and $\omega \in \mathbb{S}^{d-1}$, where $\{Y_{n,i}\}_{i=1}^{N_{n,d}}$ is the $L^2(\sigma)$-orthonormal basis of $\mathbb{Y}_n^d$ from the proof of Lemma \ref{lemma:rkhslowerboundspherical} for $\sigma$ the uniform probability measure on the sphere, and $L_\ell^{(\alpha)}$ are the generalized Laguerre polynomials. This is the standard Laguerre-spherical harmonic representation of the Hermite eigenspaces, see the $\mu=0$ case of \cite[Definition 2.1 and Theorem 2.3]{sheng2021nontensorial} after rescaling their Gaussian weight $e^{-r^2}$ to $e^{-r^2/2}$ and normalizing the Gaussian and spherical measures to be probability measures.
    For $n+2\ell \geq 1$ we set
    \begin{equation}
        E_{n,\ell, i} := \frac{1}{\sqrt{n + 2\ell}} \nabla P_{n,\ell, i}
    \end{equation}
    so that $\{E_{n,\ell, i}\}_{i=1}^{N_{n,d}}$ forms an orthonormal basis of the Gaussian gradient fields (i.e. the image of $\h_{n,\ell}$ under the isometry $f \mapsto \frac{1}{\sqrt{n+2\ell}}\nabla f$, equipped with the $L^2(\gamma_d; \r^d)$ inner product).
    For each $n \in \n$, we define the operator $C_n: \r^{N_{n,d}} \to \ell^2(\n \cup \{0\})$ by the matrix elements
    \begin{equation}
        (C_n)_{\ell, i} := \la w, E_{n,\ell, i}\ra_{L^2(\gamma_d; \r^d)}
    \end{equation}
    with $\ell \in \n \cup \{0\}$ and $i \in \{1, \ldots, N_{n,d}\}$. For $n = 0$, we do the same but with $\ell \in \n$ and $N_{0,d} = 1$ to get $C_0: \r \to \ell^2(\n)$. Finally, we set
    \begin{equation}
    \label{eqn:proofappendixdefinitionofDtildew}
        \widetilde{\mathcal{D}}(w) := \sum_{n=0}^\infty \sqrt{N_{n,d}} \operatorname{tr} \left(\sqrt{C_nC_n^\top}\right).
    \end{equation}
    We will first obtain matching lower and upper bounds to show that $Q = \widetilde{\mathcal{D}}(w)$ if we define $\nu_\epsilon$ using the more general flow $\dot{X}_t = w(X_t)$; the lower bound will hold for globally-Lipschitz $w$, while the upper bound will require extra smoothness. Afterward, we specialize to $w(x) = \sigma u\phi(\la x, \theta\ra)$ and relate $\widetilde{\mathcal{D}}(w)$ to our explicit quantity $\mathcal{D}_\phi(c)$. 
    
  To show that $Q$ is lower bounded by $\widetilde{\mathcal{D}}(w)$ for any globally-Lipschitz $w$, we use the same Schur's lemma trace calculation from the proof of Lemma \ref{lemma:rkhslowerboundspherical}.
   Fix any $k \in \mathcal{K}_{\operatorname{rot}}$ satisfying Assumption \ref{assumption:kernel} with $\int_{\r^d} \operatorname{tr}k(x, x) d\gamma_d(x) \in (0, \infty)$. 
    We know that for a fixed $n \in \n$ the spaces $(\h_{n,\ell})_{\ell=0}^\infty$ are all isomorphic to $\mathbb{Y}_n^d$ and transform in the same way under rotations (as do their gradient spaces). By Schur's lemma and rotational equivariance of $\mathcal{T}_{k, \gamma_d}$, there is some matrix $A_n \succeq 0$ for which
    \begin{equation}
    \label{eqn:proofappendixgaussianlowerboundschurlemma}
        \la E_{n,\ell, i}, \mathcal{T}_{k,\gamma_d} E_{n', r, j}\ra_{L^2(\gamma_d; \r^d)} = \1_{n=n'} (A_n)_{\ell,r} \1_{i=j} \quad (\forall \ell, r \in \n \cup \{0\})
    \end{equation}
    for all $i \in \{1, \ldots, N_{n,d}\}$ and $j \in \{1, \ldots, N_{n', d}\}$, while if $n = 0$ the same is true for $\ell, r \in \n$. By summing over the orthonormal basis of the gradient subspace of $L^2(\gamma_d; \r^d)$ and applying (\ref{eqn:proofappendixgaussianlowerboundschurlemma}), we find that
    \begin{align}
        \operatorname{tr}(\mathcal{T}_{k, \gamma_d}) &\geq \sum_{n=0}^\infty \sum_{\substack{\ell = 0\\n+2\ell \geq 1}}^\infty  \sum_{i=1}^{N_{n,d}} \la E_{n,\ell, i}, \mathcal{T}_{k,\gamma_d} E_{n,\ell, i}\ra_{L^2(\gamma_d; \r^d)}
        \\&= \sum_{n=0}^\infty N_{n,d} \sum_{\substack{\ell = 0\\ n+2\ell \geq 1}}^\infty (A_n)_{\ell,\ell} \\
        &= \sum_{n=0}^\infty N_{n,d} \operatorname{tr}(A_n).
    \end{align}
    In particular, each $A_n$ is trace-class.
    Now, fix $L \in \n$ and truncate $C_{n,L} : \r^{N_{n,d}} \to \r^{L+1}$ (and $C_{0, L} : \r \to \r^{L}$).
If we take any finite sequence of matrices $(Z_n)_{n=0}^N$ of the same shapes as $C_{n,L}$ supported on $\ell \leq L$ and define
    \begin{equation}
        \varphi_Z := \sum_{n = 0}^N \sum_{\substack{\ell = 0\\ n+2\ell \geq 1}}^L \frac{1 }{\sqrt{n+2\ell}}\sum_{i=1}^{N_{n,d}} (Z_n)_{\ell, i} P_{n,\ell,i}
    \end{equation}
   then we have
    \begin{equation}
        \nabla \varphi_Z = \sum_{n = 0}^N \sum_{\substack{\ell = 0\\n+2\ell \geq 1}}^L \sum_{i=1}^{N_{n,d}} (Z_n)_{\ell, i} E_{n,\ell,i}.
    \end{equation}
Furthermore, by definition of $(C_n)_n$ we know that
\begin{align}
    \la \nabla \varphi_Z, w\ra_{L^2(\gamma_d; \r^d)} &= \sum_{n = 0}^N \sum_{\substack{\ell = 0\\ n+2\ell \geq 1}}^L\sum_{i=1}^{N_{n,d}}  \la w, E_{n,\ell, i}\ra_{L^2(\gamma_d; \r^d)} \la \nabla \varphi_Z, E_{n,\ell, i}\ra_{L^2(\gamma_d; \r^d)}\\
    &= \sum_{n=0}^N \operatorname{tr}\left(Z_n^\top C_{n,L}\right).
\end{align}
On the other hand, by using (\ref{eqn:proofappendixgaussianlowerboundschurlemma}) and the same orthonormal basis we know that if we truncate $A_{n,L}: \r^{L+1} \to \r^{L+1}$ (with $A_{0,L}: \r^L \to \r^L$) then
    \begin{align}
        \la \nabla \varphi_Z, \mathcal{T}_{k, \gamma_d} \nabla \varphi_Z\ra_{L^2(\gamma_d; \r^d)} &= \sum_{n=0}^N \operatorname{tr}\left(Z_n^\top A_{n,L} Z_n\right).
    \end{align}
    If we take a cutoff $\chi_R \in C^\infty_c(\r^d)$ which is 1 on $B_R$ and 0 outside $B_{2R}$ with $\|\nabla \chi_R\| \lesssim 1/R$ and denote $\varphi_{Z,R} := \chi_R \varphi_Z$, then we know $\varphi_{Z,R} \in C^\infty_c$ and $\nabla \varphi_{Z,R} \to \nabla \varphi_Z$ in $L^2(\gamma_d; \r^d)$. By boundedness of $\mathcal{T}_{k,\gamma_d}$ we also know $\la \nabla \varphi_{Z,R}, \mathcal{T}_{k, \gamma_d} \nabla \varphi_{Z,R}\ra_{L^2(\gamma_d; \r^d)} \to \la \nabla \varphi_{Z}, \mathcal{T}_{k, \gamma_d} \nabla \varphi_{Z}\ra_{L^2(\gamma_d; \r^d)}$. Hence, we may apply Lemma \ref{lemma:epsilonfirstorder} to $\varphi_{Z,R}$ and take $R \to \infty$ to get
    \begin{equation}
        \liminf_{\epsilon \downarrow 0} \frac{D_{\h_k}(\gamma_d, \nu_\epsilon)}{\epsilon} \geq \frac{ \la \nabla \varphi_Z, w\ra_{L^2(\gamma_d; \r^d)}}{\sqrt{ \la \nabla \varphi_Z, \mathcal{T}_{k, \gamma_d} \nabla \varphi_Z\ra_{L^2(\gamma_d; \r^d)}}} = \frac{\sum_{n=0}^N \operatorname{tr}\left(Z_n^\top C_{n,L}\right)}{\left({\sum_{n=0}^N \operatorname{tr}\left(Z_n^\top A_{n,L}Z_n\right)}\right)^{1/2}}.
    \end{equation}
    This holds for any $(Z_n)_{n=0}^N$ and any $N \in \n$, i.e.
    \begin{equation}
    \label{eqn:proofappendixlowerboundtraceZ_nsequence}
         \liminf_{\epsilon \downarrow 0} \frac{D_{\h_k}(\gamma_d, \nu_\epsilon)}{\epsilon} \geq  \sup_{\substack{N, L \in \n,\\(Z_n)_{n=0}^N}} \frac{\sum_{n=0}^N \operatorname{tr}\left(Z_n^\top C_{n,L}\right)}{\left({\sum_{n=0}^N \operatorname{tr}\left(Z_n^\top A_{n,L}Z_n\right)}\right)^{1/2}}. \end{equation}
    We may suppose without loss of generality that $\operatorname{im}(C_{n,L}) \subset \operatorname{im}(A_{n,L})$; if not, then we can choose a test direction in $\operatorname{ker}(A_{n,L})$ with nonzero numerator to make the right hand side of (\ref{eqn:proofappendixlowerboundtraceZ_nsequence}) infinite, and so the result is trivial. 
    
    To evaluate this supremum, we fix $N, L \in \n$ and first assume $A_{n,L}$ is invertible for each $n \leq N$. Put $W_n := A_{n,L}^{1/2}Z_n$ to see that $\operatorname{tr}(Z_n^\top C_{n,L}) = \la W_n, A_{n,L}^{-1/2}C_{n,L}\ra_F$ while $\operatorname{tr}(Z_n^\top A_{n,L} Z_n) = \|W_n\|_F^2$. Hence, this is simply the dual norm in the Frobenius inner product, i.e.
    \begin{align}
        \sup_{\substack{(Z_n)_{n=0}^N}} \frac{\sum_{n=0}^N \operatorname{tr}\left(Z_n^\top C_{n,L}\right)}{\left({\sum_{n=0}^N \operatorname{tr}\left(Z_n^\top A_{n,L}Z_n\right)}\right)^{1/2}} &= \left(\sum_{n=0}^N \left\|A_{n,L}^{-1/2} C_{n,L}\right\|_F^2\right)^{1/2}\\& = \left(\sum_{n=0}^N \operatorname{tr}\left(A_{n,L}^{-1}C_{n,L}C_{n,L}^\top\right)\right)^{1/2}.
    \end{align}
    If $A_{n,L}$ is not invertible, the same holds with the pseudoinverse since $\operatorname{im}(C_{n,L}) \subset \operatorname{im}(A_{n,L})$. Hence, we always have
    \begin{equation}
        \liminf_{\epsilon \downarrow 0} \frac{D_{\h_k}(\gamma_d, \nu_\epsilon)}{\epsilon} \geq  \sup_{N,L \in \n}\left(\sum_{n=0}^N \operatorname{tr}\left(A_{n,L}^\dagger C_{n,L}C_{n,L}^\top\right)\right)^{1/2}
    \end{equation}
    and since $\operatorname{tr}(\mathcal{T}_{k,\gamma_d}) \geq \sum_{n=0}^\infty N_{n,d}\operatorname{tr}(A_n) \geq \sum_{n=0}^N N_{n,d}\operatorname{tr}(A_{n,L}) $ for all $N,L$ we know
    \begin{align}
        \sqrt{\operatorname{tr}(\mathcal{T}_{k,\gamma_d})} \liminf_{\epsilon \downarrow 0} \frac{D_{\h_k}(\gamma_d, \nu_\epsilon)}{\epsilon} &\geq \left(\sum_{n=0}^N N_{n,d}\operatorname{tr}\left(A_{n,L}\right)\right)^{1/2} \cdot \left(\sum_{n=0}^N \operatorname{tr}\left(A_{n,L}^\dagger C_{n,L}C_{n,L}^\top\right)\right)^{1/2} \\
        &\geq \sum_{n=0}^N \sqrt{N_{n,d}} \sqrt{\operatorname{tr}\left(A_{n,L}\right) \operatorname{tr}\left(A_{n,L}^\dagger C_{n,L}C_{n,L}^\top\right)},
    \end{align}
    where the second line uses Cauchy-Schwarz $\sqrt{(\sum_n a_n)(\sum_n b_n)} \geq \sum_n \sqrt{a_nb_n}$. For $A, B \succeq 0$ with $\operatorname{im}(B) \subset \operatorname{im}(A)$ we know 
    \begin{equation}
        \operatorname{tr}(\sqrt{B}) = \operatorname{tr}(A^{1/2}A^{\dagger / 2} \sqrt{B}) \leq \|A^{1/2}\|_F \|A^{\dagger / 2}\sqrt{B}\|_F = \sqrt{\operatorname{tr}(A)} \sqrt{\operatorname{tr}(A^\dagger B)}.
    \end{equation}
    Taking $A = A_{n,L}$ and $B = C_{n,L}C_{n,L}^\top$ yields
    \begin{equation}
         \sqrt{\operatorname{tr}(\mathcal{T}_{k,\gamma_d})} \liminf_{\epsilon \downarrow 0} \frac{D_{\h_k}(\gamma_d, \nu_\epsilon)}{\epsilon} \geq \sum_{n=0}^N \sqrt{N_{n,d}} \operatorname{tr}\left(\sqrt{C_{n,L}C_{n,L}^\top}\right).
    \end{equation}
    We first take $N,L \to \infty$ using monotone convergence and then an infimum over $k$ to arrive at the lower bound $Q \geq \widetilde{\mathcal{D}}(w)$.

    For the upper bound, assume further that $w \in C^{d+2}(\r^d; \r^d)$ with $\|Dw\|_{C^{d+1}} < \infty$. Suppose that $\widetilde{\mathcal{D}}(w) < \infty$, since otherwise there is nothing to prove. Let us define the Gaussian gradient subspace
    \begin{equation}
        \mathcal{G} := \overline{\left\{\nabla \varphi:\; \varphi \in C^\infty_c(\r^d)\right\}}^{L^2(\gamma_d; \r^d)}
    \end{equation}
    with $\Pi_{\mathcal{G}}: L^2(\gamma_d; \r^d) \to \mathcal{G}$ its orthogonal projection, and denote $g := \Pi_{\mathcal{G}}w$. Since each $E_{n,\ell,i}$ is a gradient field we know that
    \begin{equation}
        \la w, E_{n,\ell,i}\ra_{L^2(\gamma_d; \r^d)} = \la g, E_{n,\ell,i}\ra_{L^2(\gamma_d; \r^d)} = (C_n)_{\ell,i}.
    \end{equation}
    Let us denote by $(\Phi_t)_{t \geq 0}$ the flow of the ODE driven by $w$ and $\nu_t := (\Phi_t)_\# \gamma_d$ the resulting trajectory of measures, and so $D\Phi_t(x)w(x) = w(\Phi_t(x))$. If we define
    \begin{equation}
    \label{eqn:proofappendixdefinitionofg_tflowupperbound}
        g_t(x) := D\Phi_t(\Phi_t^{-1}(x))g(\Phi_t^{-1}(x)),
    \end{equation}
    then writing $y = \Phi_t(x)$ implies $w(y) - g_t(y) = D\Phi_t(x) [w(x) - g(x)]$.
    We claim that $(g_t, \nu_t)_{t \geq 0}$ satisfies the continuity equation. Indeed, for fixed $\varphi \in C^\infty_c(\r^d)$ we have that
    \begin{align}
        \int_{\r^d} \la \nabla \varphi(y), w(y) - g_t(y)\ra d\nu_t(y) &= \int_{\r^d} \la \nabla \varphi(\Phi_t(x)), D\Phi_t(x) (w(x) - g(x))\ra d\gamma_d(x) \\
        &= \int_{\r^d} \la \nabla (\varphi \circ \Phi_t), w-g\ra d\gamma_d \\&= 0
    \end{align}
    since $\nabla (\varphi \circ \Phi_t) \in \mathcal{G}$. This yields
    \begin{equation}
        \partial_t \nu_t + \nabla \cdot (\nu_t g_t) = 0,
    \end{equation}
    and so the drifts $(g_t)_{t \geq 0}$ generate the same trajectory of measures. In particular, for any $k$ satisfying Assumption \ref{assumption:kernel} if $\|g_t- g\|_{\h_k} \to 0$ as $t \downarrow 0$ then
    \begin{equation}
    \label{eqn:proofappendixthehtingtoproveforupperboundnormofg}
        \limsup_{\epsilon \downarrow 0} \frac{D_{\h_k}(\gamma_d, \nu_\epsilon)}{\epsilon} \leq \limsup_{\epsilon \downarrow 0} \frac{1}{\epsilon} \int_0^\epsilon \|g_t\|_{\h_k} dt = \|g\|_{\h_k}.
    \end{equation}
    Note that $\|g\|_{W^{d+1, 4}(\gamma_d; \r^d)} \lesssim \|w\|_{W^{d+1, 4}(\gamma_d; \r^d)} < \infty$ since $\Pi_{\mathcal{G}}$ is a bounded operator on $W^{d+1,4}(\gamma_d; \r^d)$. We claim that our smoothness assumptions imply
    \begin{equation}
    \label{eqn:appendixproofWd12convergenceofg_t}
    \lim_{t \downarrow 0} \|g_t - g\|_{W^{d+1, 2}(\gamma_d; \r^d)} = 0.
    \end{equation}
    To see this, by Grönwall we know that if $B_t := \Phi_t^{-1}$ and $A_t := D\Phi_t \circ B_t$ then
    \begin{equation}
    \label{eqn:proofappendixboundednessofderivativeshwnenneeded}
        \sup_{t \in [0, t_0]} \left\{ \|DB_t\|_{C^{d}} + \|A_t\|_{C^{d+1}}\right\} < \infty
    \end{equation}
    for some $t_0$. For any $f \in C^\infty_c(\r^d; \r^d)$, the chain and product rules therefore imply that if $U_t f := A_t \cdot (f \circ B_t) $ then
    \begin{equation}
    \label{eqn:proofappendixstupidgaussianpullbackbullshit}
        \sum_{|\alpha| \leq d+1} \|D^\alpha U_t f(x)\| \leq C \sum_{|\beta| \leq d+1} \|D^\beta f(B_t(x))\| \quad (\forall x \in \r^d)
    \end{equation}
    for a constant $C$ uniform in $x$ and $t \in [0, t_0]$. If we change variables $y = B_t(x)$ (i.e. $x = \Phi_t(y)$) and define 
    \begin{equation}
        R_t(y) := |\operatorname{det} D\Phi_t(y)| \exp\left(\frac{\|y\|^2 - \|\Phi_t(y)\|^2}{2}\right),
    \end{equation}
then Lipschitzness of $w$ implies that $R_t(y) \leq e^{C' t(1+|y|^2)}$ for some constant $C'  \in (0, \infty)$. 
In particular, after shrinking $t_0$ we know that there is some $C'' \in (0, \infty)$ for which
\begin{equation}
    \sup_{t \in [0, t_0]} \|R_t\|_{L^2(\gamma_d)} \leq C'' <  \infty.
\end{equation}
Hence,
(\ref{eqn:proofappendixstupidgaussianpullbackbullshit}) and the change of measure formula yield
    \begin{align}
        \|U_tf\|_{W^{d+1,2}(\gamma_d; \r^d)}^2 &\leq C \sum_{|\beta| \leq d+1} \int_{\r^d} \|D^\beta f(B_t(x))\|^2 d\gamma_d(x)
        \\ &\leq C \sum_{|\beta| \leq d+1} \int_{\r^d} \|D^\beta f(y)\|^2 R_t(y)d\gamma_d(y) \\
        &\leq C  \|f\|_{W^{d+1, 4}(\gamma_d; \r^d)}^2 \cdot \|R_t\|_{L^2(\gamma_d)} \\
        &\leq CC'' \|f\|_{W^{d+1, 4}(\gamma_d; \r^d)}^2,
    \end{align}
    where the second to last line applies Hölder's inequality.
    Redefining constants, this implies
    \begin{equation}
        \sup_{t \in [0, t_0]}\|U_tf\|_{W^{d+1, 2}(\gamma_d; \r^d)} \leq C \|f\|_{W^{d+1,4}(\gamma_d; \r^d)} \quad (\forall f \in C^\infty_c(\r^d; \r^d)) 
    \end{equation}
    and so the linear operators $U_t: W^{d+1,4}(\gamma_d; \r^d) \to W^{d+1,2}(\gamma_d; \r^d)$ are uniformly bounded for small enough $t$. Furthermore, for $f \in C^\infty_c(\r^d; \r^d)$ we know $U_t f \to f$ as $t \downarrow 0$ in $C^{d+1}_{\operatorname{loc}}(\r^d; \r^d)$ by the product rule and (\ref{eqn:proofappendixboundednessofderivativeshwnenneeded}). The boundedness allows us to extend this convergence to general $W^{d+1,4}(\gamma_d; \r^d)$ maps, which we apply to $g$ to prove (\ref{eqn:appendixproofWd12convergenceofg_t}).

    Let us now fix $N, L \in \n$ and define the matrices $\hat{A}_{n,L} := \frac{1}{\sqrt{N_{n,d}}} \sqrt{C_{n,L}C_{n,L}^\top}$ and the kernel
    \begin{equation}
        k_{N,L}(x, y) := \sum_{n=0}^N \sum_{\substack{\ell,r = 0 \\ n + 2\ell \geq 1 \\ n + 2 r \geq 1}}^L (\hat{A}_{n,L})_{\ell,r} \sum_{i=1}^{N_{n,d}}E_{n,\ell, i}(x) E_{n,r,i}(y)^\top.
    \end{equation}
    This is jointly-continuous since $N,L$ are finite, it is rotation-equivariant due to the sum over $i$, and since $\hat{A}_{n,L} \succeq 0$ it is a positive kernel. Furthermore, we compute
    \begin{equation}
        \int_{\r^d}\operatorname{tr}k_{N,L}(x, x)d\gamma_d(x) = \sum_{n=0}^N{N_{n,d}} \operatorname{tr}(\hat{A}_{n,L}) = \sum_{n=0}^N \sqrt{N_{n,d}} \operatorname{tr}\left(\sqrt{C_{n,L}C_{n,L}^\top}\right). 
    \end{equation}
    Moreover, if we let
    \begin{equation}
        g_{N,L} := \sum_{n=0}^N \sum_{\substack{\ell = 0\\ n+2\ell \geq 1}}^L \sum_{i=1}^{N_{n,d}} (C_{n,L})_{\ell, i} E_{n,\ell, i}
    \end{equation}
    be the truncation of $g$ then by the same calculations as before we have
    \begin{equation}
        \|g_{N,L}\|_{\h_{k_{N,L}}}^2 = \sum_{n=0}^N \operatorname{tr}\left(\hat{A}_{n,L}^\dagger C_{n,L}C_{n,L}^\top\right) = \sum_{n=0}^N \sqrt{N_{n,d}} \operatorname{tr}\left(\sqrt{C_{n,L} C_{n,L}^\top}\right).
    \end{equation}
    In particular, we find a sequence $(k_{N,N})_{N \in \n} \subset \mathcal{K}_{\operatorname{rot}}$ of jointly-continuous rotation-equivariant kernels for which
     \begin{equation}
     \label{eqn:proofappendixthingneededtoexhibitkNforupperbound}
        \lim_{N \to \infty} \sqrt{T_N} \cdot \|g_{N,N}\|_{\h_{k_{N,N}}} = \widetilde{\mathcal{D}}(w), \quad T_N := \int_{\r^d}\operatorname{tr} k_{N,N}(x, x) d\gamma_d(x) < \infty.
    \end{equation}
Let us define the jointly-continuous rotation-invariant scalar Sobolev kernel \begin{equation}q_{\operatorname{Sob}}(x, y) := \sum_{\alpha \in (\n \cup \{0\})^d} (1+|\alpha|)^{-(d+1)} h_\alpha (x) h_\alpha(y), \quad \tau_{\operatorname{Sob}} := d\int_{\r^d} q_{\operatorname{Sob}}(x, x) d\gamma_d(x) < \infty\end{equation}
    with $\{h_\alpha\}_{\alpha \in (\n \cup \{0\})^d}$ the normalized multivariate Hermite basis of $L^2(\gamma_d)$. Importantly, $\mathcal{T}_{q_{\operatorname{Sob}}, \gamma_d} = (\1 - \Delta_\gamma)^{-(d+1)}$ for $\Delta_\gamma = \Delta - x \cdot \nabla$ the OU generator and therefore
    \begin{equation}
        \|\cdot\|_{\h_{q_{\operatorname{Sob}}}} \asymp \|\cdot \|_{W^{d+1,2}(\gamma_d)}
    \end{equation} 
    by the Gaussian Meyer inequalities. We denote $k_{\operatorname{Sob}}(x, y) := \frac{1}{\tau_{\operatorname{Sob}}} q_{\operatorname{Sob}}(x, y) \1_{d}$ and
    \begin{equation}
        \delta_N := \|g_{N,N} - g\|_{\h_{k_{\operatorname{Sob}}}}^2,
    \end{equation}
    and we note that $\delta_N \to 0$ since $g \in W^{d+1,2}(\gamma_d; \r^d)$ and the Hermite truncation diagonalizes $\Delta_\gamma$. We define the modified kernel
    \begin{equation}
        \widetilde{k}_{N}(x, y) := k_{N,N}(x, y) + \left(\sqrt{\delta_N} + \frac{1}{N}\right) k_{\operatorname{Sob}}(x, y)
    \end{equation}
    and compute that
    \begin{equation}
        \int_{\r^d} \operatorname{tr}\widetilde{k}_N(x, x) d\gamma_d(x) = T_N + \sqrt{\delta_N} + \frac{1}{N}.
    \end{equation}
    Furthermore, by the RKHS sum formula we know 
    \begin{align}
        \|g\|_{\h_{\widetilde{k}_N}}^2 &\leq \|g_{N,N}\|_{\h_{k_{N,N}}} ^2 + \|g - g_{N,N}\|^2_{ \h_{(\sqrt{\delta_N} + \frac{1}{N})k_{\operatorname{Sob}}}}
        \\ &= \|g_{N,N}\|_{\h_{k_{N,N}}} ^2 + \frac{\delta_N}{\sqrt{\delta_N} + \frac{1}{N}}.
    \end{align}
   Putting these together, it holds that
   \begin{align}
       \limsup_{N \to \infty} {\int_{\r^d} \operatorname{tr}\widetilde{k}_N(x, x) d\gamma_d(x)} \cdot \|g\|_{\h_{\widetilde{k}_N}}^2 &\leq \limsup_{N \to \infty} \left(T_N + \sqrt{\delta_N} + \frac{1}{N}\right)  \cdot \left(\|g_{N,N}\|_{\h_{k_{N,N}}}^2 + \sqrt{\delta_N}\right) \\&= \limsup_{N \to \infty} T_N \|g_{N,N}\|_{\h_{k_{N,N}}}^2
       \\ &= \widetilde{\mathcal{D}}(w)^2.
   \end{align}
   Lastly, since $\|\cdot\|_{\h_{\widetilde{k}_N}} \lesssim \|\cdot\|_{\h_{k_{\operatorname{Sob}}}} \lesssim \|\cdot\|_{W^{d+1,2}(\gamma_d; \r^d)}$ for each fixed $N$ we know $\lim_{t \downarrow 0}\|g_t - g\|_{\h_{\widetilde{k}_N}} = 0$ by (\ref{eqn:appendixproofWd12convergenceofg_t}). The upper bound then follows from (\ref{eqn:proofappendixthehtingtoproveforupperboundnormofg}).

   Thus far, we have shown that for general flows driven by $w: \r^d \to \r^d$ we have
   \begin{equation}
       Q = \widetilde{\mathcal{D}}(w),
   \end{equation}
   where the $\geq$ direction holds if $w$ is globally-Lipschitz and the $\leq$ direction also requires $w \in C^{d+2}(\r^d; \r^d)$ with $\|Dw\|_{C^{d+1}} < \infty$. Furthermore, by the definition in (\ref{eqn:proofappendixdefinitionofDtildew}) we have that
   \begin{equation}
   \label{eqn:proofapendixalternativerepresentationofDtildew}
       \widetilde{\mathcal{D}}(w) := \sum_{n=0}^\infty \sqrt{N_{n,d}} \cdot \|C_n\|_{\mathcal{S}_1},
   \end{equation}
   where $\|\cdot\|_{\mathcal{S}_1}$ denotes the sum of singular value magnitudes. 
   To finish up, we specialize to the single-index case $w(x) = \sigma u \phi(\la x, \theta\ra)$ and denote $c := \la u, \theta\ra$. We take $v \in \mathbb{S}^{d-1}$ s.t. $\la v, \theta\ra = 0$ and $u = c\theta + \sqrt{1-c^2}v$. This implies that for any $n,\ell \in \n \cup \{0\}$ with $m := n + 2\ell \geq 1$ and $i \in \{1, \ldots, N_{n,d}\}$, if we denote $Z := \la \xi, \theta\ra$ for $\xi \sim \gamma_d$ and $s := \sqrt{1-c^2}$ then we have
   \begin{align}
       (C_n)_{\ell, i} &= \la w, E_{n,\ell,i}\ra_{L^2(\gamma_d; \r^d)} = \frac{\sigma}{\sqrt{m}} \mathbb{E}\left[\phi(Z) \la u, \nabla P_{n,\ell,i}(\xi)\ra\right] \\&= \frac{\sigma \hat{\phi}_{m-1}}{\sqrt{m}(m-1)!} \mathbb{E}\left[\operatorname{He}_{m-1}(Z) \la u, \nabla P_{n,\ell,i}(\xi)\ra\right],
   \end{align}
   where in the second line we used $\phi = \sum_{j=0}^\infty \frac{\hat{\phi}_j}{j!} \operatorname{He}_j$ and that $x \mapsto \la u, \nabla P_{n,\ell ,i}(x)\ra$ belongs to the $(m-1)^{th}$ Hermite chaos. For any $f \in W^{1,2}(\gamma_d)$ we have the Gaussian integration by parts identity
   \begin{equation}
       \mathbb{E}_{\xi \sim \gamma_d}\left[\operatorname{He}_{m-1}(\la \xi, \theta\ra) \la u, \nabla f(\xi)\ra\right] = \la u, \theta\ra \mathcal{A}_m(\theta) + \frac{1}{m}\la u, \nabla_{\mathbb{S}^{d-1}} \mathcal{A}_m(\theta)\ra,
   \end{equation}
   where we define $\mathcal{A}_m(\theta) := \mathbb{E}_{\xi \sim \gamma_d}[\operatorname{He}_m(\la \xi, \theta\ra) f(\xi)]$. In the case $f = P_{n,\ell,i}$, there is the identity
   \begin{equation}
      \frac{1}{\sqrt{m!}} \mathbb{E}_{\xi \sim \gamma_d}[\operatorname{He}_m(\la \xi, \theta\ra) P_{n,\ell,i}(\xi)] = \beta_{n,\ell,d} Y_{n,i}(\theta), \quad \beta_{n,\ell,d} := \sqrt{\frac{m! \Gamma(d/2)}{2^m \ell! \Gamma(n+\ell+d/2)}}.
   \end{equation}
 To see this, we start from (\ref{eqn:proofappendixdefinitionofthe}) and note the Laguerre polynomials have leading coefficient $L_\ell^{(\alpha)}(a) = \frac{(-1)^\ell}{\ell!} a^\ell + \text{terms of degree $<\ell$}$. Hence, the degree-$m$ homogeneous part of $P_{n,\ell,i}$, denoted $[P_{n,\ell,i}]_m$, is given by
 \begin{equation}
[P_{n,\ell,i}]_m(x) = \frac{1}{2^\ell \ell!} \sqrt{\frac{\ell! \Gamma(d/2)}{2^n \Gamma(n+\ell + d/2)}} \cdot \|x\|^m Y_{n,i}(x / \|x\|).
 \end{equation}
Applying Gaussian integration by parts $m$ times yields
 \begin{equation}
     \mathbb{E}_{\xi \sim\gamma_d}[\operatorname{He}_m(\la \xi, \theta\ra) P_{n,\ell,i}(\xi)] = \mathbb{E}_{\xi \sim \gamma_d}[(\theta \cdot \nabla)^m P_{n,\ell,i}(\xi)] = \frac{m!}{2^\ell\ell!} \sqrt{\frac{\ell! \Gamma(d/2)}{2^n\Gamma(n+\ell+d/2)}} Y_{n,i}(\theta)
 \end{equation}
 which implies the claimed identity.
 We find that
   \begin{align}
       (C_n)_{\ell,i} &= \frac{\sigma \hat{\phi}_{m-1} \sqrt{m!} \beta_{n,\ell,d}}{\sqrt{m}(m-1)!} \left(\la u, \theta\ra Y_{n,i}(\theta) + \frac{1}{m} \la u, \nabla_{\mathbb{S}^{d-1}}Y_{n,i}(\theta)\ra\right)
       \\ &= \frac{\sigma \hat{\phi}_{m-1} \sqrt{m!} \beta_{n,\ell,d}}{\sqrt{m}(m-1)!} \left(c Y_{n,i}(\theta) + \frac{s}{m} \la v, \nabla_{\mathbb{S}^{d-1}}Y_{n,i}(\theta)\ra\right),
       \label{eqn:appendixproofformulaforCnelliwhichshowsitsrank2}
   \end{align}
   where we used that $\nabla_{\mathbb{S}^{d-1}}Y_{n,i}(\theta)$ is tangent to the sphere at $\theta$ and therefore orthogonal to $\theta$. By differentiating the addition theorem for spherical harmonics (see \cite[Theorem 2.9]{atkinson2012spherical}), we have
   \begin{align}
       \sum_{i=1}^{N_{n,d}} Y_{n,i}(\theta)^2 = N_{n,d},
\quad \sum_{i=1}^{N_{n,d}} Y_{n,i}(\theta) \la v, \nabla_{\mathbb{S}^{d-1}}Y_{n,i}(\theta)\ra = 0, \\ \sum_{i=1}^{N_{n,d}} \la v, \nabla_{\mathbb{S}^{d-1}} Y_{n,i}(\theta)\ra^2 = N_{n,d} \frac{n(n+d-2)}{d-1}.
   \end{align}
   In particular, for fixed $n,\ell$ we find that  
   \begin{equation}
       \sum_{i=1}^{N_{n,d}}(C_n)_{\ell, i}^2 = \frac{\sigma^2 |\hat{\phi}_{m-1}|^2 \beta_{n,\ell,d}^2 N_{n,d}}{(m-1)!} \left(c^2  + \frac{s^2}{m^2} \frac{n(n+d-2)}{d-1}\right)
   \end{equation}
   By comparing with (\ref{eqn:appendixdefinitionofDpphicforthegaussianbounds}), we find that
   \begin{equation}
       \sigma \mathcal{D}_\phi(\la u, \theta\ra) = \sum_{n = 0}^\infty \sqrt{N_{n,d}} \cdot \|C_n\|_F.
   \end{equation}
   The final claim then follows from (\ref{eqn:proofapendixalternativerepresentationofDtildew}) and the fact that $\|C_n\|_{F} \leq \|C_n\|_{\mathcal{S}_1} \leq \sqrt{2}\|C_n\|_{F}$ since each $C_n$ has rank at most $2$, see (\ref{eqn:appendixproofformulaforCnelliwhichshowsitsrank2}).
\end{proof}

Lastly, we show the following perturbative lower bound on the Barron transport metric with Gaussian source:

\begin{lemma}[First-order $D_{\b_\phi}$ lower bound]
\label{lemma:appendixbarronfirstorderlowerbound}
    Let $\phi \in C^1(\r)$ satisfy Assumption \ref{assumption:activationfunciton} with $M := \max\{\|\phi\|_\infty, \|\phi'\|_\infty\}  \in(0, \infty)$. Fix $d \geq 2$ and $u, \theta \in \mathbb{S}^{d-1}$, and for each $\epsilon > 0$ define $\nu_\epsilon := \operatorname{Law}(X_1)$ according to the flow 
    \begin{equation}
        \dot{X}_t = \epsilon w(X_t), \quad w(x) := u \phi(\la x, \theta\ra), \quad X_0 \sim \gamma_d.
    \end{equation}
    Then, with $c := \la u, \theta \ra$ we have that
    \begin{equation}
        \liminf_{\epsilon \downarrow 0} \frac{D_{\b_\phi}(\gamma_d, \nu_\epsilon)}{\epsilon} \geq \frac{1}{M} \sup_{a,b \in \r}\frac{\left|(a \sqrt{1-c^2}+bc)\phi(b) - c \phi'(b)\right|}{1+\sqrt{a^2 + b^2}}.
    \end{equation}
    If $\phi \in \{\tanh, \sin\}$ then the right hand side is $\geq 1$ for all $c \in [-1, 1]$.
\end{lemma}
\begin{proof}
    Note that since $\|\phi\|_\infty, \|\phi'\|_\infty \leq M$ we know
\begin{equation}
    \|u\|_\infty, \|\operatorname{div} u\|_\infty \leq M\|u\|_{\b_{\phi}} \quad (\forall u \in \b_{\phi}).
\end{equation}
Now, let $(v_t, \rho_t)_{t \in [0, 1]}$ be an admissible transport from $\gamma_d$ to $\nu_\epsilon$  with \begin{equation}
    L := \int_0^1 \|v_t\|_{\b_{\phi}} dt < \infty.
\end{equation}
The prescribed flow has a $C^1$ bi-Lipschitz endpoint map (since $\phi$ is $C^1$ and Lipschitz), so $\nu_\epsilon$ has a continuous positive density. Now, if we take $(Y_t^x)_{t \in [0, 1]}$ to be a characteristic ending at some $Y_1^x = x \in \r^d$ then there is the change of variables formula
\begin{equation}
    \log \frac{\nu_\epsilon}{\gamma_d}(x) = -\frac{\|Y_0^x\|^2 - \|x\|^2}{2} - \int_0^1 \operatorname{div} v_t(Y_t^x)dt;
\end{equation}
at first this holds for a.e. $x$ by Lipschitz regularity, which we extend by continuity to hold for every $x$ (note that each $\operatorname{div}v_t$ is continuous since $\phi$ is $C^1$).
Since $\|Y_0^x - x\| \leq \int_0^1 \|v_t\|_\infty dt \leq ML$ we know $\left|\frac{\|Y_0^x\|^2 - \|x\|^2}{2}\right| \leq  \frac{M^2L^2}{2}+ML \|x\| $, and so
\begin{equation}
\label{eqn:proofsphericaltaubarronuperbounddensityratio}
    \left|\log \frac{\nu_\epsilon}{\gamma_d}(x)\right| \leq \frac{M^2L^2}{2} +M L(1+\|x\|).
\end{equation}
However, for the flow $\dot{X}_t = \epsilon w(X_t)$ which generates $\nu_\epsilon$ we know that $X_1 = X_0 + \epsilon w(X_0) + o(\epsilon)$ and so $\nu_\epsilon = \gamma_d - \epsilon \operatorname{div}(\gamma_d w) + o(\epsilon)$.
In particular, using $\log(1+a) = a + o(a)$ we find that the log density must satisfy 
\begin{align}
  \log   \frac{\nu_\epsilon}{\gamma_d}(x) &=\epsilon\left( \la x, w(x)\ra - \operatorname{div}w(x)\right) + o_x(\epsilon) \\
  &= \epsilon \left(\la x, u\ra \phi(\la x, \theta\ra) - \la u, \theta\ra \phi'(\la x, \theta\ra)\right) + o_x(\epsilon).
\end{align}
Combining with (\ref{eqn:proofsphericaltaubarronuperbounddensityratio}), we find that
\begin{equation}
   \epsilon \left|\la x, u\ra \phi(\la x, \theta\ra) - \la u, \theta\ra \phi'(\la x, \theta\ra)\right| + o_x(\epsilon)  \leq \frac{M^2L^2}{2}+ML(1+\|x\|)
\end{equation}
and therefore
\begin{equation}
    L \geq \frac{\sqrt{(1+\|x\|)^2 +  2\epsilon \left|\la x, u\ra \phi(\la x, \theta\ra) - \la u, \theta\ra \phi'(\la x, \theta\ra)\right| + o_x(\epsilon)} - (1+\|x\|)}{M}
\end{equation}
Taking an infimum over admissible paths followed by $\epsilon \downarrow 0$ with $x$ fixed implies that
\begin{equation}
    \liminf_{\epsilon \downarrow 0}\frac{D_{\b_\phi}(\gamma_d, \nu_\epsilon)}{\epsilon} \geq \frac{ \left|\la x, u\ra \phi(\la x, \theta\ra) - \la u, \theta\ra \phi'(\la x, \theta\ra)\right|}{M(1+\|x\|)}.
\end{equation}
This holds for any $x \in \r^d$, and so we may take a supremum over $x$. It suffices to consider $x \in \operatorname{span}\{u, \theta\}$ since otherwise the denominator is strictly suboptimal, and so we write $c = \la u, \theta\ra$,  $u = c\theta + \sqrt{1-c^2} v$ for a unit vector $v$ orthogonal to $\theta$, and we express
$x = av + b\theta$. In this case the first claim follows due to
\begin{equation}
    \frac{ \left|\la x, u\ra \phi(\la x, \theta\ra) - \la u, \theta\ra \phi'(\la x, \theta\ra)\right|}{1+\|x\|} = \frac{\left|(a\sqrt{1-c^2} + bc) \phi(b) - c \phi'(b)\right|}{1 + \sqrt{a^2 + b^2}}
\end{equation}
For the second claim, note first that if $\phi \in \{\tanh, \sin\}$ then $M = 1$.
Now, if $c = 0$ then the expression becomes $\frac{|a \phi(b)|}{1 + \sqrt{a^2 + b^2}}$, which goes to $|\phi(b)|$ as we take $a \to \infty$. Hence, the right hand side equals $\sup_b |\phi(b)| = 1$. If instead $c \neq 0$ then we take $x = R \sqrt{1-c^2}v + R c \theta = Ru$ for $R > 0$ to make the expression equal \begin{equation}
\label{eqn:proofappendixbarronfirstorderlowerboundsintanh}
\frac{|R\phi(Rc) - c \phi'(Rc)|}{1+R}.\end{equation}
For $\tanh$ we let $R \to \infty$ to find that $|\tanh(Rc)| \to 1$ while $\tanh'(Rc) = \operatorname{sech}(Rc)^2 \to 0$. For $\sin$ we take $R_j = \frac{\pi / 2 + 2 \pi j}{|c|}$ so that $|\sin(R_j c)| = 1$ while $\cos(R_j c) = 0$. In either case, (\ref{eqn:proofappendixbarronfirstorderlowerboundsintanh}) tends to 1 and so the result follows.
\end{proof}

\subsection{Feature learning comparisons}
\label{subsection:generativemodelingproofs}
In this section we derive the results we present in the main paper regarding comparison of $D_{\b_\phi}$ and rotation-equivariant $D_{\h_k}$. We first apply the spherical harmonic lower bound of Lemma \ref{lemma:rkhslowerboundspherical} to control the case of single-index flows with ReLU activation:

 \begin{proof}[Proof of Example \ref{example:nonlinearshearrkhs}]
Let $w(x) := q\phi(\la x, u\ra)$ with $q, u \in \mathbb{S}^{d-1}$ s.t. $c := \la q, u\ra \neq 0$ (we relabel the input and output features for notational simplicity).
We start from the ReLU-specific observation that $\1 + \epsilon w$ is the time-1 map of the flow
\begin{equation}
\label{eqn:proofsphericalODEformulation}
    \dot{X}_t = \frac{\log(1+c\epsilon)}{c} w(X_t),
\end{equation}
where $1+c\epsilon > 0$ for small enough $\epsilon$.
To see this, note that the induced dynamics of $a_t := \la X_t, u\ra$ are \begin{equation}\dot{a}_t = {\log(1+c\epsilon)} \max\{0, a_t\}\end{equation}
and therefore $a_t = e^{\log(1+c\epsilon) t}a_0$ if $a_0 > 0$ and $a_t = a_0$ if $a_0 \leq 0$. Hence,
\begin{equation}X_1 = X_0 + \frac{1+c\epsilon-1}{c}\max\{0, a_0\}q = (\1 + \epsilon w)(X_0)\end{equation}
and so (\ref{eqn:proofsphericalODEformulation}) is verified. 
Since $\|w\|_{\b_\phi} \leq 1$ by construction (see Definition \ref{def:barronspaceshomogeneous}), and $\frac{\log(1+c\epsilon)}{c\epsilon} \to 1$, we get the Barron upper bound \begin{equation}
    \label{eqn:proofsphericalbarronupperbound}
  \limsup_{\epsilon \downarrow 0} \frac{D_{\b_\phi}(\mu, \nu_\epsilon)}{\epsilon} \leq 1 .
\end{equation}

    For the Stein lower bound, we will show for odd $n \geq d$ that $A_n \gtrsim n^{-1}$ (which is $\gtrsim n^{1-d/2}$ whenever $d \geq 4$), after which the result will follow from Lemma \ref{lemma:rkhslowerboundspherical}. To this end, 
    let $P_{n, d}: [-1, 1] \to \r$ be the Legendre polynomials defined in (2.19) of \cite{atkinson2012spherical}. By the addition theorem for spherical harmonics (see \cite[Theorem 2.9]{atkinson2012spherical}), if $\{Y_{n,j}:\; 1 \leq j \leq N_{n,d}\}$ is an orthonormal basis of $\mathbb{Y}_n^d$ w.r.t. the normalized uniform probability measure $\sigma$ on $\mathbb{S}^{d-1}$ then 
\begin{equation}
    \sum_{j=1}^{N_{n,d}} Y_{n,j}(\xi) {Y_{n,j}(\eta)} = N_{n,d} P_{n,d}(\la \xi, \eta\ra) \quad (\forall \xi, \eta \in \mathbb{S}^{d-1}). 
\end{equation}
Let us take our input feature $u \in \mathbb{S}^{d-1}$ and define $Z_{n,u}: \mathbb{S}^{d-1} \to \r$ by
\begin{equation}
    \label{eqn:proofexampledefofZnu}Z_{n,u}(\theta) := \sqrt{N_{n,d}} P_{n,d}(\la \theta, u\ra),
\end{equation} and so $Z_{n, u} \in \mathbb{Y}_n^d$. By equation (2.40) of \cite{atkinson2012spherical} we know $\|Z_{n,u}\|_{L^2(\sigma)} = 1$. We will take $Y = Z_{n,u}$ as the comparator to lower bound $A_n$.

We claim that if $\phi: \r \to \r$ is positively one-homogeneous then with $w(x) = q \phi(\la u, x\ra) $ and $\varphi_n(x) = |x|^2 Z_{n,u}(\frac{x}{|x|})$ we have
\begin{equation}
    \label{eqn:proofsphericalclaim1}
\int_{\r^d} \la \nabla \varphi_n,  w\ra d\mu = cM_2(\mu)^2\left(2 + \frac{n(n+d-2)}{2}\right) \cdot \beta_n(\phi),
\end{equation}
where we denote $M_2(\mu) := \sqrt{\int_{\r^d} \|x\|^2 d\mu(x)}$ and we define the coefficient of $\phi$ by
\begin{equation}
\label{eqn:proofsphericalbetaintegral}
    \beta_n(\phi) := \left(\phi(1) + (-1)^{n+1} \phi(-1)\right)\int_{\{\la \theta, u\ra > 0\}} \la \theta, u\ra^2 Z_{n, u}(\theta) d\sigma(\theta).
\end{equation}
To see this, we start from
\begin{equation}\nabla \varphi_n(r\theta) = 2r Z_{n,u}(\theta) \theta + \frac{r^2}{r} \nabla_{\mathbb{S}^{d-1}}Z_{n,u}(\theta),\end{equation}
after which we apply positive homogeneity to find
\begin{align}
\int_{\r^d} \la \nabla \varphi_n,  w\ra d\mu &=  \mathbb{E}_{R, \Theta} \left[2R^2Z_{n,u}(\Theta)\phi(\la \Theta, u\ra) \la \Theta, q\ra + R^2 \phi(\la \Theta, u\ra) \la q, \nabla_{\mathbb{S}^{d-1}}Z_{n,u}(\Theta)\ra\right] \\
   &= cM_2(\mu)^2 \left[2\int_{\mathbb{S}^{d-1}} t\phi(t) Z_{n,u}d\sigma +  \int_{\mathbb{S}^{d-1}} \phi(t) \la u, \nabla_{\mathbb{S}^{d-1}}Z_{n,u}\ra d\sigma\right],
\end{align}
where in the second line we denote $t(\Theta) := \la \Theta, u\ra$ and apply independence of $R,\Theta$ and symmetry about $u$ (which replaces $q$ with $cu$). Now, since $\nabla_{\mathbb{S}^{d-1}}t(\theta) = u - t(\theta)\theta$ and $\nabla_{\mathbb{S}^{d-1}}Z_{n,u}(\theta)$ is orthogonal to $\theta$ we find that
\begin{equation}\la u, \nabla_{\mathbb{S}^{d-1}}Z_{n,u}\ra= \la \nabla_{\mathbb{S}^{d-1}}t, \nabla_{\mathbb{S}^{d-1}}Z_{n,u}\ra.\end{equation}
We now use that $(t\phi(t))' = 2\phi(t)$ a.e. for positively one-homogeneous functions to get
\begin{align}
    \int_{\mathbb{S}^{d-1}} \phi(t) \la u, \nabla_{\mathbb{S}^{d-1}}Z_{n,u}\ra d\sigma &= \int_{\mathbb{S}^{d-1}} \frac{(t\phi(t))'}{2}  \la \nabla_{\mathbb{S}^{d-1}}t, \nabla_{\mathbb{S}^{d-1}}Z_{n,u}\ra d\sigma \\
    &= \frac{1}{2}\int_{\mathbb{S}^{d-1}}  \la \nabla_{\mathbb{S}^{d-1}}(t \phi(t)), \nabla_{\mathbb{S}^{d-1}}Z_{n,u}\ra d\sigma\\
    &= \frac{n(n+d-2)}{2}\int_{\mathbb{S}^{d-1}} t \phi(t) Z_{n, u} d\sigma ,
\end{align}
where the second line uses the chain rule and the third uses equation (3.20) of \cite{atkinson2012spherical}. Thus,
\begin{equation}\int_{\r^d} \la \nabla \varphi_n, w\ra d\mu = cM_2(\mu)^2\left(2 + \frac{n(n+d-2)}{2}\right) \int_{\mathbb{S}^{d-1}} t\phi(t) Z_{n,u}d\sigma.\end{equation}
Now, if we let $s_+ := \phi(1)$ and $s_- := -\phi(-1)$ be the slopes then
\begin{equation}t \phi(t) = \begin{cases}
    s_+t^2, & t > 0,\\
    s_- t^2, & t < 0,
\end{cases}\end{equation}
which we combine with $Z_{n,u}(-\theta) = (-1)^n Z_{n,u}(\theta)$ to get that
\begin{align}\int_{\mathbb{S}^{d-1}}t\phi(t)Z_{n,u}d\sigma &= s_+ \int_{\{t > 0\}} t^2 Z_{n,u}d\sigma + s_- \int_{\{t<0\}}t^2Z_{n,u}d\sigma\\
&= (s_+ + (-1)^ns_-) \int_{\{t>0\}} t^2 Z_{n,u}d\sigma.
\end{align}
Thus, the equality in (\ref{eqn:proofsphericalclaim1}) is proved and it remains to lower bound the integral in (\ref{eqn:proofsphericalbetaintegral}).

To this end, we claim that if $n \geq d \geq 3$ and $n$ is odd then
    \begin{equation}
    \label{eqn:proofsphericalhomogeneouslowerbound}
        \left| \int_{\{\la \theta, u\ra > 0\}} \la \theta, u\ra^2 Z_{n,u}(\theta) d\sigma(\theta)\right| \geq \frac{d^{1/4}n^{-3}}{6}.
    \end{equation}
    For $d\geq 3$ and $n \geq 3$ with $n$ odd
    we have
\begin{align}
   \left| \int_{\{\la \theta, u\ra > 0\}} \la \theta, u\ra^2 Z_{n,u}(\theta) d\sigma(\theta)\right| &= \frac{\Gamma(d/2)}{4\pi} \sqrt{\frac{(2n+d-2)\Gamma(n+d-2)}{\Gamma(n+1)\Gamma(d-1)}} \frac{\Gamma((n-2)/2)}{\Gamma((n+d+2)/2)}\\
   &= \frac{\pi^{-3/4}}{4\sqrt{2}}\sqrt{2n+d-2} \left(\frac{\Gamma(d/2)}{\Gamma((d-1)/2)}\right)^{1/2} \cdot \frac{\Gamma((n-2)/2)}{\sqrt{\Gamma((n+1)/2) \Gamma((n+2)/2)} }\\&\qquad  \cdot \frac{\sqrt{\Gamma((n+d-2)/2)\Gamma((n+d-1)/2)}}{\Gamma((n+d+2)/2)} \\
   &\geq \frac{\pi^{-3/4}}{4\sqrt{2}}\sqrt{2n+d-2} \cdot 2^{-3/4}d^{1/4} \cdot \frac{2^{7/4}}{n^{7/4}} \cdot \left(\frac{3}{4}\right)^{1/4} \frac{2^{7/4}}{(n+d)^{7/4}}
   \\ &\geq \frac{3^{1/4}}{2^{1/4}\pi^{3/4}} \cdot \frac{d^{1/4}}{n^{7/4}(n+d)^{5/4}}
\end{align}
where the first line is \href{https://dlmf.nist.gov/18.17#E37}{DLMF 18.17.37} with $\lambda = (d-2)/2$, the second uses the duplication formula, the third uses Wendel's inequality $(\frac{x}{x+s})^{1-s} \leq \frac{\Gamma(x+s)}{x^s \Gamma(x)} \leq 1$ for $x > 0$, $s \in (0, 1)$ on all three $\Gamma$ fraction terms along with $(n+d-2)/2 \geq 3/2$, and the fourth uses $\sqrt{2n+d-2} \geq \sqrt{n+d}$. In particular, if we take $n \geq d$ then
\begin{equation}\left| \int_{\{\la \theta, u\ra > 0\}} \la \theta, u\ra^2 Z_{n,u}(\theta) d\sigma(\theta)\right| \geq \frac{3^{1/4}}{2^{3/2}\pi^{3/4}} \cdot d^{1/4}n^{-3} \geq \frac{d^{1/4}n^{-3}}{6},\end{equation}
and so (\ref{eqn:proofsphericalhomogeneouslowerbound}) is proved.

Combining (\ref{eqn:proofsphericalclaim1}) and (\ref{eqn:proofsphericalhomogeneouslowerbound}), we use our particular choice $Y = Z_{n,u}$ and $\varphi_n = |x|^2 Z_{n,u}(\frac{x}{|x|})$ to find that for any odd $n \geq d $ and any positively one-homogeneous $\phi$ we have
\begin{align}A_n &\geq \left|\int_{\r^d} \la \nabla \varphi_n, w\ra d\mu\right| \\&\geq |c|M_2(\mu)^2 \left(2 + \frac{n(n+d-2)}{2}\right) \left|\phi(1) + \phi(-1)\right| \cdot \frac{d^{1/4}n^{-3}}{6} \\
&\geq |c|\frac{M_2(\mu)^2d^{1/4}}{12n} |\phi(1) + \phi(-1)|  .\end{align}
Thus, $A_n \gtrsim n^{-1}$ for odd $n \geq d$ and any positively one-homogeneous $\phi$ which is not a scalar multiple of the identity. Therefore,
\begin{equation}\sqrt{a_{n,d}}A_n \gtrsim \frac{d^{1/4}}{\sqrt{(d-2)!}} \cdot n^{(d-4)/2 - 1},\end{equation}
which is not summable in $n$ for $d \geq 4$. Hence, the right hand side of the lower bound in (\ref{eqn:sphericalclaimquadratic}) can be made as large as desired.
\end{proof}

This lack of summability in any $d \geq 4$ in the ReLU case is also visible from Lemma 
\ref{lemma:appendixupperandlowerboundsforsingleindexgaussianflows} when $\mu = \gamma_d$. We will instead  derive the correct scaling in the $\tanh$ case:

\begin{lemma}
\label{lemma:appendixtanhhermitescaling}
    With $\phi = \tanh$, as $d \to \infty$ we have that
    \begin{equation}
        \log \mathcal{D}_{\tanh}(c) = \frac{d}{4} \log d + O(d)
    \end{equation}
    uniformly for $c \in [-1, 1]$.
\end{lemma}
\begin{proof}
    Since $\tanh$ is odd, the coefficients $\hat{\phi}_{m-1}$ vanish when $m-1$ is even (i.e. when $m$ is odd). In the other cases $m=2q$, we claim that as $q \to \infty$ there is the asymptotic
    \begin{equation}
    \label{eqn:proofappendixasymptoticsfortanhhermitecoefficients}
        \log |\hat{\phi}_{2q-1}|=q \log(2q) - q - \frac{\pi}{\sqrt{2}}\sqrt{q} + O(\log q).
    \end{equation}
    Before proving this, we show how it implies the main claim; from here on, universal constants are labeled $C_j$ and can change from line to line. 
Recall from (\ref{eqn:proofsphericalharmonicdimscaling}) that
\begin{equation}
    N_{n,d} = \frac{2n+d-2}{d-2} {n+d-3 \choose d-3} ,
\end{equation}
and since ${n + d - 3 \choose d- 3} \leq (\frac{e(n+d)}{d})^d$ we find that $N_{n,d} \leq C_0^d (1 + \frac{n}{d})^d$ for some $C_0 < \infty$. Moreover, 
\begin{equation}
\label{eqn:proofappendixtanhhermiteblahblahblah2}
N_{n,d} \geq {n+d-3 \choose d-3} \geq \frac{n^{d-3}}{(d-3)!} \geq \left(\frac{n}{d}\right)^{d-3}       \quad (\forall n \geq d).
\end{equation}
Next, with $m = n + 2\ell$ even we claim that
\begin{equation}
\label{eqn:proofappendixtanhhermiteblahblahblah1}
    |\hat{\phi}_{m-1}| \left(\frac{m\Gamma(d/2)}{2^m\ell! \Gamma(n+\ell+d/2)}\right)^{1/2} \leq (d+m)^{C_1} \left(\frac{d}{m}\right)^{d/4} \exp\left(-\frac{n^2}{4m} - C_2 \sqrt{m}\right)
\end{equation}
for some $C_1, C_2 \in (0, \infty)$. Indeed, by (\ref{eqn:proofappendixasymptoticsfortanhhermitecoefficients}) and Stirling we have for even $m$ that
\begin{align}
    \log \left( |\hat{\phi}_{m-1}| \left(\frac{m\Gamma(d/2)}{2^m\ell! \Gamma(n+\ell+d/2)}\right)^{1/2} \right) &= \log|\hat{\phi}_{m-1}| + \frac{1}{2}\log m + \frac{1}{2}\log \Gamma(d/2) - \frac{m}{2}\log 2 \\&\quad - \frac{1}{2}\log \Gamma\left(\frac{m-n}{2}+1\right) - \frac{1}{2}\log \Gamma\left(\frac{m+n+d}{2}\right)
    \\ &= \frac{m}{2} \log m - \frac{m}{2} - \frac{\pi}{2}\sqrt{m} - \frac{m}{2}\log 2 + \frac{d}{4}\log \frac{d}{2} - \frac{d}{4} \\&\quad - \frac{m-n}{4}\log\left(\frac{m-n}{2}\right) + \frac{m-n}{4} \\&\quad - \frac{m+n+d}{4} \log\left(\frac{m+n+d}{2}\right) + \frac{m+n+d}{4} \\&\quad + O(\log(d+m))
    \\
    &= \frac{d}{4} \log \frac{d}{m} - \frac{m-n}{4} \log\left(1-\frac{n}{m}\right)
    \label{eqn:proofappendixtanhhermiteblahblahblah3}
    \\&\quad - \frac{m+n+d}{4} \log\left(1 + \frac{n+d}{m}\right) - \frac{\pi}{2}\sqrt{m}\\&\quad + O(\log(d+m)),\label{eqn:proofappendixtanhhermiteblahblahblah5}
\end{align}
where we use the convention $0 \log 0 = 0$. Therefore,
\begin{align}
    \log \left( |\hat{\phi}_{m-1}| \left(\frac{m\Gamma(d/2)}{2^m\ell! \Gamma(n+\ell+d/2)}\right)^{1/2} \right) &\leq \frac{d}{4} \log \frac{d}{m} - \frac{\pi}{2}\sqrt{m} + O(\log(d+m)) \\&\quad - \frac{m}{4}\left(\left(1 - \frac{n}{m}\right) \log\left(1 - \frac{n}{m}\right) + \left(1 + \frac{n}{m}\right) \log \left(1 + \frac{n}{m}\right)\right).
\end{align}
Since $(1-x) \log(1-x) + (1+x) \log(1+x) \geq x^2$ for $|x| \leq 1$, taking $x = n/m$ implies
    \begin{equation}
        \log \left( |\hat{\phi}_{m-1}| \left(\frac{m\Gamma(d/2)}{2^m\ell! \Gamma(n+\ell+d/2)}\right)^{1/2} \right) \leq \frac{d}{4} \log \frac{d}{m} - \frac{n^2}{4m} - \frac{\pi}{2}\sqrt{m} + O(\log(d+m)) 
    \end{equation}
    which proves (\ref{eqn:proofappendixtanhhermiteblahblahblah1}). 

    For the upper bound, we note that $c^2 + (1-c^2) \frac{n(n+d-2)}{m^2(d-1)} \leq 1$ and that $\sqrt{\sum_\ell x_\ell} \leq \sum_\ell \sqrt{x_\ell}$, and so applying our upper bound on $N_{n,d}$ and (\ref{eqn:proofappendixtanhhermiteblahblahblah1}) yields
    \begin{align}
        \mathcal{D}_{\tanh}(c) &\leq \sum_{n=0}^\infty N_{n,d} \sum_{\substack{\ell = 0 \\ n + 2\ell \geq 1}}^\infty |\hat{\phi}_{m-1}| \left(\frac{m\Gamma(d/2)}{2^m \ell! \Gamma(n+\ell+d/2)}\right)^{1/2} \\
        &\leq C_0^d \sum_{m=1}^\infty (d+m)^{C_1} \left(\frac{d}{m}\right)^{d/4} e^{-C_2\sqrt{m}} \cdot \sum_{n=0}^m \left(1 + \frac{n}{d}\right)^d e^{-\frac{n^2}{4m}} \\
        &\leq  C_0^d \sum_{m=1}^\infty (d+m)^{C_1} \left(\frac{d}{m}\right)^{d/4} e^{-C_2\sqrt{m}} \cdot \sum_{n=0}^\infty\left(1 + \frac{n}{d}\right)^d e^{-\frac{n^2}{4m}}
    \end{align}
    after reorganizing the sums. Splitting the inner sum, if $n \leq d$ then $(1 + \frac{n}{d})^d \leq 2^d$ and so $\sum_{n=0}^d \left(1 + \frac{n}{d}\right)^d e^{-\frac{n^2}{4m}} \leq (d+1)2^d$. If $n > d$ then we have
    \begin{equation}
        \sum_{n=d+1}^\infty \left(1 + \frac{n}{d}\right)^d e^{-\frac{n^2}{4m}} \leq \left(\frac{2}{d}\right)^d \sum_{n = 1}^\infty n^d e^{-\frac{n^2}{4m}} \leq \left(\frac{2}{d}\right)^d (C_4 dm)^{d/2} \sqrt{m} = C_5^d \left(\frac{m}{d}\right)^{d/2} \sqrt{m},
    \end{equation}
 where the second inequality comes from comparison with the integral and Stirling. In particular,
    \begin{equation}
        \sum_{n=0}^\infty \left(1 + \frac{n}{d}\right)^d e^{-\frac{n^2}{4m}} \leq C_6^d (d+m)^{C_7} \left(1 + \left(\frac{m}{d}\right)^{d/2}\right)
    \end{equation}
    and therefore
    \begin{equation}
        \mathcal{D}_{\tanh}(c) \leq C_8^d \sum_{m=1}^\infty (d+m)^{C_9} \left(\left(\frac{d}{m}\right)^{d/4} + \left(\frac{m}{d}\right)^{d/4}\right)e^{-C_2 \sqrt{m}}.
    \end{equation}
    We now split this sum into $m \leq d$ and $m > d$. If $m \leq d$ then $(\frac{d}{m})^{d/4} \leq d^{d/4}$ and $(\frac{m}{d})^{d/4} \leq 1$, and hence 
    \begin{equation}
    \label{eqn:proofappendixtanhupperboundhermiteblablah}
         \sum_{m=1}^d (d+m)^{C_9} \left(\left(\frac{d}{m}\right)^{d/4} + \left(\frac{m}{d}\right)^{d/4}\right)e^{-C_2 \sqrt{m}} \leq C_{10}^dd^{d/4} .
    \end{equation}
    For $m > d$ we know $\left(\frac{d}{m}\right)^{d/4} + \left(\frac{m}{d}\right)^{d/4} \leq 2\left(\frac{m}{d}\right)^{d/4}$ and hence
    \begin{align}
         \sum_{m=d+1}^\infty (d+m)^{C_9} \left(\left(\frac{d}{m}\right)^{d/4} + \left(\frac{m}{d}\right)^{d/4}\right)e^{-C_2 \sqrt{m}} &\leq C_{11}^d d^{-d/4} \sum_{m=d+1}^\infty m^{d/4 + C_{12}} e^{-c\sqrt{m}} \\
         &\leq C_{13}^d d^{-d/4}\Gamma\left(\frac{d}{2} + 2C_{12} + 2\right) \\
         &\leq C_{14}^d d^{d/4},
    \end{align}
    where the second line compares with the integral over $x = \sqrt{m}$ and the third uses Stirling. Combining this with (\ref{eqn:proofappendixtanhupperboundhermiteblablah}) proves that uniformly in $c$,
    \begin{equation}
        \log \mathcal{D}_{\tanh}(c) \leq \frac{d}{4} \log d + O(d).
    \end{equation}
    For the lower bound, if we take $m \asymp d^2$ and $n \asymp d^{3/2}$ both even then
    \begin{equation}
    \label{eqn:proofappendixtanhhermiteblahblahblah6}
        N_{n,d} \geq \left(\frac{n}{d}\right)^{d-3} = \exp\left(\frac{d}{2}\log d - O(d)\right)
    \end{equation}
    by (\ref{eqn:proofappendixtanhhermiteblahblahblah2}), and the formula of (\ref{eqn:proofappendixtanhhermiteblahblahblah3})-(\ref{eqn:proofappendixtanhhermiteblahblahblah5}) simplifies to
    \begin{equation}
        \log \left( |\hat{\phi}_{m-1}| \left(\frac{m\Gamma(d/2)}{2^m\ell! \Gamma(n+\ell+d/2)}\right)^{1/2} \right) = \frac{d}{4}\log\left(\frac{d}{m}\right) + O(d)
    \end{equation}
    by expanding out the logs. Since $m \asymp d^2$, this yields
    \begin{equation}
    \label{eqn:proofappendixtanhhermiteblahblahblah7}
        |\hat{\phi}_{m-1}| \left(\frac{m\Gamma(d/2)}{2^m\ell! \Gamma(n+\ell+d/2)}\right)^{1/2} \geq \exp\left(-\frac{d}{4}\log d - O(d)\right).
    \end{equation}
    Lastly, since $c^2 + (1-c^2) \frac{n(n+d-2)}{m^2(d-1)} \geq \frac{n(n+d-2)}{m^2(d-1)} \gtrsim d^{-2}$ we may multiply (\ref{eqn:proofappendixtanhhermiteblahblahblah6}) and (\ref{eqn:proofappendixtanhhermiteblahblahblah7}) to lower bound this single term in the sum. Hence, the lower bound
    \begin{equation}
        \log \mathcal{D}_{\tanh}(c) \geq \frac{d}{4}\log d - O(d)
    \end{equation}
    holds uniformly in $c$.

We now prove the Hermite coefficient asymptotics (\ref{eqn:proofappendixasymptoticsfortanhhermitecoefficients}). Applying Gaussian integration by parts $2q-1$ times gives that if $Z \sim \mathcal{N}(0, 1)$ then
\begin{equation}
    \hat{\phi}_{2q-1} = \mathbb{E}[\tanh(Z)\operatorname{He}_{2q-1}(Z)] = \mathbb{E}[\tanh^{(2q-1)}(Z)]. 
\end{equation}
Observe that $\tanh' = \operatorname{sech}^2$ and that the Fourier transform satisfies $\widehat{\operatorname{sech}^2}(\xi) = \frac{\pi\xi}{\sinh(\pi \xi / 2)}$ (see (3.982.1) of \cite{gradshteyn2014table}), and so
\begin{align}
    \hat{\phi}_{2q-1} &= \frac{1}{\sqrt{2\pi}} \int_\r (\operatorname{sech}^2)^{(2q-2)}(x) e^{-x^2/2}dx \\&= \frac{1}{(2\pi)^{3/2}} \int_\r \widehat{(\operatorname{sech}^2)^{(2q-2)}}(\xi) \widehat{e^{-x^2/2}}(\xi) d\xi \\&= \frac{1}{2}\int_\r (i\xi)^{2q-2} \frac{\xi}{\sinh(\pi \xi / 2)} e^{-\xi^2/2}d\xi \\ 
    &= \frac{(-1)^{\frac{2q}{2}-1}}{2}\int_\r \frac{\xi^{2q-1}e^{-\xi^2/2}}{\sinh(\pi \xi / 2)} d\xi,
\end{align}
where the second line uses Parseval's identity and the fourth uses that $2q$ is even. In particular,
\begin{align}
\label{eqn:proofsphericallowerboundonLn}
    |\hat{\phi}_{2q-1}| &= \int_0^\infty \frac{\xi^{2q-1}e^{-\xi^2/2}}{\sinh(\pi \xi / 2)} d\xi = 2\int_0^\infty \frac{\xi^{2q-1}\exp\left(- \frac{\xi^2}{2} - \frac{\pi\xi}{2}\right)}{1 - e^{-\pi \xi}}d\xi.
\end{align}
Hence, since the region $\xi \in [0, 1]$ contributes $O(1)$ to this integral we find that 
\begin{equation}
    \log |\hat{\phi}_{2q-1}| = \log\left(\int_1^\infty e^{F(\xi)}d\xi\right) + O(1)
\end{equation}
where we denote $F(\xi) := (2q-1) \log \xi - \frac{\xi^2}{2} - \frac{\pi \xi}{2} $. We compute that
\begin{equation}
    F'(\sqrt{2q}) = -\frac{\pi}{2} - \frac{1}{\sqrt{2q}} = O(1), \quad \sup_{\xi>0}F''(\xi) = \sup_{\xi>0}\left\{-\frac{2q-1}{\xi^2} - 1\right\} = -1.
\end{equation}
By Taylor's theorem, for every $\xi \geq 1$ we have $F(\xi)  \leq F(\sqrt{2q}) + O(1) - \frac{1}{4}(\xi - \sqrt{2q})^2$ and therefore
\begin{equation}
    \log\left(\int_1^\infty e^{F(\xi)} d\xi \right) \leq F(\sqrt{2q}) + O(1).
\end{equation}
For the lower bound, we note that
\begin{equation}
    \sup_{\xi \in [\sqrt{2q}, \sqrt{2q}+1]}\left\{F(\sqrt{2q}) - F(\xi)\right\} = O(1)
\end{equation}
and therefore that
\begin{equation}
    \log\left(\int_1^\infty e^{F(\xi)} d\xi \right) \geq   \log\left(\int_{\sqrt{2q}}^{\sqrt{2q}+1} e^{F(\xi)} d\xi \right) \geq F(\sqrt{2q}) - O(1).
\end{equation}
Hence,
(\ref{eqn:proofappendixasymptoticsfortanhhermitecoefficients}) follows from the computation
\begin{align}
    F(\sqrt{2q}) = (2q-1) \log(\sqrt{2q}) - q - \frac{\pi}{2}\sqrt{2q} 
    = q \log(2q) - q - \frac{\pi}{\sqrt{2}}\sqrt{q} - \frac{1}{2}\log(2q).
\end{align}

\end{proof}

We can do the same for other activations of interest, such as $\sin$:

\begin{lemma}
\label{lemma:appendixtsinhermitescaling}
    With $\phi = \sin$, as $d \to \infty$ we have that
    \begin{equation}
        \log \mathcal{D}_{\sin}(c) = \sqrt{d}+ o(\sqrt{d})
    \end{equation}
    uniformly for $c \in [-1, 1]$.
\end{lemma}
\begin{proof}
    First, we use the fact that
    \begin{equation}
        \mathbb{E}_{\xi \sim \mathcal{N}(0, 1)}[e^{it\xi} \operatorname{He}_j(\xi)] = (it)^j e^{-t^2/2}
    \end{equation}
    and take imaginary parts at $t=1$ to see that
    \begin{equation}
        \mathbb{E}_{\xi \sim \mathcal{N}(0, 1)}[\sin(\xi)\operatorname{He}_j(\xi)] = e^{-1/2} \operatorname{Im}(i^j).
    \end{equation}
    Hence, the Hermite coefficients satisfy
    \begin{equation}
        |\hat{\phi}_j|^2 = \begin{cases}
            0 & j \text{ even}
            \\ 1/e & j \text{ odd}
        \end{cases}
    \end{equation}
    and so only even $m = n + 2\ell$ contribute (and therefore only even $n$).

    Now, for $n \geq 2$ we denote
    \begin{equation}
    \label{eqn:proofappendixsinhermiteboundsdefinitionofTn}
        T_{n,d} := N_{n,d}\left(\frac{n\Gamma(d/2)}{2^n \Gamma(n+d/2)}\right)^{1/2}.
    \end{equation}
    For $m = n + 2\ell$ we know that
    \begin{equation}
        \frac{\frac{m\Gamma(d/2)}{2^m\ell! \Gamma(n+\ell+d/2)}}{\frac{n\Gamma(d/2)}{2^n\Gamma(n+d/2)}} = \frac{m}{n} \frac{1}{4^\ell \ell!} \frac{\Gamma(n+d/2)}{\Gamma(n+\ell+d/2)} \leq \frac{m}{n} \frac{1}{4^\ell \ell!}  \left(\frac{2}{d}\right)^\ell.
    \end{equation}
    In particular, the sum over $\ell$ becomes
    \begin{equation}
        \sum_{\ell = 0}^\infty \frac{(n+2\ell)\Gamma(d/2)}{2^{n+2\ell}\ell! \Gamma(n+\ell+d/2)} = (1 + O(d^{-1})) \frac{n\Gamma(d/2)}{2^n\Gamma(n+d/2)},
    \end{equation}
    i.e. it is dominated by the $\ell = 0$ term. Moreover, we have
    \begin{equation}
        0 \leq c^2 + (1-c^2) \frac{n(n+d-2)}{(n+2\ell)^2(d-1)} \leq 1
    \end{equation}
    while at $\ell =0$ we have
    \begin{equation}
        c^2 + (1-c^2) \frac{n+d-2}{n(d-1)} \geq \frac{1}{n}.
    \end{equation}
    Lastly, the $n=0$ term has $\frac{\Gamma(d/2)}{ \Gamma(d/2 + 1)} \asymp d^{-1}$ and so it contributes $O(d^{-1/2})$ uniformly in $c$ after the square root.
    Therefore, uniformly in $c$ we have
    \begin{equation}
        e^{-1/2} \sum_{\substack{n \geq 2 \\ n \text{ even}}} \frac{T_{n,d}}{\sqrt{n}} \leq \mathcal{D}_{\sin}(c) \leq O(d^{-1/2}) + e^{-1/2} (1 + O(d^{-1})) \sum_{\substack{n \geq 2 \\ n \text{ even}}} T_{n,d}.
    \end{equation}
    For $n \leq d^{2/3}$ we know that
    \begin{equation}
        \log N_{n,d} = n \log d - \log(n!) + O\left(\frac{n^2}{d}+1\right)
    \end{equation}
    and that
    \begin{equation}
        \frac{1}{2} \log \left(\frac{n\Gamma(d/2)}{2^n\Gamma(n+d/2)}\right) = -\frac{n}{2}\log d + \frac{1}{2}\log n + O\left(\frac{n^2}{d}\right).
    \end{equation}
    Combining these two and using Stirling,
    \begin{align}
        \log T_{n,d} &= n\left(1 - \log \frac{n}{\sqrt{d}}\right) + O\left(\frac{n^2}{d }+1\right) \\
        &= n\left(1 - \log \frac{n}{\sqrt{d}}\right) + o(\sqrt{d})
        \label{eqn:proofappendixsinhermitescalingoflogTnd}
    \end{align}
    uniformly over $n\leq d^{2/3}$. 

    For the upper bound, we split
    \begin{align}
        \sum_{\substack{n \geq 2 \\ n \text{ even}}} T_{n,d} = \sum_{\substack{2 \leq n \leq d^{2/3} \\ n \text{ even}}} T_{n,d} + \sum_{\substack{n > d^{2/3} \\ n \text{ even}}} T_{n,d}.
    \end{align}
    In the first sum, we use that $x (1-\log x) \leq 1$ with $x = n/\sqrt{d}$ to get that $T_{n,d} \leq \exp(\sqrt{d} + o(\sqrt{d}))$ for $n \leq d^{2/3}$. Since there are only $d^{2/3}$ many terms, we have
    \begin{equation}
         \sum_{\substack{2 \leq n \leq d^{2/3} \\ n \text{ even}}} T_{n,d} \leq  \exp(\sqrt{d} + o(\sqrt{d})).
    \end{equation}
    For the second sum, we use the definition (\ref{eqn:proofappendixsinhermiteboundsdefinitionofTn}) of $T_{n,d}$ to see that
    \begin{equation}
        \frac{T_{n+1,d}}{T_{n,d}} = \frac{(n+d-2) \sqrt{2n+d}}{(2n+d-2) \sqrt{n(n+1)}}.
    \end{equation}
    If $d^{2/3} < n \leq d$ then this is $\leq \frac{\sqrt{3d}}{n} \leq \sqrt{3}d^{-1/6}$, and if $n > d$ then this is $\leq \sqrt{\frac{3}{n}} \leq \sqrt{3}d^{-1/3}$. In any case, for sufficiently large $d$ we have $\frac{T_{n+1,d}}{T_{n,d}} \leq \frac{1}{2}$ for $n > d^{2/3}$, and so this sum is geometric. Plugging these in proves the upper bound.
    For the lower bound, let $n$ be the nearest even integer to $\sqrt{d}$ and use (\ref{eqn:proofappendixsinhermitescalingoflogTnd}) to find that
    \begin{equation}
        \log T_{n,d} = \sqrt{d} + o(\sqrt{d}).
    \end{equation}
\end{proof}

These results describe a superpolynomial separation between the Barron and rotation-equivariant Stein transport metrics along single-index flows in the perturbatively small regime.
To summarize, we have shown that if $\nu_\epsilon := \operatorname{Law}(X_\epsilon)$ for the time-invariant single-index flow
\begin{equation}
    \dot{X}_t = u \phi(\la X_t, \theta\ra), \quad X_0 \sim \gamma_d,
\end{equation}
then for any $\phi \in \{\operatorname{ReLU}, \tanh, \sin\}$ and any rotation-equivariant $k \in \mathcal{K}_{\operatorname{rot}}$ satisfying Assumption \ref{assumption:kernel} with $\int_{\r^d} \operatorname{tr}k(x, x) d\gamma_d(x) \in (0, \infty)$ we apply these results along with Lemma \ref{lemma:appendixbarronfirstorderlowerbound} to find that
\begin{equation}
\label{eqn:appendixperturbativeseparationsummarystatement}
    \sqrt{\int_{\r^d} \operatorname{tr}k(x, x) d\gamma_d(x)} \cdot \liminf_{\epsilon \downarrow 0} \frac{D_{\h_k}(\gamma_d, \nu_\epsilon)}{D_{\b_\phi}(\gamma_d, \nu_\epsilon)} \gtrsim \begin{cases}
        +\infty & \phi = \operatorname{ReLU}
        \\ 
        d^{\frac{d}{4}(1-o(1))} & \phi = \tanh,\\
        e^{\sqrt{d}(1-o(1))} & \phi = \sin,
    \end{cases}
\end{equation}
if $d \geq 4$ and $\la u, \theta\ra \neq 0$, and that these scalings are sharp in the sense that there are rotation-equivariant kernels matching them.
We emphasize that in this perturbative setting we are effectively constraining transports to use time-independent drifts, thereby linearizing the problem and reducing to control of the RKHS norms of single-index functions. 

This result makes it tempting to claim that ``for small $\epsilon$, there is a superpolynomial-in-$d$ separation between rotation-equivariant $D_{\h_k}$ and Barron $D_{\b_\phi}$ in terms of the representation of $\epsilon$-length single-index flows". By (\ref{eqn:appendixperturbativeseparationsummarystatement}) this is true as $\epsilon \downarrow 0$ with $d$ fixed. However, this has the potential to be misleading: the superpolynomial separation may only kick in when $\epsilon$ is smaller than any $\frac{1}{\operatorname{poly}(d)}$.

In fact, it turns out that this is the case. More precisely, the next result shows that \begin{equation}
    \epsilon \gtrsim \frac{1}{\operatorname{poly}(d)} \implies \sqrt{\int_{\r^d} \operatorname{tr}k(x, x) d\gamma_d(x)} \cdot \frac{D_{\h_k}(\gamma_d, \nu_\epsilon)}{D_{\b_\phi}(\gamma_d, \nu_\epsilon)} \lesssim \operatorname{poly}(d)
\end{equation} 
for some rotation-equivariant $k \in \mathcal{K}_{\operatorname{rot}}$ satisfying Assumption \ref{assumption:kernel}. 
Put differently, we can construct an alternate flow which matches the endpoint map of a time-invariant single-index flow by paying a one-time cost which is $\operatorname{poly}(d)$ and independent of $\epsilon$. We demonstrate this mechanism in the simplest setting of shears, i.e. single-index flows with $\la u, \theta\ra = 0$:

\begin{proposition}[Nonperturbative shortcuts in rotation-equivariant RKHS]
\label{prop:appendixnonperturbativefeaturelearning}
    Let $d \geq 2$ and $\phi \in C^\infty_b(\r)$. 
    Fix $Q, R \in (0, \infty)$ and denote
    \begin{equation}
        \beta := 1 + \log(1+Q) + d\log(d+2+R+Q\|\phi\|_\infty) + \log(1 + \|\phi\|_{C^{\lceil d/2\rceil}}). 
    \end{equation}
    For every $\rho \in \p(\r^d)$ we also denote its $L^d$-radius by
    \begin{equation}
        \operatorname{rad}_d(\rho) := \left(\int_{\r^d}\|x\|^d d\rho(x)\right)^{1/d}.
    \end{equation}
    Then, there exists a smooth rotation-equivariant positive kernel $k_{d,\phi, Q, R}$ with the following property:
    
    For every $u, \theta \in \mathbb{S}^{d-1}$ with $\la u, \theta\ra = 0$, every $|q| \leq Q$, and every $\rho_0 \in \p(\r^d)$ satisfying $\operatorname{rad}_d(\rho_0) \leq R$, there is a family of drifts $(v_t)_{t \in [0, 1]} \subset \h_{k_{d,\phi, Q, R}}$ s.t. the following hold:
    \begin{enumerate}
        \item[(i)] Each $v_t$ is globally-Lipschitz with
        \begin{equation}
            \sup_{t \in [0, 1]} \operatorname{Lip}(v_t) \leq 12 \beta.
        \end{equation}
        These drifts are in three phases, i.e. $t \mapsto v_t$ is piecewise-constant with three pieces. 
        \item[(ii)] If $(\Phi_t)_{t \in [0, 1]}$ denote the flow maps generated by $(v_t)_{t \in [0, 1]}$, then
        \begin{equation}
            \Phi_1(x) = x + qu\phi(\la x, \theta\ra).
        \end{equation}
        \item[(iii)] If $\rho_t = (\Phi_t)_\# \rho_0$ denote the marginals along the flow (and so $(v_t, \rho_t)_{t \in [0, 1]}$ satisfy the continuity equation), then
        \begin{equation}
            \sup_{t \in [0, 1]} \operatorname{rad}_d(\rho_t) \leq R + |q| \|\phi\|_\infty.
        \end{equation}
        \item[(iv)] There is the trace bound along the flow
        \begin{equation}
            \sup_{t \in [0, 1]} \int_{\r^d} \operatorname{tr} k_{d,\phi, Q, R}(x, x) d\rho_t(x) \leq 1 + d(R + |q| \|\phi\|_\infty)^2.
        \end{equation}
        \item[(v)] The RKHS length of this transport satisfies
        \begin{equation}
            \int_{0}^1 \|v_t\|_{\h_{k_{d,\phi, Q, R}}} dt \leq 8\beta.
        \end{equation}
    \end{enumerate}
    In particular, if  $\|\phi\|_\infty, \log (1+\|\phi\|_{C^{\lceil d/2\rceil}}) \lesssim \operatorname{poly}(d)$ then there is a smooth rotation-equivariant $k_{d,\phi} \in \mathcal{K}_{\operatorname{rot}}$ s.t. for any $T \leq d^{10000}$, any $\rho_0 \in \p(\r^d)$ with $\operatorname{rad}_d(\rho_0) \leq d^{10000}$, and any single-index shear
    \begin{equation}
        \nu_T := \operatorname{Law}(X_T), \quad \dot{X}_t = u \phi(\la X_t, \theta\ra), \quad X_0 \sim \rho_0, \quad \la u, \theta \ra = 0,
    \end{equation}
there exists a transport $(v_t, \rho_t)_{t \in [0, 1]}$ from $\rho_0$ to $\nu_T$ using three phases of $\operatorname{poly}(d)$-Lipschitz drifts with an RKHS cost of
    \begin{equation}
        \sup_{t \in [0, 1]} \sqrt{\int_{\r^d} \operatorname{tr}k_{d,\phi} (x, x) d\rho_t(x)} \cdot \int_0^1 \|v_t\|_{\h_{k_{d,\phi}}} dt \lesssim \operatorname{poly}(d).
    \end{equation}
\end{proposition}
\begin{proof}
For notation, we define
\begin{equation}
    m := \left \lceil \frac{d}{2}\right \rceil, \quad \|\phi\|_{C^m} := \max_{0 \leq j \leq m}\|\phi^{(j)}\|_\infty, \quad C_d := \left(1 + \frac{d(d-1)}{2}m^2\right)^m \leq d^{4d}.
\end{equation}
Let $H^{d/2}(\mathbb{S}^{d-1})$ be the spectral Sobolev space of index $d/2$ w.r.t. the normalized uniform measure $\sigma$ on $\mathbb{S}^{d-1}$. More precisely, if $\{Y_{n,j}:\; 1 \leq j \leq N_{n,d},\; n \in \n \cup \{0\}\}$ is the orthonormal basis of $L^2(\sigma)$ composed of spherical harmonics (with $N_{n,d} = \dim \mathbb{Y}_n^d$, see (\ref{eqn:proofsphericalharmonicdimscaling})), then
\begin{equation}
    -\Delta_{\mathbb{S}^{d-1}}Y_{n,j} = \lambda_n  Y_{n,j}, \quad \lambda_n := n(n+d-2)
\end{equation}
and we may identify the Sobolev norm by
\begin{equation}
    \|f\|_{H^{d/2}(\mathbb{S}^{d-1})}^2 = \sum_{n = 0}^\infty \sum_{j=1}^{N_{n,d}}(1+\lambda_n)^{d/2} |\la f, Y_{n,j}\ra_{L^2(\sigma)}|^2.
\end{equation}
By the addition formula \cite[Theorem 2.9]{atkinson2012spherical} we know $\sum_{j=1}^{N_{n,d}}|Y_{n,j}(\eta)|^2 = N_{n,d}$ for every $\eta \in \mathbb{S}^{d-1}$, and so the squared norm of point evaluation $f \mapsto f(\eta)$ on $H^{d/2}(\mathbb{S}^{d-1})$ is
\begin{equation}
    E_d = \sum_{n=0}^{\infty}\frac{N_{n,d}}{(1+\lambda_n)^{d/2}} \leq 1 + \sum_{n = 1}^\infty \frac{2d^{d-2}n^{d-2}}{n^{d}} = 1 + \frac{\pi^2}{3}d^{d-2} \leq 2d^d,
\end{equation}
where we used that $N_{n,d} \leq 2(n+d-1)^{d-2}\leq 2d^{d-2}n^{d-2}$ and $1+\lambda_n \geq n^2$ for $n \geq 1$ and $d \geq 2$. 

Now, for each $x \in \r^d$ we define
\begin{equation}
    \Psi_x(\omega) := \phi(\la x, \omega\ra) \quad (\forall \omega \in \mathbb{S}^{d-1})
\end{equation}
and we claim the Sobolev bound
\begin{equation}
\label{eqn:proofappendixnonperturbativesobolevbound}
    \|\Psi_x\|_{H^{d/2}(\mathbb{S}^{d-1})}^2 \leq C_d (1+\|\phi\|_{C^m})^2 (1+\|x\|)^d.
\end{equation}
To show (\ref{eqn:proofappendixnonperturbativesobolevbound}), we decompose $\Delta_{\mathbb{S}^{d-1}} = \sum_{1 \leq i < j \leq d} X_{ij}^2$ for the rotation fields $X_{ij} = \omega_i \partial_{\omega_j} - \omega_j \partial_{\omega_i}$. Since $\Delta_{\mathbb{S}^{d-1}}$ commutes with every $X_{ij}$, we can expand $(\1 - \Delta_{\mathbb{S}^{d-1}})^m$ to get
\begin{equation}
    \|f\|_{H^m(\mathbb{S}^{d-1})}^2 = \|f\|_{L^2(\sigma)}^2 + \sum_{a = 1}^m {m \choose a} \sum_{1 \leq i_1 < j_1 \leq d} \cdots \sum_{1 \leq i_a < j_a \leq d} \|X_{i_1,j_1} \cdots X_{i_a,j_a}f\|_{L^2(\sigma)}^2.
\end{equation}
We have $\|\Psi_x\|_{L^2(\sigma)}^2 \leq \|\phi\|_\infty^2$ and by the chain rule we know that
\begin{equation}
    \|X_{i_1,j_1} \cdots X_{i_a,j_a}\Psi_x\|_{\infty} \leq a^a (1+\|\phi\|_{C^m}) (1+\|x\|)^a,
\end{equation}
since there are at most $a^a$ partitions of $\{1,\ldots, a\}$ and we bound the factors $\la x, X_{i_1,j_1}\cdots X_{i_a,j_a}\omega\ra$ by $\|x\|$. Since there are $(d(d-1)/2)^a$ such terms, we find that
\begin{align}
    \|\Psi_x\|_{H^m(\mathbb{S}^{d-1})}^2 &\leq (1+\|\phi\|_{C^m})^2 \left(1 + \sum_{a=1}^m {m \choose a}  \left(\frac{d(d-1)}{2}\right)^aa^{2a} (1+\|x\|)^{2a}\right) \\
    &\leq (1+\|\phi\|_{C^m})^2 (1+\|x\|)^{2m} \left(1 + \sum_{a=1}^m {m \choose a}  \left(m^2\frac{d(d-1)}{2}\right)^a \right) \\
    &=  (1+\|\phi\|_{C^m})^2 (1+\|x\|)^{2m}\left(1 + m^2\frac{d(d-1)}{2}\right)^m\\
    &= C_d (1+\|\phi\|_{C^m})^2 (1+\|x\|)^{2m}.
\end{align}
Denoting $\nu := \frac{d}{2m} \in (0, 1]$ and using the interpolation inequality
\begin{equation}
    \|f\|_{H^{d/2}(\mathbb{S}^{d-1})} \leq \|f\|_{L^2(\sigma)}^{1-\nu} \|f\|_{H^m(\mathbb{S}^{d-1})}^{\nu},
\end{equation}
we obtain
\begin{align}
    \|\Psi_x\|_{H^{d/2}(\mathbb{S}^{d-1})}^2 &\leq (1+\|\phi\|_{C^m})^{2(1-\nu)} \left[C_d(1+\|\phi\|_{C^m})^2 (1+\|x\|)^{2m} \right]^\nu \\
    &= C_d^{\nu} (1+\|\phi\|_{C^m})^2 (1+\|x\|)^{2m\nu}.
\end{align}
Since $2m \nu = d$ and $C_d \geq 1$, we get (\ref{eqn:proofappendixnonperturbativesobolevbound}).

Using these ingredients, we define the scalar kernel $\kappa_\phi: \r^d \times \r^d \to \r$ by
\begin{equation}
    \kappa_\phi(x, y) := \la \Psi_x, \Psi_y\ra_{H^{d/2}(\mathbb{S}^{d-1})}.
\end{equation}
This is positive semidefinite and since $\Psi_{Ux}(\omega) = \Psi_x(U^\top \omega)$ it is rotation-invariant. Also, since $\partial_x^\alpha \Psi_x(\omega) = \omega^\alpha \phi^{(|\alpha|)}(\la x, \omega\ra)$ the same Sobolev estimate shows $x \mapsto \Psi_x$ is smooth as an $H^{d/2}$-valued map; thus, $\kappa_\phi$ is smooth.
For any $\eta \in \mathbb{S}^{d-1}$, we know that
\begin{equation}
    \phi(\la x, \eta\ra) = \Psi_x(\eta) = \la \Psi_x, e_\eta\ra_{H^{d/2}(\mathbb{S}^{d-1})},
\end{equation}
and so by the feature map construction of $\kappa_\phi$ we have
\begin{equation}
    \|\phi(\la \cdot, \eta\ra)\|_{\h_{\kappa_\phi}}^2 \leq \|e_\eta\|^2_{H^{d/2}(\mathbb{S}^{d-1})} = E_d \leq 2d^d.
\end{equation}
Now, we set $M := (1+\|\phi\|_{C^m})^2 \left(1 + 2d^{d+1}C_d(1+R+Q\|\phi\|_\infty)^d\right)$ and define the kernel
\begin{equation}
    k_{d, \phi, Q, R}(x, y) := \la x, y\ra \1_d + \frac{2d^d}{1+M} \kappa_\phi(x, y) \1_d.
\end{equation}
This kernel is smooth, positive semidefinite, and rotation-equivariant. Furthermore, by construction we know $x \mapsto u \la x, u\ra$ has $\h_{k_{d,\phi,Q,R}}$ norm at most 1 while $x \mapsto u \phi(\la x, \theta\ra)$ has $\h_{k_{d,\phi,Q,R}}$ norm $\leq \sqrt{1+M}$. Hence, if we define $r := \max\left\{1, \; \frac{|q|\sqrt{1+M}}{2}\right\}$ and
\begin{equation}
\label{eqn:proofappendixnonperturbativefeaturelearningdefinitionofdrifts}
    v_t(x) := \begin{cases}
        -3 \log r u \la x, u\ra & 0 \leq t < \frac{1}{3} \\
        \frac{3q}{r} u \phi(\la x, \theta\ra) & \frac{1}{3} \leq t < \frac{2}{3},\\
        3\log r u \la x, u\ra & \frac{2}{3} \leq t \leq 1.
    \end{cases}
\end{equation}
then it holds that
\begin{equation}
\label{eqn:proofappendixnonperturbativefeaturelearningRKHSbound}
    \int_0^1 \|v_t\|_{\h_{k_{d, \phi, Q, R}}} dt \leq 2\log r + \frac{|q| \sqrt{1+M}}{r} \leq 2 + 2 \log(1+|q|\sqrt{1+M}),
\end{equation}
where the second inequality follows from definition of $r$. Next, since $\sqrt{1+M} \geq \|\phi\|_{C^m} \geq \|\phi'\|_\infty$ we know that $\operatorname{Lip}(\frac{3q}{r}u\phi(\la \cdot, \theta\ra))\leq 6$ and so
\begin{equation}
\label{eqn:proofappendixnonperturbativefeaturelearningLipschitzbound}
    \sup_{t \in [0, 1]} \operatorname{Lip}(v_t) \leq 6 + 3 \log(1 + |q| \sqrt{1+M}).
\end{equation}
By definition of $M$ and $C_d \leq d^{4d}$ we know that
\begin{equation}
    \sqrt{1+M} \leq 2(1+\|\phi\|_{C^m}) d^{(5d+1)/2}(1 + R + Q\|\phi\|_\infty)^{d/2}
\end{equation}
from which we can bound
\begin{align}
    \log(1 + |q|\sqrt{1+M}) &\leq 1 + \log(1+|q|) + \log(1 + \|\phi\|_{C^m}) + 4d\log(d+2+R+Q\|\phi\|_\infty).
\end{align}
Combining this with (\ref{eqn:proofappendixnonperturbativefeaturelearningRKHSbound}) and (\ref{eqn:proofappendixnonperturbativefeaturelearningLipschitzbound}) yields (v) and (i), respectively. 
Looking back at (\ref{eqn:proofappendixnonperturbativefeaturelearningdefinitionofdrifts}), we see that the first phase multiplies the $u$ coordinate by $1/r$. The second phase adds $\frac{q}{r} u \phi(\la x, \theta\ra)$ and preserves $\la x, \theta\ra$ since $u \perp \theta$. Lastly, the third phase multiplies the resulting $u$ coordinate by $r$, which proves (ii).

It remains to prove the moment and trace estimates. Write
\begin{equation}
    x = x_\perp + \la x, u\ra u, \quad \la x_\perp, u\ra = 0.
\end{equation}
During the first phase, the absolute value of the $u$ coordinate decreases while $x_\perp$ is unchanged. During the middle phase, the displacement has magnitude at most $\frac{|q|}{r}\|\phi\|_\infty$. At the beginning of the last phase, the $u$ coordinate equals \begin{equation}
    \frac{\la x, u\ra + q \phi(\la x, \theta\ra)}{r}.
\end{equation}
During the final phase this coordinate is multiplied by a factor which starts at 1 and ends at $r$, and so its magnitude never exceeds $|\la x, u\ra + q \phi(\la x, \theta\ra)|$. Thus,
\begin{equation}
    \|\Phi_t(x)\| \leq \|x\| + |q|\|\phi\|_\infty \quad (\forall t \in [0, 1],\; \forall x \in \r^d).
\end{equation}
Applying the triangle inequality in $L^d(\rho_0)$ implies $\operatorname{rad}_d(\rho_t) \leq R+|q|\|\phi\|_\infty$, proving (iii). To finish, for the linear part of the kernel we know
\begin{equation}
\label{eqn:proofappendixnonperturbativetracebound1}
    d \int_{\r^d} \|x\|^2 d\rho_t(x) \leq d \operatorname{rad}_d(\rho_t)^2 \leq d(R+|q|\|\phi\|_\infty)^2.
\end{equation}
For the nonlinear part, we have that
\begin{align}
    \int_{\r^d} \kappa_\phi(x, x) d\rho_t(x) &= \int_{\r^d} \|\Psi_x\|^2_{H^{d/2}(\mathbb{S}^{d-1})} \\&\leq C_d (1+\|\phi\|_{C^m})^2 \int_{\r^d} (1+\|x\|)^d d\rho_t(x) \\
    &\leq C_d (1+\|\phi\|_{C^m})^2 (1+\operatorname{rad}_d(\rho_t))^d \\
    &\leq C_d (1+\|\phi\|_{C^m})^2 (1+R+Q\|\phi\|_\infty)^d
\end{align}
which implies that 
\begin{align}
\label{eqn:proofappendixnonperturbativetracebound2}
    \frac{2d^d}{1+M}\int_{\r^d}\operatorname{tr} (\kappa_\phi(x, x) \1_d) d\rho_t(x) \leq \frac{2d^{d+1}C_d (1+\|\phi\|_{C^m})^2 (1+R+Q\|\phi\|_\infty)^d}{1+M} \leq 1
\end{align}
by definition of $M$. Combining (\ref{eqn:proofappendixnonperturbativetracebound1}) and (\ref{eqn:proofappendixnonperturbativetracebound2}) yields (iv).

The final ``in particular" follows by taking $k_{d, \phi} := k_{d,\phi, d^{10000, d^{10000}}}$, observing that $\frac{d}{dt}\la X_t, \theta\ra = 0$ and so $X_T = X_0 + Tu\phi(\la X_0, \theta\ra)$, and applying the result with $q=T$.
\end{proof}

\begin{remark}
    This shortcut construction uses the same compress-transport-expand strategy that the optimal transports used in the Gaussian endpoints and linear drift case, see Remark \ref{remark:steinbarronlineargaussiantogaussianmeanmovement}. In that setting it was optimal up to universal constants, but in this nonlinear setting we expect there to be more efficient shortcuts.
\end{remark}

\subsection{Reverse OU (sub)optimality for single-index flows and tilts}
\label{subsec:proofofreverseOUsingleindexflowsandtilts}

We first show the flow lower bound. The Barron norm will turn out to be $+\infty$ for the examples we consider, but for now we define an abstract quantity which captures this lower bound.

\begin{lemma}[Single-index flows]
\label{lemma:appendixreverseOUsingleindexflows}
    Let $\phi: \r \to \r$ be globally-Lipschitz, nonzero, and satisfying Assumption \ref{assumption:activationfunciton}. Let $u, \theta \in \mathbb{S}^{d-1}$ and $\sigma \in (0, \infty)$, and for each $\epsilon > 0$ define
    \begin{equation}
        \pi_\epsilon^{\operatorname{flow}} := \operatorname{Law}(X_1), \quad \dot{X}_t = \epsilon\sigma u \phi(\la X_t, \theta\ra), \quad X_0 \sim \gamma_d.
    \end{equation}
    Then, it holds that
    \begin{equation}
        \liminf_{\epsilon \downarrow 0} \frac{\mathcal{E}_{\b_\phi}\left({
        \pi_\epsilon^{\operatorname{flow}}
        }\right)}{D_{\b_\phi}(\gamma_d, \pi_\epsilon^{\operatorname{flow}})} \geq \frac{\mathfrak{F}_\phi^{\operatorname{flow}}(c)}{2}
    \end{equation}
    where $c := \la u, \theta\ra$ and we define the quantity
    \begin{equation}
        \label{eqn:proofmathfrakflow}\mathfrak{F}_\phi^{\operatorname{flow}}(c) := \int_0^1 \left\|u Q_r(\la \cdot, \theta\ra)+ \theta \la \cdot, u\ra Q_r'(\la \cdot, \theta\ra) - c \theta Q_r''(\la \cdot, \theta\ra)\right\|_{\b_\phi} dr
    \end{equation}
    for the function 
    \begin{equation}
    \label{eqn:proofdefofQr}
        Q_r(t) := \mathbb{E}_{\xi \sim \mathcal{N}(0, 1)}\left[\phi\left(rt + \sqrt{1-r^2}\xi\right)\right].
    \end{equation}
\end{lemma}
\begin{remark}
    Note that by rotational invariance, the expression in (\ref{eqn:proofmathfrakflow}) depends only on $c = \la u, \theta\ra$.
\end{remark}
\begin{proof}
   Let us denote
   \begin{equation}
       w(x) := \sigma u\phi(\la x, \theta\ra), \quad f_\epsilon := \frac{\pi_\epsilon^{\operatorname{flow}}}{\gamma_d}, \quad c := \la u, \theta\ra,
   \end{equation}
   where we know that $\pi_\epsilon^{\operatorname{flow}}$ is absolutely continuous since the flow generated by $w$ is bi-Lipschitz (due to Lipschitzness of $w$). By the continuity equation, for each $\varphi \in C^\infty_c(\r^d)$ we have
   \begin{equation}
       \left.\frac{d}{d\epsilon}\right|_{\epsilon = 0} \int_{\r^d} \varphi d\pi_\epsilon^{\operatorname{flow}} = \int_{\r^d} \la \nabla \varphi, w\ra d\gamma_d = \int_{\r^d} \varphi h d\gamma_d,
   \end{equation}
   where we use Gaussian integration by parts and denote the function 
   \begin{equation}
   \label{eqn:proofappendixdefofhforflowlowerbound}
       h(x) := \la x, w(x)\ra - \operatorname{div}w(x) = \sigma \left(\la x, u\ra \phi(\la x, \theta\ra) - c\phi'(\la x, \theta\ra)\right).
   \end{equation}
   Hence, $f_\epsilon = 1 + \epsilon h + o(\epsilon)$ in the sense of distributions. For each fixed $\tau > 0$, the linear growth of $w$ and OU regularization\footnote{Letting $K_\tau$ denote the Mehler kernel (which is smooth with Gaussian decay of all its derivatives) and $T_\epsilon$ the flow map for time $\epsilon$, since $P_\tau f_\epsilon = \int_{\r^d} K_\tau (\cdot,T_\epsilon(y)) d\gamma_d(y) $ and $T_\epsilon(y) = y + \epsilon w(y) + o(\epsilon)$, we may differentiate under the integral to get local uniform convergence of derivatives.} turn this into $P_\tau f_\epsilon = 1 + \epsilon P_\tau h + o(\epsilon)$ in $C^1_{\operatorname{loc}}$, or in other words
   \begin{equation}
       \lim_{\epsilon \downarrow 0}\frac{1}{\epsilon} \nabla \log P_\tau f_\epsilon = \nabla P_\tau h \quad \text{uniformly on compact sets}.
   \end{equation}
   By lower-semicontinuity of the Barron norm w.r.t. local uniform convergence (Lemma \ref{lemma:appendixbarronnormlscunderlocaluniformconvergence}) and Fatou's lemma, we find
   \begin{equation}
       \liminf_{\epsilon \downarrow 0} \frac{\mathcal{E}_{\b_\phi}(\pi_\epsilon^{\operatorname{flow}})}{\epsilon} = \liminf_{\epsilon \downarrow 0} \int_0^\infty \left\|\frac{1}{\epsilon} \nabla \log P_\tau f_\epsilon\right\|_{\b_\phi} d\tau \geq \int_0^\infty \|\nabla P_\tau h\|_{\b_\phi}d\tau.
   \end{equation}
   We now compute the right hand side. With $r = e^{-\tau}$, by (\ref{eqn:proofappendixdefofhforflowlowerbound}) and the Mehler formula we have
    \begin{align}
        (P_\tau h)(x) &= \mathbb{E}_{\xi \sim \gamma_d} \left[h\left(rx + \sqrt{1-r^2}\xi\right)\right] \\
        &= \sigma r \la x, u\ra \mathbb{E}\left[\phi\left(r \la x, \theta\ra + \sqrt{1-r^2}\la \xi, \theta\ra\right)\right] \\&\quad + \sigma \sqrt{1-r^2} \mathbb{E}\left[\la \xi, u\ra \phi\left(r\la x, \theta\ra + \sqrt{1-r^2}\la \xi, \theta\ra\right)\right]\\&\quad  - \sigma c \mathbb{E}\left[\phi'(r\la x, \theta\ra + \sqrt{1-r^2}\la \xi, \theta\ra)\right].
    \end{align}
    Now, by Gaussian integration by parts we know
    \begin{equation}
        \mathbb{E}[\la \xi, u\ra F(\la \xi, \theta\ra)] = \la u, \theta\ra \mathbb{E}[F'(\la \xi, \theta\ra)] \quad (\forall F \in W^{1,2}(\gamma_1)).
    \end{equation}
    We apply this with $F(z) = \phi(r \la x, \theta\ra + \sqrt{1-r^2}z)$ and $F'(z) = \sqrt{1-r^2}\phi'(r\la x, \theta\ra + \sqrt{1-r^2}z)$ to get that
    \begin{equation}
        \mathbb{E}\left[\la \xi, u\ra \phi\left(r\la x, \theta\ra + \sqrt{1-r^2}\la \xi, \theta\ra\right)\right] = c \sqrt{1-r^2} \mathbb{E}\left[\phi'(r\la x, \theta\ra + \sqrt{1-r^2}\la \xi, \theta\ra)\right],
    \end{equation}
    which we plug into our earlier expression to find
    \begin{align}
        \frac{1}{\sigma} (P_\tau h)(x) &= r \la x, u\ra Q_r(\la x, \theta\ra) -cr^2 \mathbb{E}\left[\phi'(r\la x, \theta\ra + \sqrt{1-r^2}\la \xi, \theta\ra)\right] \\
        &= r \la x, u\ra Q_r(\la x, \theta\ra) - cr Q_r'(\la x, \theta\ra).
    \end{align}
   Differentiating, we find that
   \begin{equation}
       (\nabla P_\tau h)(x) = \sigma r \left(u Q_r(\la x, \theta\ra) + \theta \la x, u\ra Q'_r(\la x, \theta\ra) - c\theta Q_r''(\la x, \theta\ra)\right)
   \end{equation}
   and therefore that
   \begin{equation}
       \int_0^\infty \|\nabla P_\tau h\|_{\b_\phi} d\tau = \sigma \mathfrak{F}_\phi^{\operatorname{flow}}(c).
   \end{equation}
    Since the flow defining $\pi_\epsilon^{\operatorname{flow}}$ is admissible we know that $D_{\b_\phi}(\gamma_d, \pi_\epsilon^{\operatorname{flow}}) \leq 2\sigma \epsilon$ (where the factor of 2 accounts for the weight $1 + \|\theta\|$ in the bounded-activation Barron definition), yielding the claim.
\end{proof}

Before moving on, we fill in the claim that the Barron norm is lower semicontinuous under locally uniform convergence. The proof uses a standard compactification trick:
\begin{lemma}
\label{lemma:appendixbarronnormlscunderlocaluniformconvergence}
Let $\phi: \r \to \r$ satisfy Assumption \ref{assumption:activationfunciton}, and 
    let $(w_n)_{n \in \n} \subset \b_\phi$ and $w \in C_{\operatorname{loc}}(\r^d; \r^d)$ be s.t. $w_n \to w$ uniformly on compact sets. Then,
    \begin{equation}
        \|w\|_{\b_\phi} \leq \liminf_{n \to \infty}\|w_n\|_{\b_\phi}.
    \end{equation}
\end{lemma}
\begin{proof}
   Set $\hat{\Theta} := \r^d \times \r$ to be the parameter space. If $\phi$ is bounded, let $\Theta := \hat{\Theta} \cup \{\infty\}$ be its one-point compactification and define
   \begin{equation}
       \psi_{a,b}(x) := \frac{\phi(\la a, x\ra + b)}{1 + \|a\| + |b|}, \quad \psi_\infty(x) := 0.
   \end{equation}
   The map $\theta \mapsto \psi_\theta$ is continuous from $\Theta$ into $C_{\operatorname{loc}}(\r^d)$: for $\theta \in \hat{\Theta}$ we get continuity at $\theta$ from continuity of $\phi$, while for continuity at $\theta = \infty$ we use that for each compact $K$ we have 
   \begin{equation}
       \sup_{x \in K}|\psi_{a,b}(x)| \leq \frac{\|\phi\|_\infty}{1 + \|a\| + |b|} \to 0
   \end{equation}
   as $\|a\| + |b| \to \infty$ . Similarly, if $\phi$ is positively $s$-homogeneous we define
   \begin{equation}
       \Theta := \{(a, b) \in \hat{\Theta}:\; \|a\| + |b| = 1\}, \quad \psi_{a, b}(x) := \phi(\la a, x\ra + b).
   \end{equation}
   This $\Theta$ is compact, and again $\theta \mapsto \psi_\theta$ is a continuous map from $\Theta$ into $C_{\operatorname{loc}}(\r^d)$. Hence, in either case the Barron norm admits the equivalent representation
   \begin{equation}
       \|u\|_{\b_\phi} = \inf\left\{\|\nu\|_{\operatorname{TV}}:\; u(x) = \int_\Theta \psi_\theta(x) d\nu(\theta)\right\} \quad (\forall u \in \b_\phi)
   \end{equation}
   with $\nu$ ranging over finite $\r^d$-valued measures on the compact Polish space $\Theta$.

   With this, let $L := \liminf_{n \to \infty}\|w_n\|_{\b_\phi}$ and assume without loss of generality that $L < \infty$. Passing to a subsequence, assume $\|w_n\|_{\b_\phi} \to L$ and choose $\nu_n$ s.t.
   \begin{equation}
       w_n(x) = \int_\Theta \psi_\theta(x) d\nu_n(\theta), \quad \|\nu_n\|_{\operatorname{TV}} \leq \|w_n\|_{\b_\phi} + \frac{1}{n}.
   \end{equation}
   Since $\Theta$ is compact and Polish, $C(\Theta; \r^d)$ is separable and $\mathcal{M}(\Theta; \r^d) = C(\Theta; \r^d)^*$ is the space of signed vector-valued measures equipped with the TV norm.
Since $(\nu_n)_{n \in \n}$ is bounded in $\mathcal{M}(\Theta; \r^d)$, by Banach-Alaoglu we may pass to a subsequence satisfying $\nu_n \rightharpoonup \nu$ weak-* for some $\nu \in \mathcal{M}(\Theta; \r^d)$. Since $\theta \mapsto \psi_\theta(x)$ is bounded continuous for every $x$, we know
   \begin{equation}
       w(x) = \lim_{n \to \infty}w_n(x) = \int_{\Theta} \psi_\theta(x) d\nu(\theta)
   \end{equation}
   and so $\nu$ represents $w$. Since $\|\nu\|_{\operatorname{TV}} \leq \liminf_{n \to \infty}\|\nu_n\|_{\operatorname{TV}} \leq L$ by lower-semicontinuity of the dual norm under weak-* convergence, the claim follows.
\end{proof}

Next, we show the tilt upper bound in terms of a second abstract quantity:

\begin{lemma}[Single-index tilts]
\label{lemma:appendixreverseOUsingleindextilts}
Let $\phi: \r \to \r$ be globally-Lipschitz, nonconstant, bounded below, and suppose it satisfies Assumption \ref{assumption:activationfunciton} and $\mathbb{E}_{\xi \sim \mathcal{N}(0, 1)}[\phi'(\xi)] \neq 0$.  Fix $\theta \in \mathbb{S}^{d-1}$ and define
\begin{equation}
    \pi_\epsilon^{\operatorname{tilt}} := (1 + \epsilon h_\phi) \gamma_d , \quad h_\phi(x) := {\phi}(\la x, \theta\ra) - \mathbb{E}_{\xi \sim \mathcal{N}(0, 1)}[\phi(\xi)]
\end{equation}
for $\epsilon > 0$ small enough that $1 + \epsilon h_\phi \geq 0$. Then, it holds that
\begin{equation}
    \limsup_{\epsilon \downarrow 0} \frac{\mathcal{E}_{\b_\phi}\left({\pi_\epsilon^{\operatorname{tilt}}}\right)}{D_{\b_\phi}(\gamma_d, \pi_\epsilon^{\operatorname{tilt}})} \leq \mathfrak{F}_\phi^{\operatorname{tilt}} \cdot \frac{\min\left\{\|\phi\|_\infty, \; \max\left\{|\phi(0)|, \; \operatorname{Lip}(\phi)\right\}\right\}}{\left|\mathbb{E}_{\xi \sim \mathcal{N}(0, 1)}[\phi'(\xi)]\right|}
\end{equation}
where $Q_r$ is as in (\ref{eqn:proofdefofQr}) and we define the quantity
\begin{equation}
    \label{eqn:proofmathfraktilt}\mathfrak{F}_\phi^{\operatorname{tilt}} := \limsup_{\epsilon \downarrow 0}\int_0^1 \left\|\theta \frac{Q_r'(\la \cdot, \theta\ra)}{r(1 + \epsilon Q_r(\la \cdot, \theta\ra) - \epsilon \mathbb{E}_{\xi \sim \mathcal{N}(0, 1)}[\phi(\xi)])}\right\|_{\b_\phi}dr.
\end{equation}
    
\end{lemma}
\begin{proof}
    Let us denote 
    \begin{equation}
        m_\phi := \mathbb{E}_{\xi \sim \mathcal{N}(0, 1)}[\phi(\xi)], \quad f_\epsilon := \frac{\pi_\epsilon^{\operatorname{tilt}}}{\gamma_d} = 1 + \epsilon h_\phi.
    \end{equation}
    For $r = e^{-\tau}$, by the Mehler formula we know
    \begin{equation}
        (P_\tau f_\epsilon)(x) = 1 + \epsilon (Q_r(\la x, \theta\ra) - m_\phi)
    \end{equation}
    and therefore that for $r \in (0, 1)$ we have
    \begin{equation}
        (\nabla \log P_\tau f_\epsilon)(x) = \epsilon \theta \frac{Q_r'(\la x, \theta\ra)}{1 + \epsilon Q_r(\la x, \theta\ra) - \epsilon m_\phi}.
    \end{equation}
    Using $d\tau = -dr / r$ and integrating,
    \begin{equation}
    \label{eqn:proofappendixinsgleindextiltSteinupperbound}
        \limsup_{\epsilon \downarrow 0} \frac{\mathcal{E}_{\b_\phi}\left({\pi_\epsilon^{\operatorname{tilt}}}\right)}{\epsilon} = \mathfrak{F}_\phi^{\operatorname{tilt}}.
    \end{equation}
    
    For the Barron transport distance lower bound, we use an argument similar to that of Lemma \ref{lemma:appendixwassersteinandKLcontrol} adapted for unbounded $\phi$. Denote the quantities
    \begin{equation}
        L := \operatorname{Lip}(\phi), \quad C_0 := \max\{|\phi(0)|, \; L\}, \quad M := \|\phi\|_\infty, \quad C_\phi := \min\{M, C_0\}
    \end{equation}
    with $M = +\infty$ for unbounded $\phi$. We claim that for every $w \in \b_\phi$ and every $\rho \in \p_1(\r^d)$ we have
    \begin{equation}
    \label{eqn:proofappendixW1estimatetoprove}
        \int_{\r^d}\|w(x)\| d\rho(x) \leq (C_0 + LW_1(\gamma_d, \rho)) \|w\|_{\b_\phi}.
    \end{equation}
    To prove this, note that since $\phi$ is $L$-Lipschitz then
    \begin{equation}
    \label{eqn:proofappendixinequalitycausephiisLlipschitz}
        \mathbb{E}_{\xi \sim \gamma_d} \left|\phi\left(\la a, \xi\ra + b\right)\right| \leq |\phi(0)| + L \mathbb{E}_{\xi \sim \gamma_d}|\la a, \xi\ra + b| \leq |\phi(0)| + L(\|a \| + |b|).
    \end{equation}
    Since $\phi(0) = 0$ for positively one-homogeneous $\phi$, if we define
    \begin{equation}
        \omega(a, b) := \begin{cases}
            1 + \|a\| + |b| & \phi \text{ bounded},
            \\ \|a\| + |b| & \phi \text{ positively one-homogeneous}
        \end{cases}
    \end{equation}
    then (\ref{eqn:proofappendixinequalitycausephiisLlipschitz}) implies that
    \begin{equation}
       \int_{\r^d}\left|\phi\left(\la a, x\ra + b\right)\right|d\gamma_d(x) \leq C_0 \omega(a, b).
    \end{equation}
    Since $x \mapsto |\phi(\la a, x\ra + b)|$ is $L\|a\|$-Lipschitz and $\|a\| \leq \omega(a, b)$, by Kantorovich-Rubinstein we have
    \begin{align}
        \int_{\r^d} |\phi(\la a, x\ra + b)|d\rho(x)  &\leq \int_{\r^d} |\phi(\la a, x\ra + b)|d\gamma_d(x) + L\|a\|W_1(\gamma_d, \rho) \\
        &\leq \left(C_0 + LW_1(\gamma_d, \rho)\right) \omega(a, b).
    \end{align}
    Integrating this over any vector measure representing $w$ and taking the infimum over such representations proves (\ref{eqn:proofappendixW1estimatetoprove}). 

    We now claim that 
    \begin{equation}
    \label{eqn:proofappendixinsgleindextiltDblowerbound}
        \liminf_{\epsilon \downarrow 0} \frac{D_{\b_\phi}(\gamma_d, \pi_\epsilon^{\operatorname{tilt}})}{\epsilon} \geq \frac{\left| \mathbb{E}_{\xi \sim \mathcal{N}(0, 1)} \phi'(\xi)\right|}{C_\phi},
    \end{equation}
    which would yield the main result after combining with (\ref{eqn:proofappendixinsgleindextiltSteinupperbound}).
   To prove this, let $\epsilon_n \downarrow 0$ be any sequence realizing the $\liminf$, let $(v_t^n, \rho_t^n)_{t \in [0, 1]}$ be any admissible transport from $\gamma_d$ to $\pi_{\epsilon_n}^{\operatorname{tilt}}$ with
   \begin{equation}
       \ell_n := \int_0^1 \|v_t^n\|_{\b_\phi}dt \leq D_{\b_\phi}(\gamma_d, \pi_{\epsilon_n}^{\operatorname{tilt}}) + \epsilon_n^2,
   \end{equation}
   and define
    \begin{equation}
        A_n(t) := \int_0^t \int_{\r^d} \|v_s^n(x)\| d\rho_s^n(x) ds.
    \end{equation}
    Assume that $\lim_{n \to \infty }D_{\b_\phi}(\gamma_d, \pi_{\epsilon_n}^{\operatorname{tilt}}) / \epsilon_n < \infty$  (since (\ref{eqn:proofappendixinsgleindextiltDblowerbound}) holds trivially otherwise), 
in which case $\ell_n \leq \sqrt{\epsilon_n}$ for all large enough $n$.
    By the standard continuity equation estimate (i.e. the length of an absolutely continuous curve in $(\p_1(\r^d), W_1)$ upper bounds the $W_1$ distance) we know
$
        W_1(\gamma_d, \rho_t^n) \leq A_n(t)
$. Applying (\ref{eqn:proofappendixW1estimatetoprove}) yields that
\begin{equation}
    A'_n(t) \leq (C_0 + LA_n(t)) \|v_t^n\|_{\b_\phi}
\end{equation}
for a.e. $t$, after which Grönwall's inequality implies
\begin{equation}
    A_n(1) \leq \frac{C_0}{L}\left(e^{L\ell_n} - 1\right),
\end{equation}
where $L > 0$ since $\phi$ is nonconstant. If $\phi$ is bounded, then we may instead use that $W_1(\gamma_d, \pi_{\epsilon_n}^{\operatorname{tilt}}) \leq M D_{\b_\phi}(\gamma_d, \pi_{\epsilon_n}^{\operatorname{tilt}}) \leq M\ell_n$ (see Proposition \ref{prop:wassersteinkllowerbound}). Hence, in either case we find
\begin{equation}
\label{eqn:proofappendixW1estimatetoprove234}
    W_1(\gamma_d, \pi_{\epsilon_n}^{\operatorname{tilt}}) \leq \ell_n \cdot \min\left\{M, \; C_0 \frac{e^{L\ell_n} -1}{L\ell_n}\right\}.
\end{equation}
On the other hand, applying Kantorovich-Rubinstein with the $1$-Lipschitz function $x \mapsto \la x, \theta\ra$ implies 
\begin{align}
W_1(\gamma_d, \pi_{\epsilon_n}^{\operatorname{tilt}}) &\geq \left|\int_{\r^d} \la x, \theta \ra d\pi_{\epsilon_n}^{\operatorname{tilt}}(x) - \int_{\r^d} \la x, \theta \ra d\gamma_d(x)\right| = \left|\epsilon_n\int_{\r^d} \la x, \theta\ra h_\phi(x) d\gamma_d(x)\right| \\
&= \left|\epsilon_n \mathbb{E}_{\xi \sim \mathcal{N}(0, 1)} \left[\xi (\phi(\xi) - m_\phi)\right]\right| = \left|\epsilon_n \mathbb{E}_{\xi \sim \mathcal{N}(0, 1)} \left[\xi \phi(\xi) \right]\right| \\
&= \epsilon_n\left| \mathbb{E}_{\xi \sim \mathcal{N}(0, 1)} \phi'(\xi)\right|
\end{align}
by Gaussian integration by parts. Combining this with (\ref{eqn:proofappendixW1estimatetoprove234}) yields
\begin{equation}
    \frac{\ell_n}{\epsilon_n} \geq \frac{\left| \mathbb{E}_{\xi \sim \mathcal{N}(0, 1)} \phi'(\xi)\right|}{\min\left\{M, \; C_0 \frac{e^{L\ell_n} - 1}{L\ell_n }\right\}}.
\end{equation}
Taking an infimum over transports from $\gamma_d$ to $\pi_{\epsilon_n}^{\operatorname{tilt}}$ and using that $\ell_n \leq \sqrt{\epsilon_n}$ for all large enough $n$ implies that
\begin{equation}
    \frac{D_{\b_\phi}(\gamma_d, \pi_{\epsilon_n}^{\operatorname{tilt}})}{\epsilon_n} \geq \frac{\left| \mathbb{E}_{\xi \sim \mathcal{N}(0, 1)} \phi'(\xi)\right|}{\min\left\{M, \; C_0 \frac{e^{L\sqrt{\epsilon_n}} - 1}{L\sqrt{\epsilon_n} }\right\}} - \epsilon_n
\end{equation}
for all $n$ large enough. In particular, as $n \to \infty$ we know 
\begin{equation}
    \min\left\{M, \; C_0 \frac{e^{L\sqrt{\epsilon_n}} - 1}{L\sqrt{\epsilon_n} }\right\} \to \min \{M, \; C_0\} = C_\phi \quad \text{since $\epsilon_n \downarrow 0$}
\end{equation}
and so (\ref{eqn:proofappendixinsgleindextiltDblowerbound}) is proved.
\end{proof}

We now specialize to some common activations and compute the quantities $\mathfrak{F}_\phi^{\operatorname{flow}}$ and $\mathfrak{F}_\phi^{\operatorname{tilt}}$. First, we show a simpler form of $\mathfrak{F}_\phi^{\operatorname{tilt}}$ for the $\sin$ and $\tanh$ cases:

\begin{lemma}
\label{lemma:appendixtiltdenominatorremoval}
    If $\lambda \in (0, \infty)$ and $\phi(z) = \sin(\lambda z)$ or $\phi(z) = \tanh(\lambda z)$, then 
    \begin{equation}
        \mathfrak{F}_\phi^{\operatorname{tilt}} = \int_0^1 \frac{1}{r}\left\|\theta Q_r'(\la \cdot, \theta\ra)\right\|_{\b_\phi}dr.
    \end{equation}
\end{lemma}

\begin{proof}
   We will show that for $\phi(z) = \sin(\lambda z)$ and $\phi(z) = \tanh(\lambda z)$ it holds that if
   \begin{equation}
       F_{r,\epsilon}(t) := \frac{Q_r'(t)}{r(1 + \epsilon Q_r(t))} - \frac{Q'_r(t)}{r} = -\epsilon \frac{Q_r(t)Q_r'(t)}{r(1+\epsilon Q_r(t))} = \frac{1}{r} \frac{d}{dt}\left[\frac{\log(1 + \epsilon Q_r)}{\epsilon} - Q_r\right],
   \end{equation}
   then there is the scalar Barron norm estimate
   \begin{equation}
       \lim_{\epsilon \downarrow 0}\int_0^1 \left\| F_{r,\epsilon} \right\|_{\b_\phi} dr = 0,
   \end{equation}
   after which the main claim follows by the reverse triangle inequality.

   For $\phi(z) = \sin(\lambda z)$, we may use $\sin(a+b) = \sin(a) \cos(b) + \cos(a) \sin(b)$ to find
   \begin{align}
       Q_r(t) &= \mathbb{E}_{\xi \sim \mathcal{N}(0, 1)} \left[\sin\left(\lambda r t + \lambda \sqrt{1-r^2}\xi\right)\right] \\
       &= \sin(\lambda rt) \mathbb{E}_{\xi \sim \mathcal{N}(0, 1)} \left[\cos\left(\lambda \sqrt{1-r^2}\xi\right)\right] + \cos(\lambda rt)\mathbb{E}_{\xi \sim \mathcal{N}(0, 1)} \left[\sin\left(\lambda \sqrt{1-r^2}\xi\right)\right] 
       \\ &= \sin(\lambda rt) \mathbb{E}_{\xi \sim \mathcal{N}(0, 1)} \left[\cos\left(\lambda \sqrt{1-r^2}\xi\right)\right]
       \\&= A_r \sin(\lambda r t),
   \end{align}
   where $A_r := \exp(-\frac{\lambda^2}{2}(1-r^2)) \leq 1$. In particular,
   \begin{equation}
       F_{r,\epsilon} (t) = -\epsilon \lambda A_r^2 \frac{\sin(\lambda rt) \cos(\lambda r t)}{1 + \epsilon A_r \sin(\lambda rt)} = -\epsilon \lambda A_r^2\sum_{j=0}^\infty (-\epsilon A_r)^j \sin(\lambda r t)^{j+1} \cos(\lambda r t),
   \end{equation}
   where for the second equality we take $\epsilon < 1$ and use geometric series $\frac{1}{1-x} = \sum_{j=0}^\infty x^j $. Denoting 
   \begin{equation}
       h_j(x) := \sin(x)^{j+1}\cos(x) = \frac{1}{2^{j+2}i^{j+1}} (e^{ix} - e^{-ix})^{j+1} (e^{ix} + e^{-ix}) = \sum_{n = -(j+2)}^{j+2} a_{j,n}\sin(nx + b_{j,n})
   \end{equation}
   where $|b_{j,n}| \leq \frac{\pi}{2}$ and $\sum_{n}|a_{j,n}| \leq 1$. Hence,
   \begin{equation}
       \|h_j(\lambda r \cdot)\|_{\b_\phi} \leq \sum_{n=-(j+2)}^{j+2} |a_{j,n}| \left(1 + r|n| + \frac{|b_{j,n}|}{\lambda}\right) \leq 1 + r(j+2) + \frac{\pi}{2\lambda}.
   \end{equation}
   Since $F_{r,\epsilon}(t) = -\epsilon \lambda A_r^2 \sum_{j=0}^\infty (-\epsilon A_r)^j h_j(\lambda r t)$, we find
   \begin{align}
       \|F_{r,\epsilon}\|_{\b_\phi} &\leq \epsilon \lambda A_r^2 \sum_{j=0}^\infty (\epsilon A_r)^j \left(1 + r(j+2) + \frac{\pi}{2\lambda}\right) \\
       &\leq \epsilon \lambda  \sum_{j=0}^\infty \epsilon^j \left(j+3 + \frac{\pi}{2\lambda}\right) \\
       &\leq (8\lambda + \pi) \epsilon
   \end{align}
   for all $r \in (0, 1)$ and $\epsilon \in (0, 1/2)$. Hence, $\int_0^1 \|F_{r,\epsilon}\|_{\b_\phi} dr \to 0$.

    For $\phi(z) = \tanh(\lambda z)$ we apply a deconvolution argument with the Fourier convention $\hat{f}(\xi) = \int_\r f(x) e^{-2\pi ix\xi}dx$ (note that this is different than the rest of the paper). If we denote
    \begin{equation}
        G_{r,\epsilon}(z) := \frac{\log(1 + \epsilon Q_r(z/r))}{\epsilon} - Q_r(z/r)
    \end{equation}
    then $F_{r,\epsilon}(t) = G_{r,\epsilon}'(rt)$ and so for each $j$ we have
    \begin{equation}
    \label{eqn:proofappendixfouriertransformofFprime}
        \partial_\xi ^j \widehat{F_{r,\epsilon}'}(\xi) = r^{-j}\widehat{(-2\pi ix)^j G_{r,\epsilon}''}\left(\frac{\xi}{r}\right).
    \end{equation}
    Fix any $a \in (\frac{\pi}{8\lambda}, \frac{\pi}{2\lambda})$. We claim there are $D_{\lambda, a}, \epsilon_{\lambda, a} \in (0, \infty)$ s.t. for all $\epsilon \in (0, \epsilon_{\lambda, a})$ and all $r \in (0, 1)$ the function $G_{r,\epsilon}$ is holomorphic on an open neighborhood of $|\operatorname{Im}(z)| \leq a$ with the estimate
    \begin{equation}
    \label{eqn:proofappendixstripholomorphicsecondderivativeestimate}
      \sup_{|y| \leq a} |x+iy|^j \left|G_{r,\epsilon}''(x+iy)\right| \leq \frac{D_{\lambda, a}\epsilon}{1 + x^2} \quad (\forall x \in \r)
    \end{equation}
    for $j \in \{0, 1, 2\}$. To see this, we write $u = \lambda x$, $v = \lambda y$ and start from the identities
    \begin{equation}
        |\cosh(u+iv)|^2 = \sinh(u)^2 + \cos(v)^2, \quad |\sinh(u+iv)|^2 = \sinh(u)^2 + \sin(v)^2.
    \end{equation}
    For $z = x + iy$ with $|y| \leq a$ we know $|v| \leq \lambda a < \pi / 2$ and so
    \begin{align}
    \label{eqn:proofappendixcoshboundheedeheedee}
        |\cosh(u+iv)|^2 \geq \cos(v)^2 (\sinh(u)^2 + 1) = \cos(v)^2 \cosh(u)^2 \geq \cos(\lambda a)^2 \cosh(u)^2
    \end{align}
    and also
    \begin{equation}
        |\sinh(u+iv)|^2 \leq \sinh(u)^2 + 1 = \cosh(u)^2.
    \end{equation}
    Dividing these two, 
    \begin{equation}
        |\tanh(\lambda z)| = \frac{|\sinh(u + iv)|}{|\cosh(u+iv)|} \leq \sec(\lambda a).
    \end{equation}
    Also, taking the reciprocal of (\ref{eqn:proofappendixcoshboundheedeheedee}) yields
    \begin{equation}
        |\operatorname{sech}(\lambda z)|^2 = \frac{1}{|\cosh(u+iv)|^2}  \leq \sec(\lambda a)^2 \operatorname{sech}(\lambda x)^2 \leq 4 \sec(\lambda a)^2 e^{-2\lambda |x|},
    \end{equation}
    where we used that $\operatorname{sech}(t)^2 = \frac{4}{(e^t + e^{-t})^2} \leq 4e^{-2|t|}$. Now, we know
    \begin{equation}
       (\tanh (\lambda z))' = \lambda \operatorname{sech}(\lambda z)^2, \quad (\tanh (\lambda z))'' = -2\lambda^2 \tanh(\lambda z)\operatorname{sech}(\lambda z)^2.
    \end{equation}
    and so if we define $q_r(z) := Q_r(z/r) = \mathbb{E}_{\xi \sim \mathcal{N}(0, 1)}[\tanh(\lambda z + \lambda \sqrt{1-r^2} \xi)]$ and differentiate under the expectation then
    \begin{equation}
        |q_r(z)| \leq \sec(\lambda a), \quad |q_r'(z)| \leq 4\lambda \sec(\lambda a)^2 \mathbb{E}[e^{-2\lambda |x + \sqrt{1-r^2}\xi|}]
    \end{equation}
    and similarly
    \begin{equation}
        |q_r''(z)| \leq 8\lambda^2 \sec(\lambda a)^3\mathbb{E}[e^{-2\lambda |x + \sqrt{1-r^2}\xi|}].
    \end{equation}
Note that $|x + \sqrt{1-r^2}\xi| \geq |x| - \sqrt{1-r^2}|\xi| \geq |x| - |\xi|$ and so
\begin{equation}
    \mathbb{E}_{\mathcal{N}(0, 1)}[e^{-2\lambda |x + \sqrt{1-r^2}\xi|}] \leq e^{-2\lambda |x|} \cdot \mathbb{E}_{\xi \sim \mathcal{N}(0, 1)}[e^{2\lambda |\xi|}].
\end{equation}
    Hence, if we define the constant \begin{equation}C_{\lambda, a} := \sec(\lambda a) + \left(4\lambda \sec(\lambda a)^2 + 8 \lambda^2 \sec (\lambda a)^3\right) \cdot \mathbb{E}_{\xi \sim \mathcal{N}(0, 1)}[e^{2\lambda |\xi|}]\in (0, \infty)\end{equation} then for all $x \in \r$, $y \in [-a, a]$, and $r \in (0, 1)$ we have
    \begin{equation}
    \label{eqn:proofappendixboundsonq_rblahblah}
        |q_r(x+iy)| \leq C_{\lambda, a }, \quad |q_r'(x+iy)| + |q_r''(x+iy)| \leq C_{\lambda, a} e^{-2\lambda |x|}.
    \end{equation}
    Take $ 0 < \epsilon \leq \min \{1,\; 1/(2C_{\lambda, a})\} =: \epsilon_{\lambda, a}$. Since
    \begin{equation}
        G_{r,\epsilon} = \frac{\log(1+\epsilon q_r)}{\epsilon} - q_r,
    \end{equation}
    and throughout $|y| \leq a$ we have
$        |\epsilon q_r(x+iy)| \leq  \frac{1}{2}$, 
    we know that the principal branch of the logarithm is holomorphic on a neighborhood of the strip $|\operatorname{Im}(z)| \leq a$ and therefore so too is $G_{r,\epsilon}$. Furthermore, we compute
    \begin{equation}
        G_{r,\epsilon}'' = -\epsilon \frac{(q_r')^2 + q_rq_r''}{1 + \epsilon q_r} + \epsilon^2 \frac{q_r(q_r')^2}{(1+\epsilon q_r)^2}
    \end{equation}
    after which we apply the bound (\ref{eqn:proofappendixboundsonq_rblahblah}) to get
    \begin{align}
        |G_{r,\epsilon}''(x+iy)| &\leq 2\epsilon \left(|q'_r(x+iy)|^2 + |q_r(x+iy)||q_r''(x+iy)| \right)+ 4\epsilon^2 |q_r(x+iy)| |q_r'(x+iy)|^2 \\
        &\leq 4\epsilon (C_{\lambda, a}^2 + C_{\lambda, a}^3)  e^{-2\lambda|x|}.
    \end{align}
    If we define
    \begin{equation}
        \widetilde{C}_{\lambda, a} := \max_{j \in \{0, 1, 2\}} \sup_{x \in \r}(1+x^2)(|x| + a)^j e^{-2\lambda |x|} < \infty
    \end{equation}
    then we have shown (\ref{eqn:proofappendixstripholomorphicsecondderivativeestimate}) with $D_{\lambda, a} := (2\pi)^24(C_{\lambda, a}^2 + C_{\lambda, a}^3 )\widetilde{C}_{\lambda, a}$. Since the map $z \mapsto (-2\pi i z)^j G_{r,\epsilon}''(z)$ is holomorphic in the strip and satisfies the decay estimate (\ref{eqn:proofappendixstripholomorphicsecondderivativeestimate}), we get the exponential Fourier decay
    \begin{equation}
       \left| \widehat{(-2\pi i x)^jG''_{r,\epsilon}}(\xi)\right| \leq \pi D_{\lambda, a}  \epsilon e^{-2\pi a |\xi|} \quad (\forall \xi \in \r, \; \forall j \in \{0, 1, 2\}),
    \end{equation}
    see e.g. Theorem 2.1 of Chapter 4 in \cite{steinshakarchi2003complex}.
    In light of (\ref{eqn:proofappendixfouriertransformofFprime}), this implies that
    \begin{equation}
    \label{eqn:proofappendixboundsonFouriertransformofFprime1335}
        \left|\partial_\xi^j \widehat{F_{r,\epsilon}'}(\xi)\right| \leq r^{-j}\pi D_{\lambda, a} \epsilon e^{-2\pi a |\xi| / r} \quad (\forall \xi \in \r,\; \forall j \in \{0, 1, 2\})
    \end{equation}
    for all $r \in (0, 1)$ and $\epsilon \in (0,  \epsilon_{\lambda, a}]$. Let us now define
    \begin{equation}
        k_r(t) := \tanh(4\lambda r t), \quad \widehat{k_r'}(\xi) = \frac{\pi^2 \xi / (2\lambda r)}{\sinh(\pi^2 \xi / (4\lambda r))},
    \end{equation}
    which implies the estimate
    \begin{equation}
    \label{eqn:proofappendixboundsonFouriertransformofFprime1336}
        \left|\partial_\xi^j  \frac{1}{\widehat{k_r'}(\xi)}\right| \leq \frac{1}{2}\left(\frac{\pi^2}{4\lambda r}\right)^j e^{\pi^2 |\xi| / (4\lambda r)} \quad (\forall \xi \in \r,\; \forall j \in \{0, 1, 2\})
    \end{equation}
    To see this, note that $\frac{\sinh(x/2)}{x} = \frac{1}{2}\int_0^1 \cosh(\frac{tx}{2}) dt$ and so differentiating under the integral yields
    \begin{equation}
        \left|\frac{d^j}{dx^j} \frac{\sinh(x/2)}{x}\right|  \leq \frac{1}{2^{j+1}} \int_0^1 t^j e^{|x|/2}dt \leq 2^{-j-1}e^{|x|/2}.
    \end{equation}
If we define the function
\begin{equation}
\label{eqn:proofappendixfouriernnonsesedefinitionformreps}
    \widehat{m_{r,\epsilon}}(\xi) := \frac{\widehat{F'_{r,\epsilon}}(\xi)}{\widehat{k_r'}(\xi)} \quad (\forall \xi \in \r),
\end{equation}
then by (\ref{eqn:proofappendixboundsonFouriertransformofFprime1335}), (\ref{eqn:proofappendixboundsonFouriertransformofFprime1336}) and the product rule, if we denote $M_{\lambda, a} :=\frac{\pi}{2} D_{\lambda, a} \max_{j \in \{0, 1, 2\}} (1 + \frac{\pi^2}{4\lambda})^j$ then
\begin{equation}
    \left|\partial_\xi^j \widehat{m_{r,\epsilon}}(\xi)\right| \leq M_{\lambda, a} \epsilon r^{-j} e^{-\left(2\pi a - \frac{\pi^2}{4\lambda}\right) |\xi|/r} \quad (\forall \xi \in \r, \; \forall j \in \{0, 1, 2\}).
\end{equation} Note that $2\pi a - \frac{\pi^2}{4\lambda} > 0$ since $a > \frac{\pi}{8\lambda}$. The important consequence is that if we denote $c_{\lambda, a} := \frac{M_{\lambda, a}}{\sqrt{2\pi a - \frac{\pi^2}{4\lambda}}} $ then we have
\begin{equation}
    \left\|\partial_\xi^j\widehat{m_{r,\epsilon}}\right\|_{L^2(d\xi)} \leq c_{\lambda, a} \cdot \epsilon r^{\frac{1}{2} - j} \quad (\forall j \in \{0, 1, 2\})
\end{equation}
for all $\epsilon \in (0, \epsilon_{\lambda, a})$ and $r \in (0, 1)$. Hence, we may apply Fourier inversion to define $m_{r,\epsilon}: \r \to \r$, and so by Plancherel
\begin{align}
    \left\|m_{r,\epsilon}\right\|_{L^2(dx)} &= \left\|\widehat{m_{r,\epsilon}}\right\|_{L^2(d\xi)} \leq c_{\lambda, a} \epsilon r^{1/2}, \\ \left\|x m_{r,\epsilon}\right\|_{L^2(dx)} &= \frac{1}{2\pi}\left\|\partial_\xi\widehat{m_{r,\epsilon}}\right\|_{L^2(d\xi)} \leq \frac{c_{\lambda, a}}{2\pi} \epsilon r^{-1/2}, \\ \left\|x^2m_{r,\epsilon}\right\|_{L^2(dx)} &= \frac{1}{4\pi^2}\left\|\partial_\xi^2\widehat{m_{r,\epsilon}}\right\|_{L^2(d\xi)}\leq \frac{c_{\lambda, a}}{4\pi^2} \epsilon r^{-3/2}.
\end{align}
Furthermore, by Cauchy-Schwarz
\begin{align}
    \int_\r (1 + r|x|)|m_{r,\epsilon}(x)| dx &\leq \left(\int_\r \frac{(1+r|x|)^2}{(1+r^2x^2)^2}dx\right)^{1/2} \cdot \left(\int_\r (1+r^2x^2)^2 |m_{r,\epsilon}(x)|^2 dx\right)^{1/2} \\
    &= \sqrt{\frac{\pi+2}{r}} \cdot \left(\|m_{r,\epsilon}\|_{L^2(dx)}^2 + 2r^2\|xm_{r,\epsilon}\|_{L^2(dx)}^2 + r^4 \|x^2m_{r,\epsilon}\|_{L^2(dx)}^2 \right)^{1/2}\\
    &\leq \sqrt{\frac{\pi + 2}{r}} \cdot c_{\lambda, a}\epsilon \cdot \left(r + \frac{2r^2}{4\pi^2 r} + \frac{r^4}{16 \pi^4 r^3}\right)^{1/2}
    \\ &= \sqrt{\pi + 2} \cdot \left(1 + \frac{1}{4\pi^2}\right) \cdot c_{\lambda, a}\epsilon,
\end{align}
where the second line used that $\int_\r \frac{(1+r|x|)^2}{(1+r^2x^2)^2}dx = \frac{1}{r} \int_\r \frac{(1+|y|)^2}{(1+y^2)^2}dy = \frac{\pi+2}{r}$. 
Now, by  (\ref{eqn:proofappendixfouriernnonsesedefinitionformreps}) and the convolution theorem we know $F_{r,\epsilon}' = k_r' * m_{r,\epsilon}$ and therefore that
\begin{equation}
\label{eqn:proofappendixfouriernonsenseconvolutionrepresentation}
    F_{r,\epsilon} = k_r * m_{r,\epsilon} + C
\end{equation}
for some constant $C \in \r$. Since $F_{r,\epsilon} = -\epsilon \frac{Q_r Q_r'}{r(1 + \epsilon Q_r)}$ and $\|Q_r\|_\infty < \infty$ while $\lim_{|t| \to \infty }Q_r'(t) = 0$, we know that $\lim_{|t| \to \infty} F_{r,\epsilon}(t) = 0$. Since $\|k_r\|_\infty < \infty$ and $m_{r,\epsilon} \in L^1(dx)$, by dominated convergence
\begin{equation}
    \lim_{t \to \infty} (k_r * m_{r,\epsilon})(t) = \int_\r m_{r,\epsilon}(x) dx = \widehat{m_{r,\epsilon}}(0) = \frac{\widehat{F'_{r,\epsilon}}(0)}{\widehat{k'_r}(0)} = \frac{F_{r,\epsilon}(\infty) - F_{r,\epsilon}(-\infty)}{2} = 0.
\end{equation}
Thus, $C = 0$ and so (\ref{eqn:proofappendixfouriernonsenseconvolutionrepresentation}) reads
\begin{equation}
    F_{r,\epsilon}(t) = \int_\r \phi(4rt - 4rx) m_{r,\epsilon}(x) dx. 
\end{equation}
This is a representation of $F_{r,\epsilon}$ as a mixture of neurons, and so
\begin{align}
    \|F_{r,\epsilon}\|_{\b_\phi} &\leq \int_\r (1 + 4r + 4r|x|) |m_{r,\epsilon}(x)| dx \leq 5 \int_\r (1+r|x|) |m_{r,\epsilon}(x)| dx \\&\leq 5 \sqrt{\pi + 2} \cdot \left(1 + \frac{1}{4\pi^2}\right) \cdot c_{\lambda, a}\epsilon.
\end{align}
This holds for all $\epsilon \in (0, \epsilon_{\lambda, a})$ and all $r \in (0, 1)$, which yields the claim.
\end{proof}

We may now compute the quantities of interest for the activation functions we consider:
\begin{proposition}
\label{prop:appendixreverseOUsingleindexflowsandtiltsactivations}
    Let $\mathfrak{F}_\phi^{\operatorname{flow}}$ and $\mathfrak{F}_\phi^{\operatorname{tilt}}$ be as in (\ref{eqn:proofmathfrakflow}) and (\ref{eqn:proofmathfraktilt}), respectively. 
    \begin{itemize}
        \item \textbf{(Sinusoidal activation).} If $\phi(z) = \sin(\lambda z)$ for some $\lambda \in (0, \infty)$, then it holds that
        \begin{equation}
          \mathfrak{F}_\phi^{\operatorname{flow}}(c) = +\infty  \quad (\forall c \in [-1, 1]), \quad \mathfrak{F}_\phi^{\operatorname{tilt}} \leq 1 + \sqrt{\frac{2}{\pi}} + \sqrt{\frac{\pi}{2}}.
        \end{equation}
\item \textbf{(Tanh activation).} If $\phi(z) = \tanh(\lambda z)$ for some $\lambda \in (0, \infty)$, then it holds that
        \begin{equation}
          \mathfrak{F}_\phi^{\operatorname{flow}}(c) = +\infty  \quad (\forall c \in (-1, 1)), \quad \mathfrak{F}_\phi^{\operatorname{tilt}} \leq 1 + \sqrt{\frac{2}{\pi}} + \sqrt{\frac{\pi}{2}}.
        \end{equation}
        \item \textbf{(ReLU activation).} If $\phi(z) = \max\{0, z\}$, then it holds that
        \begin{equation}
          \mathfrak{F}_\phi^{\operatorname{flow}}(c) = +\infty  \quad (\forall c \in (-1, 1)), \quad \mathfrak{F}_\phi^{\operatorname{tilt}} \leq 9 + \sqrt{\frac{2}{\pi}} + \sqrt{\frac{\pi}{2}}.
        \end{equation}
    \end{itemize}
\end{proposition}
\begin{proof}
    First we will show that $\mathfrak{F}_\phi^{\operatorname{flow}}(c) = +\infty$ in the stated cases, for which we denote
    \begin{equation}
        V_r(x) := u Q_r(\la x, \theta\ra) + \theta \la x, u\ra Q_r'(\la x, \theta\ra) - c\theta Q_r''(\la x, \theta\ra).
    \end{equation}
    If $|c| < 1$ and we take $v := \frac{u - c\theta}{\sqrt{1 - c^2}}$ then $\la v, \theta\ra = 0$ while $\la v, u\ra = \sqrt{1-c^2} > 0$. Hence,
    \begin{equation}
        V_r(av) = u Q_r(0) + a \cdot \theta \sqrt{1-c^2} Q_r'(0) - c\theta Q_r''(0) \quad (\forall a > 0).
    \end{equation}
    In the cases $\phi(z) = \sin(\lambda z)$ and $\phi(z) = \tanh(\lambda z)$ we know $\|\phi\|_\infty, \|\phi'\|_\infty, \|\phi''\|_{\infty} < \infty$ and so $\|Q_r\|_\infty, \|Q_r'\|_\infty, \|Q_r''\|_{\infty} < \infty$ for all $r \in (0, 1)$. For $\phi(z) = \sin(\lambda z)$ we know from the proof of Lemma \ref{lemma:appendixtiltdenominatorremoval} that
    \begin{equation}
        Q'_r(0) = \lambda r A_r \cos(0) > 0 \quad (\forall r \in (0, 1))
    \end{equation}
    and for $\phi(z) = \tanh(\lambda z)$ we directly compute
    \begin{equation}
        Q'_r(0) = r \lambda \mathbb{E}_{\xi \sim \mathcal{N}(0, 1)} \left[\operatorname{sech}\left(\lambda \sqrt{1-r^2}\xi\right)^2\right] > 0 \quad (\forall r \in (0, 1]).
    \end{equation}
    Hence, $a \mapsto V_r(av)$ is unbounded. Since $\|V_r\|_\infty \leq \|\phi\|_\infty \|V_r\|_{\b_\phi}$, this implies $\|V_r\|_{\b_\phi} = \infty$ for all $r \in (0, 1)$. If instead $|c| = 1$ then $u = c\theta$ in which case
    \begin{equation}
        V_r(a \theta) = u Q_r(a) + au Q_r'(a) - u  Q_r''(a) \quad (\forall a > 0).
    \end{equation}
    In the case $\phi(z) = \sin(\lambda z)$ we know $ a \mapsto a Q_r'(a) = \lambda r A_r \cdot a \cos(\lambda r a)$ is unbounded, and $\|V_r\|_{\b_\phi}=+\infty$ is true even when $|c| = 1$. Lastly, in the case $\phi(z) = \max\{0, z\}$ and $|c| < 1$ we again define $v := \frac{u - c\theta}{\sqrt{1-c^2}}$ to get
    \begin{equation}
       \la  V_r(av + b\theta), \theta\ra = c Q_r(b) + \left(a\sqrt{1-c^2} + bc\right) Q_r'(b) - cQ_r''(b) \quad (\forall a, b > 0).
    \end{equation}
    Hence, comparing with $b=0$ gives
    \begin{equation}
        \la V_r(av+b\theta) - V_r(av), \theta\ra = \left(a \sqrt{1-c^2}\right) \left(Q_r'(b) - Q_r'(0)\right) + C_{r,b,c}
    \end{equation}
    for some constant $C_{r,b,c}$ independent of $a$. Hence,
    \begin{align}
        \frac{\|V_r(av+b\theta) - V_r(av)\|}{\|(av+b\theta) - av\|} &= \frac{\|V_r(av+b\theta) - V_r(av)\|}{b} \geq \frac{\left| \la V_r(av+b\theta) - V_r(av), \theta\ra\right|}{b}  \\
        &\geq \frac{a\sqrt{1-c^2}}{b} \left|Q_r'(b) - Q'_r(0)\right| - \frac{|C_{r,b,c}|}{b}.
    \end{align}
    For ReLU, $Q_r'(b) = r\Phi(\frac{rb}{\sqrt{1-r^2}})$ is strictly increasing and so $Q_r'(b) - Q'_r(0) \neq 0$, which means this quantity can be made as large as possible (by taking $a$ large with $b$ fixed). Hence, $V_r$ cannot be globally-Lipschitz; since $\|\phi'\|_{L^\infty} < \infty$ this means that $\|V_r\|_{\b_\phi} = \infty$ as desired.

    We now bound the energy of the tilts. For any Lipschitz $\phi$ we know
    \begin{equation}
        \frac{1}{r}Q_r'(t) = \mathbb{E}_{\xi \sim \mathcal{N}(0, 1)}\left[\phi'\left(rt + \sqrt{1-r^2}\xi\right)\right] = \frac{1}{\sqrt{1-r^2}} \mathbb{E}_{\xi \sim \mathcal{N}(0, 1)}\left[\xi \phi\left(rt + \sqrt{1-r^2}\xi\right)\right]
    \end{equation}
    by Gaussian integration by parts. This is a valid Barron representation, and so
    \begin{align}
        \frac{1}{r} \|\theta Q_r'(\la \cdot, \theta\ra)\|_{\b_\phi} &\leq \frac{1}{\sqrt{1-r^2}} \mathbb{E}_{\xi \sim \mathcal{N}(0, 1)} \left[|\xi| \left(1 + r + \sqrt{1-r^2}|\xi|\right)\right] \\
        &= \sqrt{\frac{2}{\pi}}\frac{1+r }{\sqrt{1-r^2}} + 1.
    \end{align}
    Using that $\int_0^1 \frac{1}{\sqrt{1-r^2}}dr = \frac{\pi}{2}$ and $\int_0^1 \frac{r}{\sqrt{1-r^2}}dr = 1$, we find
    \begin{equation}\label{eqn:proofappendixthedenominatorfreeupperboundtilt}
        \int_0^1 \frac{1}{r} \|\theta Q_r'(\la \cdot, \theta\ra)\|_{\b_\phi} dr \leq \sqrt{\frac{2}{\pi}}\left(\frac{\pi}{2} + 1\right) + 1.
    \end{equation}
    The tilt bounds then follow for the $\phi(z) = \sin(\lambda z)$ and $\phi(z) = \tanh(\lambda z)$ cases from Lemma \ref{lemma:appendixtiltdenominatorremoval}. 
    
    For the ReLU case $\phi(z) = \max\{0, z\}$, we use the fact that if $f : \r \to \r$ is $C^2$ with \begin{equation}
        \lim_{x \to -\infty} f(x) = \lim_{x \to -\infty} f'(x) = 0, \quad \int_\r (1 + |x|) |f''(x)|dx < \infty
    \end{equation}
    then there is the representation
    \begin{equation}
        f(x) = \int_\r \max\{0, x - a\}f''(a) da.
    \end{equation}
   This comes from the identification of $\phi$ as the Green's function of the 1D Laplacian, i.e. \begin{equation}
    \partial_x^2 \phi(x - a) = \partial^2_x \left[\max\{0, x-a\}\right] = \partial_x \left[\1_{x > a}\right] = \delta_{x-a}\end{equation}
   in the sense of distributions. In particular, we have
   \begin{equation}
       \|f\|_{\b_\phi} \leq \int_\r (1 + |a|) |f''(a)| da.
   \end{equation}
    Hence, if we note that $\mathbb{E}_{\xi \sim \mathcal{N}(0, 1)} \max\{0, \xi\} = \frac{1}{\sqrt{2\pi}}$ and we define the remainder
    \begin{align}
        R_{r,\epsilon}(t) &:= \frac{Q'_r(t)}{r \left(1 + \epsilon Q_r(t) - \frac{\epsilon}{\sqrt{2\pi}}\right)} - \frac{Q_r'(t)}{r}
        \\&= \frac{Q'_r(t)}{r} \cdot \left(\frac{1}{1 + \epsilon Q_r(t) - \frac{\epsilon}{\sqrt{2\pi}}} - 1\right),
    \end{align}
    then the same argument of (\ref{eqn:proofappendixthedenominatorfreeupperboundtilt}) yields
    \begin{equation}
        \label{eqn:proofappendixfuckingbullshitRrepsnonsense}\mathfrak{F}_\phi^{\operatorname{tilt}} \leq \sqrt{\frac{\pi}{2}} + \sqrt{\frac{2}{\pi}}  + 1 + \limsup_{\epsilon \downarrow 0}\int_0^1 \int_\r(1 + |a|) |R_{r,\epsilon}''(a)|da dr,
    \end{equation}
    where we use that $\lim_{t \to -\infty}R_{r,\epsilon}(t) = \lim_{t \to -\infty}R'_{r,\epsilon}(t)  = 0$ since $Q_r(t), Q'_r(t), Q''_r(t) \to 0$ as $t \to -\infty$ by dominated convergence.
    By the product rule,
    \begin{equation}
        R'_{r,\epsilon}(t) = \frac{Q''_r(t)}{r} \cdot \left(\frac{1}{1 + \epsilon Q_r(t) - \frac{\epsilon}{\sqrt{2\pi}}} - 1\right) - \frac{Q'_r(t)}{r} \cdot \frac{\epsilon Q'_r(t)}{\left(1 + \epsilon Q_r(t) - \frac{\epsilon}{\sqrt{2\pi}}\right)^2}
    \end{equation}
    and similarly
    \begin{align}
        R''_{r,\epsilon}(t) &= \frac{Q'''_r(t)}{r} \cdot \left(\frac{1}{1 + \epsilon Q_r(t) - \frac{\epsilon}{\sqrt{2\pi}}} - 1\right)  - \frac{Q''_r(t)}{r} \cdot \frac{\epsilon Q'_r(t)}{\left(1 + \epsilon Q_r(t) - \frac{\epsilon}{\sqrt{2\pi}}\right)^2} \\ &\quad - \frac{Q''_r(t)}{r} \cdot \frac{\epsilon Q'_r(t)}{\left(1 + \epsilon Q_r(t) - \frac{\epsilon}{\sqrt{2\pi}}\right)^2}\\&\quad  - \frac{Q'_r(t)}{r} \cdot \frac{\left(1 + \epsilon Q_r(t) - \frac{\epsilon}{\sqrt{2\pi}}\right)^2 \epsilon Q''_r(t) - 2\epsilon^2 (Q_r'(t))^2 \left(1 + \epsilon Q_r(t) - \frac{\epsilon}{\sqrt{2\pi}}\right) }{\left(1 + \epsilon Q_r(t) - \frac{\epsilon}{\sqrt{2\pi}}\right)^4} \\
        &= \frac{Q'''_r(t)}{r} \cdot \left(\frac{1}{1 + \epsilon Q_r(t) - \frac{\epsilon}{\sqrt{2\pi}}} - 1\right) - \frac{3Q''_r(t)}{r} \cdot \frac{\epsilon Q'_r(t)}{\left(1 + \epsilon Q_r(t) - \frac{\epsilon}{\sqrt{2\pi}}\right)^2} \\
        &\quad + \frac{2\epsilon^2 (Q_r'(t))^3}{r\left(1 + \epsilon Q_r(t) - \frac{\epsilon}{\sqrt{2\pi}}\right)^3}.
    \end{align}
    This splits $R_{r,\epsilon}''$ into three terms.
    Using that $Q_r(t) \geq 0$ and so $1 + \epsilon Q_r(t) - \frac{\epsilon}{\sqrt{2\pi}} \geq \frac{1}{2}$ for $\epsilon \leq \sqrt{\frac{\pi}{2}}$, we find that
    \begin{align}
        \left|\frac{Q'''_r(t)}{r} \cdot \left(\frac{1}{1 + \epsilon Q_r(t) - \frac{\epsilon}{\sqrt{2\pi}}} - 1\right)\right| &= \left|\frac{Q_r'''(t)}{r} \cdot \frac{\epsilon \left(\frac{1}{\sqrt{2\pi}} - Q_r(t)\right)}{1 + \epsilon Q_r(t) - \frac{\epsilon}{\sqrt{2\pi}}}\right| \\&\leq \frac{2\epsilon}{r}\cdot  |Q_r'''(t)| \cdot \left(\frac{1}{\sqrt{2\pi}} + Q_r(t)\right)
    \end{align}
    and that 
    \begin{equation}
       \left|\frac{3Q''_r(t)}{r} \cdot \frac{\epsilon Q'_r(t)}{\left(1 + \epsilon Q_r(t) - \frac{\epsilon}{\sqrt{2\pi}}\right)^2}\right| \leq \frac{12 \epsilon}{r} \cdot \left|Q_r''(t)\right| \cdot \left|Q_r'(t)\right|
    \end{equation}
For $\phi(z) = \max\{0, z\}$ we may compute that
\begin{equation}
    Q_r'(t) = r \Phi\left(\frac{rt}{\sqrt{1-r^2}}\right), \quad Q_r''(t) = \frac{r^2}{\sqrt{1-r^2}} \varphi\left(\frac{rt}{\sqrt{1-r^2}}\right),
\end{equation}
\begin{equation}
    Q_r'''(t) = -\frac{r^4 t}{(1-r^2)^{3/2}} \varphi\left(\frac{rt}{\sqrt{1-r^2}}\right), \quad Q_r(t) \leq r|t| + \frac{\sqrt{1-r^2}}{\sqrt{2\pi}}
\end{equation}
for $\Phi$ and $\varphi$ the Gaussian CDF and PDF, respectively. 
Hence,
\begin{equation}
   \frac{1}{r} \int_\r (1+|t|) \left( |Q_r'''(t)| \cdot \left(\frac{1}{\sqrt{2\pi}} + Q_r(t)\right) + |Q_r''(t)| \cdot |Q_r'(t)| \right)dt \leq C \left(1 + r + \frac{r}{\sqrt{1-r^2}} + \sqrt{1 - r^2}\right)
\end{equation}
and since $\int_0^1 \frac{r}{\sqrt{1-r^2}}dr < \infty$ we find
    \begin{equation}
    \int_0^1\frac{1}{r}\int_\r(1+|t|)\left( |Q_r'''(t)| \cdot \left(\frac{1}{\sqrt{2\pi}} + Q_r(t)\right) + |Q_r''(t)| \cdot |Q_r'(t)| \right)dtdr < \infty.
    \end{equation}
    This implies that the first two terms in the expansion of $R''_{r,\epsilon}$ stay first order in $\epsilon$, and do not contribute to (\ref{eqn:proofappendixfuckingbullshitRrepsnonsense}), i.e.
    \begin{equation}
        \mathfrak{F}_\phi^{\operatorname{tilt}} \leq \sqrt{\frac{\pi}{2}} + \sqrt{\frac{2}{\pi}} + 1 + \limsup_{\epsilon \downarrow 0}  \int_0^1 \frac{2\epsilon^2}{r}\int_\r (1 + |a|) \cdot \left|\frac{Q_r'(a)}{1 + \epsilon Q_r(a) - \frac{\epsilon}{\sqrt{2\pi}}}\right|^3 dadr.
    \end{equation}
    Now, since $Q_r' > 0$ for $r \in (0, 1)$ and $Q_r(-\infty) = 0$, $Q_r(\infty) = \infty$, we may substitute $u = Q_r(a) - \frac{1}{\sqrt{2\pi}}$ (with $du = Q'_r(a) da$) to find that
    \begin{align}
        \int_\r \left|\frac{Q_r'(a)}{1 + \epsilon Q_r(a) - \frac{\epsilon}{\sqrt{2\pi}}}\right|^3 da &\leq r^2 \int_\r \frac{Q_r'(a) }{\left|1 + \epsilon Q_r(a) - \frac{\epsilon}{\sqrt{2\pi}}\right|^3}da \\&= r^2 \int_{ - \frac{1}{\sqrt{2\pi}}}^\infty \frac{1}{(1 + \epsilon u)^3}du
        \\&= r^2 \frac{1}{2\epsilon \left(1 - \frac{\epsilon}{\sqrt{2\pi}}\right)^2} \\
        &\leq \frac{2r^2}{\epsilon},
    \end{align}
    where we used that $Q_r'(a) \leq r$ and $\frac{\epsilon}{\sqrt{2\pi}} \leq \frac{1}{2}$.
    In particular, 
    \begin{equation}
        \limsup_{\epsilon \downarrow 0}  \int_0^1 \frac{2\epsilon^2}{r}\int_\r  \left|\frac{Q_r'(a)}{1 + \epsilon Q_r(a) - \frac{\epsilon}{\sqrt{2\pi}}}\right|^3 dadr \leq \limsup_{\epsilon \downarrow 0} \int_0^1 4\epsilon r dr = 0.
    \end{equation}
    We split the remaining integral into $a < 0$ and $a \geq 0$. Indeed, by using $0 \leq Q_r'(a) \leq r$ and lower bounding the denominator we have
    \begin{equation}
        \int_{-\infty}^0 |a| \cdot \left|\frac{Q_r'(a)}{1 + \epsilon Q_r(a) - \frac{\epsilon}{\sqrt{2\pi}}}\right|^3 da \leq -8r^2 \int_{-\infty}^0 aQ_r'(a) da = 8r^2 \int_{-\infty}^0 Q_r(a) da
    \end{equation}
    using integration by parts. We calculate
    \begin{align}
        \int_{-\infty}^0 Q_r(a) da &= \mathbb{E}_{\xi \sim \mathcal{N}(0, 1)}\left[\1_{\xi > 0}\int_{-\sqrt{1-r^2}\xi / r}^0 \left(ra + \sqrt{1-r^2}\xi\right)da\right] \\&=\frac{1-r^2}{2r} \mathbb{E}_{\xi \sim \mathcal{N}(0, 1)}\left[\xi^2 \1_{\xi > 0}\right]\leq \frac{1}{4r},
    \end{align}
    which means that 
    \begin{equation}
         \limsup_{\epsilon \downarrow 0}  \int_0^1 \frac{2\epsilon^2}{r}\int_{-\infty}^0  |a| \cdot \left|\frac{Q_r'(a)}{1 + \epsilon Q_r(a) - \frac{\epsilon}{\sqrt{2\pi}}}\right|^3 dadr \leq \limsup_{\epsilon \downarrow 0} \int_0^1 4\epsilon^2 dr = 0
    \end{equation}
    and so the only contribution comes from $a \geq 0$. 
  If $a \geq 0$ then we know $Q_r(a) \geq ra$ and so
  \begin{equation}
 |a| \cdot \left|\frac{Q_r'(a)}{1 + \epsilon Q_r(a) - \frac{\epsilon}{\sqrt{2\pi}}}\right|^3 \leq  \frac{a r^3}{\left(1 + \epsilon ra - \frac{\epsilon}{\sqrt{2\pi}}\right)^3} \leq  \frac{a r^3}{\left(\frac{1}{2} + \epsilon r a\right)^3} \leq  \frac{8ar^3}{(1+\epsilon ra)^3}
  \end{equation}
  since $0 \leq Q_r'(a) \leq r$ and $\frac{\epsilon}{\sqrt{2\pi}} \leq \frac{1}{2}$. Hence,
  \begin{align}
         \limsup_{\epsilon \downarrow 0}  \int_0^1 \frac{2\epsilon^2}{r}\int_0^\infty  |a| \cdot \left|\frac{Q_r'(a)}{1 + \epsilon Q_r(a) - \frac{\epsilon}{\sqrt{2\pi}}}\right|^3 dadr &\leq \limsup_{\epsilon \downarrow 0} \int_0^1 16\epsilon^2 r^2 \int_0^\infty \frac{a}{(1+\epsilon r a)^3}da dr \\
         &= \limsup_{\epsilon \downarrow 0} \int_0^1 16\epsilon^2 r^2 \cdot \frac{1}{2\epsilon^2r^2}dr\\
         &= 8.
    \end{align}
\end{proof}

\begin{proof}[Proof of Example \ref{example:reverseOUsingleindexflowsandtilts}]
    We combine Lemmas \ref{lemma:appendixreverseOUsingleindexflows} and \ref{lemma:appendixreverseOUsingleindextilts} with Proposition \ref{prop:appendixreverseOUsingleindexflowsandtiltsactivations} to find that if $|\la u, \theta\ra |<1$ the flow ratios are infinite and the tilt ratios are always bounded above by
    \begin{equation}
        \left(9 + \sqrt{\frac{2}{\pi}} + \sqrt{\frac{\pi}{2}}\right) \cdot \frac{1}{\left|\mathbb{E}_{\mathcal{N}(0, 1)}[\phi'(\xi)]\right|}
    \end{equation}
    since $\phi(0) = 0$ and $\operatorname{Lip}(\phi) \leq 1$ for $\phi(z) = \sin(z)$, $\phi(z) = \tanh(z)$, and $\phi(z) = \max\{0, z\}$. In either case, we may confirm that $\left|\mathbb{E}_{\mathcal{N}(0, 1)}[\phi'(\xi)]\right| \geq \frac{1}{2}$, from which the claim follows.
\end{proof}

\subsection{Reverse OU suboptimality for triangular shear measures}
\label{subsec:appendixtriangularshearmeasures}
In the main paper, we present lower bounds for the representation and approximation by two-layer neural networks of the OU scores of certain ResNet pushforwards. These proof techniques extend to any measure with a triangular shear structure, which may be of wider interest. We study the general case in this section, and in Section \ref{subsec:appendixapplicationtoResNetpushsforwards} we specialize these general lower bounds to the ResNet pushforward examples. 

\begin{definition}[Triangular shear measure]
Fix $n \in \n$ and let $\psi: \r \to \r$ and $\Psi: \r^n \to \r$ be measurable. Using coordinates $(z, u) \in \r^{n+1}$ with $z \in \r^n$ and $u \in \r$, we define the induced {\em triangular shear measure} $\pi^{\psi, \Psi} \in \p(\r^{n+1})$ by
\begin{equation}
\label{eqn:appendixshearstructurerepresentation}
    \pi^{\psi, \Psi} := \operatorname{Law}(Z, \psi(U) + \Psi(Z)), \quad (Z, U) \sim \gamma_{n+1}.
\end{equation}
\end{definition}

We recall the following preliminaries from Section \ref{subsec:prelimgenerativepaths}. The OU semigroup $(P_\tau)_{\tau \geq 0}$ and its adjoint $(P_\tau^*)_{\tau \geq 0}$ act on functions $f: \r^{n} \to \r$ and probability measures $\rho \in \p(\r^n)$, respectively, by
\begin{equation}
    (P_\tau f)(x) := \mathbb{E}_{Z \sim \gamma_n}\left[f\left(e^{-\tau}x + \sqrt{1-e^{-2\tau}}Z\right)\right],  \quad P_\tau^* \rho := \operatorname{Law}\left(e^{-\tau} X+ \sqrt{1 - e^{-2\tau}}Z\right)
\end{equation}
 with $X \sim \rho$ and $Z \sim \gamma_n$ independently. In this section we will use the same notation for the corresponding semigroups on $\r^{n+1}$, with the dimensions clear from context.

If $\rho_\tau^{\psi, \Psi} = P_{\tau}^* \pi^{\psi, \Psi} $ denotes the OU flow initialized from $\pi^{\psi, \Psi}$ and we define the time-dependent scores relative to Lebesgue and the Gaussian for $\tau > 0$ by
\begin{equation}
    s_\tau^{\psi, \Psi} := \nabla \log \rho_\tau^{\psi, \Psi}, \quad b_\tau^{\psi, \Psi} := \nabla \log \frac{\rho_\tau^{\psi, \Psi}}{\gamma_{n+1}} = s_\tau^{\psi, \Psi} + \operatorname{Id}_{n+1},
\end{equation}
respectively,
 then $(v_t^{\psi, \Psi}, \tilde{\rho}_t^{\psi, \Psi})_{t \in [0, 1]}$ solves the continuity equation with $\tilde{\rho}_0^{\psi, \Psi} = \gamma_{n+1}$, $\tilde{\rho}_1^{\psi, \Psi} = \pi^{\psi, \Psi}$ and
\begin{equation}
    v_t^{\psi, \Psi} := \frac{1}{t}b_{-\log t}^{\psi, \Psi}, \quad \tilde{\rho}_t^{\psi, \Psi} := \rho_{-\log t}^{\psi, \Psi} \quad (\forall t \in (0, 1)).
\end{equation}
Hence, $v_t^{\psi, \Psi}$ is the standard probability flow denoising generative model and $b_\tau^{\psi, \Psi}$ is a time-change of it.
Lastly, we recall our definition of the reverse OU energy for a Banach norm $\|\cdot\|_\b$ to be
\begin{equation}
    \mathcal{E}_\b(\pi^{\psi, \Psi}) = \int_0^1 \|v_t^{\psi, \Psi}\|_\b dt = \int_0^\infty \|b_\tau^{\psi, \Psi}\|_\b d\tau,
\end{equation}
where we take $\b = \b_\phi$ or $\b = \b_\phi^R$, depending on the situation.

In this section, we prove lower bounds for the representation and approximation of the generative drifts $v_t^{\psi, \Psi}$ using two-layer neural networks in terms of the properties of the maps $(\psi,\Psi)$ which generate the shear. These lower bounds will also hold for the Lebesgue and Gaussian scores $s_\tau^{\psi, \Psi}$ and $b_\tau^{\psi, \Psi}$ and therefore apply similarly to SDE-based denoising generative models, other time parameterizations, variance-exploding formulations, etc. We summarize these results here:

\begin{theorem}[Shallow representation and approximation lower bounds, quantitative]
    Fix $\psi, \Psi$ and define $s_\tau^{\psi, \Psi}, b_\tau^{\psi, \Psi}$ as above.  Then, the following hold:
    \begin{enumerate}
        \item[(i)] If $\Psi \in C^5_b(\r^n)$ and $\psi$ is increasing and equal to the identity on $[-2\|\Psi\|_\infty, 2\|\Psi\|_\infty]$, then
        \begin{equation}
            \|\partial_1^3 s_\tau^{\psi, \Psi}\|_\infty \gtrsim \|\partial_1^3 \Psi\|_\infty \quad \text{ for all $0 < \tau \lesssim \frac{\|\partial_1^3\Psi\|_\infty}{n \|\Psi\|_{C^5}}$}.
        \end{equation}
        In particular, if $\phi \in C^3_b(\r)$ then $\inf_{R > 0}R^2\|s_\tau^{\psi, \Psi}\|_{\b_\phi^R} \gtrsim \|\partial_1^3 \Psi\|_\infty$, while if $\phi$ is positively $3$-homogeneous then $\|s_\tau^{\psi, \Psi}\|_{\b_\phi}\gtrsim \|\partial_1^3 \Psi\|_\infty $. The implied constants depend only on $\|\Psi\|_\infty$.
        \item[(ii)] Let $\Psi$ be bounded and $L$-Lipschitz and $\psi$ linear with positive slope, and fix $\tau \geq 0$, $\epsilon > 0$. If $\hat{s}_\tau$ is a two-layer neural network of width $m$, then \begin{equation}\|s_\tau^{\psi, \Psi} - \hat{s}_\tau\|_{L^2(\rho_\tau^{\psi, \Psi}; \r^{n+1})} \leq \epsilon \implies 
        m \geq m_\Psi(Q(\epsilon + L^2\tau), \tau),
        \end{equation}
        where $m_\Psi$ is defined in Definition \ref{def:appendixhermitewidthenvelope} using the Hermite decomposition of $\Psi$ and $Q$ is a constant depending only on $\|\Psi\|_\infty$ and $\psi$. In such cases, for any $\phi$ satisfying Assumption \ref{assumption:activationfunciton} we have
        \begin{equation}
            \|\hat{s}_\tau\|_{\b_\phi} \gtrsim \epsilon \sqrt{m_\Psi(Q(2\epsilon + L^2\tau), \tau) -1}
        \end{equation}
        with implied constant depending on $\phi, \|\Psi\|_\infty,\psi$.
    \end{enumerate}
    These statements also hold with $s_\tau^{\psi, \Psi}$ replaced by $b_\tau^{\psi, \Psi}$.
\end{theorem}

In words, our representation result (Corollary \ref{corollary:representationlowerbound}) lower bounds the third derivatives of the scores (and therefore the Barron norm for sufficiently regular $\phi$) in terms of the derivatives of $\Psi$. The approximation result (Proposition \ref{prop:appendixdaapproximationlowerbound}) is stronger, as it leverages the spectral tails of the Hermite tensor decomposition of $\Psi$ in order to lower bound the width of any shallow $L^2$ approximation; this can be converted to a lower bound on the Barron norm of the approximation.

In both cases we allow the added flexibility of $\psi$ different from the identity: in the representation case we allow $\psi$ to have nonlinear tails (in which case the last coordinate of $\pi^{\psi, \Psi}$ has non-Gaussian tails), and in the approximation case we allow $\psi$ to be a scalar multiple of the identity (in which case the last coordinate of $\pi^{\psi, \Psi}$ is $\Psi(Z)$ smoothed by a Gaussian of any fixed variance).

To help interpret these lower bounds, we also demonstrate some settings where the corresponding upper bounds hold in a nonquantitative sense:

\begin{theorem}[Shallow representation and approximation upper bounds, nonquantitative]
     Fix $\psi, \Psi$ and define $s_\tau^{\psi, \Psi}, b_\tau^{\psi, \Psi}$ as above.  Then, the following hold:
     \begin{enumerate}
         \item[(i)] If $\Psi(z) = \sum_{r=1}^M \Phi_r(P_{E_r}z)$ for subspaces $E_r \subset \r^n$ with $\Phi_r \in C^\infty_c(E_r)$ and $\psi$ is a smooth increasing diffeomorphism with $\psi(u) \sim u^p$ outside of a compact set for some $p > 2$, then $s_\tau^{\psi, \Psi} \in \b_{\operatorname{ReLU}^s}$ for any $s \in \n$ and all $\tau \geq 0$.
         \item[(ii)] For any $\psi, \Psi$, any $\epsilon, \tau  > 0$, and any nonpolynomial activation $\phi$, there is some two-layer neural network $\hat{s}_\tau$ of finite width which uses activation $\phi$ and satisfies $\|s_\tau^{\psi, \Psi} - \hat{s}_\tau\|_{L^2(\rho_\tau^{\psi, \Psi}; \r^{n+1})} \leq \epsilon$.
     \end{enumerate}
     This first statement also holds with $s_\tau^{\psi, \Psi}$ replaced by $b_\tau^{\psi, \Psi}$, and the second statement does if $\pi^{\psi, \Psi} \in \p_2(\r^d)$.
\end{theorem}

In words, we provide conditions under which the OU-smoothed scores $s_\tau^{\psi, \Psi}, b_\tau^{\psi, \Psi}$ are (i) themselves Barron and therefore representable by infinite-width two-layer neural networks of finite path norm (Lemma \ref{lemma:appendixrepresentationupperbound}) and (ii) are approximable to arbitrary $L^2$ accuracy by finite-width two-layer neural networks (Proposition \ref{prop:appendixapproximationupperbound}). These are proved by a Sobolev-to-Barron embedding (Lemmas \ref{lemma:appendixcylindricalschwartz} and \ref{lemma:appendixsobolevtobarron}) and by the classic universal $L^2$ approximation result of Hornik \cite{hornik1991approximation}, respectively. We do not provide quantitative upper bounds on the widths or Barron norms; these results merely serve as sanity checks to complement our quantitative lower bounds.

\subsubsection{Representation lower bound via third derivative}

We first lower bound the homogeneous Sobolev seminorm $\|\cdot\|_{\dot{W}^{3,\infty}(\r^{n+1}; \r^{n+1})}$ of the drifts and their integrated value $\mathcal{E}_{\dot{W}^{3,\infty}(\r^{n+1}; \r^{n+1})}(\pi^{\psi, \Psi})$.  
The mechanism is elementary: if $\Psi$ is smooth and bounded with large derivatives and $\psi$ is the identity near the origin, then a standard OU stability argument transfers those large derivatives to $s_\tau^{\psi, \Psi}$ for small OU times.

\begin{lemma}[Third derivative representation lower bound]
\label{lemma:stupidnonsenserevised}
    Let $n \in \n$, fix $\Psi \in C^5_b(\r^{n}; \r)$, and let  $\beta, D \in (0, \infty)$ be  s.t. $D> 1$ and
    \begin{equation}
    \label{eqn:proofappendixassumptionsonPsiforrepresentation}
        \|\Psi\|_\infty \leq \beta, \quad \|\partial^\alpha \Psi\|_\infty \leq D^{|\alpha|} \quad (\forall 1 \leq |\alpha| \leq 5).
    \end{equation}
   Fix $\psi: \r \to \r$ measurable and nondecreasing s.t. $\psi(u) = u$ for all $u \in [-2\beta, 2\beta]$.
    Then, there is some $L_{\beta} \in (0, \infty)$ s.t. for every $\eta \in \{1, 2, 3\}$ we have
    \begin{equation}
    \label{eqn:appendixlemmathirdderivativelowerbound}
       \|\partial_1^\eta s_\tau^{\psi, \Psi}\|_{\infty} \geq \frac{ \|\partial_1^\eta \Psi\|_\infty}{2} \quad \left(\forall \tau \in \left(0, \;L_\beta \frac{ \|\partial_1^\eta \Psi\|_\infty}{nD^{\eta+2}}\right]\right).
    \end{equation}
    This also holds with $s_\tau^{\psi, \Psi}$ replaced by $b_\tau^{\psi, \Psi}$.
    
\end{lemma}
\begin{remark}
    Since $Z \sim \gamma_n$ is rotation-invariant, the same bound follows if every $\partial_1$ is replaced by any other coordinate derivative $\partial_i$.
\end{remark}
\begin{remark}
Such a result holds for all derivative orders $\eta$. However, in the natural case $\Psi$ has frequency $D$, i.e. $\|\partial_1^\eta\Psi\|_\infty \asymp D^\eta$, we would get a lower bound of $D^{\eta - 2}$
on the integrated score norms. Hence, only when $\eta \geq 3$ does this argument provide a growing lower bound in the regime where $D$ is large; this is why we focus on the third derivative.
\end{remark}
\begin{proof}
Denote by $\varphi_{a^2}$ the density of $\mathcal{N}(0, a^2)$ w.r.t. Lebesgue and by $q_\tau$ the density of $e^{-\tau} \psi(U) + \sqrt{1-e^{-2\tau}}H$ for $U, H \sim \gamma_1$ i.i.d. If we denote $\sigma := \sqrt{1-e^{-2\tau}}$ then
\begin{equation}
    q_\tau(w) = \int_\r \varphi_1(u) \varphi_{\sigma^2}(w - e^{-\tau}\psi(u)) du.
\end{equation}
Since $e^{-\tau} U + \sigma H \sim \gamma_1$ we know
\begin{equation}
    \varphi_1(w) = \int_\r \varphi_1(u) \varphi_{\sigma^2}(w - e^{-\tau}u) du
\end{equation}
and hence for $j \in \{0, \ldots, 4\}$ we have
\begin{equation}
    q_\tau^{(j)}(w) - \varphi_1^{(j)}(w) = \int_{|u| > 2\beta} \varphi_1(u)\left[\varphi_{\sigma^2}^{(j)}(w-e^{-\tau}\psi(u)) - \varphi_{\sigma^2}^{(j)}(w-e^{-\tau}u)\right]du.
\end{equation}
If $\tau \in (0, 1/2]$ (and so $\tau \leq \sigma^2 \leq 2\tau$) and $|t| \geq \beta / 2$ then for all $j \in \{0, \ldots, 4\}$ we claim that
\begin{align}
\label{eqn:proofappendixboundonderivativesofgaussianpdf}
    \left|\varphi^{(j)}_{{\sigma^2}}(t)\right| \leq \frac{E_\beta}{2} \tau, \quad E_\beta := 200 \left(\frac{112}{e\beta^2}\right)^{7/2}.
\end{align}
This follows from the observation $|\varphi_1^{(j)}(x)| \leq 10 (1+|x|^4)e^{-x^2/2} \leq 100 e^{-x^2/4}$ and therefore $|\varphi_{\sigma^2}^{(j)}(t)|\leq 100 \sigma^{-(j+1)}e^{-t^2/(4\sigma^2)} \leq 100\tau^{-5/2}e^{-\beta^2/(32\tau)}$. Combining this with the elementary calculus bound $\sup_{\tau \leq 1/2} \tau^{-7/2}e^{-\beta^2/(32\tau)} \leq (112/(e\beta^2))^{7/2}$
yields (\ref{eqn:proofappendixboundonderivativesofgaussianpdf}) and consequently
\begin{equation}
\label{eqn:proofappendix1dboundonqtauderivatives}
    \max_{0 \leq j \leq 4} \sup_{|w| \leq e^{-\tau}\beta} \left|q_\tau^{(j)}(w) - \varphi^{(j)}_1(w)\right| \leq E_\beta \tau \quad (\forall  \tau \in (0, 1/2]),
\end{equation}
where we note that $|w| \leq e^{-\tau}\beta$ implies that $\min\{|w-e^{-\tau}\psi(u)|, |w-e^{-\tau}u|\} \geq e^{-\tau} \beta \geq \beta / 2$ by monotonicity of $\psi$ and $|u| > 2\beta$.

Now, with $\rho_\tau^{\psi, \Psi} = P_\tau^* \pi^{\psi, \Psi} \in \p(\r^{n+1})$ the Mehler formula implies that $\rho_\tau^{\psi, \Psi} = \operatorname{Law}\left((Z_\tau, U_\tau)\right)$ for the random variables
\begin{equation}
Z_\tau := e^{-\tau}Z + \sqrt{1-e^{-2\tau}} G, \quad U_\tau := e^{-\tau}\left(\psi(U) + \Psi(Z)\right) + \sqrt{1-e^{-2\tau}} H,
\end{equation}
where $Z, G \sim \gamma_n$ and $U, H \sim \gamma_1$ are all independent. Since $\operatorname{Law}(Z_\tau) = \gamma_n$ we have the joint density
\begin{equation}
\label{eqn:proofappendixrepresentaitonlowerboundjointdensity}
    \rho_\tau^{\psi, \Psi}(z, u) = \gamma_n(z) F_\tau(z, u),
\end{equation}
where we define $F_\tau(z, \cdot)$ to be the density of $\operatorname{Law}(U_\tau \mid Z_\tau = z)$. Now, we know $Z \mid \{Z_\tau = z\} \sim \mathcal{N}(e^{-\tau}z, (1-e^{-2\tau})\1_n)$ by Gaussian regression and so
\begin{equation}
    F_\tau(z, u) = \mathbb{E}_{G \sim \gamma_n} \left[q_\tau\left(u - e^{-\tau}\Psi(e^{-\tau}z + \sqrt{1-e^{-2\tau}} G)\right)\right]
\end{equation}
since $q_\tau$ is the density of $e^{-\tau}\psi(U) + \sqrt{1-e^{-2\tau}}H$. Let us also define the Gaussian comparator
\begin{equation}
    \bar{F}_\tau(z, u) := \mathbb{E}_{G \sim \gamma_n} \left[\varphi_1\left(u - e^{-\tau}\Psi(e^{-\tau}z + \sqrt{1-e^{-2\tau}}G)\right)\right]
\end{equation}
For $\ell \in \{0, 1\}$ we have
\begin{equation}
    \left|\partial_u^\ell (F_\tau - \bar{F}_\tau)(e^{\tau}z, 0)\right| = \left|\mathbb{E}_{G \sim \gamma_n}\left[(q_\tau^{(\ell)} - \varphi_1^{(\ell)})(-e^{-\tau}\Psi(z+\sqrt{1-e^{-2\tau}}G))\right]\right|.
\end{equation}
Differentiating three times in the first coordinate of $e^{\tau}z$ and applying (\ref{eqn:proofappendixassumptionsonPsiforrepresentation}) and (\ref{eqn:proofappendix1dboundonqtauderivatives}) yield
\begin{equation}
\label{eqn:proofappendixrepresentationlowerboundblabla5}
     \sup_{z \in \r^n} \max_{\ell \in \{0, 1\}}\left|\partial_1^k\partial_u^\ell (F_\tau - \bar{F}_\tau)(e^{\tau}z, 0)\right| \leq 5E_\beta D^k \tau \quad (\forall k \in \{0, \ldots, 3\},\; \forall \tau \in (0, 1/2]).
\end{equation}
Next, for $k \in \{0, \ldots, 3\}$ and $\ell \in \{0, 1\}$, each summand of $\partial_1^k \partial_u^\ell \bar{F}_\tau(e^\tau z, 0)$ has the form
\begin{equation}
   ce^{-a\tau} \mathbb{E}_{G \sim \gamma_n} \left[f_{j,\alpha,\tau}(z + \sqrt{1 - e^{-2\tau}} G)\right], \quad f_{j,\alpha, \tau}(y) :=  \varphi_1^{(j)}\left(-e^{-\tau} \Psi(y)\right)\prod_{i=1}^m \partial_1^{\alpha_i} \Psi(y),
\end{equation}
where $0 \leq a \leq 6$, $j \in \{0, \ldots, 4\}$, $\sum_{i=1}^m \alpha_i = k$, and the sum of $|c|$ across all the summands is $\leq 5$. Note the Laplacian bound
\begin{equation}
\label{eqn:proofappendixrepresentaitonlowerboundblahblah4}
    \|\Delta f_{j,\alpha,\tau}\|_\infty \leq n B (2+2m+m^2)D^{k+2} \leq 17BnD^{k+2},
\end{equation}
where we denote $B := \max_{0 \leq j \leq 6} \|\varphi_1^{(j)}\|_\infty \leq 20000$. Consequently, the difference between corresponding summands of $\partial_1^k \partial_u^\ell \bar{F}_\tau(e^\tau z, 0)$ and $\partial_1^k \partial_u^\ell \bar{F}_0( z, 0)$ is upper bounded in absolute value by
\begin{align}
|ce^{-a\tau} \mathbb{E}_{G \sim \gamma_n} &\left[f_{j,\alpha,\tau}(z + \sqrt{1 - e^{-2\tau}} G)\right] - c f_{j,\alpha,0}(z)| \leq  \\
    &\quad |c|e^{-a\tau} \left|\mathbb{E}_{G \sim \gamma_n} \left[f_{j,\alpha,\tau}(z + \sqrt{1 - e^{-2\tau}} G)\right] - f_{j,\alpha,\tau}(z)\right| + \\&\quad |c| \left|e^{-a \tau} f_{j,\alpha,\tau}(z) - f_{j,\alpha,0}(z)\right|.
    \label{eqn:proofappendixrepresentationlowerboundblahblah3}
\end{align}
We know that
\begin{align}
     \left|e^{-a\tau}\varphi_1^{(j)} (-e^{-\tau}\Psi(z)) - \varphi_1^{(j)}(-\Psi(z))\right| &\leq (1-e^{-a\tau})\|\varphi_1^{(j)}\|_\infty  +   \left|\varphi_1^{(j)} (-e^{-\tau}\Psi(z)) - \varphi_1^{(j)}(-\Psi(z))\right| \\
     &\leq (1-e^{-a\tau}) \|\varphi_1^{(j)}\|_\infty  + \|\varphi_1^{(j+1)}\|_\infty (1-e^{-\tau}) |\Psi(z)|  \\
     &\leq B(a + \beta) \tau
\end{align}
and therefore the second term in (\ref{eqn:proofappendixrepresentationlowerboundblahblah3}) is bounded by
\begin{equation}
    \left|e^{-a \tau} f_{j,\alpha,\tau}(z) - f_{j,\alpha,0}(z)\right| \leq  B(6 + \beta) \tau D^k.
\end{equation}
For the first term, we use the heat semigroup identity 
\begin{align}
    \left|\mathbb{E}_{G \sim \gamma_n} \left[f_{j,\alpha,\tau}(z + \sqrt{1 - e^{-2\tau}} G)\right] - f_{j,\alpha,\tau}(z)\right| &= \left|\frac{1}{2} \int_0^{1-e^{-2\tau}} \mathbb{E}_{G \sim \gamma_n} \left[\Delta f_{j,\alpha,\tau}(z + \sqrt{t}G)\right]dt\right|
    \\ &\leq \frac{1-e^{-2\tau}}{2}\|\Delta f_{j,\alpha, \tau}\|_\infty \\ &\leq \frac{1-e^{-2\tau}}{2}(17BnD^{k+2}) \\
    &\leq 17BnD^{k+2}\tau,
\end{align}
where for the third line we plugged in (\ref{eqn:proofappendixrepresentaitonlowerboundblahblah4}). Summing over the summands, 
we find that
\begin{align}
    \sup_{z \in \r^n} \max_{\ell \in \{0, 1\}}\left|\partial_1^k \partial_u^\ell \bar{F}_\tau(e^\tau z, 0) - \partial_1^k \partial_u^\ell \bar{F}_0(z, 0)\right| &\leq 5 \cdot (B(6 + \beta) \tau D^k + 17BnD^{k+2}\tau) \\&\leq B(115 + 5\beta) nD^{k+2}\tau.
\end{align}
Combining with (\ref{eqn:proofappendixrepresentationlowerboundblabla5}), we find that
\begin{equation}
\label{eqn:proofappendixrepresentationlowerboundmainestimate}
     \sup_{z \in \r^n} \max_{\ell \in \{0, 1\}}\left|\partial_1^k\partial_u^\ell F_\tau(e^\tau z, 0) - \partial_1^k\partial_u^\ell F_0(z, 0)\right| \leq J_\beta nD^{k+2} \tau 
\end{equation}
for every $k \in \{0, \ldots, 3\}$ and $\tau \in (0, 1/2]$, where we define $J_\beta := B(115 + 5\beta) + 5E_\beta$ and the Gaussian comparator $F_0(z, u) := \bar{F}_0(z, u) = \varphi_1(u-\Psi(z))$.

Armed with this stability along the OU flow, we note that the last coordinate of the score satisfies
\begin{equation}
   S_\tau := (s_\tau^{\psi, \Psi})_{n+1} = \partial_u \log \rho_\tau^{\psi, \Psi} = \frac{\partial_u F_\tau}{F_\tau}
\end{equation}
due to the joint density (\ref{eqn:proofappendixrepresentaitonlowerboundjointdensity}). Denote also the comparator
$S_0(z, u) := \Psi(z) - u$. 
By the quotient rule, for any $S = \frac{\partial_u F}{F}$ we have
\begin{equation}
\label{eqn:proofappendixrepresentationlowerboundquotientrule1}
    \partial_1 S = \frac{F_{1u}}{F} - \frac{F_1F_u}{F^2}, \quad \partial_1^2 S = \frac{F_{11u}}{F} -\frac{2F_{1u}F_1 + F_uF_{11}}{F^2} + \frac{2F_uF_1^2}{F^3},
\end{equation}
\begin{equation}
\label{eqn:proofappendixrepresentationlowerboundquotientrule2}
    \partial_1^3 S = \frac{F_{111u}}{F} - \frac{3F_{11u}F_{1} + 3F_{1u}F_{11} + F_{u}F_{111}}{F^2} + \frac{6F_{1u}F_1^2 + 6F_uF_{1}F_{11}}{F^3} - \frac{6F_uF_{1}^3}{F^4}.
\end{equation}
Denoting $m_\beta := \varphi_1(\beta)$, from here on we fix $nD^2 \tau \leq \frac{m_\beta}{2J_\beta}$. Applying (\ref{eqn:proofappendixrepresentationlowerboundmainestimate}) with $k = \ell = 0$ implies that 
\begin{equation}
    F_\tau(e^\tau z, 0) \geq F_0(z, 0) - J_\beta nD^2 \tau \geq \frac{m_\beta}{2}.
\end{equation}
Moreover, since we have $F_0(z, u) = \varphi_1(u-\Psi(z))$ the chain rule implies
\begin{equation}
    \max_{\ell \in \{0, 1\}} \left|\partial_1^k \partial_u^\ell F_0(z, 0)\right| \leq 5BD^k \quad (\forall k \in \{0, \ldots, 3\}),
\end{equation}
after which another application of (\ref{eqn:proofappendixrepresentationlowerboundmainestimate}) yields
\begin{equation}
\label{eqn:proofappendixrepresentationlowerboundmainestimate2}
     \max_{\ell \in \{0, 1\}} \left|\partial_1^k \partial_u^\ell F_\tau(e^\tau z, 0)\right| \leq 5BD^k + J_\beta nD^{k+2}\tau \leq J_\beta D^k \quad (\forall k \in \{0, \ldots, 3\}).
\end{equation}
For any term in (\ref{eqn:proofappendixrepresentationlowerboundquotientrule1}), (\ref{eqn:proofappendixrepresentationlowerboundquotientrule2}) of the form $c \frac{\prod_{i=1}^m\partial_1^{\alpha_i}\partial_u^{\ell_i}F}{F^q}$ with $\sum_{i=1}^m \alpha_i = k$ and $\ell_i \in \{0, 1\}$ we have the bound
\begin{align}
    \left|\frac{\prod_{i=1}^m\partial_1^{\alpha_i}\partial_u^{\ell_i}F_\tau}{F_\tau^q} - \frac{\prod_{i=1}^m\partial_1^{\alpha_i}\partial_u^{\ell_i}F_0}{F_0^q}\right| &\leq \frac{1}{F_\tau^q} \left|\prod_{i=1}^m\partial_1^{\alpha_i}\partial_u^{\ell_i}F_\tau - \prod_{i=1}^m\partial_1^{\alpha_i}\partial_u^{\ell_i}F_0\right| \\&\quad + \left|\prod_{i=1}^m\partial_1^{\alpha_i}\partial_u^{\ell_i}F_0\right| \cdot \left|\frac{1}{F_\tau^q} - \frac{1}{F_0^q}\right|,
\end{align}
where we recall that the $F_\tau$ terms take $e^\tau z$ as their argument while the $F_0$ terms use $z$.
If we apply the telescoping bound
\begin{equation}
   \left| \prod_{i=1}^m a_i - \prod_{i=1}^m b_i\right| \leq \sum_{i=1}^m |a_i - b_i| \prod_{j \neq i} \max\{|a_j|, |b_j|\},
\end{equation}
then by applying (\ref{eqn:proofappendixrepresentationlowerboundmainestimate}) and (\ref{eqn:proofappendixrepresentationlowerboundmainestimate2}) we find
\begin{equation}
    \left|\prod_{i=1}^m\partial_1^{\alpha_i}\partial_u^{\ell_i}F_\tau - \prod_{i=1}^m\partial_1^{\alpha_i}\partial_u^{\ell_i}F_0\right|  \leq m J_\beta^m n D^{k+2}\tau.
\end{equation}
Furthermore, we may apply $|\frac{1}{a^q} - \frac{1}{b^q}| \leq \frac{q}{\min\{a,b\}^{q+1}}|a-b| $ for $a, b > 0$ to find 
\begin{equation}
    \left|\frac{1}{F_\tau^q} - \frac{1}{F_0^q}\right| \leq q \left(\frac{2}{m_\beta}\right)^{q+1} |F_\tau - F_0| \leq q J_\beta \left(\frac{2}{m_\beta}\right)^{q+1} nD^2 \tau,
\end{equation}
after which we use that $\left|\prod_{i=1}^m\partial_1^{\alpha_i}\partial_u^{\ell_i}F_0\right| \leq J_\beta^m D^k$ to find
\begin{equation}
    \left|\frac{\prod_{i=1}^m\partial_1^{\alpha_i}\partial_u^{\ell_i}F_\tau}{F_\tau^q} - \frac{\prod_{i=1}^m\partial_1^{\alpha_i}\partial_u^{\ell_i}F_0}{F_0^q}\right| \leq \left(mJ_\beta^m \left(\frac{2}{m_\beta}\right)^q + qJ_\beta^{m+1} \left(\frac{2}{m_\beta}\right)^{q+1}\right) nD^{k+2}\tau.
\end{equation}
In particular, from (\ref{eqn:proofappendixrepresentationlowerboundquotientrule1}), (\ref{eqn:proofappendixrepresentationlowerboundquotientrule2}) we see that $|c| \leq 6$ and so
\begin{equation}
   \sup_{z \in \r^n} \left|\partial_1^\eta S_\tau(e^\tau z, 0) - \partial_1^\eta S_0(z, 0)\right| \leq K_\beta n D^{\eta + 2}\tau \quad (\forall \eta \in \{1, 2, 3\})
\end{equation}
for all $\tau \leq \frac{1}{K_\beta nD^2}$, where we define the constant
\begin{equation}
    K_\beta := \frac{2J_\beta}{m_\beta} + \frac{60J_\beta^2}{m_\beta^2} + \frac{400J_\beta^3}{m_\beta^3} + \frac{960J_\beta^4}{m_\beta^4} + \frac{768J_\beta^5}{m_\beta^5} \geq \frac{2J_\beta}{m_\beta}. 
\end{equation}
Since we have $\partial_1^\eta S_0 = \partial_1^\eta \Psi$, we find that
\begin{equation}
    \|\partial_1^\eta s_\tau^{\psi, \Psi}\|_\infty \geq {\|\partial_1^\eta\Psi\|_\infty} - K_\beta nD^{\eta + 2} \tau,
\end{equation}
from which the claim follows. Since $(b_\tau^{\psi, \Psi}(z, u))_{n+1} = (s_\tau^{\psi, \Psi}(z, u))_{n+1} + u$, the $\partial_1$ derivatives of the last coordinate are unchanged if we replace $s_\tau^{\psi, \Psi}$ with $b_\tau^{\psi, \Psi}$.

It remains to find a more explicit upper bound for $K_\beta$. Since $J_\beta \geq 115B$ we know $J_\beta \geq m_\beta$, and so
\begin{equation}
    K_\beta \leq 2190 \left(\frac{J_\beta}{m_\beta}\right)^5.
\end{equation}
Since $J_\beta \leq 20000(115 + 5\beta) + 1000 \left(\frac{112}{e\beta^2}\right)^{7/2} \leq 4.5 \cdot 10^8 (1 + \beta + \beta^{-7})$ and there are the inequalities $(1+\beta + \beta^{-7}) \leq 2e^{\beta^2/2}(1+\beta^{-7})$, $(1+\beta^{-7})^5 \leq 16 (1+\beta^{-35})$, we may use $m_\beta = \frac{1}{\sqrt{2\pi}} e^{-\beta^2 / 2}$ to find that
\begin{equation}
    K_\beta \leq 10^{52}e^{5\beta^2} (1+\beta^{-35}).
\end{equation}
Defining $L_\beta := 10^{-52}e^{-5\beta^2} (1+\beta^{-35})^{-1} / 2$ yields the claim.
\end{proof}

For a Barron space which controls third derivatives, this implies a lower bound on the Barron norms of the scores:

\begin{corollary}[Shallow representation lower bound]
\label{corollary:representationlowerbound}
Let $\psi, \Psi$ be as in Lemma \ref{lemma:stupidnonsenserevised} and define $D,L_\beta$ accordingly. Then, the following hold:
\begin{enumerate}
    \item[(i)] Let $\phi \in C^3(\r)$ be bounded with $\|\phi'''\|_\infty < \infty$  and let $\b_\phi^R$ be its Barron space with input weights bounded by $R$ (see Definition \ref{def:barronspaces}). Then,
    \begin{equation}
       \inf_{R \in (0, \infty)} R^2 \mathcal{E}_{\b_\phi^R}(\pi^{\psi, \Psi}) \geq \frac{L_\beta}{2nD^5} \cdot \max_{i \in \{1, \ldots, n\}}{\|\partial_i^3 \Psi\|_\infty^2}.
    \end{equation}
    \item[(ii)] Let $\phi: \r \to \r$ be positively 3-homogeneous (e.g. $\operatorname{ReLU}^3$) and let $\b_\phi$ be its Barron space (see Definition \ref{def:barronspaceshomogeneous}). Then,
     \begin{equation}
       \mathcal{E}_{\b_\phi}(\pi^{\psi, \Psi})\geq \frac{L_\beta}{2nD^5} \cdot \max_{i \in \{1, \ldots, n\}}{\|\partial_i^3 \Psi\|_\infty^2}.
    \end{equation}
\end{enumerate}
\end{corollary}
\begin{proof}
The lower bound in Lemma \ref{lemma:stupidnonsenserevised} holds for any coordinate $i \in \{1, \ldots, n\}$.
If $\phi$ is as in (i) then any Barron function $v(x) = \sum_{j}\mu_j \phi(\la a_j, x\ra + b_j) $  has \begin{equation}|\partial_1^3 v(x)| \leq \|\phi'''\|_\infty \sum_j \|\mu_j\| \|a_j\|^3 \leq \|\phi'''\|_\infty R^2 \|v\|_{\b_\phi^R}.\end{equation}
If instead $\phi$ is $3$-homogeneous as in (ii), then the same argument works since $\|\phi'''\|_{L^\infty} < \infty$.
\end{proof}

We also verify that the scores can lie in the Barron space under sufficient smoothness and tail behavior. This is a sanity check to ensure that the result of Corollary \ref{corollary:representationlowerbound} is quantitative and not simply a disguised version of the statement that ``the scores are not Barron". First, we define a class of functions for which this is simple to verify:

\begin{lemma}[Cylindrical Schwartz functions are Barron w.r.t. $\operatorname{ReLU}^s$]
\label{lemma:appendixcylindricalschwartz}
    Let $n \in \n$ and $\lambda \in (0, \infty)$. We define $\mathcal{C}_\lambda$ to be the class of functions $f: \r^n \times \r \to \r$ which can be written as a finite sum
    \begin{equation}
        f(z, u) = \sum_{r=1}^M F_r(P_{E_r}z, u)
    \end{equation}
    where each $E_r \subset \r^n$ is a linear subspace (with orthogonal projection $P_{E_r}$) and each $F_r: E_r \times \r \to \r$ satisfies the seminorm bounds
    \begin{equation}
        |\partial_x^\alpha \partial_u^j F_r(x, u)| \lesssim_{\alpha, j, N} (1+\|x\|)^{-N} (1+|u|)^{-\lambda - j}
    \end{equation}
    for all $N \in \n$, all multi-indices $\alpha$, and all $j \in \n \cup \{0\}$. 
    
    The class $\mathcal{C}_\lambda$ is an algebra which is closed under smooth composition and OU smoothing, i.e. the following properties hold:
    \begin{enumerate}
        \item[(i)] If $f, g \in \mathcal{C}_\lambda$, then $fg \in \mathcal{C}_\lambda$ with seminorms controlled by those of $f,g$ and the angles between the involved subspaces.
        \item[(ii)] If $f \in \mathcal{C}_\lambda$ and $\eta \in C^\infty(\r)$ satisfies $\eta(0) = 0$ and all derivatives of $\eta$ are bounded, then $\eta \circ f \in \mathcal{C}_\lambda$ with seminorms controlled by those of $f$, the derivative bounds on $\eta$, and the angles between the involved subspaces.
        \item[(iii)] If $f \in \mathcal{C}_\lambda$ and $g(z, u) = \mathbb{E}_{G \sim \gamma_n}[f(az + b G, u)]$ for some $0 < a_- \leq a \leq a_+ < \infty$ and $0 \leq b \leq b_+ < \infty$, then $g \in \mathcal{C}_\lambda$ with seminorms controlled by those of $f$ with constants depending only on $a_-, a_+, b_+$, and subspace dimensions.
    \end{enumerate}
    
    Furthermore, for any $s  \in \n$ with $\phi(t) = \max\{0, t\}^s$ the $\operatorname{ReLU}^s$ activation and $\b_\phi$ its associated Barron space of functions from $\r^{n+1} \to \r$ (using the scalar version of Definition \ref{def:barronspaceshomogeneous}), we have $\mathcal{C}_\lambda \subset \b_{\phi}$ with Barron norm controlled by the $\mathcal{C}_\lambda$ seminorms.
\end{lemma}
\begin{proof}
    For any profile $F: E \times \r \to \r$, we denote the seminorm
    \begin{equation}
        [F]_{\alpha, j, N} := \sup_{x \in E,\; u \in \r} (1+\|x\|)^N (1+|u|)^{\lambda + j} |\partial_x^\alpha \partial_u^j F(x, u)|.
    \end{equation}
    
    \textbf{(i).} For this, it suffices to consider two summands
    \begin{equation}
        f(x, u) = F(P_Ex, u), \quad g(x, u) = G(P_Vx, u).
    \end{equation}
    Denote the subspace $W := E + V$ and write $H(x, u) := F(P_Ex, u) G(P_Vx, u)$. Then, $(fg)(x, u) = H(P_Wx, u)$. The linear map $W \ni x \mapsto (P_Ex, P_Vx)$ is injective, and hence by finite-dimensionality $\|x\| \leq C_{E,V} (\|P_Ex\| + \|P_Vx\|)$ and so \begin{equation}
    \label{eqn:proofappendixcylindricalschwartzlemmapart1}
    1 + \|x\| \lesssim (1+\|P_Ex\|)(1+\|P_Vx\|).\end{equation} By the product rule, $\partial_x^\alpha \partial_u^jH(x, u)$ is a finite sum of products $(\partial_x^{\alpha_1}\partial_u^{j_1}F)(P_Ex, u)(\partial_x^{\alpha_2}\partial_u^{j_2}G)(P_Vx, u)$ with $\alpha_1 + \alpha_2 = \alpha$ and $j_1 + j_2 = j$. Using (\ref{eqn:proofappendixcylindricalschwartzlemmapart1}),
    \begin{equation}
        |\partial_x^\alpha \partial_u^jH(x, u)| \lesssim_{\alpha, j, N} (1+\|x\|)^{-N}(1+|u|)^{-2\lambda - j} \lesssim (1+\|x\|)^{-N}(1+|u|)^{-\lambda - j} 
    \end{equation}
    since $\lambda > 0$. Hence, $fg \in \mathcal{C}_\lambda$ with controlled seminorms. The result extends to $f, g$ being any finite sums by linearity and the triangle inequality.

    \textbf{(ii).} Write $f(z, u) = \sum_{r=1}^M F_r(P_{E_r}z, u)$ and define $B := \sum_{r=1}^M \|F_r\|_\infty$. We use the anchored multivariate fundamental theorem of calculus: if 
    \begin{equation}
        \Theta(y_1, \ldots, y_M) := \eta\left(\sum_{r=1}^M y_r\right)
    \end{equation}
    then $\Theta(0) = \eta(0) = 0$ and 
    \begin{equation}
        \Theta(y) = \sum_{\emptyset \neq J \subset \{1, \ldots, M\}} \left(\prod_{r \in J}y_r\right) \int_{[0, 1]^{|J|}} \eta^{(|J|)} \left(\sum_{r \in J} t_r y_r\right)dt.
    \end{equation}  
    Substituting in $y_r = F_r(P_{E_r}z, u)$ gives a finite decomposition of $\eta \circ f$. Fix a nonempty $J \subset \{1, \ldots, M\}$ and define the subspace $E_J := \sum_{r \in J} E_r \subset \r^n$. The corresponding term in $\eta \circ f$, denoted $H_J$, has 
    \begin{equation}
    \label{eqn:proofappendixcylindricalschwartzlemmapart2}
        H_J(x, u) = \prod_{r \in J} F_r(P_{E_r}x, u) \int_{[0, 1]^{|J|}} \eta^{(|J|)} \left(\sum_{r \in J}t_r F_r(P_{E_r}x, u)\right)dt
    \end{equation}
    for every $x \in E_J$. Since the arguments of $\eta^{(|J|)}$ lie in $[-B, B]$, all derivatives of the integral factor are bounded due to the assumed bounds on $\eta$, Faà di Bruno, and the seminorm bounds on $F_r$. As before, the map $E_J \ni x \mapsto (P_{E_r}x)_{r\in J}$ is injective and therefore
    \begin{equation}
        1+\|x\| \lesssim_J \prod_{r \in J}(1+\|P_{E_r}x\|).
    \end{equation}
    Differentiating (\ref{eqn:proofappendixcylindricalschwartzlemmapart2}) and applying Faà di Bruno, each resulting term is a product of finitely many derivatives of $F_r(P_{E_r}x, u)$ multiplied by a bounded derivative of $\eta$. Due to the prefactor, every such term contains at least one value or derivative of $F_r(P_{E_r}x, u)$ for each $r \in J$. Hence, using the seminorm definition of each $F_r$ implies
    \begin{equation}
        |\partial_x^\alpha \partial_u^j H_J(x, u)| \lesssim_{\alpha, j, N} (1+\|x\|)^{-N}(1+|u|)^{-|J|\lambda - j}.
    \end{equation}
    Since $J$ is nonempty and $\lambda > 0$, this proves the second claim. 

    \textbf{(iii).} By linearity, it suffices to consider a single summand $f(z, u) = F(P_Ez, u)$. Since $P_EG \sim \gamma_E$, we have
    \begin{equation}
        g(z, u) = \mathbb{E}_{G \sim \gamma_n}[F(P_E(az + bG), u)] = \widetilde{F}(P_Ez, u), \quad \widetilde{F}(x, u) := \mathbb{E}_{\xi \sim \gamma_E}[F(ax + b\xi, u)].
    \end{equation}
    Hence, $g$ is cylindrical on the same subspace. Furthermore, by differentiating under the expectation we find
    \begin{equation}
    \label{eqn:proofappendixcylindricalschwartzlemmapart3}
        \partial_x^\alpha \partial_u^j \widetilde{F}(x, u) = a^{|\alpha|} \mathbb{E}_{\xi \sim \gamma_E} [\partial_x^\alpha \partial_u^jF(ax+b\xi, u)].
    \end{equation}
    Since $a \geq a_-$ and $b \leq b_+$ we know that
    \begin{equation}
        \|x\| = a^{-1}\|ax\| \leq a_{-}^{-1} \|ax + b\xi\| + a_-^{-1} b_+\|\xi\|
    \end{equation}
    by the triangle inequality, and so
    \begin{equation}
        (1+\|x\|)^N \lesssim_{N, a_-, b_+} (1+\|ax+b\xi\|)^N (1+\|\xi\|)^N.
    \end{equation}
    Combining this with (\ref{eqn:proofappendixcylindricalschwartzlemmapart3}) and $a \leq a_+$, we get
    \begin{align}
        (1+\|x\|)^N (1+|u|)^{\lambda + j}| \partial_x^\alpha \partial_u^j \widetilde{F}(x, u)| &\lesssim a_+^{|\alpha|} \mathbb{E}_{\xi \sim \gamma_E}\left[(1+\|\xi\|)^N (1+\|ax+b\xi\|)^N (1+|u|)^{\lambda + j}| \partial_x^\alpha \partial_u^j {F}(ax+b\xi, u)|\right] \\
        &\lesssim_{\alpha, N, a_-, a_+, b_+} [F]_{\alpha, j, N} \mathbb{E}_{\xi \sim \gamma_E}[(1+\|\xi\|)^N]. 
    \end{align}
    Since the Gaussian has all moments finite, this implies $g \in \mathcal{C}_\lambda$ with the claimed seminorm control. 

    \textbf{(Barron embedding).} We will apply a global Sobolev-to-Barron embedding, whose proof we defer to Lemma \ref{lemma:appendixsobolevtobarron}. By the triangle inequality, it suffices to consider a single summand $f(z, u) = F(P_Ez, u)$. Write $D = \dim(E)$ and identify $E$ with $\r^D$. Fix $N > D+s$ and $A_F := \sum_{|\alpha| + j \leq D+s+2}[F]_{\alpha, j, N}$. Choose a radial cutoff $\chi \in C^\infty_c(\r)$ equal to 1 on $[-1, 1]$ and supported in $[-2, 2]$, and set
    \begin{equation}
        \zeta(u) := \chi(u/2) - \chi(u).
    \end{equation}
    Then, $\zeta$ is supported on $1 \leq |u| \leq 4$ and 
    \begin{equation}
    \label{eqn:proofappendixbarronembeddingupperboundpartitionofunity}
        1 = \chi(u) + \sum_{q = 0}^\infty \zeta(2^{-q}u) \quad (\forall u \in \r).
    \end{equation}
    Using this, we define
    \begin{equation}
        F_q(x, u) := \zeta(2^{-q}u) F(x, u), \quad F_{-1}(x, u) = \chi(u)F(x, u) \quad (\forall q \in \n \cup \{0\}).
    \end{equation}
    For $q \in \n \cup \{0\}$ we rescale the last coordinate
    \begin{equation}
        \widetilde{F}_q(x, u) := F_q(x, 2^qu) = \zeta(u) F(x, 2^qu).
    \end{equation}
    By the product and chain rules in the $u$ coordinate,
    \begin{equation}
        \partial_x^\alpha \partial_u^j \widetilde{F}_q(x, u) = \sum_{\ell = 0}^j {j \choose \ell} \zeta^{(j - \ell)} (u)2^{q\ell}(\partial_x^\alpha \partial_u^\ell F)(x, 2^q u).
    \end{equation}
    Now, on the support of $\zeta^{(j - \ell)}$ we have $|u| \geq 1$, in which case $2^{q\ell}(1 + 2^q|u|)^{-\lambda - \ell} \leq 2^{-q \lambda}$. This means that for all $|\alpha| + j \leq D+s+2$ we have
    \begin{align}
        \left|\partial_x^\alpha \partial_u^j \widetilde{F}_q(x, u)\right| \lesssim_{D,s} A_F 2^{-q\lambda} (1+\|x\|)^{-N}
    \end{align}
    and these derivatives vanish for $|u| > 4$. Since $N > D+s$, we find
    \begin{align}
        \sum_{|\alpha| + j \leq D+s+2} \int_{\r^D \times \r} (1+\|(x, u)\|)^s \left|\partial_x^\alpha \partial_u^j \widetilde{F}_q(x, u)\right| dxdu &\lesssim_{D,s,N} A_F 2^{-q\lambda} \int_{\r^D} (1+\|x\|)^{s-N}dx \\
        &\lesssim_{D,s,N} A_F 2^{-q\lambda}.
    \end{align}
    Applying Lemma \ref{lemma:appendixsobolevtobarron} yields 
    \begin{equation}
        \|\widetilde{F}_q\|_{\b_\phi} \lesssim_{D,s,N} A_F2^{-q\lambda}.
    \end{equation}
    Now, since $F_q(x, u) = \widetilde{F}_q(x, 2^{-q}u)$ we may obtain a Barron representation of $F_q$ from that of $\widetilde{F}_q$ by replacing $\phi(\la a, x\ra + a' u + b)$ with $\phi(\la a, x\ra + 2^{-q}a'u + b)$. This can only decrease the Barron cost, and so
    \begin{equation}
        \|F_q\|_{\b_\phi} \lesssim_{D,s,N}A_F 2^{-q\lambda}. 
    \end{equation}
    A similar application of Lemma \ref{lemma:appendixsobolevtobarron} yields that
    \begin{equation}
        \|F_{-1}\|_{\b_\phi} \lesssim_{D,s,N} A_F.
    \end{equation}
    We may apply (\ref{eqn:proofappendixbarronembeddingupperboundpartitionofunity}) to find that
    \begin{equation}
        F_{-1}(x, u) + \sum_{q=0}^Q  F_q(x, u) = \chi(2^{-Q-1}u) F(x, u)
    \end{equation}
    and so they converge locally uniformly to $F$ as $Q \to \infty$. By lower-semicontinuity of the Barron norm under local uniform convergence (\ref{lemma:appendixbarronnormlscunderlocaluniformconvergence}) and the triangle inequality,
    \begin{align}
        \|F\|_{\b_\phi} \leq \|F_{-1}\|_{\b_\phi} + \sum_{q=0}^\infty \|F_q\|_{\b_\phi} \lesssim_{D,s,N} A_F \left(1 + \frac{1}{1-2^{-\lambda}}\right) < \infty.
    \end{align}
    This gives a Barron representation on $E \times \r \subset \r^n \times \r$, which extends to one on the whole space without changing the parameter norms (since the neurons only see $E \times \r$). 
\end{proof}

\begin{lemma}[Global Sobolev-to-Barron embedding for $\operatorname{ReLU}^s$]
\label{lemma:appendixsobolevtobarron}
Fix any $D,s  \in \n$ with $\phi(t) = \max\{0, t\}^s$ the $\operatorname{ReLU}^s$ activation and $\b_\phi$ its associated Barron space of functions from $\r^{D} \to \r$ (using the scalar version of Definition \ref{def:barronspaceshomogeneous}). 

Then, there is some $C_{D,s} \in (0, \infty)$ s.t. for any $f \in W^{D+s+1,1}(\r^D)$ if we choose its continuous representative we have
\begin{equation}
    \|f\|_{\b_{\phi}} \leq C_{D,s} \sum_{|\alpha| \leq D+s+1} \int_{\r^D} (1+|x|)^s |\partial^\alpha f(x)| dx.
\end{equation}
\end{lemma}
\begin{proof}
    First, suppose that $D$ is odd and $f \in C^\infty_c(\r^D)$. Write its Radon transform by
    \begin{equation}
        Rf(\theta, t) := \int_{\theta^\perp} f(t\theta + y)dy.
    \end{equation}
    By the odd-dimensional Radon inversion formula (see (27) and (28) of \cite{parhi2021banach}), we have
    \begin{equation}
        f(x) = c_D \int_{\mathbb{S}^{D-1}} (\partial_t^{D-1} Rf)(\theta, \la \theta, x\ra) d\theta, \quad c_D = \frac{(-1)^{(D-1)/2}}{2(2\pi)^{D-1}}.
    \end{equation}
    We have seen that for every $g \in C^\infty_c(\r)$, repeated integration by parts gives
    \begin{equation}
        g(v) = \frac{1}{s!}\int_\r \max\{0, v-t\}^s g^{(s+1)}(t) dt.
    \end{equation}
    Applying this to $g(\cdot) = \partial_t^{D-1}Rf(\theta, \cdot)$ gives the global representation
    \begin{equation}
        f(x) = \frac{c_D}{s!} \int_{\mathbb{S}^{D-1}} \int_\r \max\{0, \la \theta, x\ra - t\}^s (\partial_t^{D+s} Rf)(\theta, t) dtd\theta.
    \end{equation}
    By differentiating under the integral, there is the Radon transform identity
    \begin{align}
        (\partial_t^k Rf)(\theta, t) &= \int_{\theta^\perp} \partial_t^k [f(t\theta + y)] dy \\
        &= \int_{\theta^\perp} (\theta \cdot \nabla)^k f(t\theta + y) dy \\
        &= R[(\theta \cdot \nabla)^k f](\theta, t),
    \end{align}
    where $(\theta \cdot \nabla)^k f$ is the $k^{th}$ directional derivative of $f$ in direction $\theta$.
    Since $\partial_t^{D+s}Rf = R[(\theta \cdot \nabla)^{D+s}f]$, the homogeneous Barron cost of this representation is 
    \begin{align}
        \|f\|_{\b_\phi} &\leq \frac{|c_D|}{s!} \int_{\mathbb{S}^{D-1}} \int_\r (1+|t|)^s |R[(\theta \cdot \nabla)^{D+s}f](\theta, t)| dtd\theta  \\&=  
        \frac{|c_D|}{s!} \int_{\mathbb{S}^{D-1}} \int_\r (1+|t|)^s \left|\int_{\theta^\perp} (\theta \cdot \nabla)^{D+s} f(t\theta + y)dy\right|dtd\theta 
        \\
        &\leq \frac{|c_D|}{s!} \int_{\mathbb{S}^{D-1}} \int_\r \int_{\theta^\perp} (1+|t|)^s \left|(\theta \cdot \nabla)^{D+s}f(t\theta + y)\right| dy dt d\theta \\
        &= \frac{|c_D|}{s!} \int_{\mathbb{S}^{D-1}}\int_{\r^D} (1+|\la \theta, x\ra|)^s \left|(\theta \cdot \nabla)^{D+s}f(x)\right| dx d\theta,
\end{align}
    where the last line fixes $\theta$ and uses the change of coordinates $x = t\theta + y$ (which has Jacobian 1).
    By the multinomial formula, we know
    \begin{equation}
        (\theta \cdot \nabla)^{D+s}f = \sum_{|\alpha| = D+s} \frac{(D+s)!}{\prod_j \alpha_j!} \theta^\alpha \partial^\alpha f.
    \end{equation}
    Since $|\theta^\alpha| \leq 1$, $|\la \theta, x\ra| \leq \|x\|$, and $\prod_j \alpha_j! \geq 1$, we find that
    \begin{equation}
    \label{eqn:proofappendixsobolevtobarronoddD}
        \|f\|_{\b_\phi}\leq  \frac{|c_D| \cdot |\mathbb{S}^{D-1}| (D+s)!}{s!} \sum_{|\alpha| = D+s}  \int_{\r^D} (1+\|x\|)^s |\partial^\alpha f(x)|dx.
    \end{equation}
    Setting $C_{D,s} := \frac{|c_D| \cdot |\mathbb{S}^{D-1}| (D+s)!}{s!}$ yields the claim for odd $D$ and $f \in C^\infty_c(\r^D)$. 

    Letting now $D$ be even and $f \in C^\infty_c(\r^D)$, we fix $\eta \in C^\infty_c(\r)$ with $\eta(0) = 1$ and define
    \begin{equation}
        F(x, u) := f(x) \eta(u).
    \end{equation}
    Restricting any Barron representation of $F$ to $u = 0$ gives a Barron representation of $f$ with no increase in parameter norm, since each $\phi(\la a, x\ra + a' u + b)$ becomes $\phi(\la a, x\ra + b)$. Hence,
    \begin{equation}
        \|f\|_{\b_\phi} \leq \|F\|_{\b_\phi}.
    \end{equation}
    Let us define for even $D$ the constant
    \begin{equation}
        C_{D,s} := C_{D+1,s} \max_{0 \leq j \leq D+s+1}\int_\r (1+|u|)^s |\eta^{(j)}(u)| du
    \end{equation}
    for the fixed $\eta$ (which we will use for all $f: \r^D \to \r$).
    Using that $D+1$ is odd, we apply (\ref{eqn:proofappendixsobolevtobarronoddD}) along with the product rule to see that
    \begin{align}
        \|F\|_{\b_\phi} &\leq C_{D+1, s} \sum_{|\alpha| + j = D + s + 1} \int_{\r^D} \int_\r (1 + \|(x, u)\|)^s |\partial^\alpha f(x)| |\eta^{(j)}(u)| du dx \\
        &\leq C_{D+1, s} \sum_{|\alpha| + j = D+s+1} \left(\int_{\r^D} (1+\|x\|)^s |\partial^\alpha f(x)| dx\right)\left(\int_\r (1+|u|)^s |\eta^{(j)}(u)| du\right) \\
        &\leq {C}_{D,s} \sum_{|\alpha| \leq D+s+1}\int_{\r^D} (1+\|x\|)^s |\partial^\alpha f(x)| dx,
    \end{align}
    where we used that $1 + \|(x, u)\| \leq (1+\|x\|)(1+|u|)$. Hence, the claim holds for any $D\in \n$ and any $f \in C^\infty_c(\r^D)$.
    For general $f$ with finite right hand side, we take $f_j \in C^\infty_c(\r^D)$ s.t. $f_j \to f$ locally uniformly and
    \begin{equation}
        \sum_{|\alpha| \leq D+s+1} \int_{\r^D} (1+|x|)^s |\partial^\alpha (f_j - f)(x)| dx \to 0
    \end{equation}
    using cutoff and mollification. By lower-semicontinuity of the Barron norm w.r.t. local uniform convergence (Lemma \ref{lemma:appendixbarronnormlscunderlocaluniformconvergence}), the claim follows.
\end{proof}

\begin{lemma}[Shallow representation upper bound, nonquantitative]
\label{lemma:appendixrepresentationupperbound}
    Fix $s  \in \n$ and let $\phi(t) = \max\{0, t\}^s$ be the $\operatorname{ReLU}^s$ activation with $\b_\phi$ its associated Barron space (see Definition \ref{def:barronspaceshomogeneous}).
     Let $n \in \n$ and suppose $\Psi(z) = \sum_{r=1}^M \Phi_r(P_{E_r}z)$, where each $E_r \subset \r^n$ is a linear subspace (with orthogonal projection $P_{E_r}$) and $\Phi_r \in C^\infty_c(E_r)$.
     Let $\psi \in C^\infty(\r)$ be a strictly increasing diffeomorphism and choose $a_\pm, R \in (0, \infty)$, $b_\pm \in \r$, and $p > 2$ s.t.
    \begin{equation}
        \psi(u) = a_+ u^p + b_+ \quad (\forall u \geq R), \quad \psi(u) = -a_- (-u)^p + b_- \quad (\forall u \leq -R).
    \end{equation}
    Then, $s_\tau^{\psi, \Psi}, b_\tau^{\psi, \Psi} \in \b_\phi$ for all $\tau \geq 0$ with
    \begin{equation}
      \sup_{\tau \in [0, T]} \max\left\{\| s_\tau^{\psi, \Psi}\|_{\b_\phi}, \| b_\tau^{\psi, \Psi}\|_{\b_\phi}\right\} < \infty \quad (\forall T \in [0, \infty)).
    \end{equation}
\end{lemma}
\begin{proof}
Fix $T > 0$ and let $q_\tau$ be the density of $e^{-\tau}\psi(U) + \sigma H$ for $U, H \sim \gamma_1$ independent and $\sigma := \sqrt{1-e^{-2\tau}}$ (as in the proof of Lemma \ref{lemma:stupidnonsenserevised}).  Denote its score by $h_\tau := q_\tau ' / q_\tau$. Since $\psi$ is exactly a power function of exponent $p$ outside of a compact set, the change of variables formula gives that
   \begin{equation}
   \label{eqn:proofappendixrepresentationupperboundOUsmoothingpreserves1000}
       q_0(u) = \frac{1}{\sqrt{2\pi}pa_+^{1/p}} (u-b_+)^{-(p-1)/p} \exp\left(-\frac{a_+^{-2/p}}{2}(u-b_+)^{2/p}\right)
   \end{equation}
  for $u$ large and therefore if $\lambda := 1-\frac{2}{p}$ we have
   \begin{equation}
       h_0(u) = -\frac{a_+^{-2/p}}{p}(u-b_+)^{-\lambda} - \frac{p-1}{p(u-b_+)}.
   \end{equation}
   The analogous formula holds on the negative tail. Hence, for every $j \in \n \cup \{0\}$ we have
   \begin{equation}
       |h_0^{(j)}(u)| \lesssim_j (1+|u|)^{-\lambda -j}.
   \end{equation}
   We claim that this estimate persists under OU smoothing, i.e. that
\begin{equation}
\label{eqn:proofappendixrepresentationupperboundOUsmoothingpreserves}
       \sup_{\tau \in [0, T]} \sup_{u \in \r} (1+|u|)^{\lambda + j} |h_\tau^{(j)}(u)| < \infty \quad (\forall j \in \n \cup \{0\}).
   \end{equation}
Indeed, since $q_\tau(\cdot) = e^\tau q_0(e^\tau \cdot) * \varphi_{\sigma^2}$ we may denote $ \tilde{q}_\tau(u) :=  e^\tau q_0(e^\tau u)$ and write
\begin{equation}
    q_\tau(u) = \tilde{q}_\tau(u) G_\tau(u),\quad G_\tau(u) := \mathbb{E}\left[\exp\left(\ell_\tau(u - \sigma H) - \ell_\tau(u)\right)\right],  \quad \ell_\tau := \log \tilde{q}_\tau.
\end{equation}
From (\ref{eqn:proofappendixrepresentationupperboundOUsmoothingpreserves1000}) we know that
\begin{equation}
\label{eqn:proofappendixrepresentationupperboundOUsmoothingpreserves101}
    |\ell_\tau^{(j+1)}(u)| \lesssim_{j,T} (1+|u|)^{-\lambda - j}
\end{equation}
uniformly for $\tau \in [0, T]$.
Since $\|\ell_\tau'\|_\infty \leq C_T$ is uniformly bounded for $\tau \in [0, T]$ we know $G_\tau(u) \leq \mathbb{E}[e^{C_T|H|}] < \infty$. Similarly, we know $G_\tau(u) \geq e^{-C_T}\mathbb{P}(|H| \leq 1) \gtrsim_T 1$, and therefore
\begin{equation}
    G_\tau(u) \asymp_T 1
\end{equation}
uniformly for $\tau \in [0, T]$.
 On the event $|\sigma H| \leq |u|/2$, by the mean value theorem and (\ref{eqn:proofappendixrepresentationupperboundOUsmoothingpreserves101}) we have
\begin{equation}
\label{eqn:proofappendixrepresentationupperboundOUsmoothingpreservesmeanvaluenonsense}
\left|\ell_\tau^{(j)}(u-\sigma H) - \ell_\tau^{(j)}(u)\right| \lesssim_{j,T}|H|(1+|u|)^{-\lambda - j}  \quad (\forall j \in \n \cup \{0\}),
\end{equation}
 By Faà di Bruno, for $j \in \n$ on this event we know that
\begin{align}
    |\partial_u^j e^{\ell_\tau(u-\sigma H) - \ell_\tau(u)}| &\lesssim_{j,T} e^{C_T |H|} \sum_{\substack{m, r_1, \ldots, r_m \in \n \\ r_1 + \ldots + r_m = j}}|H|^m (1+|u|)^{-m\lambda - j}\\
    &\lesssim_{j,T} e^{C_T |H|} (1+|H|)^j (1+|u|)^{-\lambda - j},
\end{align}
where the second inequality uses $\lambda > 0$. On the other hand, the bound (\ref{eqn:proofappendixrepresentationupperboundOUsmoothingpreserves101}) implies that
\begin{equation}
     |\partial_u^j e^{\ell_\tau(u-\sigma H) - \ell_\tau(u)}| \lesssim_{j,T} e^{c_T|H|} (1+|H|)^j,
\end{equation}
and since
\begin{equation}
\mathbb{E}\left[e^{C_T|H|} (1+|H|)^j \1_{|\sigma H| > |u|/2}\right] \lesssim_{j,T}e^{-c_T u^2}
\end{equation}
for some constant $c_T$ we find that
\begin{equation}
    |G_\tau^{(j)}(u)| \lesssim_{j,T} (1+|u|)^{-\lambda - j} \quad (\forall j \in \n).
\end{equation}
Since $G_\tau \asymp_T 1$, we find that
\begin{equation}
\label{eqn:proofappendixrepresentationupperboundOUsmoothingpreserves102}
     |\partial^j \log G_\tau(u)| \lesssim_{j,T} (1+|u|)^{-\lambda - j} \quad (\forall j \in \n).
\end{equation}
However, since $\log q_\tau = \ell_\tau + \log G_\tau$ and so $h_\tau^{(j)} = \ell_\tau^{(j+1)} + \partial^{(j+1)} \log G_\tau$, we may apply (\ref{eqn:proofappendixrepresentationupperboundOUsmoothingpreserves102}) and (\ref{eqn:proofappendixrepresentationupperboundOUsmoothingpreserves101}) to show the claim (\ref{eqn:proofappendixrepresentationupperboundOUsmoothingpreserves}).

Moving on, (\ref{eqn:proofappendixrepresentationupperboundOUsmoothingpreserves}) implies that $(1+|u|)^s |h_\tau^{(s+1)}(u)| \lesssim (1+|u|)^{-1 - \lambda}$ and so
   \begin{equation}
   \label{eqn:proofappendixrepresentationupperboundOUsmoothingpreserves2}
       \sup_{\tau \in [0, T]} \int_\r (1+|u|)^s |h_\tau^{(s+1)}(u)|du < \infty.
   \end{equation}
   The bound (\ref{eqn:proofappendixrepresentationupperboundOUsmoothingpreserves}) also implies that $h_\tau^{(j)} \to 0$ as $u \to -\infty$, and so repeated application of the fundamental theorem of calculus yields
   \begin{equation}
       h_\tau(v) = \frac{1}{s!} \int_\r h_\tau^{(s+1)}(u) \max\{0, v-u\}^s du.
   \end{equation}
   This is a valid Barron representation, and so (\ref{eqn:proofappendixrepresentationupperboundOUsmoothingpreserves2}) yields $\sup_{\tau \in [0, T]}\|h_\tau\|_{\b_{\operatorname{ReLU}^s}}<\infty$.

Next, using the notation from the proof of Lemma \ref{lemma:stupidnonsenserevised} we have the density
   \begin{equation}
       \rho_\tau^{\psi, \Psi}(z, u) = \gamma_n(z) F_\tau(z, u), \quad F_\tau(z, u) = \mathbb{E}_{G \sim \gamma_n}[q_\tau(u - e^{-\tau}\Psi(e^{-\tau}z + \sqrt{1-e^{-2\tau}}G))].
   \end{equation}
   If we define the quantity
   \begin{equation}
       R_\tau(z, u) := \frac{F_\tau(z, u)}{q_\tau(u)}, \quad S_\tau := \nabla \log R_\tau,
   \end{equation}
   then using that
   \begin{equation}
    \log \rho_\tau^{\psi, \Psi}(z, u) = \log \gamma_n(z) + \log q_\tau(u) + \log R_\tau(z, u)
   \end{equation}
   we find that the scores satisfy
   \begin{equation}
   \label{eqn:proofappendixrepresentationupperboundOUsmoothingpreserves3}
       s_\tau^{\psi, \Psi}(z, u) = \begin{bmatrix}
           -z \\ h_\tau(u)
       \end{bmatrix} + S_\tau(z, u), \quad b_\tau^{\psi, \Psi} = \begin{bmatrix}
           0 \\ u + h_\tau(u)
       \end{bmatrix} + S_\tau(z, u).
   \end{equation}
   We claim that affine vector fields are in $\b_{\operatorname{ReLU}^s}$; to see this, note that for any $r_0, \ldots, r_s$ distinct we have that $t \mapsto (t+r_0)^s, \ldots, t\mapsto (t+r_s)^s$ span the space of polynomials of degree at most $s$ (since the coefficient matrix is Vandermonde and therefore invertible). Hence, every affine map is a linear combination of these $t \mapsto (t+r_j)^s$ polynomials. However, since
   \begin{equation}
       (t+r)^s = \max\{0, t+r\}^s + (-1)^s \max\{0, -t-r\}^s
   \end{equation}
   we can apply it coordinatewise to find that affine vector fields are Barron. In light of (\ref{eqn:proofappendixrepresentationupperboundOUsmoothingpreserves3}) it remains to show that $\sup_{\tau \in [0, T]}\|S_\tau\|_{\b_{\operatorname{ReLU}^s}} < \infty$.

To this end, if we denote $B := \sum_{r=1}^M \|\Phi_r\|_\infty$ and $K_\tau(u, c) := q_\tau(u-c) / q_\tau(u)$ then by definition we have
\begin{equation}
    R_\tau(z, u) = \mathbb{E}_{G \sim \gamma_n}[K_\tau(u, e^{-\tau} \Psi(e^{-\tau}z + \sigma G))].
\end{equation}
Differentiating under the expectation, for $\alpha \neq 0$ every term $\partial_z^\alpha \partial_u^j R_\tau(z, u)$ has the form
\begin{equation}
    Ce^{-(|\alpha| + k)\tau} \mathbb{E}_{G \sim \gamma_n}\left[(\partial_c^k \partial_u^j K_\tau) (u, e^{-\tau}\Psi(e^{-\tau}z + \sigma G)) \prod_{\ell = 1}^k \partial^{\beta_\ell}\Psi(e^{-\tau }z + \sigma G)\right]
\end{equation}
with $k \geq 1$, $|\beta_\ell| \geq 1$, and $\sum_\ell \beta_\ell = \alpha$. For $\alpha = 0$, we use instead that
\begin{equation}
    \partial_u^j (R_\tau - 1) = \mathbb{E}_{G \sim \gamma_n}[\partial_u^j (K_\tau(u, e^{-\tau}\Psi(e^{-\tau}z + \sigma G))-1)].
\end{equation}
Now, since we have
\begin{equation}
    \log K_\tau(u, c) = \log q_\tau(u - c) - \log q_\tau(u) = -\int_{u-c}^u h_\tau(v) dv,
\end{equation}
the estimate (\ref{eqn:proofappendixrepresentationupperboundOUsmoothingpreserves}) on $h_\tau$ implies that $K_\tau(u, c) \asymp_T 1$ uniformly in $|c| \leq B$ and furthermore
\begin{equation}
\label{eqn:proofappendixderivativeestimatesonK_tau}
    |\partial_c^k \partial_u^j K_\tau(u, c)| \lesssim_{k,j,T} (1+|u|)^{-\lambda - j} \quad (\forall k + j \geq 1)
\end{equation}
uniformly in $|c| \leq B$.
Since $K_\tau(u, 0) = 1$, the fundamental theorem of calculus in the $c$ coordinate implies that for all $j \in \n \cup \{0\}$ we have
\begin{equation}
\label{eqn:proofappendixrepresentationupperboundestimatesonKtau}
    |\partial_u^j (K_\tau(u, c) - 1)| \lesssim_{j,T} |c|(1+|u|)^{-\lambda - j}.
\end{equation}
We define the new function $A_\tau: \r^n \times \r \to \r$ by
\begin{equation}
    A_\tau(z, u) := K_\tau(u, e^{-\tau}\Psi(z)) - 1.
\end{equation}
Apply the anchored fundamental theorem of calculus, as in the proof of Lemma \ref{lemma:appendixcylindricalschwartz}(ii), to $(y_1, \ldots, y_M) \mapsto K_\tau(u, e^{-\tau}\sum_{r=1}^M y_r) - 1$. This function vanishes at $y=0$ (since $K_\tau(u, 0) = 1$) and so
\begin{equation}
    A_{\tau}(z, u) = \sum_{\emptyset \neq J \subset \{1, \ldots, M\}} H_{\tau, J}(P_{E_J}z, u), \quad E_J := \sum_{r \in J}E_r,
\end{equation}
where for $x \in E_J$ we define
\begin{equation}
    H_{\tau, J}(x, u) := e^{-|J|\tau} \left(\prod_{r \in J}\Phi_r(P_{E_r}x)\right) \int_{[0, 1]^{|J|}} (\partial_c^{|J|}K_\tau)\left(u, e^{-\tau}\sum_{r\in J} t_r \Phi_r(P_{E_r}x)\right)dt.
\end{equation}
For each $J$ the support of $H_{\tau, J}$ in the $x$ coordinate is contained in the set
\begin{equation}
\{x \in E_J:\; P_{E_r}x \in \operatorname{spt}(\Phi_r) \; \forall r \in J\},
\end{equation}
which is compact (since $\|x\| \lesssim_J \sum_{r \in J}\|P_{E_r}x\|$). Since we know $|e^{-\tau}\sum_{r \in J}t_r \Phi_r(P_{E_r}x)| \leq B$ (and so the estimate (\ref{eqn:proofappendixderivativeestimatesonK_tau}) applies), by the product and chain rule each derivative of $H_{\tau, J}$ is a finite sum of products of values/derivatives of $\Phi_r$ with those of $K_\tau$ with the $u$ derivatives only applying to $K_\tau$ and therefore
\begin{equation}
    |\partial_x^\alpha \partial_u^j H_{\tau, J}(x, u)| \lesssim_{\alpha, j,J,T} (1+|u|)^{-\lambda - j}.
\end{equation}
Since $H_{\tau, J}$ is compactly-supported in $x$, we find that $A_\tau \in \mathcal{C}_\lambda$ uniformly over $\tau \in [0, T]$.

 Next, since $R_\tau(z, u) - 1 = \mathbb{E}_{G \sim \gamma_n}[A_{\tau}(e^{-\tau}z + \sigma G, u)]$ we apply Lemma \ref{lemma:appendixcylindricalschwartz}(iii) to find that $r_\tau := R_\tau - 1 \in \mathcal{C}_\lambda$ uniformly over $\tau \in [0, T]$. Next, since $K_\tau(u, c) \asymp_T 1$ uniformly in $|c| \leq B$ we also have $R_\tau(z, u) \asymp_T 1$. Hence, if $\eta \in C^\infty_c(\r)$ is s.t. $\eta(0) = 0$ and
\begin{equation}
    \eta(t) := \frac{1}{t+1}- 1 \quad
\end{equation}
for all $t$ in a neighborhood of the image of $r_\tau$ for all $\tau \in [0, T]$ (which is possible since $1+r_\tau = R_\tau$ is bounded above and away from 0 uniformly for $\tau \in [0, T]$), then by Lemma \ref{lemma:appendixcylindricalschwartz}(ii) we know $\eta \circ r_\tau \in \mathcal{C}_\lambda$. Furthermore, each coordinate of $\nabla r_\tau$ is also in $\mathcal{C}_\lambda$ since taking derivatives only helps with the $u$ decay. Finally, since
\begin{equation}
    S_\tau = \nabla \log R_\tau = \frac{\nabla R_\tau}{R_\tau} = \nabla r_\tau + \nabla r_\tau \left(\frac{1}{R_\tau} - 1\right) = \nabla r_\tau + \nabla r_\tau (\eta \circ r_\tau),
\end{equation}
by the algebra property in Lemma \ref{lemma:appendixcylindricalschwartz}(i) we find that each coordinate of $S_\tau$ is in $\mathcal{C}_\lambda$, uniformly in $\tau \in [0, T]$. The claim follows by applying the Barron embedding statement of Lemma \ref{lemma:appendixcylindricalschwartz} coordinatewise.
\end{proof}

\subsubsection{Approximation lower bound via Hermite spectral tails}
Throughout this subsection we will take $\psi(u) = \sigma u$ for some $\sigma > 0$ (i.e. $\psi(U)$ is centered Gaussian) and focus on the dependence on $\Psi$.
We will now lower bound the widths and Barron norms of two-layer neural network \textit{approximations} to the scores $s_\tau^{\psi, \Psi}/ b_\tau^{\psi, \Psi}$, as opposed to exact representations. This approach is more robust, leveraging high Hermite rank of $\Psi$ to directly lower bound the width of any sum of ridge functions approximating the score. The technique borrows largely from the recent paper \cite{bruna2026multiscale}, and we adapt the general lower bound method to the reverse OU application. 

Let us first define a class of functions which contains two-layer neural networks of width $m$. Since the Hermite method controls width in a general sense, the neurons can be very general ridge functions which need not share an activation function.

\begin{definition}[Shallow function class]
    For each $n, m \in \n$, we define the class $\mathcal{R}_m^{(n)}$ of sums of at most $m$ square-integrable ridge functions on $\r^n$, i.e.
    \begin{equation}
        \mathcal{R}_m^{(n)} := \left\{g: \r^n \to \r:\; g(x) = \sum_{j=1}^m f_j(\la x, \theta_j\ra) \text{ with } f_j \in L^2(\gamma_1) \text{ and } \theta_j \in \mathbb{S}^{n-1}\right\} \subset L^2(\gamma_n).
    \end{equation}
    The functions $f_j: \r \to \r$ encode both the activations and the output weights of each neuron.
\end{definition}

We will make use of some notation from Gaussian/Hermite analysis; for references, see \cite[Definition 2.1, Example 3.18, Theorem 4.1]{janson1997gaussian} of the textbook of Janson.
Fix $n \in \n$ and let $\Psi: \r^n \to \r$ be s.t. $\Psi \in L^2(\gamma_n)$. For each $k \in \n \cup \{0\}$, denote by $T_k(\Psi) \in \operatorname{Sym}^k(\r^n)$ its degree-$k$ Hermite coefficient tensor, normalized s.t.
    \begin{equation}
        \|T_k(\Psi)\|_F = \|\Pi_k \Psi\|_{L^2(\gamma_n)}
    \end{equation}
    for $\Pi_k$ the orthogonal projection in $L^2(\gamma_n)$ onto the $k^{th}$ Hermite chaos. In other words, $T_k(\Psi) := \mathbb{E}_{\xi \sim \gamma_n}[\Psi(\xi) \mathbf{He}_k(\xi) / \sqrt{k!}]$ where $\mathbf{He}_k: \r^n \to \operatorname{Sym}^k(\r^n)$ are the standard monic probabilist's Hermite polynomial tensors which satisfy $\mathbb E\left[
\langle A,\mathbf{He}_k(\xi)\rangle_F
\langle B,\mathbf{He}_k(\xi)\rangle_F
\right]=
k!\langle A,B\rangle_F$ for $A, B \in \operatorname{Sym}^k(\r^n)$. Let $M_k(\Psi) \in \r^{n^{\lfloor k/2\rfloor }\times n^{\lceil k/2\rceil}}$ be the balanced matricization of $T_k(\Psi)$, and denote its singular values by
    \begin{equation}
\sigma_1(M_k(\Psi)) \geq \ldots \geq \sigma_{n^{\lfloor k/2\rfloor}}(M_k(\Psi)) \geq 0.
    \end{equation}
 Since $T_k(\Psi)$ is symmetric, the singular values of its balanced matricization are independent of which partition of the $k$ indices into groups of sizes \(\lfloor k/2\rfloor\) and \(\lceil k/2\rceil\) we use. Also, note that $\|M_k(\Psi)\|_F = \|T_k(\Psi)\|_F$, and so $\sum_j \sigma_j(M_k(\Psi))^2 = \|\Pi_k\Psi\|_{L^2(\gamma_n)}^2$.

To state our general approximation result for the reverse OU transports to triangular shear measures, we make the following definition:

    \begin{definition}[Hermite width envelope]
    \label{def:appendixhermitewidthenvelope}
    Let $\Psi: \r^n \to \r$ be s.t. $\Psi \in L^2(\gamma_n)$.
         Define the function $a_\Psi: \n \cup \{0\} \times [0, \infty) \to [0, \infty)$ by the envelope
     \begin{equation}
        a_\Psi(m, \tau) :=  \left(\sum_{k=0}^\infty e^{-2(k+1)\tau}\sum_{j={m+1}}^{n^{\lfloor k/2\rfloor}} \sigma_j(M_k(\Psi))^2\right)^{1/2}.
     \end{equation}
    It is nonincreasing in its first argument, and we define $m_\Psi: (0, \infty) \times [0, \infty) \to \n \cup \{0,\infty\}$ to be its generalized inverse, i.e.
     \begin{equation}
         m_\Psi(\delta, \tau) := \inf \{m \in \n \cup \{0\}:\; a_\Psi(m, \tau) \leq \delta \}
     \end{equation}
     with the convention that $\inf \emptyset = \infty$.
     We call $m_\Psi$ the {\em Hermite width envelope} of $\Psi$.
    \end{definition}

    Using this definition, we get a particularly clean lower bound on the width of shallow approximations to the OU-smoothed scores of the triangular shear measure generated by $(\psi, \Psi)$ in the case $\psi$ is linear:

    \begin{proposition}[Hermite tail shallow approximation lower bound]
    \label{prop:appendixdaapproximationlowerbound}
     Let $n \in \n$, fix $\Psi: \r^n \to \r$ bounded and Lipschitz, and let $\beta, L \in (0, \infty)$ be s.t.
     \begin{equation}
         \|\Psi\|_\infty \leq \beta,\quad \operatorname{Lip}(\Psi) \leq L.
     \end{equation}
     Fix $\psi(u) = \sigma u$ for some $\sigma \in (0, \infty)$.
     
     Fix any $\tau \in [0, \infty)$ and $\epsilon \in (0, \infty)$, and suppose that there is some $\hat{s}_\tau: \r^{n+1} \to \r^{n+1}$ for which $\la e_{n+1}, \hat{s}_\tau(\cdot)\ra \in \mathcal{R}_m^{(n+1)}$ for some width $m \in \n$ (such as a two-layer neural network of width $m$) and
     \begin{equation}\|s_\tau^{\psi, \Psi} - \hat{s}_\tau\|_{L^2(\rho_\tau^{\psi, \Psi}; \r^{n+1})} \leq \epsilon.\end{equation} 
     Then, there is some $Q_{\beta, \sigma} \in (0, \infty)$ depending only on $\beta, \sigma$ s.t.
    \begin{equation}
         m \geq m_\Psi\left(Q_{\beta, \sigma}  (\epsilon + L^2 \tau),\; \tau\right).
    \end{equation}
    Moreover, for any nonzero $\phi: \r \to \r$ satisfying Assumption \ref{assumption:activationfunciton} there is some $ K_{\phi, \beta, \sigma} \in (0, \infty)$ s.t.
    \begin{equation}
        \|\hat{s}_\tau\|_{\b_\phi} \geq K_{\phi, \beta, \sigma}  \cdot  \epsilon\sqrt{\max\left\{0,\; m_\Psi(Q_{\beta, \sigma} (2\epsilon + L^2\tau), \;\tau) - 1\right\}}
    \end{equation}
    These statements also hold with $s_\tau^{\psi, \Psi}$ replaced by $b_\tau^{\psi, \Psi}$.
    \end{proposition}

    In words, if $\hat{s}_\tau$ is a two-layer neural network for which $\|s_\tau^{\psi, \Psi} - \hat{s}_\tau\|_{L^2(\rho_\tau^{\psi, \Psi}; \r^{n+1})} \leq \epsilon$, then its width must be at least $m_\Psi(Q_{\beta, \sigma}(\epsilon + L^2\tau), \tau)$. The rest of this subsection is devoted to the proof of this result.

    We start with a general approximation lower bound in terms of the spectral tails of Hermite tensors, which is a direct extension of \cite[Lemma 4.14]{bruna2026multiscale}. Informally, functions of high Hermite rank cannot be approximated using sums of ridge functions without sufficient width. This is quantified by $a_\Psi$ in a way which is robust to OU smoothing, yielding:

\begin{lemma}[Shallow approximation lower bound in $L^2(\gamma_n)$ via Hermite tail]
\label{lemma:appendixhermiteranklowerbound}
For every $\Psi \in L^2(\gamma_n)$ and every $m \in \n$ and $\tau \in [0, \infty)$, we have
    \begin{equation}
        \inf_{g \in \mathcal{R}_m^{(n)}} \|e^{-\tau} P_\tau \Psi - g\|_{L^2(\gamma_n)} \geq a_\Psi(m, \tau).
    \end{equation}
\end{lemma}
\begin{proof}
    Fix some $g(\cdot) = \sum_{j=1}^m f_j (\la \cdot, \theta_j\ra) \in \mathcal{R}_m^{(n)}$, where each $f_j \in L^2(\gamma_1)$. With $h_k := \operatorname{He}_k / \sqrt{k!}$ the orthonormalized Hermite basis of $L^2(\gamma_1)$, write $\hat{f}_j(k) := \la f_j, h_k\ra_{L^2(\gamma_1)}$ so that
    \begin{equation}
        f_j = \sum_{k=0}^\infty \hat{f}_j(k) h_k \text{ in $L^2(\gamma_1)$} \quad (\forall j \in \{1, \ldots, m\}).
    \end{equation}
    Since $\theta_j \in \mathbb{S}^{n-1}$, the defining property of the multivariate Hermite tensors gives
    \begin{equation}
        h_k(\la x, \theta_j\ra) = \la \frac{\mathbf{He}_k(x)}{\sqrt{k!}}, \theta_j^{\otimes k}\ra_F.
    \end{equation}
    Hence, $T_k(f_j(\la\cdot, \theta_j\ra)) = \hat{f}_j(k) \theta_j^{\otimes k}$ and therefore
    \begin{equation}
        T_k(g) = \sum_{j=1}^m \hat{f}_j(k) \theta_j^{\otimes k}.
    \end{equation}
    In particular, $\operatorname{rank}(M_k(g)) \leq m$ for every $k \in \n \cup \{0\}$. Since the OU semigroup $P_\tau$ acts on the $k^{th}$ Wiener chaos as multiplication by $e^{-k\tau}$, we find that
    $T_k(e^{-\tau}P_\tau \Psi) = e^{-(k+1)\tau} T_k(\Psi)$ and so
    \begin{align}
        \|e^{-\tau}P_\tau \Psi - g\|_{L^2(\gamma_n)}^2 &= \sum_{k=0}^\infty \|e^{-(k+1)\tau} T_k(\Psi) - T_k(g)\|_F^2 \\
        &= \sum_{k=0}^\infty \|e^{-(k+1)\tau}M_k(\Psi) - M_k(g)\|^2_F \\
        &\geq \sum_{k=0}^\infty e^{-2(k+1)\tau} \sum_{j=m+1}^{n^{\lfloor k/2\rfloor}}\sigma_j(M_k(\Psi))^2 \\
        &= a_\Psi(m, \tau)^2,
    \end{align}
    where the inequality is due to Eckart-Young and the fact $\operatorname{rank}(M_k(g)) \leq m$. 
\end{proof}

The above result operates in $L^2(\gamma_n)$ in order to deploy the methods of Hermite analysis. However, our object of interest is the approximation of OU scores in $L^2(\rho_\tau^{\psi, \Psi}; \r^{n+1})$ (where $\rho_\tau^{\psi, \Psi} = P_\tau^* \pi^{\psi, \Psi}$ are the OU marginals). In order to apply Lemma \ref{lemma:appendixhermiteranklowerbound}, we define the following operator which averages over the non-Gaussian coordinate:

\begin{definition}[Truncated Gaussian averaging operator]
Fix $R > 0$.
    Let $F: \r^{n+1} \to \r^{n+1}$ be a measurable vector field. Using coordinates $(z, u) \in \r^{n+1}$ with $z \in \r^n$ and $u \in \r$, assume that $F_{n+1}(z, \cdot)$ is integrable w.r.t. $\gamma_1$ on $[-R, R]$ for a.e. $z$, and
    define $\mathcal{A}_RF: \r^{n} \to \r$ by
    \begin{equation}
        (\mathcal{A}_RF)(z) := \mathbb{E}_{U \sim \gamma_1}[F_{n+1}(z, U) \mid |U| \leq R].
    \end{equation}
    We call $\mathcal{A}_R$ the {\em averaging operator with truncation $R$}.
\end{definition}

For our triangular shear measures of interest, all other coordinates are Gaussian. This means that using some estimates on the averaging operator, we can lower bound $L^2(\rho_\tau^{\psi, \Psi}; \r^{n+1})$ score approximation in terms of $L^2(\gamma_n)$ function approximation:

\begin{lemma}
\label{lemma:appendixaveragingoperator}
    Fix $\sigma, \beta, L , R\in (0, \infty)$, let $\psi(u) = \sigma u$, and let $\Psi: \r^n \to \r$ satisfy $\|\Psi\|_\infty \leq \beta$ and $\operatorname{Lip}(\Psi) \leq L$.
    Then, for every $\tau \in [0, \infty)$ and every measurable $F: \r^{n+1} \to \r^{n+1}$ for which $\mathcal{A}_RF$ is defined and in $L^2(\gamma_n)$, it holds that
    \begin{align}
        \| s_\tau^{\psi, \Psi} - F\|_{L^2(\rho_\tau^{\psi, \Psi}; \r^{n+1})} \geq  \alpha_{R,\beta, \sigma} \cdot \left(\|e^{-\tau}P_\tau \Psi - c_{\sigma, \tau}\mathcal{A}_RF\|_{L^2(\gamma_n)} - C_{R,\beta, \sigma} L^2 \tau\right)
    \end{align}
    where we define the explicit constants
    \begin{align}
        \alpha_{R, \beta, \sigma} &:= \sqrt{\gamma_1([-R, R])} \max\{1, \sigma\}^{-5/2} \exp\left(-\frac{R\beta + \beta^2/2 + R^2\max\{0, 1-\sigma^2\}/2}{2\min\{1, \sigma\}^2}\right),
        \\  C_{R, \beta, \sigma}  &:= \frac{2(R+\beta)}{\min\{1, \sigma\}^2} \exp\left(\frac{(R+\beta)^2}{2\min\{1, \sigma\}^2}\right),\\
        c_{\sigma, \tau} &:= e^{-2\tau} \sigma^2 + 1 - e^{-2\tau}.
    \end{align}
    The same is true with $s_\tau^{\psi, \Psi}$ replaced by $b_\tau^{\psi, \Psi}$.
\end{lemma}
\begin{proof}
The claim follows directly from the following two estimates on $\mathcal{A}_R$:
\begin{equation}
\label{eqn:proofappendixaveraginglemmasecondthingtoprove}
    \|e^{-\tau}P_\tau \Psi - c_{\sigma, \tau}\mathcal{A}_R s_\tau^{\psi, \Psi}\|_{\infty} \leq C_{R, \beta,\sigma} L^2 \tau.
\end{equation}
\begin{equation}
\label{eqn:proofappendixaveraginglemmafirstthingtoprove}
    \|s_\tau^{\psi, \Psi} - F\|_{L^2(\rho_\tau^{\psi, \Psi}; \r^{n+1})} \geq \alpha_{R,\beta, \sigma} c_{\sigma, \tau}\|\mathcal{A}_R s_\tau^{\psi, \Psi} - \mathcal{A}_R F\|_{L^2(\gamma_n)},
\end{equation}
The same will be true for $b_\tau^{\psi, \Psi}$, after which the result will follow from the reverse triangle inequality.

To show these claims, let us denote $q_- := \min\{1, \sigma\}$, $q_+ := \max\{1, \sigma\}$, $c = c_{\sigma, \tau}$ and observe that $q_-^2 \leq c \leq q_+^2$. For each $z \in \r^n$ we define
\begin{equation}
    \eta_z := e^{-\tau}\Psi(e^{-\tau}z + \sqrt{1-e^{-2\tau}}X), \quad X \sim \gamma_n.
\end{equation}
Note that $|\eta_z| \leq \beta$ and $\mathbb{E}[\eta_z] = e^{-\tau}P_\tau \Psi(z)$. If we denote
\begin{equation}
    \widetilde{\eta}_z(u) := \frac{\mathbb{E}\left[\eta_z \exp\left(-\frac{(u-\eta_z)^2}{2c}\right)\right]}{\mathbb{E}\left[\exp\left(-\frac{(u-\eta_z)^2}{2c}\right)\right]}
\end{equation}
then we claim that
\begin{equation}
    (s_\tau^{\psi, \Psi}(z, u))_{n+1} = \frac{\widetilde{\eta}_z(u) - u}{c}.
\end{equation}
Indeed, as was the case in the proof of Lemma \ref{lemma:stupidnonsenserevised}, by Gaussian regression we know that conditionally on the first $n$ coordinates being equal to $z$ the last coordinate under $\rho_\tau^{\psi, \Psi}$ has the same law as
\begin{equation}
    \sqrt{c}W + \eta_z, \quad W \sim \gamma_1 \text{ independently of $\eta_z$}.
\end{equation}
The conditional density of the last coordinate is therefore $p_z(u) = \frac{1}{\sqrt{2\pi c}} \mathbb{E}\left[\exp\left(-\frac{(u-\eta_z)^2}{2c}\right)\right]$, and so
\begin{align}
    (s_\tau^{\psi, \Psi}(z, u))_{n+1} = \partial_u \log p_z(u) = \frac{\mathbb{E}\left[-\frac{u-\eta_z}{c} \exp\left(-\frac{(u-\eta_z)^2}{2c}\right)\right]}{\mathbb{E}\left[\exp\left(-\frac{(u-\eta_z)^2}{2c}\right)\right]} = \frac{\widetilde{\eta}_z(u) - u}{c}.
\end{align}
This also implies that $(b_\tau^{\psi, \Psi}(z, u))_{n+1} = \frac{\widetilde{\eta}_z(u) - u}{c}+u$, and since the standard Gaussian conditioned on $[-R, R]$ is centered we find
\begin{equation}
\label{eqn:proofappendixvalueoftheaveragingoperator}
    (c\mathcal{A}_R s_\tau^{\psi, \Psi})(z) = (c\mathcal{A}_R b_\tau^{\psi, \Psi})(z) = \mathbb{E}_{U \sim \gamma_1}[\widetilde{\eta}_z(U) \mid |U| \leq R],
\end{equation}
and in particular $\mathcal{A}_R s_\tau^{\psi, \Psi} = \mathcal{A}_R b_\tau^{\psi, \Psi}$. We now compare $\widetilde{\eta}_z(u)$ and $\mathbb{E}[\eta_z]$. For $u, y \in \r$ we let
\begin{equation}
    w_u(y) := \exp\left(-\frac{(u-y)^2}{2c}\right)
\end{equation}
to see that $\widetilde{\eta}_z(u) = \mathbb{E}[\eta_z w_u(\eta_z)] / \mathbb{E}[w_u(\eta_z)]$ and therefore
\begin{equation}
\label{eqn:proofappendixaveraginglemmacovarianceequality}
    \widetilde{\eta}_z(u) - \mathbb{E}[\eta_z] = \frac{\operatorname{Cov}(\eta_z, w_u(\eta_z))}{\mathbb{E}[w_u(\eta_z)]}.
\end{equation}
Now, since $|\eta_z| \leq \beta$ and $|u| \leq R$ we know that
$\mathbb{E}[w_u(\eta_z)] \geq \exp\left(-\frac{(R+\beta)^2}{2c}\right)$.
Furthermore, for $y \in [-\beta, \beta]$ we know
$|w'_u(y)| = \frac{|u-y|}{c} w_u(y) \leq \frac{R+\beta}{c}$. 
Hence, the covariance inequality $|\operatorname{Cov}(Y, f(Y))| \leq \operatorname{Lip}(f) \operatorname{Var}(Y)$ implies that
\begin{equation}
\label{eqn:proofappendixaveraginglemmaupperboundintermsofvariance}
    \sup_{u \in [-R, R]}\left|\widetilde{\eta}_z(u) - \mathbb{E}[\eta_z]\right| \leq \frac{R+\beta}{c} \exp\left(\frac{(R+\beta)^2}{2c}\right) \operatorname{Var}(\eta_z).
\end{equation}
However, the function $x \mapsto e^{-\tau}\Psi(e^{-\tau}z + \sqrt{1-e^{-2\tau}}x)$ is $Le^{-\tau}\sqrt{1-e^{-2\tau}}$-Lipschitz and so the Gaussian Poincaré inequality yields
\begin{equation}
    \operatorname{Var}(\eta_z) \leq L^2 e^{-2\tau} (1-e^{-2\tau}) \leq 2L^2 \tau.
\end{equation}
Combining this with (\ref{eqn:proofappendixaveraginglemmaupperboundintermsofvariance}) and $c \geq q_-^2$ and using $\mathbb{E}[\eta_z] = e^{-\tau}P_\tau \Psi(z)$, we find
\begin{equation}
    \sup_{z \in \r^n}\sup_{u \in [-R, R]}  \left|\widetilde{\eta}_z(u) - e^{-\tau}P_\tau \Psi(z)\right| \leq C_{R, \beta, \sigma}L^2 \tau.
\end{equation}
Averaging over $u$ and using (\ref{eqn:proofappendixvalueoftheaveragingoperator}) gives the estimate (\ref{eqn:proofappendixaveraginglemmasecondthingtoprove}) for both $s_\tau^{\psi, \Psi}$ and $b_\tau^{\psi, \Psi}$.

To prove the second estimate (\ref{eqn:proofappendixaveraginglemmafirstthingtoprove}), we note that the density of $\rho_\tau^{\psi, \Psi}$ is given by
\begin{equation}
    \rho_{\tau}^{\psi, \Psi}(z, u) = \gamma_n(z) p_z(u)
\end{equation}
since the first $n$ coordinates have marginal law $\gamma_n$ and the last coordinate has density $p_z$ conditioned on the event where the first $n$ coordinates equal $z$. Therefore,
\begin{align}
    \|s_{\tau}^{\psi, \Psi} - F\|_{L^2(\rho_\tau^{\psi, \Psi}; \r^{n+1})}^2 &\geq \int_{\r^n} \int_{-\infty}^\infty \left|(s_\tau^{\psi, \Psi}(z, u))_{n+1} - F_{n+1}(z, u)\right|^2 p_z(u) du d\gamma_n(z) \\
    &\geq \kappa_{R,\beta, \sigma} \int_{\r^n} \int_{-R}^R \left|(s_\tau^{\psi, \Psi}(z, u))_{n+1} - F_{n+1}(z, u)\right|^2 d\gamma_1(u)d\gamma_n(z),
\end{align}
where we use that
\begin{equation}
    \frac{p_z(u)}{\gamma_1(u)} = \frac{1}{\sqrt{c}} \mathbb{E}\left[\exp\left(\frac{c-1}{2c}u^2 + \frac{u\eta_z}{c} - \frac{\eta_z^2}{2c}\right)\right] \geq \kappa_{R, \beta, \sigma}
\end{equation}
for all $|u| \leq R$ and we denote
\begin{equation}
    \kappa_{R, \beta, \sigma} := \frac{1}{q_+} \exp\left(-\frac{R\beta + \beta^2 / 2 + R^2\max\{0, 1-\sigma^2\}/2}{q_-^2}\right).
\end{equation}
Hence, by applying Jensen's inequality under the conditional law $\gamma_1(\cdot \mid [-R, R])$ we find
\begin{align}
    \|s_{\tau}^{\psi, \Psi} - F\|_{L^2(\rho_\tau^{\psi, \Psi}; \r^{n+1})}^2 &\geq \kappa_{R, \beta, \sigma} \gamma_1([-R, R]) \int_{\r^n}\left|(\mathcal{A}_Rs_\tau^{\psi, \Psi})(z) - (\mathcal{A}_R F)(z)\right|^2 d\gamma_n(z)  \\
    &= \kappa_{R, \beta, \sigma} \gamma_1([-R, R])\| \mathcal{A}_R s_\tau^{\psi, \Psi} - \mathcal{A}_RF\|_{L^2(\gamma_n)}^2.
\end{align}
Using $\kappa_{R,\beta,\sigma}\gamma_1([-R, R]) / c^2 \geq \alpha_{R, \beta, \sigma}^2$, we get (\ref{eqn:proofappendixaveraginglemmafirstthingtoprove}). The same argument applies to $b_\tau^{\psi, \Psi}$.
\end{proof}

Hence, for any candidate vector field $F: \r^{n+1} \to \r^{n+1}$ which attempts to approximate $s_\tau^{\psi, \Psi}$ or $b_\tau^{\psi, \Psi}$ in $L^2(\rho_\tau^{\psi, \Psi})$, if $\mathcal{A}_RF \in \mathcal{R}_m^{(n)}$ then combining Lemmas \ref{lemma:appendixhermiteranklowerbound} and \ref{lemma:appendixaveragingoperator} yields an approximation lower bound. We next show that $\mathcal{A}_R$ preserves the ridge width constraint:

\begin{lemma}
\label{lemma:appendixeaveragingoperatorpreserveswidth}
    Fix $m \in \n$ and let $F : \r^{n+1} \to \r^{n+1}$ be measurable s.t. its last coordinate satisfies $F_{n+1} \in \mathcal{R}_m^{(n+1)}$. Then, $\mathcal{A}_RF \in \mathcal{R}_m^{(n)}$ for all $R > 0$.
\end{lemma}
\begin{proof}
   Let us express
   \begin{equation}
       F_{n+1}(z, u) = \sum_{j=1}^m f_j(\la z, v_j\ra + b_j u)
   \end{equation}
   for $f_j \in L^2(\gamma_1)$ and $\|v_j\|^2 + |b_j|^2 = 1$. Choose $\theta_j \in \mathbb{S}^{n-1}$ s.t. $v_j = \|v_j\| \theta_j$ (where if $\|v_j\| = 0$ then we choose arbitrary $\theta_j$) and define
   \begin{equation}
       \widetilde{f}_j(t) := \mathbb{E}_{U \sim \gamma_1}[f_j(\|v_j\|t + b_j U) \mid |U| \leq R].
   \end{equation}
    Naturally, this means that
    \begin{equation}
        (\mathcal{A}_R F)(z)  = \sum_{j=1}^m \widetilde{f}_j(\la z, \theta_j\ra).
    \end{equation}
    It remains to check that $\widetilde{f}_j \in L^2(\gamma_1)$. With $Y, U \sim \gamma_1$ i.i.d. we have
    \begin{align}
        \|\widetilde{f}_j\|_{L^2(\gamma_1)}^2 &\leq \mathbb{E}[f_j(\|v_j\|Y + b_j U)^2 \mid |U| \leq R] \\&\leq \frac{1}{\gamma_1([-R, R])} \mathbb{E}[f_j(\|v_j\|Y + b_j U)^2 ]\\
        &= \frac{1}{\gamma_1([-R, R])} \|f_j\|_{L^2(\gamma_1)}^2,
    \end{align}
    where the first line is Jensen and the third is because $\operatorname{Law}(\|v_j\|Y + b_jU ) = \gamma_1$.
\end{proof}

We may combine these statements to yield a proof of Proposition \ref{prop:appendixdaapproximationlowerbound}:

\begin{proof}[Proof of Proposition \ref{prop:appendixdaapproximationlowerbound}]
Letting $q_-,q_+$ be as before, we choose $R := \frac{q_-^2}{\beta + q_-}$, and so $R \leq q_- \leq 1$, $\frac{\beta R}{q_-^2} \leq 1$, and $\frac{R^2}{q_-^2} \leq 1$. We first show that
\begin{equation}
    Q_{\beta, \sigma} := 9 (1+\beta) \frac{\max\{1, \sigma\}^{5/2}}{\min\{1, \sigma\}^2} \exp\left(\frac{\beta^2}{2\min\{1, \sigma\}^2}\right)   \geq \max\{\alpha_{R, \beta, \sigma}^{-1}, C_{R, \beta, \sigma}\}.
\end{equation}
Since $R \leq 1$ we know $\gamma_1([-R, R]) \geq 2Re^{-1/2}/\sqrt{2\pi} = \sqrt{\frac{2}{\pi e}} R$, and so
\begin{align}
    \alpha_{R, \beta, \sigma}^{-1} &= \gamma_1([-R, R])^{-1/2} q_+^{5/2} \exp\left(\frac{R\beta + \beta^2 / 2 + R^2 \max\{0, 1-\sigma^2\}/2}{2q_-^2}\right)\\
    &\leq e\left(\frac{\pi}{2}\right)^{1/4} R^{-1/2} q_+^{5/2} \exp\left(\frac{\beta^2}{4q_-^2}\right) \\
    &\leq  e\left(\frac{\pi}{2}\right)^{1/4} \frac{1+\beta}{q_-^2} q_+^{5/2} \exp\left(\frac{\beta^2}{2q_-^2}\right) \\
    &\leq Q_{\beta, \sigma}.
\end{align}
Similarly, we have
\begin{align}
    C_{R, \beta, \sigma} 
    &= \frac{2(R+\beta)}{q_-^2} \exp\left(\frac{\beta^2}{2q_-^2} + \frac{\beta R}{q_-^2} + \frac{R^2}{2q_-^2}\right) \\
    &\leq 2e^{3/2} \frac{1+\beta}{q_-^2} \exp\left(\frac{\beta^2}{2q_-^2}\right)\\
    &\leq Q_{\beta, \sigma}.
\end{align}
We now prove the width lower bound. Applying Lemma \ref{lemma:appendixaveragingoperator} to $F = \hat{s}_\tau$ gives
\begin{equation}
    \epsilon \geq \alpha_{R, \beta, \sigma} \left(\|e^{-\tau}P_\tau \Psi - c_{\sigma, \tau} \mathcal{A}_R \hat{s}_\tau\|_{L^2(\gamma_n)} - C_{R, \beta, \sigma}L^2\tau\right)
\end{equation}
and therefore
\begin{equation}
    \|e^{-\tau}P_\tau \Psi - c_{\sigma, \tau} \mathcal{A}_R \hat{s}_\tau\|_{L^2(\gamma_n)} \leq \alpha_{R, \beta, \sigma}^{-1}\epsilon + C_{R, \beta, \sigma}L^2\tau \leq Q_{\beta, \sigma}(\epsilon + L^2 \tau).
\end{equation}
Since $(\hat{s}_\tau)_{n+1} \in \mathcal{R}_m^{(n+1)}$, Lemma \ref{lemma:appendixeaveragingoperatorpreserveswidth} ensures that $c_{\sigma, \tau}\mathcal{A}_R \hat{s}_\tau \in \mathcal{R}_m^{(n)}$. By Lemma \ref{lemma:appendixhermiteranklowerbound}, we know that
\begin{equation}
    a_\Psi(m, \tau) \leq Q_{\beta, \sigma}(\epsilon + L^2 \tau)
\end{equation}
and therefore $m \geq m_\Psi(Q_{\beta, \sigma}(\epsilon + L^2\tau), \tau)$. 

It remains to prove the Barron lower bound.
Let $\h := L^2(\rho_\tau^{\psi, \Psi})$, denote
\begin{equation}
    \kappa_s := (\mathbb{E}_{G \sim \gamma_1}|G|^{2s})^{1/(2s)} = \left(\frac{2^s \Gamma(s+1/2)}{\sqrt{\pi}}\right)^{1/(2s)},
\end{equation}
and define the constant
\begin{equation}
    \Lambda_{s,\beta,\sigma} := \begin{cases}
        \left((\kappa_sq_+)^{2s} + \max\{1,\beta\}^{2s}\right)^{1/2} & 0 < s < 1/2,\\
        \max\{\kappa_s q_+ + \beta, 1\}^s & s \geq 1/2.
    \end{cases}
\end{equation}
Finally, set
\begin{equation}
    A_{\phi, \beta, \sigma} := \begin{cases}
\|\phi\|_\infty & \text{$\phi$ bounded},
\\ \max\{|\phi(1)|, |\phi(-1)|\} \Lambda_{s, \beta, \sigma} & \text{$\phi$ positively $s$-homogeneous.}
    \end{cases}
\end{equation}
We show that every normalized neuron has $\h$-norm at most $A_{\phi, \beta, \sigma}$. If $\phi$ is bounded then obviously
\begin{equation}
    \left\|\frac{\phi(\la \cdot, a\ra + b)}{1 + \|a\| + |b|}\right\|_{\h} \leq \frac{\|\phi\|_\infty}{1 + \|a\| + |b|} \leq A_{\phi, \beta, \sigma}.
\end{equation}
Suppose now that $\phi$ is positively $s$-homogeneous. 
Let $X_\tau \sim \rho_\tau^{\psi, \Psi}$ so that 
\begin{equation}
    X_\tau = e^{-\tau} (Z, \sigma U + \Psi(Z)) + \sqrt{1-e^{-2\tau}} G
\end{equation}
for $Z \sim \gamma_n$, $U \sim \gamma_1$, $G \sim \gamma_{n+1}$ independent. For parameters $a = (v, t) \in \r^{n} \times \r$ and $b \in \r$, write
\begin{equation}
    \la X_\tau, a\ra + b = H_a + D_{a,b}
\end{equation}
where
\begin{equation}
    H_a := e^{-\tau} (\la Z, v\ra + \sigma t U) + \sqrt{1-e^{-2\tau}} \la G, a\ra
\end{equation}
is centered Gaussian and 
\begin{equation}
    D_{a,b} := e^{-\tau}t\Psi(Z) + b.
\end{equation}
Since we have
\begin{equation}
    \mathbb{E}[H_a^2] = \|v\|^2 + c_{\sigma, \tau} t^2 \leq q_+^2 \|a\|^2
\end{equation}
we find that $\|H_a\|_{L^{2s}} \leq \kappa_s q_+ \|a\|$. Moreover,
\begin{equation}
    |D_{a,b}| \leq |t|\beta + |b| \leq \beta \|a\| + |b|.
\end{equation}
Suppose first that $s \geq 1/2$. Since $|\phi(w)| \leq \max\{|\phi(1)|, |\phi(-1)|\} |w|^s$ we have
\begin{align}
    \|\phi(\la X_\tau, a\ra + b)\|_{L^2} &\leq \max\{|\phi(1)|, |\phi(-1)|\} \cdot  \|H_a + D_{a,b}\|_{L^{2s}}^s
    \\ &\leq \max\{|\phi(1)|, |\phi(-1)|\} \cdot (\kappa_s q_+ \|a\| + \beta\|a\| + |b|)^s 
    \\&\leq A_{\phi, \beta, \sigma}(\|a\| + |b|)^s.
\end{align}
If we instead have $0 < s < 1/2$ we know $|x+y|^{2s} \leq |x|^{2s} + |y|^{2s}$ and so
\begin{align}
    \mathbb{E}|H_a + D_{a,b}|^{2s} &\leq \mathbb{E}|H_a|^{2s} + \mathbb{E}|D_{a,b}|^{2s} \\
    &\leq (\kappa_s q_+)^{2s}\|a\|^{2s} + (\beta \|a\| + |b|)^{2s}
    \\&\leq \Lambda_{s, \beta, \sigma}^2 (\|a\|+|b|)^{2s}.
\end{align}
Hence, in either case we know that
\begin{equation}
    \left\|\frac{\phi(\la \cdot, a\ra + b)}{(\|a\| + |b|)^{s}}\right\|_\h \leq A_{\phi, \beta, \sigma}.
\end{equation}
We finish by noting that if $\mathcal{N}_{\phi,k}^{(n+1, n+1)}$ denotes the class of vector-valued two-layer neural networks of width at most $k$ with activation $\phi$, then
\begin{equation}
\label{eqn:proofappendixpropositionapproximationtheBarronstep}
    \inf_{G \in \mathcal{N}_{\phi,k}^{(n+1, n+1)}} \|\hat{s}_\tau - G\|_\h \leq \frac{A_{\phi, \beta, \sigma} \|\hat{s}_\tau\|_{\b_\phi}}{\sqrt{k}} \quad (\forall k \in \n).
\end{equation}
Indeed, given a Barron representation $\hat{s}_\tau(\cdot) = \int_{\r^{n+1} \times \r} \phi(\la \cdot, a\ra + b) d\mu(a, b)$ with \begin{equation}B_\mu := \int_{\r^{n+1} \times \r} w(a, b) d\mu(a, b) \leq \|\hat{s}_\tau\|_{\b_\phi} + \delta, \quad w(a, b) := \begin{cases}
    1 + \|a\| + |b| & \text{$\phi$ bounded},
    \\
    (\|a\| + |b|)^s & \text{$\phi$ $s$-homogeneous},
\end{cases}\end{equation}
we normalize $\mu$ to a probability measure, sample $k$ i.i.d. normalized neurons $Y_j := \frac{\phi(\la \cdot, a_j\ra + b_j)}{w(a_j, b_j)}$, and take
\begin{equation}
    \hat{G}(\cdot) := \frac{B_\mu}{k}\sum_{j=1}^k Y_j.
\end{equation}
We may calculate that
\begin{equation}
    \mathbb{E}\|\hat{s}_\tau - \hat{G}\|_\h^2 = \frac{B_\mu^2}{k} \mathbb{E}\|Y_1 - \mathbb{E}Y_1\|^2_\h \leq \frac{B_\mu^2A_{\phi, \beta, \sigma}^2}{k} \leq \frac{A_{\phi, \beta, \sigma}^2 (\|\widehat{s}_\tau\|_{\b_\phi} + \delta)^2}{k},
\end{equation}
after which taking $\delta \downarrow 0$ proves (\ref{eqn:proofappendixpropositionapproximationtheBarronstep}). Define
\begin{equation}
    M := m_\Psi(Q_{\beta, \sigma}(2\epsilon + L^2\tau), \tau).
\end{equation}
If $M \leq 1$ then the claim is trivial, so we suppose $M \geq 2$. For every $k < M$, the width lower bound we already proved implies
\begin{equation}
\label{eqn:proofappendixpropositionapproximationtheBarronstep1}
    \inf_{G \in \mathcal{N}_{\phi,k}^{(n+1, n+1)}} \|s_\tau^{\psi, \Psi} - G\|_\h \geq 2\epsilon
\end{equation}
since the last coordinate satisfies $G_{n+1} \in\mathcal{R}^{(n+1)}_k$ for all $G \in \mathcal{N}_{\phi,k}^{(n+1, n+1)}$. On the other hand, by the triangle inequality and (\ref{eqn:proofappendixpropositionapproximationtheBarronstep}) we have
\begin{equation}
\label{eqn:proofappendixpropositionapproximationtheBarronstep2}
    \inf_{G \in \mathcal{N}_{\phi,k}^{(n+1, n+1)}} \|s_\tau^{\psi, \Psi} - G\|_\h \leq \|s_\tau^{\psi, \Psi} - \hat{s}_\tau\|_\h + \inf_{G \in \mathcal{N}_{\phi,k}^{(n+1, n+1)}} \|\hat{s}_\tau - G\|_\h \leq \epsilon + \frac{A_{\phi, \beta, \sigma}\|\hat{s}_\tau\|_{\b_\phi}}{\sqrt{k}}
\end{equation}
for every $k < M$.
Combining (\ref{eqn:proofappendixpropositionapproximationtheBarronstep1}) and (\ref{eqn:proofappendixpropositionapproximationtheBarronstep2}) implies that
\begin{equation}
    \|\hat{s}_\tau\|_{\b_\phi} \geq \frac{\epsilon \sqrt{k}}{A_{\phi, \beta, \sigma}} \quad (\forall k < M).
\end{equation}
Taking $k = M-1$ (or $k \uparrow \infty$ if $M = \infty$) and setting $K_{\phi, \beta, \sigma} = 1/A_{\phi, \beta, \sigma}$ yields the claim.
\end{proof}

There is also a nonquantitative upper bound in the sense that for every $\epsilon, \tau, n$ there exists a two-layer neural network $\hat{s}_\tau$ of finite width and Barron norm for which $\|s_\tau^{\psi, \Psi} - \hat{s}_\tau\|_{L^2(\rho_\tau^{\psi, \Psi}; \r^{n+1})} \leq \epsilon$. This is nothing more than the standard universal $L^2$ approximation statement for two-layer networks, due originally to Hornik in \cite{hornik1991approximation}. For completeness, we record a precise formulation below:

\begin{proposition}[Shallow approximation upper bound, nonquantitative]
\label{prop:appendixapproximationupperbound}
    Let $n \in \n$ and $\psi: \r \to \r$, $\Psi: \r^n \to \r$ be measurable, and let $\phi: \r \to \r$ be nonpolynomial and satisfy Assumption \ref{assumption:activationfunciton}. 
    Then, for any $\epsilon, \tau \in (0, \infty)$ there is some $m \in \n$ and neural network $\hat{s}_\tau: \r^{n+1} \to \r^{n+1}$ of the form $\hat{s}_\tau(x) = \sum_{j=1}^m u_j \phi(\la x, a_j\ra + b_j)$ with $u_j, a_j \in \r^{n+1}$ and $b_j \in \r$ for which $\|\hat{s}_\tau\|_{\b_\phi}<\infty$ and
    \begin{equation}
        \|s_\tau^{\psi, \Psi} - \hat{s}_\tau\|_{L^2(\rho_\tau^{\psi, \Psi}; \r^{n+1})} \leq \epsilon.
    \end{equation}
    If $\pi^{\psi, \Psi} \in \p_2(\r^{n+1})$, then the same statement holds with $s_\tau^{\psi, \Psi}$ replaced by $b_\tau^{\psi, \Psi}$. 
\end{proposition}
\begin{proof}
Let us write the OU interpolant by \begin{equation}
    X_\tau := e^{-\tau} X + \sqrt{1-e^{-2\tau}} Z, \quad X \sim \pi^{\psi, \Psi}, \; Z \sim \gamma_{n+1}
\end{equation}
with $X, Z$ independent. Fix $\tau > 0$.
    By the Tweedie formula (\ref{eqn:prelimtweedie}) we have $s_\tau^{\psi, \Psi}(x) = -\frac{1}{\sqrt{1-e^{-2\tau}}} \mathbb{E}[Z \mid X_\tau = x]$. Conditional Jensen implies that
    \begin{align}
        \|s_\tau^{\psi, \Psi}\|_{L^2(\rho_\tau^{\psi, \Psi}; \r^{n+1})}^2 = \frac{1}{1-e^{-2\tau}} \mathbb{E}\left[\|\mathbb{E}[Z \mid X_\tau]\|^2\right] \leq \frac{\mathbb{E}\|Z\|^2}{1-e^{-2\tau}} = \frac{n+1}{1-e^{-2\tau}} < \infty
    \end{align}
    and hence $s_\tau^{\psi, \Psi} \in L^2(\rho_\tau^{\psi, \Psi}; \r^{n+1}) $. If $\phi$ is nonconstant and bounded, then \cite[Theorem 1]{hornik1991approximation} directly implies the result. If $\phi$ is instead positively $s$-homogeneous for $s \in \n $, then the $(s+1)^{th}$ discrete derivative of $\phi$ is bounded, nonconstant (since $\phi$ is nonpolynomial), and representable using $s+2$ neurons, so the same universal approximation statement applies at the cost of increasing the width. Lastly, if $\pi^{{\psi, \Psi}}$ has finite second moment then $b_\tau^{\psi, \Psi} = s_\tau^{\psi, \Psi} + \operatorname{Id}$ is also in $L^2(\rho_\tau^{\psi, \Psi}; \r^{n+1})$, and so we may apply the same argument.
\end{proof}

\subsubsection{Early-stopping lower bounds}
\label{subsec:appendixearlystoppinglowerboundstriangular}
The presented lower bounds describe the difficulty of score-matching using shallow neural networks for OU times in a certain window $\tau \in [0, \tau_0]$. However, in practice it is common to avoid modeling scores all the way to $\tau \downarrow 0$ by applying a cutoff. Qualitatively speaking, to approximate the target distribution $\pi$ to accuracy $\delta$ it suffices to fix $\tau_1 > 0$ for which the distance between $P_{\tau_1}^*\pi$ and $\pi$ is $\lesssim \delta$ and to only learn approximate scores $\|s_\tau - \hat{s}_\tau\|_{L^2(\rho_\tau)} \lesssim \delta$ for $\tau \in [\tau_1, \infty)$, see e.g. \cite{chen2023sampling}. Since we have shown that approximating scores on $[0, \tau_0]$ with shallow networks is hard, it remains to show that choosing $\tau_1 > \tau_0$ introduces a lot of early-stopping error. We will provide some general tools for triangular shears here, which we apply to our ResNet pushforward examples in Section \ref{subsec:appendixapplicationtoResNetpushsforwards}.

We start with a lower bound on total variation and $W_1$ distances which uses only the triangular shear structure. For $\mu, \nu \in \p(\r^d)$, we set $\operatorname{TV}(\mu, \nu) = \sup |\mu(A)- \nu(A)|$ over measurable sets $A$.

\begin{lemma}[Early-stopping error, $\operatorname{TV}$ and $W_1$]
\label{lemma:appendixtriangularearlystoppingerrorTVW1}
    Let $n \geq 2$ and $\Psi\in L^1(\gamma_n)$ and define
    \begin{equation}
        \pi := \pi^{\operatorname{Id}, \Psi} = \operatorname{Law}(Z, U + \Psi(Z)), \quad \rho_\tau := P_\tau^* \pi,
    \end{equation}
    with $Z \sim \gamma_n$ and $U \sim \gamma_1$ independent. If we denote $M_\Psi := \mathbb{E}_{Z \sim \gamma_n}[\Psi(Z)^2 e^{-\Psi(Z)^2/4}]$ then
    \begin{equation}
        \min\left\{\operatorname{TV}(\pi, \rho_\tau), \; W_1(\pi, \rho_\tau)\right\} \geq \frac{M_\Psi}{4\sqrt{2}} (1-e^{-2\tau}) \quad (\forall \tau \geq 0).
    \end{equation}
\end{lemma}
\begin{proof}
Since $\Psi \in L^1(\gamma_n)$, both $\pi$ and $\rho_\tau$ have finite first moments. Consider
\begin{equation}
    f(z, u) := e^{-u^2/2}.
\end{equation}
We know $0 \leq f \leq 1$ and $\|\nabla f\|_\infty = \sup_{u \in \r} |u|e^{-u^2/2} \leq 1$. So, by the dual characterizations of total variation and $W_1$, we know
\begin{equation}
     \min\left\{\operatorname{TV}(\pi, \rho_\tau), \; W_1(\pi, \rho_\tau)\right\}  \geq \int_{\r^{n+1}} f d\rho_\tau - \int_{\r^{n+1}} f d\pi.
\end{equation}
The last-coordinate marginal of $\rho_\tau$ has law equal to $\operatorname{Law}(U + e^{-\tau}\Psi(W))$ for $U \sim \gamma_1$ independent of $W \sim \gamma_n$. Using the fact that
\begin{equation}
    \mathbb{E}_{U \sim \gamma_1} \left[e^{-(U+a)^2/2}\right] = \frac{e^{-a^2/4}}{\sqrt{2}},
\end{equation}
we obtain that
\begin{equation}
    \int_{\r^{n+1}} f d\rho_\tau - \int_{\r^{n+1}} f d\pi = \frac{1}{\sqrt{2}} \mathbb{E}_{Z \sim \gamma_n}\left[e^{-e^{-2\tau}\Psi(Z)^2/4} - e^{-\Psi(Z)^2/4}\right].
\end{equation}
Since $e^x - 1 \geq x$ for $x \geq 0$, pointwise we have
\begin{equation}
    e^{-e^{-2\tau} \Psi(z)^2 / 4} - e^{-\Psi(z)^2/4} = e^{-\Psi(z)^2/4} \left(e^{(1-e^{-2\tau}) \Psi(z)^2/4} - 1\right) \geq \frac{1-e^{-2\tau}}{4} \Psi(z)^2 e^{-\Psi(z)^2/4},
\end{equation}
from which the claim follows by taking an expectation.
\end{proof}

By Pinsker's inequality, this $\operatorname{TV}$ lower bound implies a lower bound on both forward and reverse $\sqrt{\operatorname{KL}}$. However, we can get a sharper lower bound in terms of the Hermite mass of $\Psi$:

\begin{lemma}[Early-stopping error, ${\operatorname{KL}}$]
\label{lemma:appendixtriangularearlystoppingerrorKL}
Let $\Psi, \pi, \rho_\tau$ be as in Lemma \ref{lemma:appendixtriangularearlystoppingerrorTVW1}, and assume further that $\|\Psi\|_\infty \leq \beta$. Let $D_\tau$ denote either $\operatorname{KL}(\rho_\tau \| \pi)$ or $\operatorname{KL}(\pi \| \rho_\tau)$. Then,
\begin{equation}
    \sqrt{D_\tau} \geq \frac{1}{\sqrt{2(1+\beta^2)}} \max_{m \in \n \cup \{0\}} \left\{ (1-e^{-\ell_n(m) \tau}) a_\Psi(m, 0)\right\} \quad (\forall \tau \geq 0),
\end{equation}
where we denote $\ell_n(m) := 1 + 2\lceil \log(m+1)/\log(n)\rceil$. 
\end{lemma}
\begin{proof}
The claim is immediate for $\tau = 0$, so fix $\tau > 0$. 
Both $\pi$ and $\rho_\tau$ have first-coordinate marginal $\gamma_n$, and conditional on the first coordinate being $z \in \r^n$ the last coordinate under $\pi$ has law $\mathcal{N}( \Psi(z), 1)$ and under $\rho_\tau$ has law
\begin{equation}
    \operatorname{Law}\left(W + e^{-\tau} \Psi\left(e^{-\tau}z + \sqrt{1-e^{-2\tau}}G\right)\right)
\end{equation}
with $W \sim \gamma_1$, $G \sim \gamma_n$ independent; this can be shown via Gaussian regression, see the proof of Lemma \ref{lemma:stupidnonsenserevised}. In particular, both conditional laws are a standard Gaussian plus an independent shift supported on $[-\beta, \beta]$. Thus, with $U$ a random variable drawn from the conditional law of the last coordinate under either $\pi$ or $\rho_\tau$ we know
\begin{equation}
    \log \mathbb{E} e^{t(U - \mathbb{E}U)} \leq \frac{1+\beta^2}{2}t^2 \quad (\forall t \in \r),
\end{equation}
where the bounded shift contributes $\beta^2 t^2/2$ by Hoeffding's lemma and the Gaussian contributes $t^2/2$. Now, we know the variational inequality that for probability measures $P, Q$ on $\r$ we have
\begin{align}
    \operatorname{KL}(P \| Q) &\geq \sup_{f \text{ Lipschitz}}\left\{ \mathbb{E}_P[f] - \log \mathbb{E}_Q\left[e^f\right]\right\}
    \\ &\geq \sup_{t \in \r} \left\{t(\mathbb{E}_PY - \mathbb{E}_QY) - \log \mathbb{E}_Q e^{t(Y - \mathbb{E}_QY)}\right\} ,
\end{align}
where we apply the first line with $f(y) = t(y - \mathbb{E}_Q[y])$ to get the second. Thus, with $P$ and $Q$ the distributions of the last coordinate under $\pi$ and $\rho_\tau$ conditioned on the first being $z$, we have
\begin{equation}
    \operatorname{KL}(P \| Q) \geq \sup_{t \in \r} \left\{t(\mathbb{E}_PY - \mathbb{E}_QY) - \frac{1+\beta^2}{2}t^2\right\} = \frac{\left|\mathbb{E}_PY - \mathbb{E}_QY\right|^2}{2(1+\beta^2)} = \frac{|\Psi(z) - e^{-\tau}P_\tau \Psi(z)|^2}{2(1+\beta^2)}.
\end{equation}
The same argument works with $P$ and $Q$ swapped. Since the marginals of the first coordinates coincide, by the chain rule for relative entropy we find
\begin{equation}
    D_\tau \geq \frac{\|\Psi - e^{-\tau} P_\tau \Psi\|_{L^2(\gamma_n)}^2}{2(1+\beta^2)} = \frac{1}{2(1+\beta^2)} \sum_{k=0}^\infty (1 - e^{-(k+1)\tau})^2 \|M_k(\Psi)\|_F^2,
\end{equation}
where we expand in Hermite chaoses. For each $m \in \n \cup \{0\}$, by definition of $\ell_n(m)$ we know $k+1 < \ell_n(m) \implies n^{\lfloor k/2\rfloor} \leq m$. Hence,
\begin{align}
    D_\tau &\geq \frac{1}{2(1+\beta^2)} \sum_{k=0}^\infty (1-e^{-(k+1)\tau})^2 \sum_{j=m+1}^{n^{\lfloor k/2\rfloor}} \sigma_j(M_k(\Psi))^2 \\ &\geq \frac{(1-e^{-\ell_n(m)\tau})^2}{2(1+\beta^2)} \sum_{k=0}^\infty \sum_{j=m+1}^{n^{\lfloor k/2\rfloor}} \sigma_j(M_k(\Psi))^2 \\&= \frac{(1-e^{-\ell_n(m)\tau})^2}{2(1+\beta^2)} a_\Psi(m, 0)^2.
\end{align}
Taking square roots and maximizing over $m$ proves the claim. Since $a_\Psi(m, 0) \to 0$ as $m \to \infty$, this maximum is attained at a finite $m$.
\end{proof}

\subsection{Reverse OU suboptimality for ResNet pushforwards}
\label{subsec:appendixapplicationtoResNetpushsforwards}

In the previous section, we dealt with the general class of Gaussian triangular shear measures
\begin{equation}
    \pi^{\psi, \Psi} = \operatorname{Law}(Z, \psi(U) + \Psi(Z)) \in \p(\r^{n+1}), \quad (Z, U) \sim \gamma_{n+1}
\end{equation}
for $\psi: \r \to \r$ and $\Psi: \r^n \to \r$. 
For this class, we developed some tools to get lower and upper bounds for the representation and approximation of the reverse OU scores $s_\tau^{\psi, \Psi}, b_\tau^{\psi, \Psi}$ by two-layer neural networks. In this section, we will (1) describe a very natural subclass of $\pi^{\psi, \Psi}$ given by pushforwards of $\gamma_{n+1}$ by certain ResNets and (2) verify that the induced shear maps $(\psi, \Psi)$ satisfy the assumptions needed to apply the general lower and upper bounds.

Our class of ResNets will have blocks which each perform a shear operation (i.e. $x \mapsto x+ u \phi(\la x, \theta\ra)$ with $\la u, \theta\ra = 0$). In words, the architecture is:
\begin{enumerate}
    \item First, there are $L$ layers which separately perform computation on the first $n$ coordinates (labeled $z$) and on the last coordinate (labeled $u$). We denote the input by $(z^{(0)}, u^{(0)}) \in \r^{n+1}$, and after layer $\ell$ we denote the hidden state by $(z^{(\ell)}, u^{(\ell)}) \in \r^{n+1}$. The computations in the $z$ and $u$ channels are performed independently. After these $L$ layers, the value in the $u$ coordinate is $u^{(L)} = \psi(u^{(0)})$.
    \item Next, there is a ``readout" layer which uses the state $z^{(L)}$ of the $z$ coordinate to compute $\Psi$ and add its output to the $u$ coordinate. After this layer, the last coordinate contains $u^{(L+1)} = \psi(u^{(0)}) + \Psi(z^{(0)})$ while the $z$ coordinate remains $z^{(L+1)} = z^{(L)}$. 
    \item Lastly, there are another $L$ layers which only act on the $z$ coordinate and leave the $u$ coordinate unchanged. These layers implement the inverses of the original $L$ layers in reverse order. In particular, $u^{(2L+1)} = u^{(L+1)}$ and  $z^{(2L+1)} = z^{(0)}$. 
\end{enumerate}
Combining these three stages produces a neural network of depth $2L+1$ whose input-output map $F: \r^{n+1} \to \r^{n+1}$ is given by $F(z^{(0)}, u^{(0)}) = (z^{(0)}, \psi(u^{(0)}) + \Psi(z^{(0)}))$. Hence, $F_\# \gamma_{n+1} = \pi^{\psi, \Psi}$. This architecture allows us to represent $\pi^{\psi, \Psi}$ as a ResNet pushforward of the Gaussian assuming that we can (1) implement $\Psi$ using $L$ layers of single-index shears in the $z$ channel along with a readout layer and (2) implement $\psi$ using $L$ layers in the $u$ channel. The shear structure of the $z$ channel is crucial, since it allows us to invert the entire stage. As an added bonus, it will let us represent each layer using a continuous-time flow.

We now introduce notation for this construction. For an activation $\varphi: \r \to \r$ and a width $m \in \n$, we define a class of functions from $\r \to \r$ by
\begin{equation}
    \mathcal{N}_{\varphi, m}:= \left\{ x \mapsto c_0 + \sum_{j=1}^m c_j \varphi(\alpha_j x  + \beta_j);\quad (c_j)_{j=0}^m, (\alpha_j)_{j=1}^m,(\beta_j)_{j=1}^m\subset \r\right\}.
\end{equation}
These are the functions which can be implemented by width-$m$ neural networks with activation $\varphi$. Any $ \widetilde{\varphi} \in \mathcal{N}_{\varphi, m}$ can be treated as a compatible activation, in the sense that any layer of width $w$ with activation $\widetilde{\varphi}$ can be implemented using a layer of width $wm$ with activation $\varphi$.

\begin{definition}[Triangular ResNet]
\label{definition:triangularresnet}
Fix an activation $\varphi: \r \to \r$ and $n,W, L \in \n$. 
Choose scalar functions $g, \chi \in \mathcal{N}_{\varphi, W}$. For each $\ell \in \{1, \ldots, L\}$, choose $ h_\ell \in \mathcal{N}_{\varphi, W}$ and input/output features $ a_\ell, b_\ell \in \r^n$ with $\la a_\ell, b_\ell\ra = 0$. Choose also a readout direction $v \in \mathbb{S}^{n-1}$.
Using these choices, we define a ResNet:
\begin{enumerate}
    \item With $(z^{(0)}, u^{(0)}) \in \r^{n+1}$ the input, the first stage of $L$ layers is
    \begin{equation}
\begin{aligned}[c]
z^{(\ell)}
  &= z^{(\ell-1)}
     + b_\ell g\left(
       \la z^{(\ell-1)}, a_\ell\ra
     \right), \\
u^{(\ell)}
  &= u^{(\ell-1)}
     + h_\ell\left(u^{(\ell-1)}\right),
\end{aligned}
\qquad
(\ell \in \{1, \ldots, L\}).
\end{equation}

\item The middle readout stage is 
    \begin{equation}
\begin{aligned}[c]
z^{(L+1)}
  &= z^{(L)}, \\
u^{(L+1)}
  &= u^{(L)}
     + \chi\left(\la z^{(L)}, v\ra\right).
\end{aligned}
\end{equation}

\item Denoting $\ell := L+1-j$, the third stage of $L$ layers is
    \begin{equation}
\begin{aligned}[c]
z^{(L+j+1)}
  &= z^{(L+j)}
     - b_\ell g\left(
       \la z^{(L+j)}, a_\ell\ra
     \right), \\
u^{(L+j+1)}
  &= u^{(L+j)},
\end{aligned}
\qquad
(j \in \{1, \ldots, L\}).
\end{equation}

\end{enumerate}
Denoting the parameters by $\theta := \left(g, \chi, (h_\ell, a_\ell,b_\ell)_{\ell=1}^L,v\right)$, we write
$\operatorname{TResNet}_{\theta}: \r^{n+1} \to \r^{n+1}$ to represent the mapping sending $(z^{(0)}, u^{(0)}) \mapsto (z^{(2L+1)}, u^{(2L+1)})$. 

We also denote the Gaussian pushforward by $\pi^\theta := (\operatorname{TResNet}_\theta)_\# \gamma_{n+1}$, the OU flow by $\rho_\tau^\theta := P_\tau^* \pi^\theta$, and the Lebesgue and Gaussian scores respectively by $s_\tau^\theta := \nabla \log \rho_\tau^\theta$ and $b_\tau^\theta := \nabla \log \frac{\rho_\tau^\theta}{\gamma_{n+1}}$.
\end{definition}

This is a standard ResNet of depth $2L+1$ using blocks which each have one hidden layer of $2W$ neurons with activation $\varphi$. Throughout this section, we will use $\varphi$ to denote the base activation that the neural network is implemented with. We use the notation ``TResNet" to emphasize the triangular structure, since we compute the $z$ and $u$ coordinates separately, combine their outputs, and roll the $z$ computation back to its initial state. In the $(\psi, \Psi)$ notation from earlier, we have:

\begin{lemma}[Triangular structure]
    Let $\theta = \left(g, \chi, (h_\ell, a_\ell,b_\ell)_{\ell=1}^L,v\right)$. Then,
    \begin{equation}
        \operatorname{TResNet}_\theta(z, u) = (z, \psi_\theta(u) + \Psi_\theta(z))
    \end{equation}
    where we define $\psi_\theta: \r \to \r$ by
    $\psi_\theta := (\operatorname{Id} + h_L) \circ \cdots \circ (\operatorname{Id} + h_1)$
    and $\Psi_\theta: \r^n \to \r$ by
    \begin{equation}
        \Psi_\theta(z) := \chi(\la G_L(z), v\ra), \quad G_\ell(z) := G_{\ell-1}(z) + b_\ell g \left(\la G_{\ell - 1}(z), a_\ell\ra\right), \quad G_0(z) := z.
    \end{equation}
    In particular, $\pi^\theta = \pi^{\psi_\theta, \Psi_\theta}$, $s_\tau^\theta = s_\tau^{\psi_\theta, \Psi_\theta}$, and $b_\tau^\theta = b_\tau^{\psi_\theta, \Psi_\theta}$.
\end{lemma}
\begin{proof}
    Defining the map $S_\ell(x) := x + b_\ell g(\la x, a_\ell\ra)$, since $\la a_\ell, b_\ell\ra = 0$ we know that
    \begin{equation}
        S_\ell^{-1}(x) = x - b_\ell g(\la x, a_\ell\ra). 
    \end{equation}
    Since $G_\ell = S_\ell \circ \cdots \circ S_1$, the claim follows by construction.
\end{proof}

\subsubsection{Representation lower and upper bounds}
We first apply Lemma \ref{lemma:stupidnonsenserevised} for a shallow representation lower bound of the OU scores $s_\tau^{\theta},b_\tau^{\theta}$. The mechanism is simple: composing nonlinearities can amplify derivatives exponentially. We will first present a general condition, followed by a concrete construction. 

\begin{lemma}[Shallow representation lower bound]
\label{lemma:appendixrepresentationlowerboundresnets}
Fix $\omega \in C^5_b(\r) \cap \mathcal{N}_{\varphi, W}$ s.t.
\begin{equation}
\label{eqn:proofappendixrepresentationconditionsonomega}
    \omega(0) = 0, \quad \omega'(0) > 0, \quad \omega''(0) = 0, \quad \omega'''(0)\neq 0.
\end{equation}
Then, there is a constant $\Lambda_\omega \in (0, \infty)$ depending only on $\omega$ for which the following holds. 
    Let $\theta = \left(g, \chi, (h_\ell, a_\ell,b_\ell)_{\ell=1}^L,v\right)$ satisfy the following properties:
    \begin{enumerate}
    \item[(i)] We have $g = \Lambda_\omega \omega$ and $\chi = \omega$.
        \item[(ii)] For every $\ell \in \{1, \ldots, L\}$, $\operatorname{Id} + h_\ell$ is increasing with $h_\ell = 0$ on $[-2\|\omega\|_\infty, 2\|\omega\|_\infty]$.
        \item[(iii)] For every $\ell \in \{1, \ldots, L\}$ we have $\|a_\ell\| = \|b_\ell\| = 1$ and $\la a_\ell, b_\ell\ra = 0$. Also, $a_{\ell + 1} = b_\ell$ for every $\ell \in \{1, \ldots, L-1\}$  along with $b_L = v$. 
    \end{enumerate}
    Then, there are constants $B_\omega \in (0, \infty)$, $c_\omega, C_\omega \in (1, \infty)$ depending only on $\omega$ s.t. $C_\omega / c_\omega \geq 2$ and \begin{equation}
       6\|s_\tau^{\theta}\|_{\b_{\operatorname{ReLU}^3}}  \geq  \|\partial_{a_1}^3 s_\tau^{\theta}\|_{\infty} \geq B_\omega C_\omega^{L} \quad \left(\forall \tau \in \left(0, {B_\omega}{c_\omega^{- L}}\right]\right).
    \end{equation}
    The same statement holds for $b_\tau^{\theta}$. In particular, $\mathcal{E}_{\b_{\operatorname{ReLU}^3}}(\pi^{\theta}) \geq B_\omega^2 2^L$ and if $\phi \in C^3(\r)$ satisfies Assumption \ref{assumption:activationfunciton} with $\|\phi\|_\infty, \|\phi'''\|_\infty < \infty $ then $\inf_{R > 0} R^2 \mathcal{E}_{\b_\phi^R}(\pi^\theta) \geq B_\omega^2 2^L / \|\phi'''\|_\infty$.
    
\end{lemma}
\begin{remark}
    In the case $\{a_1, \ldots, a_L, v\} \subset \mathbb{S}^{n-1}$ is an orthonormal set, due to the chaining in assumption (iii) this construction has the particularly clean form
    \begin{align}
        \Psi_\theta(z) &= \omega\left(\la z, v\ra + \Lambda_\omega \omega\left(\la z, a_L\ra + \Lambda_\omega \omega\left( \cdots + \Lambda_\omega \omega\left(\la z, a_1\ra\right)\right)\right)\right).
    \end{align}
    One can choose $(h_\ell)_{\ell = 1}^L$ to produce $\psi_\theta = (\operatorname{Id} + h_L) \circ \cdots \circ (\operatorname{Id} + h_1)$ with complicated behavior outside $[-2\|\omega\|_\infty, 2\|\omega\|_\infty]$, but inside this region $\psi_\theta$ is the identity by assumption (ii).
\end{remark}
\begin{proof}
    Let us denote $\kappa_1 := \omega'(0)$, $K_\omega := 52(1 + \|\omega\|_{C^5})$, and
    \begin{equation}
        \Lambda_\omega := \max\left\{1,\; \frac{6}{\kappa_1}, \; \frac{1458K_\omega^5}{\kappa_1^6}\right\}.
    \end{equation}
    We also set $C_\omega := (\kappa_1 \Lambda_\omega-2)^3$ and $c_\omega := 2K_\omega^5 (1+\Lambda_\omega)^5 / C_\omega$ as well as
    \begin{equation}
        B_\omega := |\omega'''(0)| \min\left\{\frac{1}{2},\; \frac{L_\beta}{K_\omega^5}\right\}
    \end{equation}
    where $\beta := \|\omega\|_\infty$ and $L_\beta$ is the constant from Lemma \ref{lemma:stupidnonsenserevised}. Since $\kappa_1\Lambda_\omega \geq 6$ and $\Lambda_\omega \geq 1$ we know 
    $ \kappa_1\Lambda_\omega - 2 \geq \frac{2}{3}\kappa_1 \Lambda_\omega$ and $1 + \Lambda_\omega \leq 2\Lambda_\omega$. Hence,
    \begin{equation}
        \frac{C_\omega}{c_\omega} = \frac{(\kappa_1 \Lambda_\omega - 2)^6}{2K_\omega^5(1+\Lambda_\omega)^5} \geq \frac{\kappa_1^6 \Lambda_\omega}{729K_\omega^5} \geq 2.
    \end{equation}
    Also, $K_\omega \geq \kappa_1$ and so $K_\omega(1+\Lambda_\omega) > \kappa_1\Lambda_\omega > \kappa_1 \Lambda_\omega - 2 \geq 4$. Thus,
    \begin{equation}
        c_\omega = \frac{2K_\omega^5 (1+\Lambda_\omega)^5}{(\kappa_1\Lambda_\omega - 2)^3} > 2(\kappa_1\Lambda_\omega - 2)^2 \geq 32
    \end{equation}
    and so $1 < c_\omega \leq C_\omega / 2$. Now, choose an orthonormal basis of $\r^n$ whose first vector is $a_1$ and set
    \begin{equation}
        D := K_{\omega }^{L+1}(1+\Lambda_\omega)^L.
    \end{equation}
    By Faà di Bruno applied inductively to the recursion defining $G_\ell$, we have
    \begin{equation}
        \|\partial^\alpha G_\ell\|_\infty \leq (K_\omega (1 + \Lambda_\omega))^{\ell |\alpha|} \quad (1 \leq |\alpha| \leq 5,\; \forall \ell \in \{1, \ldots, L\}),
    \end{equation}
    since the number of partitions of a set of at most 5 elements is $\leq 52$. Applying this formula once more to $\Psi_\theta(z) = \omega(\la G_L(z), v\ra)$ gives
    \begin{equation}
        \|\Psi_\theta\|_\infty \leq \beta, \quad \|\partial^\alpha \Psi_\theta\|_\infty  \leq D^{|\alpha|} \quad (1 \leq |\alpha| \leq 5).
    \end{equation}
    It remains to lower bound $\|\partial_{a_1}^3 \Psi_\theta\|_\infty$. For this, we denote
    \begin{equation}
        w_0 := a_1, \quad w_\ell := b_\ell,\quad J_\ell := \1_n + \kappa_1 \Lambda_\omega w_\ell w_{\ell - 1}^\top
    \end{equation}
    for $\ell \in \{1, \ldots, L\}$. For $0 \leq i < j \leq L$ we define
    \begin{equation}
        P_{i,j} := \la w_j, J_j \cdots J_{i+1}w_i\ra, \quad P_{i,i} := 1.
    \end{equation}
    We claim that
    \begin{equation}
    \label{eqn:proofappendixresnetpushforwardrepresentationinductionpropagation}
        P_{i,j} \geq (\kappa_1 \Lambda_\omega - 2)^{j-i}.
    \end{equation}
    Indeed, starting from $u_i := w_i$ we recursively set $u_k := J_k u_{k-1}$ for $k \in \{i+1, \ldots, j\}$ and $A_k := \la w_{k-1}, u_{k-1}\ra$ for $k \in \{i+1, \ldots, j+1\}$. We prove inductively that $A_k > 0$ and $\|u_{k-1}\| \leq 2A_k$. For $k = i+1$ this is trivial, and since
    \begin{equation}
        u_k = u_{k-1} + \kappa_1 \Lambda_\omega A_k w_k
    \end{equation}
    we have that
    \begin{align}
        A_{k+1} &= \la w_k, u_k\ra = \la w_k, u_{k-1}\ra + \kappa_1 \Lambda_\omega A_k \geq -\|u_{k-1}\| + \kappa_1 \Lambda_\omega A_k \geq (\kappa_1 \Lambda_\omega - 2) A_k > 0.
    \end{align}
    Moreover, 
    \begin{align}
        \|u_k\| \leq \|u_{k-1}\| + \kappa_1 \Lambda_\omega A_k \leq (\kappa_1\Lambda_\omega + 2)A_k \leq 2(\kappa_1 \Lambda_\omega - 2)A_k \leq 2A_{k+1},
    \end{align}
    where the second to last inequality uses that $\kappa_1 \Lambda_\omega \geq 6$. This closes the induction, and hence $(\kappa_1\Lambda_\omega - 2)A_k \leq A_{k+1}$ for all $k$. Since $A_{j+1} = P_{i,j}$, (\ref{eqn:proofappendixresnetpushforwardrepresentationinductionpropagation}) follows. 

    Now, let us define \begin{equation}
        F(t) := \la w_L, G_L(tw_0)\ra.
    \end{equation}
From the recursion defining $G_\ell$, we know that
    \begin{equation}
        G_\ell(tw_0) = G_{\ell - 1}(tw_0) + \Lambda_\omega w_\ell \omega(\la w_{\ell - 1}, G_{\ell - 1}(tw_0)\ra). 
    \end{equation}
    Since $G_\ell(0) = 0$ (due to $\omega(0) = 0$), differentiating once at $t = 0$ yields
    \begin{equation}
        \left.\frac{d}{dt}G_\ell(tw_0)\right|_{t=0} = J_\ell \left.\frac{d}{dt}G_{\ell - 1}(tw_0)\right|_{t=0}
    \end{equation}
    which we iterate to find that
    \begin{equation}
        F'(0) = \la w_L, J_L \cdots J_1 w_0\ra = P_{0,L}.
    \end{equation}
    Similarly, differentiating the recursion twice yields
    \begin{equation}
        \left.\frac{d^2}{dt^2}G_\ell(tw_0)\right|_{t=0} = J_\ell \left.\frac{d^2}{dt^2}G_{\ell - 1}(tw_0)\right|_{t=0} + \Lambda_\omega w_\ell \omega''(0) \la w_{\ell - 1}, \left.\frac{d}{dt}G_{\ell - 1}(tw_0)\right|_{t=0}\ra^2.
    \end{equation}
    Since $\omega''(0) = 0$ and $G_0(tw_0) = tw_0$ has zero second derivative at $t=0$, by induction we find $ \left.\frac{d^2}{dt^2}G_\ell(tw_0)\right|_{t=0} = 0$ for each $\ell$ and therefore $F''(0) = 0$. Differentiating a third time, all terms involving $\omega''(0)$ or a second derivative of $G_{\ell - 1}(tw_0)$ vanish. Thus,
    \begin{align}
         \left.\frac{d^3}{dt^3}G_\ell(tw_0)\right|_{t=0} &= J_\ell \left.\frac{d^3}{dt^3}G_{\ell - 1}(tw_0)\right|_{t=0} + \Lambda_\omega w_\ell \omega'''(0) \la w_{\ell - 1}, \left.\frac{d}{dt}G_{\ell - 1}(tw_0)\right|_{t=0}\ra^3 \\
         &= J_\ell \left.\frac{d^3}{dt^3}G_{\ell - 1}(tw_0)\right|_{t=0} + \Lambda_\omega w_\ell \omega'''(0) P_{0,\ell - 1}^3.
    \end{align}
    By iterating this and then projecting onto $w_L$, we find that
    \begin{equation}
        F'''(0) = \Lambda_\omega \omega'''(0) \sum_{\ell = 1}^LP_{0,\ell - 1}^3 P_{\ell, L}
    \end{equation}
    since the cubic term $w_\ell P_{0,\ell-1}^3$ is propagated by $J_L \cdots J_{\ell + 1}$. Finally, since $\Psi_\theta(tw_0)  = \omega(F(t))$ we apply the chain rule to find
    \begin{equation}
        \frac{d^3}{dt^3} \omega(F(t)) = \omega'''(F(t))F'(t)^3 + 3\omega''(F(t)) F'(t) F''(t) + \omega'(F(t)) F'''(t).
    \end{equation}
    At $t = 0$ we have $F(0) = F''(0) = \omega''(0) = 0$ and $\omega'(0) = \kappa_1$, and so
    \begin{equation}
        \partial_{a_1}^3 \Psi_\theta(0) = \omega'''(0) F'(0)^3 + \kappa_1 F'''(0) = \omega'''(0) \left(P_{0,L}^3 + \kappa_1 \Lambda_\omega \sum_{\ell = 1}^L P_{0, \ell-1}^3 P_{\ell,L}\right).
    \end{equation}
    Since all terms in the parentheses are positive, there is no cancellation and so
    \begin{equation}
        \|\partial_{a_1}^3 \Psi_\theta\|_\infty \geq |\partial_{a_1}^3 \Psi_\theta(0)| \geq |\omega'''(0)| P_{0,L}^3 \geq |\omega'''(0)| (\kappa_1 \Lambda_\omega - 2)^{3L} = |\omega'''(0)|  C_\omega^L,
    \end{equation}
    where we applied (\ref{eqn:proofappendixresnetpushforwardrepresentationinductionpropagation}). By assumption (ii) we know that $\psi_\theta$ is increasing and the identity on $[-2\beta, 2\beta]$.
    By Lemma \ref{lemma:stupidnonsenserevised}, we have that \begin{equation}\|\partial_{a_1}^3 s_\tau^\theta\|_\infty \geq \frac{|\omega'''(0)|}{2}C_\omega^L \geq B_\omega C_\omega^L\end{equation} whenever $\tau \leq L_\beta \frac{|\omega'''(0)| C_\omega^L}{(L+1)D^5}$. Note that $\Psi_\theta$ depends on a subspace of dimension at most $L+1$ and so we replace $n $ with $L+1$ for this application (since the OU semigroup tensorizes).
    Since $L+1 \leq 2^L$, we find that
    \begin{align}
         L_\beta \frac{|\omega'''(0)| C_\omega^L}{(L+1)D^5}  &= L_\beta \frac{|\omega'''(0)|}{(L+1)K_\omega^5} \cdot \left(\frac{C_\omega}{K_\omega^5 (1+\Lambda_\omega)^5}\right)^L \\&\geq L_\beta \frac{|\omega'''(0)| }{K_\omega^5} \cdot \left(\frac{C_\omega}{2K_\omega^5 (1+\Lambda_\omega)^5}\right)^L 
         \\ &= L_\beta \frac{|\omega'''(0)|}{K_\omega^5} \cdot c_\omega^{-L} \\&\geq B_\omega c_\omega^{-L}.
    \end{align}
    The rest of the result follows from the proof of Corollary \ref{corollary:representationlowerbound}.
\end{proof}
\begin{remark}
    Note that this obstruction is proved only for exponentially-small OU times. We create small but very rapid variations, which amplify the third derivative but become hidden under a small amount of noise. 
\end{remark}

We may also ensure that the Barron norms of the OU scores are finite for the examples produced in Lemma \ref{lemma:appendixrepresentationlowerboundresnets}. By Lemma \ref{lemma:appendixrepresentationupperbound}, it is sufficient to show that $\Psi_\theta$ is a finite sum of smooth compactly-supported cylindrical functions and $\psi_\theta$ has polynomial tails. Both of these can be achieved using a suitable base activation $\varphi$:

\begin{lemma}[Shallow representation upper bound]
\label{lemma:appendixrepresentationupperboundresnets}
    Assume the setting of Lemma \ref{lemma:appendixrepresentationlowerboundresnets}. Assume further that $\omega \in C^\infty_c \cap \mathcal{N}_{\varphi, W}$ and that $\psi := (\operatorname{Id} + h_L) \circ \cdots \circ (\operatorname{Id} + h_1)$ is a $C^\infty$ diffeomorphism with
    \begin{equation}
    \label{eqn:proofappendixresnetrepresentationupperboundtailcondition}
       \psi(u) = \begin{cases}
            -a_- (-u)^p + b_- & u \leq -R \\
            a_+ u^p + b_+ & u \geq R
        \end{cases}
    \end{equation}
    for some $a_\pm \in (0, \infty)$, $b_\pm \in \r$, $R > 2\|\omega\|_\infty$, and $p > 2$. Then, $s_\tau^\theta \in \b_{\operatorname{ReLU}^s}$ for all $\tau \geq 0$ and $s \in \n$ with the uniform bound
    \begin{equation}
        \sup_{\tau \in [0, T]}\|s_\tau^\theta\|_{\b_{\operatorname{ReLU}^s}} < \infty \quad (\forall T \in (0, \infty)).
    \end{equation}
    The same statements hold with $s_\tau^\theta$ replaced by $b_\tau^\theta$.
\end{lemma}\begin{proof}
We know that $\psi_\theta = \psi$. By the assumed tail behavior, we know that $\psi_\theta$ satisfies the assumptions of Lemma \ref{lemma:appendixrepresentationupperbound}. It remains to verify the assumptions on $\Psi_\theta$. We first show that if $f(z) = F(P_Ez)$ is a cylindrical function with $F \in C^\infty_c(E)$ and $S(z) := z + bg(\la a, z\ra)$ is a single-index field (with $a, b \in \r^n$) then $f \circ S$ is a sum of two cylindrical functions which are $C^\infty_c$ on their active subspaces. Indeed, with $V := E + \operatorname{span}\{a\}$ we define for $x \in V$ the function
\begin{equation}
    H(x) := F(P_Ex + P_E b g(\la a, x\ra)) - F(P_Ex)
\end{equation}
and so $f(S(z)) = F(P_Ez) + H(P_Vz)$. Clearly, $H$ is smooth, and since $\|x\| \lesssim_{E, a} \|P_Ex\| + |\la a, x\ra|$ we find that $H \in C^\infty_c(V)$. 
By linearity, if $f$ is a finite sum of cylindrical functions, each of which is $C^\infty_c$ on its active subspace, then the same is true for $f \circ S$.

Now, in the present construction we have $g, \chi \in C^\infty_c(\r)$ and $\Psi_\theta = f_0 \circ S_L \circ \cdots \circ S_1$ with $f_0(z) = \omega(\la v, z\ra)$. Since shears preserve the finite structure of being a sum of smooth cylindrical functions, $\Psi_\theta$ satisfies the assumptions of Lemma \ref{lemma:appendixrepresentationupperbound}.
\end{proof}

We will now instantiate these so that the lower and upper bounds both apply to the same teacher ResNet. We choose $\phi = \operatorname{ReLU}^3$ for the student activation, and we must take $\varphi$ to be a smooth mollification of $\phi$ to ensure smoothness of $\omega$ and $\psi$:

\begin{proposition}[Shallow representation bounds for $\operatorname{ReLU}^3$]
\label{prop:appendixrepresentaitonupperlowerbounds}
Let  $\phi(t) = \max\{0, t\}^3$ be the $\operatorname{ReLU}^3$ activation, and define $\varphi = \phi * \eta$ using some nonnegative even $\eta \in C^\infty_c(\r)$ with $\int_\r \eta = 1$ which is supported in $[-\delta, \delta]$ for some $\delta \in (0, 1/4)$. 

Then, for each $d \geq 4$ there exists a $\operatorname{TResNet}_\theta: \r^d \to \r^d$ of depth $2d-3$ and width $16$ with two-layer blocks using activation $\varphi$ s.t. 
\begin{equation}
    \mathcal{E}_{\b_\phi}(\pi^\theta) \gtrsim 2^d, \quad D_{\b_\phi}(\gamma_d, \pi^\theta) \lesssim d, \quad \text{and } s_\tau^\theta, b_\tau^\theta \in \b_\phi \quad (\forall \tau \geq 0),
\end{equation}
where the implied constants depend only on $\eta$.
\end{proposition}
\begin{proof}
By mollification, if we denote $m_2 := \int_\r r^2 \eta(r) dr \leq \delta^2$ then $\varphi(t) = 0$ for all $t \leq -\delta$, while for $t \geq \delta$ we have
\begin{align}
    \varphi(t) &= \int_\r (t-r)^3 \eta(r) dr \\&= t^3 \int_\r \eta(r) dr - 3t^2 \int_r r \eta(r) dr + 3t \int_\r r^2\eta(r) dr - \int_\r r^3 \eta(r) dr \\
    &= t^3 + 3tm_2
\end{align}
since $\eta$ is even. 
We define the scalar function $\omega \in \mathcal{N}_{\varphi, 10}$ using 10 neurons in the following way:
    \begin{equation}
        \omega(t) := B\left(t - \frac{1}{2}\right) - B\left(t + \frac{1}{2}\right), \quad B(t) := \frac{1}{6}\sum_{j=0}^4 (-1)^j {4 \choose j} \varphi(t + 2 - j).
    \end{equation}
    Note that $\omega \in C_c^\infty(\r)$ (since $B(\cdot)$ is a mollification of the discrete fourth derivative of $\operatorname{ReLU}^3$, which is compactly supported), and in particular $\|\omega\|_\infty < \infty$. Furthermore, since the unsmoothed discrete derivative equals $\frac{5}{4}t - t^3$ near zero, convolution with the even mollifier implies that 
    \begin{equation}
        \omega(t) = \left(\frac{5}{4} - 3m_2\right)t - t^3 \quad \left(\forall |t| < \frac{1}{2}- \delta\right). 
    \end{equation}
Thus, $\omega(0) = 0$, $\omega'(0) = \frac{5}{4} - 3m_2 > 0$, $\omega''(0) = 0$, and $\omega'''(0) = -6 \neq 0$, i.e. $\omega$ satisfies (\ref{eqn:proofappendixrepresentationconditionsonomega}). 

 Now, take $R := 2\|\omega\|_\infty + 1> 2\|\omega\|_\infty + \delta$ and set
 \begin{equation}
     \sigma := \sqrt{\frac{23}{3}R^2 - 4m_2}, \quad r = \frac{R+\sigma}{2}.
 \end{equation}
 Define the scalar function $h \in \mathcal{N}_{\varphi, 6}$ using 6 neurons in the following way:
 \begin{equation}
 \label{eqn:proofappendixdefinitionofh1forconcreterepresentationresnet}
     h(u) := \frac{G(u) - G(-u)}{3R\sigma}, \quad G(u) := 2\varphi(u-r) - 3\varphi(u-r-R) + \varphi(u-r-2R).
 \end{equation}
 Since $\sigma > R$ we have $r > R$, which means for every $|u| \leq 2 \|\omega\|_\infty$ we know $G(u) = 0$ and so $h(u) = 0$. Also, since $G'$ is the mollification of \begin{equation}3(2\max\{0, u-r\})^2 - 3\max\{0, u-r-R\}^2 + \max\{0, u-r-2R\}^2),\end{equation} 
 which is everywhere nonnegative, we know $G' \geq 0$. For $u \geq r + 2R + \delta$ we know $G(-u) = 0$ and we may expand the cubic tails to find that if $v = u-r$ then
 \begin{align}
     G(u) &= 2v^3 + 6vm_2 - 3(v-R)^3 - 9(v-R)m_2 + (v-2R)^3 + 3(v-2R)m_2 \\
     &= 2v^3 - 3(v-R)^3 + (v-2R)^3 + 3m_2 (2v - 3(v-R) + (v-2R)) \\
     &= 3Rv^2 + 3R^2v - 5R^3 + 3m_2 R\\
     &= 3R\left((u-r)^2 + R(u-r) - \frac{5}{3}R^2 + m_2\right)
     \\&= 3R(u^2 - \sigma u),
 \end{align}
 where we used that $2r-R = \sigma$ and $r^2 -Rr = \frac{5}{3}R^2 - m_2$. By oddness, this implies
 \begin{equation}
    u + h(u) = \frac{u|u|}{\sigma}  \quad (\forall |u| \geq r+2R + \delta).
 \end{equation}

    We take $n = d-1$, $L = d-2$, and choose ResNet parameters $\theta = (g, \chi, (h_\ell, a_\ell, b_\ell)_{\ell = 1}^L, v)$ according to the following:
    \begin{enumerate}
        \item[(i)] We take $g = \Lambda_\omega \omega$ and $\chi = \omega$, where $\Lambda_\omega$ is defined in Lemma \ref{lemma:appendixrepresentationlowerboundresnets}. Note that $\omega \in C^\infty_c(\r)$ satisfies (\ref{eqn:proofappendixrepresentationconditionsonomega}) and can be made with 10 neurons with activation $\varphi$, i.e. $\omega \in \mathcal{N}_{\varphi, 10}$.
        \item[(ii)] For each $\ell \in \{3, \ldots, L\}$ we take $h_\ell = 0$, and we take $h_1 = h_2 = h$ as defined in (\ref{eqn:proofappendixdefinitionofh1forconcreterepresentationresnet}); this can be made with 6 neurons with activation $\varphi$, i.e. $h_1,h_2 \in \mathcal{N}_{\varphi, 6}$. Note that $(\operatorname{Id} + h_2)\circ (\operatorname{Id} + h_1)$ is an increasing $C^\infty$ diffeomorphism which satisfies the tail condition (\ref{eqn:proofappendixresnetrepresentationupperboundtailcondition}) with $p=4$.
        \item[(iii)] For each $\ell \in \{1, \ldots, L\}$ we take $a_\ell = e_\ell$ and $b_\ell = e_{\ell +1}$, along with $v = e_{L+1}$.
    \end{enumerate}
    The resulting map $\operatorname{TResNet}_\theta: \r^d \to \r^d$ is a ResNet of depth $2L+1 = 2d-3$ which uses two-layer blocks with activation $\varphi$: the first two have width 16, and all subsequent ones have width 10. The assumptions of Lemmas \ref{lemma:appendixrepresentationlowerboundresnets} and \ref{lemma:appendixrepresentationupperboundresnets} are both satisfied, and so $\mathcal{E}_{\b_\phi}(\pi^\theta) \geq B_\omega^2 2^L = \frac{B_\omega^2}{4}2^d \gtrsim 2^d$ while $s_\tau^\theta, b_\tau^\theta \in \b_\phi$ for all $\tau \geq 0$ (uniformly on compact OU time intervals). 

    We finish by constructing a Barron flow transporting $\gamma_d$ to $\pi^\theta$ which realizes $D_{\b_\phi}(\gamma_d, \pi^\theta) \lesssim d$. Since the ResNet layers in the $z$ coordinate which compute $\Psi^\theta$ are orthogonal shears (as is the readout layer), we can represent them by an ODE with drifts given by the layers themselves at cost $\lesssim d$; this flow computes $\Psi^\theta$, writes the readout, and uncomputes $\Psi^\theta$. Since we can prepend a Barron flow on the $u$ coordinate before this, the claim follows if we can show that
    \begin{equation}
        D_{\b_\phi}(\gamma_1, \psi_\# \gamma_1) < \infty, \quad \psi := (\operatorname{Id} + h) \circ (\operatorname{Id} + h).
    \end{equation}
    It suffices to exhibit a finite-cost flow implementing $F = \operatorname{Id} + h$, which we will run twice. To start, note that the ODE 
    \begin{equation}
    \label{eqn:proofappendixrepresentationupperlowerboundsflowdefiningC}
        \dot{u} = -\frac{1}{2}u^3
    \end{equation}
    has a time-1 map $C: \r \to (-1, 1)$ given by the smooth increasing diffeomorphism
    \begin{equation}
        C(u) = \frac{u}{\sqrt{1+u^2}}.
    \end{equation}
    If we choose a smooth cutoff $0 \leq \zeta \leq 1$ which vanishes on $(-\infty, 1/4]$ and equals 1 on $[1/2, \infty)$ and define the velocity
    \begin{equation}
    \label{eqn:proofappendixrepresentationupperlowerboundsflowdefiningE}
        v(u) := u^2 (\zeta(u) - \zeta(-u)),
    \end{equation}
    then the ODE $\dot{u} = v(u)$ has a time-1 map $E: (-1, 1) \to \r$ which is an odd increasing diffeomorphism with
    \begin{equation}
        E(u) = \frac{u}{1-u} \quad (\forall u \in [1/2, 1)).
    \end{equation}
    Indeed, on this region the ODE is $\dot{u} = u^2$ which is well-posed (since the blowup time is $1/u > 1$), and for other $u \in [0, 1/2)$ comparison with $\dot{u} = u^2$ ensures existence through time 1. By oddness, the same is true for negative initial conditions. Now, we define the smooth increasing diffeomorphism $K := E^{-1} \circ F \circ C^{-1}: (-1, 1) \to (-1, 1)$. For $u$ sufficiently close to 1, the quadratic tails of $F$ give
    \begin{equation}
        F(C^{-1}(u)) = \frac{u^2}{\sigma(1-u^2)}, \quad K(u) = \frac{u^2}{\sigma + (1-\sigma)u^2}. 
    \end{equation}
    Hence, $K$ extends smoothly to 1 with $K(1) = 1$ and $K'(1) = 2\sigma > 0$. By oddness, the same holds at $-1$. Thus, $K(\pm 1) = \pm 1$ and $\min_{[-1, 1]}K' > 0$. Defining the linear interpolation
    \begin{equation}
        K_t(u) := (1-t)u + tK(u), \quad w_t(u) := (K - \operatorname{Id})(K_t^{-1}(u)),
    \end{equation}
    we find that $\partial_t K_t(u) = w_t(K_t(u))$ and that $w_t(u)$ is jointly smooth in $(t, u)$ with fixed compact support. In particular, using the $\operatorname{ReLU}^3$ representation
    \begin{equation}
    \label{eqn:proofappendixrepresentationupperlowerboundsflowdefiningstuff}
        f(u) = \frac{1}{6}\int_\r \phi(u-a) f^{(4)}(a) da
    \end{equation}
    for $f$ which vanish on some $(-\infty, u_0]$ with compactly supported fourth derivative, we may extend $w_t$ to all of $\r$ in a way that
    \begin{equation}
        \sup_{t \in [0, 1]}\|w_t\|_{\b_\phi} \leq \frac{1}{6}\sup_{t \in [0, 1]} \int_\r (1+|u|)^3 |\partial_u^4 w_t(u)|du < \infty.
    \end{equation}
    The flow (\ref{eqn:proofappendixrepresentationupperlowerboundsflowdefiningC}) defining $C$ has the $\operatorname{ReLU}^3$ representation $\frac{\phi(-u) - \phi(u)}{2}$. Lastly, the flow (\ref{eqn:proofappendixrepresentationupperlowerboundsflowdefiningE}) defining $E$ has $v(u) = q(u) - q(-u)$ with $q(u) := u^2 \zeta(u)$. Therefore, $q(u) =  0$ for $u \leq 1/4$ and $q(u) = u^2$ for $u \geq 1/2$, which means $q^{(4)}$ is smooth and supported on $[1/4, 1/2]$. Using the representation (\ref{eqn:proofappendixrepresentationupperlowerboundsflowdefiningstuff}), this implies that $q$ has finite Barron norm which means $v$ does too. Since $F = E \circ K \circ C$, the result follows by concatenating these Barron flows.
\end{proof}

\subsubsection{Approximation lower bounds}
\label{subsec:appendixresnetapproximation}

We will apply Proposition \ref{prop:appendixdaapproximationlowerbound} for a shallow approximation lower bound of the OU scores $s_\tau^{\theta}, b_\tau^{\theta}$. To do so, we require a lower bound on the Hermite width envelope $m_{\Psi_\theta}$, which is defined in terms of spectral tails of the Hermite modes (see Definition \ref{def:appendixhermitewidthenvelope}). 

First, we demonstrate that three-layer neural networks of the form (\ref{eqn:proofappendixapproximationresnetthreelayernetworksofthetype}) produce heavy spectral tails in high Hermite degrees. The mechanism is Hermite branching: the nonlinear Hermite modes of $g$ generate many independent tensor factors along the orthogonal feature directions, producing high-rank Hermite matricizations. A heavy Hermite tail of $\chi$ ensures that these high-rank obstructions occur at high Hermite degrees, where the rank can be large. 
Recalling the definition of $a_\Psi$ from Definition \ref{def:appendixhermitewidthenvelope} and that $\operatorname{He}_j$ denote the monic probabilist's Hermite polynomials, a precise statement is given below:

\begin{lemma}[Hermite branching lemma]
\label{lemma:appendixhermitebranchinglemma}
    Let $g: \r \to \r$ be Lipschitz, centered (i.e. $\mathbb{E}_{G \sim \gamma_1} [g(G)] = 0$), and non-affine in $L^2(\gamma_1)$ (i.e. $\|g\|_{L^2(\gamma_1)} > |\mathbb{E}_{G \sim \gamma_1} [Gg(G)]|$). Let $\chi: \r \to \r$ be bounded and Lipschitz. Fix $L \in \n$ and an orthonormal basis $\{e_0, \ldots, e_L\}$ of $\r^{L+1}$, and define $F_L: \r^{L+1} \to \r$ by
    \begin{equation}
    \label{eqn:proofappendixapproximationresnetthreelayernetworksofthetype}
        F_L(z) := \chi\left(z_0 + \frac{1}{\sqrt{L}} \sum_{\ell = 1}^L g(z_\ell)\right).
    \end{equation}
    For positive integers $K_1 \leq K_2$ we define the quantity
   \begin{equation}
    \mathcal{V}_{\chi, g}(K_1, K_2) := \sum_{j=K_1}^{K_2} \frac{1}{j!}\mathbb{E}_{G \sim \gamma_1} \left[\chi(\sqrt{\lambda_g}G) \operatorname{He}_j(G)\right]^2, \quad \lambda_g := {1 + \|g\|_{L^2(\gamma_1)}^2}.
   \end{equation}
   Then, there are some $b_g, c_g \in (0, \infty)$ and $K_g \in \n$ (depending only on $g$) s.t. for any $K_g \leq K_1 \leq K_2$ there is some $L_0 = L_0(g, \chi, K_1, K_2)$ s.t.
   \begin{equation}
       a_{F_L}(m, \tau) \geq \frac{1}{2}e^{-b_g K_2\tau} \sqrt{\mathcal{V}_{\chi, g} (K_1, K_2)}
   \end{equation}
   for every $L \geq L_0$, every $\tau \geq 0$, and every integer $0 \leq m \leq L^{c_g K_1}$.
\end{lemma}

We defer the proof of Lemma \ref{lemma:appendixhermitebranchinglemma}, which uses a probabilistic construction to control the crossing behavior of matricized Hermite tensors, to the end of this subsection. Combining this result with Proposition \ref{prop:appendixdaapproximationlowerbound} yields a robust approximation lower bound for the scores $s_\tau^{\operatorname{Id}, F_L}, b_\tau^{\operatorname{Id}, F_L}$ along the OU flow initialized from the triangular shear measure $\pi^{\operatorname{Id}, F_L} \in \p(\r^{L+2})$.

Note that $\Psi_\theta = F_L$ and $\psi_\theta = \operatorname{Id}$ (up to a rotation) in the case where $n = L+1$ and the parameters $\theta = (g, \chi, (h_\ell, a_\ell, b_\ell)_{\ell = 1}^L, v)$ are s.t. $\{a_1, \ldots, a_L, v\}$ is an orthonormal basis of $\r^n$, $b_\ell = \frac{1}{\sqrt{L}}v$, and $h_\ell = 0$. For such $\theta$, we have that $\pi^\theta, s^\theta_\tau, b^\theta_\tau$ equal $\pi^{\operatorname{Id}, F_L}, s_\tau^{\operatorname{Id}, F_L}, b_\tau^{\operatorname{Id}, F_L}$. Hence, even though the map $F_L$ is expressed in (\ref{eqn:proofappendixapproximationresnetthreelayernetworksofthetype}) in a parallelized form as a three-layer network, this same construction can be viewed as a deep narrow triangular ResNet in the sense of Definition \ref{definition:triangularresnet}.

\begin{corollary}[Shallow approximation width lower bound]
\label{corollary:approximationlowerbound}
    Let $g, \chi: \r \to \r$ satisfy the assumptions of Lemma \ref{lemma:appendixhermitebranchinglemma}. 
   Suppose further that there are constants $c, K_0 \in (1, \infty)$ and a function $V: \n \to (0, \infty)$ for which
   \begin{equation}
       \mathcal{V}_{\chi, g}(K, \lfloor cK\rfloor) \geq V(K) \quad (\forall K \geq K_0).
   \end{equation}
    For every $\epsilon > 0$, we denote the quantity
    \begin{equation}
        K(\epsilon) := \max\left\{K \geq K_0:\; V(K) \geq C_{\chi, g} \epsilon^2\right\},
    \end{equation}
    where $C_{\chi, g} := 64 Q_{\|\chi\|_\infty, 1}^2 \left(1 + \operatorname{Lip}(\chi)^2 (1 + \operatorname{Lip}(g)^2)\right)^2$ and our convention here is that $\max \emptyset = 0$.
   
   Then, there are constants $\epsilon_0, c_1, c_2 \in (0, \infty)$ depending only on $g, \chi,V,c,K_0$ s.t. the following holds: for each $\epsilon \in (0, \epsilon_0]$ there is some $L_0 = L_0(g, \chi, \epsilon)$ s.t. for every $L \geq L_0$, every $\tau \in [0, c_1 \min\{\epsilon, 1/K(\epsilon)\}]$, and every two-layer neural network $\hat{s}_\tau: \r^{L+2} \to \r^{L+2}$ of width $m$ with
   \begin{equation}\|s_{\tau}^{\operatorname{Id}, F_L} - \hat{s}_\tau\|_{L^2\left(\rho_\tau^{\operatorname{Id}, F_L}; \r^{L+2}\right)} \leq \epsilon\end{equation}  we must have \begin{equation}
       m \geq L^{c_2 K(\epsilon)} \quad \text{and} \quad \|\hat{s}_\tau\|_{\b_\phi} \geq D_{\phi, \chi} \epsilon L^{c_2 K(\epsilon)/2} \quad \forall \phi \text{ satisfying Assumption \ref{assumption:activationfunciton}},
   \end{equation} 
   where $D_{\phi, \chi}>0$ is a constant depending only on $\phi$ and $\chi$.
   
   The same statements hold with $s_\tau^{\operatorname{Id}, F_L}$ replaced by $b_\tau^{\operatorname{Id}, F_L}$. Moreover, if we set $n=L+1$ and $\theta = (g, \chi, (h_\ell, a_\ell, b_\ell)_{\ell = 1}^L, v)$ is s.t.
   \begin{equation}
   \label{eqn:proofappendixcorollarylowerboundconstraintsontheta}
       \{a_1, \ldots, a_L, v\} \subset \r^n \text{ is orthonormal}, \quad b_\ell = \frac{1}{\sqrt{L}}v, \quad h_\ell = 0 \quad (\forall \ell \in \{1, \ldots, L\}),
   \end{equation}
  then the same statements hold with $s_\tau^{\operatorname{Id}, F_L}$ replaced by either of $s_\tau^\theta$, $b_\tau^\theta$ and $\rho_\tau^{\operatorname{Id}, F_L}$ replaced by $\rho_\tau^\theta$.
\end{corollary}
\begin{proof}
Since $\Psi_\theta = F_L$ and $\psi_\theta = \operatorname{Id}$ under (\ref{eqn:proofappendixcorollarylowerboundconstraintsontheta}), it suffices to prove the claim for $F_L$.
    Denote $Q := Q_{\|\chi\|_\infty, 1}$ for $Q_{\beta, \sigma}$ from Proposition \ref{prop:appendixdaapproximationlowerbound} along with $\Lambda := \operatorname{Lip}(\chi) \sqrt{1 + \operatorname{Lip}(g)^2}$. Then, for every $L$ by the chain rule we have 
    \begin{equation}
        \|F_L\|_\infty \leq \|\chi\|_\infty, \quad \operatorname{Lip}(F_L) \leq \Lambda. 
    \end{equation}
    Choose $\epsilon_0$ sufficiently small that $K(\epsilon) \geq \max\{K_g, \lceil K_0\rceil\}$ for all $\epsilon \in (0, \epsilon_0]$, which is possible since $V$ is strictly positive. For any $\epsilon \in (0, \epsilon_0]$ by definition of $K(\epsilon)$ we have
    \begin{equation}
        \mathcal{V}_{\chi, g}(K(\epsilon), \lfloor cK(\epsilon)\rfloor) \geq V(K(\epsilon)) \geq C_{\chi, g} \epsilon^2 = 64 Q^2 (1+\Lambda^2)^2 \epsilon^2.
    \end{equation}
    Applying Lemma \ref{lemma:appendixhermitebranchinglemma} with $K_1 = K(\epsilon)$ and $K_2 = \lfloor cK(\epsilon)\rfloor$, we obtain some $L_0 = L_0(g, \chi, \epsilon)$ s.t. for all $L \geq L_0$ and all integers $0 \leq r \leq L^{c_g K(\epsilon)}$ we have
    \begin{equation}
        a_{F_L}(r, \tau) \geq \frac{1}{2}e^{-b_g cK(\epsilon) \tau}\sqrt{\mathcal{V}_{\chi, g}(K(\epsilon), \lfloor cK(\epsilon)\rfloor)} \geq 4Q(1+\Lambda^2) e^{-b_g cK(\epsilon)\tau}\epsilon.
    \end{equation}
    Choosing $c_1 \leq \min\{1, \log(2)/(b_g c)\}$, if $\tau \in [0, c_1 \min\{\epsilon, 1/K(\epsilon)\}]$ then $e^{-b_gcK(\epsilon)\tau} \geq 1/2$ and $\tau \leq \epsilon$. This means that
    \begin{equation}
        a_{F_L}(r, \tau) \geq 2Q(1+\Lambda^2) \epsilon.
    \end{equation}
    Since $Q(2\epsilon + \Lambda^2 \tau) \leq 2Q(1+\Lambda^2)\epsilon$, it follows from the definition of the Hermite width envelope (see Definition \ref{def:appendixhermitewidthenvelope}) that
    \begin{equation}
        m_{F_L}\left(Q(\epsilon + \Lambda^2 \tau), \; \tau\right), m_{F_L}\left(2Q(\epsilon + \Lambda^2 \tau), \; \tau\right) \geq  \lfloor L^{c_g K(\epsilon)}\rfloor.
    \end{equation}
    Applying Proposition \ref{prop:appendixdaapproximationlowerbound} with $\psi = \operatorname{Id}$, $\Psi = F_L$, and $n = L+1$, we find that the width lower bound holds with $c_2 = c_g$ and that for any $\phi$ satisfying Assumption \ref{assumption:activationfunciton} we have
    \begin{equation}
        \|\hat{s}_\tau\|_{\b_\phi} \geq K_{\phi, \|\chi\|_\infty, 1} \epsilon \sqrt{ \lfloor L^{c_g K(\epsilon)}\rfloor} 
    \end{equation}
    with the same holding for $b_\tau^{\psi, \Psi}$.
    Increasing $L_0$ if needed, we may take $\lfloor L^{c_g K(\epsilon)}\rfloor \geq L^{c_g K(\epsilon)}/2$, in which case the Barron norm claim holds with $D_{\phi, \chi} =  K_{\phi, \|\chi\|_\infty, 1} / \sqrt{2}$.
\end{proof}

We finish these approximation lower bounds by getting concrete scalings of $V(K)$ for our activations of interest. Importantly, $V$ depends only on $\chi$ and on the Gaussian variance of $g$:

\begin{lemma}[Hermite decay for ReLU difference and tanh]
\label{lemma:appendixhermitedecayforapproximation}
Let $g: \r \to \r$ be centered with $\lambda_g := 1 + \|g\|_{L^2(\gamma_1)}^2 < \infty$. Then, there is some $K_0 \in (0, \infty)$ depending only on $\lambda_g$ s.t. $\mathcal{V}_{\chi, g}(K, 2K) \geq V(K)$ for all $K \geq K_0$, where we define $V: \n \to (0, \infty)$ as follows:
\begin{enumerate}
    \item[(i)] If $\chi(t) = \frac{\max\{0, t\} - \max\{0, t-a\}}{a}$ for some fixed $a \in (0, \infty)$ then we take $V(K) = bK^{-3/2}$ for some $b \in (0, \infty)$ depending on $\lambda_g, a$.
    \item[(ii)] If $\chi(t) = \tanh(t)$ then we take $V(K) = be^{-\pi \sqrt{K / \lambda_g}}$ for some $b \in (0, \infty)$ depending on $\lambda_g$.
\end{enumerate}
\end{lemma}
\begin{remark}
    In particular, for (i) we get $K(\epsilon) \asymp \epsilon^{-4/3}$ while for (ii) we have $K(\epsilon) \asymp \log(1/\epsilon)^2$.
\end{remark}
\begin{proof}
Fix $\lambda > 0$ (which we take as $\lambda = \lambda_g$), and let us denote $\alpha_j := \mathbb{E}_{G \sim \gamma_1}[\chi(\sqrt{\lambda }G) \operatorname{He}_j(G)] / \sqrt{j!}$. 

For the ReLU difference, we simply apply \cite[Lemma A.3]{fu2023singleattention} to see that 
\begin{equation}
    \frac{1}{\sqrt{j!}}\mathbb{E}_{G \sim \gamma_1} \left[\max\{0, G-w\} \operatorname{He}_j(G)\right] = \frac{e^{-w^2/2} \operatorname{He}_{j-2}(w)}{\sqrt{2\pi j!}} \quad (\forall j \geq 2)
\end{equation}
for any $w \in \r$. By homogeneity and linearity, for every $j \geq 2$ 
this gives
\begin{equation}
\label{eqn:proofappendixReLUdifferencehermitescaling}
    \alpha_j = \frac{\sqrt{\lambda} }{a\sqrt{2\pi j!}} \left[\operatorname{He}_{j-2}(0) - e^{-a^2/(2\lambda)} \operatorname{He}_{j-2}\left(\frac{a}{\sqrt{\lambda}}\right)\right].
\end{equation}
If we denote $u(t) := e^{-t^2/4} \frac{\operatorname{He}_n(t)}{\operatorname{He}_n(0)}$ for an even $n \in \n \cup \{0\}$, then the Hermite equation $\operatorname{He}_n'' - t \operatorname{He}_n' + n \operatorname{He}_n = 0$ implies that
\begin{equation}
    u''(t) + \left(n + \frac{1}{2}-\frac{t^2}{4}\right)u(t) = 0, \quad u(0) = 1, \quad u'(0) = 0.
\end{equation}
Consequently, for all $t \geq 0$ we have
\begin{align}
    \frac{d}{dt} \left[u'(t)^2 + \left(n + \frac{1}{2} - \frac{t^2}{4} \right)u(t)^2\right] &= 2 u'(t) \left(u''(t) + \left(n + \frac{1}{2} - \frac{t^2}{4}\right) u(t)\right)- \frac{t}{2}u(t)^2 \\&= -\frac{t}{2}u(t)^2 \leq 0.
\end{align}
Hence,
\begin{equation}
   \left(n + \frac{1}{2} - \frac{t^2}{4}\right)u(t)^2  \leq  u'(t)^2 + \left(n + \frac{1}{2} - \frac{t^2}{4}\right)u(t)^2 \leq n + \frac{1}{2}
\end{equation}
and so whenever $n + 1/2> t^2/4$ we have 
\begin{equation}
    \left|\frac{e^{-t^2/2} \operatorname{He}_{n}(t)}{\operatorname{He}_{n}(0)}\right| \leq e^{-t^2/4} \sqrt{\frac{n+\frac{1}{2}}{n+\frac{1}{2}-\frac{t^2}{4}}}.
\end{equation}
Therefore, as $n = j-2\to \infty$ we find that
\begin{equation}
    \limsup_{\substack{j \to \infty\\ j \text{ even}}} \left|\frac{e^{-a^2/(2\lambda)} \operatorname{He}_{j-2}(a/\sqrt{\lambda})}{\operatorname{He}_{j-2}(0)}\right| \leq e^{-a^2/(4\lambda)} < 1.
\end{equation}
In particular, as $j \to \infty$ with $j = 2r + 2$ even we have the asymptotic
\begin{equation}
    \alpha_j^2 \asymp_{\lambda, a} \frac{\operatorname{He}_{j-2}(0)^2}{j!} = \frac{{2r \choose r}}{4^r (2r + 1) (2r+2)} \asymp \frac{1}{\sqrt{\pi r} (2r+1)(2r+2)} \asymp j^{-5/2},
\end{equation}
where we used that ${2r \choose r} / 4^r \asymp 1/\sqrt{\pi r}$ as $r \to \infty$. Hence, $\alpha_j^2 \asymp j^{-5/2}$ for $j$ large and even. Summing over a band gives
\begin{equation}
    \mathcal{V}_{\chi, g}(K, 2K) \gtrsim_{\lambda, a} \sum_{\substack{K \leq j \leq 2K \\ j \text{ even}}} j^{-5/2} \asymp K^{-3/2}.
\end{equation}
Note that by (\ref{eqn:proofappendixReLUdifferencehermitescaling}) we know that $|\alpha_j| > 0$ for all $j \geq 2$ even, and so by decreasing $b$ our result holds even up to $K_0 =2$.

    For $\tanh$, we replicate the argument of Lemma \ref{lemma:appendixtanhhermitescaling}. Note that $\alpha_j = 0$ for even $j$. By the same Fourier transform formula as in Lemma \ref{lemma:appendixtanhhermitescaling} we have
    \begin{equation}
        \widehat{(\tanh(\sqrt{\lambda} \cdot))'}(\xi) = \frac{\pi \xi / \sqrt{\lambda}}{\sinh(\pi \xi / (2 \sqrt{\lambda}))},
    \end{equation}
    and so the same integration by parts and Parseval argument implies that for odd $j$
    \begin{equation}
        \alpha_j = \frac{(-1)^{(j-1)/2}}{\sqrt{\lambda j!}} \int_0^\infty \frac{\xi^j e^{-\xi^2/2}}{\sinh(\pi \xi / (2\sqrt{\lambda}))}d\xi.
    \end{equation}
    We again set $F(\xi) := j \log \xi - \frac{\xi^2}{2} - \frac{\pi \xi}{2\sqrt{\lambda}}$. Splitting at $\xi = 1$, we find that as odd $j \to \infty$ there is the asymptotic
    \begin{equation}
        \log |\alpha_j| = -\frac{1}{2}\log(\lambda j!) + \log \int_1^\infty e^{F(\xi)}d\xi + O_\lambda(1).
    \end{equation}
    The same Taylor estimates as in Lemma \ref{lemma:appendixtanhhermitescaling} (with $2q$ replaced by $j$) apply because $F'(\sqrt{j}) = -\frac{\pi}{2\sqrt{\lambda}}$ and $F''(\xi) = -\frac{j}{\xi^2} - 1 \leq -1$ for all $\xi$. In particular, \begin{equation}
        \log\left(\int_1^\infty e^{F(\xi)}d\xi\right) = F(\sqrt{j}) + O_\lambda(1)
    \end{equation}
    and consequently
    \begin{equation}
        \log |\alpha_j| = \frac{j}{2}\log j - \frac{j}{2} - \frac{\pi}{2\sqrt{\lambda}}\sqrt{j} - \frac{1}{2}\log(\lambda j!) + O_\lambda(1).
    \end{equation}
    Applying Stirling, we get
    \begin{equation}
        \log |\alpha_j| = -\frac{\pi}{2\sqrt{\lambda}}\sqrt{j} - \frac{1}{4}\log j + O_\lambda(1)
    \end{equation}
    or in other words that
    \begin{equation}
        \alpha_j^2 \asymp_\lambda j^{-1/2} e^{-\pi \sqrt{j/\lambda}} \quad (j\to \infty  \text{ odd}).
    \end{equation}
    We sum over the band to find that
    \begin{equation}
        \mathcal{V}_{\tanh, g}(K, 2K) \asymp_\lambda \sum_{\substack{K \leq j \leq 2K \\ j \text{ odd}}} j^{-1/2} e^{-\pi \sqrt{j/\lambda}} \asymp_\lambda e^{-\pi \sqrt{K/\lambda}},
    \end{equation}
    where the last asymptotic is due to the integral equality
    \begin{equation}
        \int_K^{2K} x^{-1/2}e^{-\pi \sqrt{x/\lambda}}dx = \frac{2\sqrt{\lambda}}{\pi} \left(e^{-\pi \sqrt{K/\lambda}} - e^{-\pi \sqrt{2K/\lambda}}\right).
    \end{equation}
    
\end{proof}

\begin{remark}
    Note that for any $\epsilon, \tau > 0$ and any $\theta$ there is an approximate score network $\hat{s}_\tau$ s.t. $\|s_\tau^\theta - \hat{s}_\tau\|_{L^2(\rho_\tau^\theta; \r^{n+1})} \leq \epsilon$, see Proposition \ref{prop:appendixapproximationupperbound}. The lower bounds in this subsection say that any such approximate score network must have large width and Barron norm.
\end{remark}

We can also apply the lower bounds from Section \ref{subsec:appendixearlystoppinglowerboundstriangular} to verify that early-stopping cannot sidestep our approximation lower bound: one must model the scores in the window of OU times $\tau \in [0, c_1 \min\{\epsilon, 1/K(\epsilon)]\}$ in which Corollary \ref{corollary:approximationlowerbound} applies. 

\begin{lemma}[Early-stopping lower bound]
\label{lemma:appendixearlystoppingforresnets}
    Assume the setting of Corollary \ref{corollary:approximationlowerbound}. Then, for every $a \in (0, \infty)$ there exist constants $\epsilon_0, b_1, b_2 \in (0, \infty)$ s.t. for every $\epsilon \in (0, \epsilon_0]$ there is an $L_0(\epsilon)$ for which the following holds:

    For every $L \geq L_0(\epsilon)$ and every $\tau_0 \geq a \min\{\epsilon, 1/K(\epsilon)\}$, if we denote $\pi = \pi^{\operatorname{Id}, F_L}$ and $\rho_{\tau_0} = \rho_{\tau_0}^{\operatorname{Id}, F_L}$ then we have the lower bounds
    \begin{align}
        \min\left\{\operatorname{TV}(\pi, \rho_{\tau_0}), \; W_1(\pi, \rho_{\tau_0})\right\} &\geq b_1  \min\{\epsilon, 1/K(\epsilon)\},\\
        \min\left\{\sqrt{\operatorname{KL}(\pi \|\rho_{\tau_0})}, \; \sqrt{\operatorname{KL}( \rho_{\tau_0} \| \pi)}\right\} &\geq b_2 \epsilon.
    \end{align}
    The constants depend only on $a, g, \chi, c, K_0, V$.
\end{lemma}
\begin{remark}
    In light of Lemma \ref{lemma:appendixhermitedecayforapproximation}, if $\chi$ is the difference of two ReLUs then $\operatorname{TV}, W_1 \gtrsim \epsilon^{4/3}$ while if $\chi = \tanh$ then $\operatorname{TV}, W_1 \gtrsim \epsilon$. In both cases, $\sqrt{\operatorname{KL}} \gtrsim \epsilon$.
\end{remark}
\begin{proof}
    Choose $\epsilon_0 \leq 1$ sufficiently small that $K(\epsilon) \geq K_g$. Applying Lemma \ref{lemma:appendixhermitebranchinglemma} at time 0, both to the band $[K(\epsilon), \lfloor cK(\epsilon)\rfloor]$ and to one fixed admissible band with positive mass, gives
    \begin{equation}
        a_{F_L}(\lfloor L^{c_gK(\epsilon)}\rfloor, 0) \gtrsim \epsilon, \quad a_{F_L}(0, 0) = \|F_L\|_{L^2(\gamma_{L+1})} \gtrsim 1
    \end{equation}
    for $L \geq L_0(\epsilon)$. Here and below, the implied constants are independent of $\epsilon, L, \tau_0$. Since $\|F_L\|_\infty \leq \|\chi\|_\infty$, we have
    \begin{equation}
        M_{F_L} \geq e^{-\|\chi\|^2_\infty / 4}\|F_L\|_{L^2(\gamma_{L+1})}^2 \gtrsim 1,
    \end{equation}
    and for $L \geq 2$ we know
    \begin{equation}
        \ell_{L+1}(\lfloor L^{c_gK(\epsilon)}\rfloor) \geq 2c_g K(\epsilon) \frac{\log L}{\log(L+1)} \geq c_g K(\epsilon).
    \end{equation}
    Lemma \ref{lemma:appendixtriangularearlystoppingerrorTVW1} therefore implies that
    \begin{align}
        \min\{\operatorname{TV}(\pi, \rho_{\tau_0}), W_1(\pi, \rho_{\tau_0})\} &\gtrsim 1-e^{-2\tau_0} \\&\geq 1-e^{-2 a \min\{\epsilon, 1/K(\epsilon)\}} \\&\geq (1-e^{-2a}) \min\{\epsilon, 1/K(\epsilon)\}.
    \end{align}
    Similarly, Lemma \ref{lemma:appendixtriangularearlystoppingerrorKL} implies that with $D_{\tau_0}$ as the KL in either direction,
    \begin{align}
        \sqrt{D_{\tau_0}} &\gtrsim \max\left\{1-e^{-\tau_0},\; \epsilon\left(1-e^{-c_g K(\epsilon) \tau_0}\right)\right\}.
    \end{align}
    If $\epsilon \leq 1/K(\epsilon)$ then $1-e^{-\tau_0} \geq 1-e^{-a\epsilon} \geq (1-e^{-a})\epsilon$, while if $\epsilon > 1/K(\epsilon)$ then we have that $1-e^{-c_g K(\epsilon) \tau_0} \geq 1-e^{-c_ga}$. In either case, $\sqrt{D_{\tau_0}} \gtrsim \epsilon$.
\end{proof}

We now turn to the proof of Lemma \ref{lemma:appendixhermitebranchinglemma}. In a few words, we perform a probabilistic counting argument using the number of crossings of the Hermite coefficient tensors of $F_L$. This allows us to lower bound the Hermite rank in a way which translates to our width lower bounds.

\begin{proof}[Proof of Lemma \ref{lemma:appendixhermitebranchinglemma}]
Let us denote $V_g := \|g\|_{L^2(\gamma_1)}^2$ and $\beta_j := \mathbb{E}_{G \sim \gamma_1}[g(G) h_j(G)]$ for $j \in \n \cup \{0\}$, where $h_j := \operatorname{He}_j / \sqrt{j!}$ are the $L^2(\gamma_1)$-orthonormal Hermite polynomials. Because $g$ is centered we know $\beta_0 = 0$ and therefore $V_g = \sum_{j \in \n} \beta_j^2$. Moreover, since $\beta_1 = \mathbb{E}_{G \sim \gamma_1}[Gg(G)]$ then
\begin{equation}
    0 < \mathfrak{v}_g := \frac{\sum_{j \geq 2} \beta_j^2}{8\lambda_g} \leq \frac{1}{8}, \quad \lambda_g := 1+V_g,
\end{equation}
since $g$ is non-affine in $L^2(\gamma_1)$. The heart of this proof is to show that for all $ \lceil 4/\mathfrak{v}_g\rceil \leq K_1 \leq K_2$, we have the estimate 
\begin{equation}
\label{eqn:proofappendixthehermitebranchinglemmamainestimate}
    \liminf_{L \to \infty} \sum_{k=K_1}^{\lceil B_g K_2\rceil} \;\sum_{j > L^{\lfloor \mathfrak{v}_g K_1\rfloor-1}} \sigma_j (M_k(F_L))^2 \geq \frac{13}{48} \mathcal{V}_{\chi, g}(K_1, K_2)
\end{equation}
where $M_k(F_L)$ are the balanced matricizations of the Hermite tensor coefficients of $F_L$ (see Definition \ref{def:appendixhermitewidthenvelope}) and we denote
\begin{equation}
    B_g := \frac{8 (1 + \operatorname{Lip}(g))^2}{\lambda_g}.
\end{equation}
Indeed, from (\ref{eqn:proofappendixthehermitebranchinglemmamainestimate}) we find an $L_0=L_0(g, \chi, K_1, K_2)$ s.t. 
\begin{equation}
    \sum_{k=K_1}^{\lceil B_g K_2\rceil} \;\sum_{j > L^{\lfloor \mathfrak{v}_g K_1\rfloor-1}} \sigma_j (M_k(F_L))^2 \geq \frac{1}{4} \mathcal{V}_{\chi, g}(K_1, K_2) \quad (\forall L \geq L_0).
\end{equation}
Moreover, $K_1 \geq \lceil 4/\mathfrak{v}_g\rceil$ implies $\lfloor \mathfrak{v}_g K_1\rfloor - 1 \geq  \mathfrak{v}_g  K_1 - 2 \geq c_g K_1$ with $c_g :=  \mathfrak{v}_g /2$, and so for every $m \leq L^{c_g K_1}$ and $\tau \geq 0$ we find
\begin{equation}
    a_{F_L}(m, \tau)^2 \geq \frac{1}{4}e^{-2(\lceil B_g K_2\rceil + 1)\tau} \mathcal{V}_{\chi, g}(K_1, K_2).
\end{equation}
Taking $b_g := B_g + 1$ yields the claim.
The proof of (\ref{eqn:proofappendixthehermitebranchinglemmamainestimate}) proceeds in the following high-level steps:
\begin{enumerate}
    \item For each $k \in \n$, we fix a balanced cut $S_k \subset \{1, \ldots, k\}$ with $|S_k| = \lfloor k / 2\rfloor$ to define the matricization $M_k(F_L)$. 
    We show that our Hermite spectral tail of interest is controlled by the elements of the Hermite tensor coefficients whose indices cross the cut. More precisely, for any integer $R > 1$ we show that
     \begin{equation}
        \liminf_{L \to \infty} \sum_{j > L^{R-1}} \sigma_j (M_k(F_L))^2 \geq \liminf_{L \to \infty}\sum_{\substack{\mathbf{p} \in \mathcal{I}_k \\ C_{S_k}(\mathbf{p}) \geq R}} |d_{\mathbf{p}, L}|^2 \quad (\forall k \in \n),
    \end{equation}
    where $\mathcal{I}_k \subset \{0, \ldots, k\}^k$ is a certain collection of $k$-tuples, $C_{S_k}(\mathbf{p})$ denotes the number of distinct positive entries of $\mathbf{p}$ appearing on both sides of the cut $S_k$, and $d_{\mathbf{p}, L}$ is proportional to the Hermite coefficient tensor entry $T_k(F_L)[\mathbf{p}]$. 

    \item By tensor symmetry, the fraction of entries satisfying the crossing condition can be computed using a uniformly random balanced cut. Using this, we show that
    \begin{equation}
        \lim_{L \to \infty}\sum_{\substack{\mathbf{p} \in \mathcal{I}_k \\ C_{S_k}(\mathbf{p}) \geq R}} |d_{\mathbf{p}, L}|^2 = \sum_{J =1}^k \alpha_J^2 \cdot \mathbb{P}_J(N=k, \; C_S \geq R) \quad (\forall k \in \n)
    \end{equation}
    with the notation
    \begin{equation}
         \alpha_J := \mathbb{E}_{G \sim \gamma_1}\left[\chi(\sqrt{\lambda_g}G)h_J(G)\right].
    \end{equation}
    Here, $\mathbb{P}_J$ is an auxiliary probability law describing the Hermite degrees and uniformly random balanced cut, and under $\mathbb{P}_J$ we denote by $N$ the total degree and by $C_S$ the crossing count for a uniformly random balanced cut $S$. Note that $\mathcal{V}_{\chi, g}(K_1, K_2)  = \sum_{J=K_1}^{K_2}\alpha_J^2$.
    
    \item  By simple probabilistic estimates and plugging in our quantities, we show that \begin{equation}\mathbb{P}_J(N \leq  B_g K_2 , \;C_S\geq  \mathfrak{v}_g K_1)\geq\frac{13}{48} \quad (\forall K_1 \leq J \leq K_2).\end{equation}
    Combining this with the prior steps proves the main estimate (\ref{eqn:proofappendixthehermitebranchinglemmamainestimate}). 
\end{enumerate}
\textbf{(Step 1).} Fix $k \in \n$ and let $L > k$. Choose a subset $S_k \subset \{1, \ldots, k\}$ of size $|S_k| = \lfloor k/2\rfloor$; $S_k$ defines the row indices of the balanced matricization and $S_k^c$ defines the column indices. 
    For each tuple $\mathbf{i} = (i_1, \ldots, i_k) \in \{0, \ldots, L\}^k$, we denote $\mathbf{i}_{S} = (i_r)_{r \in S}$ and observe that by symmetry of $T_k(F_L)$ we have
    \begin{equation}
         M_k(F_L)[\mathbf{i}_{S_k}, \mathbf{i}_{S_k^c}] = T_k(F_L)[\mathbf{i}] = \sqrt{\frac{\prod_{\ell = 0}^L \nu_\ell(\mathbf{i})!}{k!}} \mathbb{E}_{Z \sim \gamma_{L+1}}\left[F_L(Z) \prod_{\ell = 0}^L h_{\nu_\ell(\mathbf{i})}(Z_\ell)\right],
    \end{equation}
    where we denote $\nu_\ell(\mathbf{i}) = \#\{r:\; i_r = \ell\}$.
    We define the crossing quantity
    \begin{equation}
        C_S(\mathbf{i}) := \left|\{i_r:\; r \in S\}\cap \{i_r:\; r \in S^c\} \cap \{1,\ldots, L\}\right|
    \end{equation}
    to be the number of distinct coordinate labels (other than 0) which appear in both the row and column indices for a balanced cut $S$. Now, let $\mathcal{I}_k\subset\{0,\ldots,k\}^k$ be the set of tuples whose distinct positive labels, listed in order of first appearance, are $\{1, 2, \ldots, s\}$ for some $s \leq k$. Equivalently, we require that the distinct nonzero values of $\mathbf{p} \in \mathcal{I}_k$ are $\{1, 2, \ldots, s\}$ and 
    \begin{equation}
        \min\{r:\; p_r = 1\} < \min\{r:\; p_r = 2\} < \ldots < \min\{r:\; p_r = s\}.
    \end{equation}
    In particular, a positive entry may repeat a previously introduced label or introduce the next unused label, but it cannot skip a label.\footnote{By this, we mean tuples like $(0, 1, 1, 2, 0, 2, 3)$ or $(0, 1, 2, 1, \ldots)$ but not $(0, 2, \ldots)$.} Note that $s$ depends on $\mathbf{p}$. For every $\mathbf{p} \in \mathcal{I}_k$ we define two matrices of the same dimensions as $M_k(F_L)$ by
    \begin{equation}
        U_{\mathbf{p}, L}[\mathbf{i}_{S_k}, \mathbf{i}_{S_k^c}] := L^{-s/2} \sum_{\ell_1, \ldots, \ell_s = 1}^L \;\prod_{r = 1}^k \1_{i_r = \ell_{p_r}}, \quad U_{\mathbf{p}, L}^{\neq}[\mathbf{i}_{S_k}, \mathbf{i}_{S_k^c}] := L^{-s/2} \sum_{\substack{\ell_1, \ldots, \ell_s = 1\\ \ell_1, \ldots, \ell_s \text{ distinct}}}^L \;\prod_{r = 1}^k \1_{i_r = \ell_{p_r}},
    \end{equation}
    where we denote $\ell_0 := 0$. Thus, $U_{\mathbf{p},L}^{\neq}$ enforces exactly the coordinate equalities specified by $\mathbf{p}$ whereas $U_{\mathbf{p}, L}$ also permits different positive labels of $\mathbf{p}$ to receive the same coordinate. 
    We set
    \begin{equation}
        d_{\mathbf{p}, L} := L^{s/2} T_k(F_L)[\mathbf{p}].
    \end{equation}
    Because $F_L$ is invariant under permutations of the coordinates $1, \ldots, L$, tuples whose nonzero entries are equivalent up to relabeling have the same tensor coefficients. Hence,
    \begin{equation}
        M_k(F_L) = \sum_{\mathbf{p} \in \mathcal{I}_k} d_{\mathbf{p}, L} U_{\mathbf{p}, L}^{\neq}.
    \end{equation}
    Fix an integer $R > 1$. We decompose $M_k(F_L) = A_{k,L} + H_{k,L} + E_{k,L}$ using the matrices
    \begin{equation}
        A_{k,L} := \sum_{\substack{\mathbf{p} \in \mathcal{I}_k \\ C_{S_k}(\mathbf{p}) < R}}d_{\mathbf{p}, L} U_{\mathbf{p}, L}, \quad H_{k,L} := \sum_{\substack{\mathbf{p} \in \mathcal{I}_k \\ C_{S_k}(\mathbf{p}) \geq R}}d_{\mathbf{p}, L} U_{\mathbf{p}, L}, \quad E_{k,L} := \sum_{\substack{\mathbf{p} \in \mathcal{I}_k}}d_{\mathbf{p}, L}( U_{\mathbf{p}, L}^{\neq} -  U_{\mathbf{p}, L}).
    \end{equation}
    We will now prove some bounds on these matrices. 
    If we denote $a_t := \#\{r\in S_k:\; p_r = t\}$ and $b_t := \#\{r \in S_k^c:\; p_r = t\}$, then we find
    \begin{equation}
        U_{\mathbf{p},L} = \left(e_0^{\otimes a_0} (e_0^{\otimes b_0})^\top\right) \otimes \bigotimes_{t=1}^s D_t, \quad D_t := \frac{1}{\sqrt{L}} \sum_{\ell = 1}^L e_\ell^{\otimes a_t} (e_\ell^{\otimes b_t})^\top
    \end{equation}
    up to row and column permutations,
    with $e_\ell^{\otimes 0} := 1$. These permutations preserve rank and singular values. Now, if $a_t, b_t \geq 1$ then both families $\{e_\ell^{\otimes a_t}\}_{\ell = 1}^L$ and $\{e_\ell^{\otimes b_t}\}_{\ell = 1}^L$ are orthonormal, and so the displayed expression for $D_t$ is an SVD with all singular values equal to $1/\sqrt{L}$. If either $a_t = 0$ or $b_t = 0$, then $D_t$ is a row or column vector of norm 1 (since its squared norm is $\frac{1}{L}\sum_{\ell = 1}^L 1 = 1$). It therefore has one nonzero singular value equal to 1, as does the term $e_0^{\otimes a_0} (e_0^{\otimes b_0})^\top$.
    We also have
    \begin{equation}
        C_{S_k}(\mathbf{p}) = \#\{t \in \{1, \ldots, s\}:\; a_t,b_t > 0\}.
    \end{equation}
    Since the nonzero singular values of a Kronecker product are all the possible products of those of the factors, we find that $U_{\mathbf{p},L}$ has exactly $L^{C_{S_k}(\mathbf{p})}$ nonzero singular values which are all equal to $L^{-C_{S_k}(\mathbf{p})/2}$. Hence,
    \begin{equation}
        \operatorname{rank}( U_{\mathbf{p}, L}) = L^{C_{S_k}(\mathbf{p})}, \quad \|U_{\mathbf{p}, L}\|_{\operatorname{op}} = L^{-C_{S_k}(\mathbf{p})/2}.
    \end{equation}
    Next, we claim that 
    \begin{equation}
    \label{eqn:proofappendixapproximationHermitebranchinglemmaboundondpL}
        |d_{\mathbf{p}, L}| \leq \|\chi\|_\infty V_g^{s/2}  \leq \|\chi\|_\infty \max\{1, V_g^{k/2}\} =: B_k.
    \end{equation}
    To show this, denote $j := \nu_0(\mathbf{p})$ and $q_t := \nu_t(\mathbf{p}) \geq 1$. We start from the fact that if $Z = (Z_0, \ldots, Z_L) \sim \gamma_{L+1}$ then
    \begin{equation}
        d_{\mathbf{p}, L} = L^{s/2} \sqrt{\frac{j! \prod_{t=1}^s q_t!}{k!}} \mathbb{E}\left[F_L(Z) h_j(Z_0)\prod_{t = 1}^sh_{q_t}(Z_t) \right].
    \end{equation}
    If we define $Q_j(y) := \mathbb{E}_{G \sim \gamma_1}[\chi(G+y) h_j(G)]$ and differentiate the convolution, we find
    \begin{equation}
        Q_j^{(s)}(y) = \sqrt{\frac{(j+s)!}{j!}} \mathbb{E}_{G \sim \gamma_1}[\chi(G+y)h_{j+s}(G)]
    \end{equation}
    and therefore $\|Q_j^{(s)}\|_\infty \leq \|\chi\|_\infty \sqrt{\frac{(j+s)!}{j!}} $. If we define $(\Delta_uQ)(y) := Q(y+u) - Q(y)$, then repeated application of the fundamental theorem of calculus yields
    \begin{equation}\label{eqn:proofappendixapproximationHermitebranchinglemmaboundondpL10000}
        (\Delta_{u_1} \cdots \Delta_{u_s}Q)(y) = \left(\prod_{t=1}^s u_t\right) \int_{[0,1]^s} Q^{(s)}\left(y + \sum_{t=1}^su_tv_t\right)dv,
    \end{equation}
    which we apply to $Q_j$ to find that
    \begin{equation}
        \|\Delta_{u_1} \cdots \Delta_{u_s}Q_j\|_\infty \leq \|Q_j^{(s)}\|_\infty \prod_{t=1}^s |u_t| \leq \|\chi\|_\infty \sqrt{\frac{(j+s)!}{j!}} \prod_{t=1}^s |u_t|.
    \end{equation}
    On the other hand, denoting $Y_L := \frac{1}{\sqrt{L}}\sum_{\ell = s+1}^L g(Z_\ell)$ and integrating over $Z_0$ yields
    \begin{align}
        d_{\mathbf{p}, L} &= L^{s/2} \sqrt{\frac{j! \prod_{t=1}^s q_t!}{k!}} \mathbb{E}\left[\chi\left(Z_0 + Y_L + \sum_{t=1}^s \frac{g(Z_t)}{\sqrt{L}} \right)h_j(Z_0) \prod_{t=1}^s h_{q_t}(Z_t)\right] \\
        &= L^{s/2} \sqrt{\frac{j! \prod_{t=1}^s q_t!}{k!}} \mathbb{E}\left[Q_j\left(Y_L + \sum_{t=1}^s \frac{g(Z_t)}{\sqrt{L}}\right)\prod_{t=1}^s h_{q_t}(Z_t)\right].
        \label{eqn:proofappendixapproximationlowerbounddpLnonsense2}
    \end{align}
    Now, use the finite differences identity $(\Delta_{u_1} \cdots \Delta_{u_s}Q)(y) = \sum_{A \subset \{1, \ldots, s\}} (-1)^{s-|A|} Q(y + \sum_{t \in A}u_t)$ along with the fact
    \begin{align}
        \mathbb{E}\left[Q_j\left(Y_L + \sum_{t \in A} \frac{g(Z_t)}{\sqrt{L}}\right)\prod_{t=1}^s h_{q_t}(Z_t)\right] = 0 \text{ if $A \subsetneq \{1, \ldots, s\}$},
    \end{align}
    which is true since for any $t \notin A$ we may integrate over $Z_t$ and use $\mathbb{E}[h_{q_t}(Z_t)] = 0$. So, only the $A = \{1, \ldots, s\}$ term of the finite differences identity survives, which we plug into (\ref{eqn:proofappendixapproximationlowerbounddpLnonsense2}) to find that
    \begin{align}
    \label{eqn:proofappendixapproximationlowerbounddpLnonsense3}
        d_{\mathbf{p}, L} &= L^{s/2} \sqrt{\frac{j! \prod_{t=1}^s q_t!}{k!}} \mathbb{E}\left[\left(\Delta_{g(Z_1)/\sqrt{L}}\cdots \Delta_{g(Z_s)/\sqrt{L}}Q_j\right)(Y_L) \prod_{t=1}^s h_{q_t}(Z_t)\right]
        \end{align}
        and therefore
        \begin{align}
        | d_{\mathbf{p}, L}|
       & \leq \|\chi\|_\infty  \sqrt{\frac{j! \prod_{t=1}^s q_t!}{k!}}  \sqrt{\frac{(j+s)!}{j!}} \prod_{t=1}^s \mathbb{E}\left[|g(Z_t)h_{q_t}(Z_t)|\right]  \\
        &\leq \|\chi\|_\infty  \sqrt{\frac{(j+s)! \prod_{t=1}^s q_t!}{k!}}  \|g\|_{L^2(\gamma_1)}^s \\
        &\leq \|\chi\|_\infty V_g^{s/2},
    \end{align}
    where we use Cauchy-Schwarz and that $\|h_{q_t}\|_{L^2(\gamma_1)} = 1$ in the second line and that $\sqrt{\frac{(j+s)! \prod_{t=1}^s q_t!}{k!}} \leq 1$ in the third. This proves (\ref{eqn:proofappendixapproximationHermitebranchinglemmaboundondpL}).

    Now, since $|\mathcal{I}_k| \leq (k+1)^k$ we find that
    \begin{equation}
        \operatorname{rank}(A_{k,L}) \leq (k+1)^k L^{R - 1}, \quad \|H_{k,L}\|_{\operatorname{op}} \leq (k+1)^kB_k L^{-R/2}.
    \end{equation}
    Also, since $U_{\mathbf{p},L} - U_{\mathbf{p},L}^{\neq}$'s nonzero entries come from exactly the assignments with a repeated coordinate label we find
    \begin{equation}
    \label{eqn:proofappendixappromxiationResnetfrobenisunormboundneq}
        \|U_{\mathbf{p},L} - U_{\mathbf{p},L}^{\neq}\|_F^2 = L^{-s} \left(L^s - L(L-1)\cdots (L-s+1)\right) \leq \frac{{s \choose 2}}{L} \leq \frac{{k \choose 2}}{L}.
    \end{equation}
   In particular, $\|E_{k,L}\|_F^2 \leq (k+1)^{2k}B_k^2 {k \choose 2} /L = O_k({L^{-1}})$. If we define $d_m(B) := \sqrt{\sum_{j > m} \sigma_j(B)^2}$ and observe that this is a distance to the set of matrices of rank at most $m$ (w.r.t. Frobenius norm), then by the triangle inequality we have that
   \begin{equation}
      d_m(B)^2 \leq (1+\eta)d_m(B+E)^2 + (1+\eta^{-1})\|E\|_F^2.
   \end{equation}
   Therefore, taking $B = A_{k,L} + H_{k,L}$ and $E = E_{k,L}$ gives
\begin{equation}
    \sum_{j > L^{R - 1}} \sigma_j(M_k(F_L))^2 \geq \frac{1}{1+\eta}\sum_{j > L^{R - 1}} \sigma_j(A_{k,L} + H_{k,L})^2 - O_{k,\eta}(L^{-1}).
\end{equation}
    Using the elementary inequality
    \begin{equation}
        \sum_{j > m} \sigma_j(A + H)^2 \geq \|H\|_F^2 - (m+\operatorname{rank}(A)) \|H\|_{\operatorname{op}}^2 \quad \left(\forall A, H \in \r^{M \times M'}, \; \forall M,M' \in \n\right),
    \end{equation}
 if we take $m = L^{R - 1}$ along with $A = A_{k,L}$ and $H = H_{k,L}$ then
    \begin{align}
        \sum_{j > L^{R - 1}} \sigma_j(M_k(F_L))^2 \geq \frac{1}{1+\eta}\|H_{k,L}\|_F^2 - O_{k,\eta}(L^{-1}).
    \end{align}
   Taking $L \to \infty$, we find that
    \begin{equation}
    \label{eqn:proofappendixappromxiationResnetfrobenisunormstep1}
         \liminf_{L \to \infty} \sum_{j > L^{R-1}} \sigma_j (M_k(F_L))^2  \geq \liminf_{L \to \infty} \frac{1}{1+\eta}\|H_{k,L}\|_F^2.
    \end{equation}
     By (\ref{eqn:proofappendixappromxiationResnetfrobenisunormboundneq}), replacing $U_{\mathbf{p},L}$ by $U_{\mathbf{p},L}^{\neq}$ in $H_{k,L}$ changes the matrix by at most $O_k(L^{-1/2})$ in Frobenius norm. Since the matrices $U_{\mathbf{p},L}^{\neq}$ have disjoint entry supports for distinct $\mathbf{p}$, we find
    \begin{equation}
    \label{eqn:proofappendixapproximationlowerbounddpLnonsense500}
        \|H_{k,L}\|_F^2 = \sum_{\substack{\mathbf{p} \in \mathcal{I}_k \\ C_{S_k}(\mathbf{p}) \geq R}} |d_{\mathbf{p}, L}|^2 \frac{L(L-1)\cdots (L-s+1)}{L^s} + o(1)
    \end{equation}
    with $s = \max_r p_r$. Combining this with (\ref{eqn:proofappendixappromxiationResnetfrobenisunormstep1}), taking $\eta \downarrow 0$, and noting that $L(L-1) \cdots (L-s+1) / L^s \to 1$ yields
    \begin{equation}
    \label{eqn:proofappendixappromxiationResnetfrobenisunormstep1NEW}
        \liminf_{L \to \infty} \sum_{j > L^{R-1}} \sigma_j (M_k(F_L))^2 \geq \liminf_{L \to \infty}\sum_{\substack{\mathbf{p} \in \mathcal{I}_k \\ C_{S_k}(\mathbf{p}) \geq R}} |d_{\mathbf{p}, L}|^2 .
    \end{equation}
    \textbf{(Step 2).} 
We first fix $\mathbf{p}$ and compute the limit of $d_{\mathbf{p}, L}$. Recalling our notation $j = \nu_0(\mathbf{p})$, $q_t = \nu_t(\mathbf{p})$, and $Y_L = \frac{1}{\sqrt{L}}\sum_{\ell = s+1}^L g(Z_\ell)$ for $Z = (Z_0, \ldots, Z_L) \sim \gamma_{L+1}$, we start from (\ref{eqn:proofappendixapproximationlowerbounddpLnonsense3}) and the finite difference integral representation (\ref{eqn:proofappendixapproximationHermitebranchinglemmaboundondpL10000}) to get that
    \begin{align}
        d_{\mathbf{p}, L} &= \sqrt{\frac{j! \prod_{t=1}^s q_t!}{k!}} \mathbb{E}\left[\left(\int_{[0, 1]^s} Q^{(s)}_j \left(Y_L + \frac{1}{\sqrt{L}}\sum_{t=1}^s g(Z_t)v_t\right)dv\right)\prod_{t=1}^s g(Z_t)h_{q_t}(Z_t)\right].
    \end{align}
    Noting that $Q_j^{(s+1)}$ is bounded (and so $Q_j^{(s)}$ is Lipschitz), $g$ is Lipschitz, and the $h_{q_t}$ are polynomials of fixed degree, we find that
    \begin{align}
 d_{\mathbf{p}, L} &= \sqrt{\frac{j! \prod_{t=1}^s q_t!}{k!}} \cdot \mathbb{E}\left[\left(\int_{[0, 1]^s} Q^{(s)}_j \left(Y_L\right)dv\right)\prod_{t=1}^s g(Z_t)h_{q_t}(Z_t)\right] + O_k(L^{-1/2}) \\
         &= \sqrt{\frac{j! \prod_{t=1}^s q_t!}{k!}} \cdot \mathbb{E}\left[Q^{(s)}_j \left(Y_L\right)\right] \cdot \prod_{t=1}^s \beta_{q_t} + O_k(L^{-1/2}), 
    \end{align}
    where the second line uses that $Y_L$ is independent of $\sigma(Z_1, \ldots, Z_s)$. 
    Since $g$ is centered with finite Gaussian moments, by the central limit theorem we know that $\operatorname{Law}(Y_L) \to \mathcal{N}(0, V_g)$ narrowly as $L \to \infty$. Since $Q_j^{(s)}$ is bounded and continuous, we find that
\begin{align}
    \lim_{L \to \infty} d_{\mathbf{p}, L} = \sqrt{\frac{j! \prod_{t=1}^s q_t!}{k!}} \cdot \mathbb{E}_{G \sim \gamma_1} \left[ Q_j^{(s)} (\sqrt{V_g}G )\right] \cdot \prod_{t=1}^s \beta_{q_t}.
\end{align}
Convolving the two Gaussians (whose variances add to $\lambda_g = 1 + V_g$) implies that
\begin{equation}
    \mathbb{E}_{G \sim \gamma_1} \left[ Q_j^{(s)} (\sqrt{V_g}G )\right] = \frac{1}{\sqrt{j!}} \left.\frac{d^{j+s}}{dy^{j+s}} \mathbb{E}_{G \sim \gamma_1}[\chi(\sqrt{\lambda_g}G + y)]\right|_{y=0} = \sqrt{\frac{(j+s)!}{j!}} \lambda_g^{-(j+s)/2} \alpha_{j+s},
\end{equation}
and therefore with $J := j+s$ we have
\begin{equation}
\label{eqn:proofappendixresnetapproximationbranchinglemmastep2limofdpL}
    \lim_{L \to \infty} d_{\mathbf{p}, L} = 
    \alpha_J \lambda_g^{-J/2}
    \sqrt{\frac{J! \prod_{t=1}^s q_t!}{k!}}\prod_{t=1}^s \beta_{q_t}.
\end{equation}
Now, for each integer $J \in \n$ we assign to integer configurations $j \in \n \cup \{0\}$, $(m_q)_{q \in \n} \subset \n \cup \{0\}$ satisfying $j + \sum_{q \in \n}m_q = J$ the probabilities
\begin{equation}
\label{eqn:proofappendixapproximationloweroundhermitebranchingpdfofJ}
    \mathbb{P}_J\left(j, (m_q)_{q \in \n}\right) := \frac{J!}{j! \prod_{q \in \n}m_q!} \left(\frac{1}{\lambda_g}\right)^j \prod_{q \in \n}\left(\frac{\beta_q^2}{\lambda_g}\right)^{m_q}.
\end{equation}
These sum to 1 by the multinomial formula, since $\frac{1}{\lambda_g} + \sum_{q \in \n} \frac{\beta_q^2}{\lambda_g} = \frac{1+V_g}{\lambda_g} = 1$. For each configuration drawn from $\mathbb{P}_J$, we set \begin{equation}
    s := \sum_{q \in \n}m_q, \quad N := j + \sum_{q \in \n}qm_q,
\end{equation}
choose $q_1 \leq \ldots \leq q_s$ with exactly $m_q$ occurrences of each $q$, and form the tuple 
\begin{equation}
    \mathbf{i} := (\underbrace{0, \ldots, 0}_{j}, \underbrace{1, \ldots, 1}_{q_1}, \ldots, \underbrace{s, \ldots, s}_{q_s}) \in \mathcal{I}_N.
\end{equation}
Choose $S \subset \{1, \ldots, N\}$ uniformly at random among subsets of size $\lfloor N /2 \rfloor$ and define
\begin{equation}
    C_S := C_S(\mathbf{i}) = |\{i_r:\; r \in S\} \cap \{i_r:\; r \in S^c\} \setminus \{0\}|.
\end{equation}
To identify (\ref{eqn:proofappendixappromxiationResnetfrobenisunormstep1NEW}) probabilistically, fix any configuration $j, (m_q)_{q \in \n}$ with $N = k$. There are
\begin{equation}
    \frac{k!}{j! \prod_{q \in \n}(q!)^{m_q}}
\end{equation}
distinct permutations of the tuple $\mathbf{i}$. Each canonical tuple $\mathbf{p} \in \mathcal{I}_k$ with these multiplicities arises from exactly $\prod_{q \in \n}m_q!$ such permutations, as labels with equal multiplicities can be arbitrarily permuted without changing their canonical representative $\mathbf{p}$. Hence, the number of canonical $k$-tuples with the fixed multiplicities $(m_q)_{q \in \n}$ is equal to
\begin{equation}
        \frac{k!}{j! \prod_{q \in \n}(q!)^{m_q} m_q!}.
\end{equation}
Let $\pi$ be a uniformly random permutation of $\{1, \ldots, k\}$. 
By the above counting argument, the canonical representative of $(i_{\pi(1)}, \ldots, i_{\pi(k)})$ produces a uniformly random canonical tuple $\mathbf{p}_\pi \in \mathcal{I}_k$ with the given multiplicities.
Note that relabeling preserves crossing counts, i.e.
\begin{equation}
  C_{S_k}(\mathbf{p}_\pi) = C_{S_k}(i_{\pi(1)}, \ldots, i_{\pi(k)}) = C_{\pi(S_k)}(\mathbf{i}).
\end{equation}
Since $\pi(S_k)$ is a uniformly random balanced cut, the fraction of canonical tuples $\mathbf{p}_\pi$ satisfying $C_{S_k}(\mathbf{p}_\pi) \geq R$ is exactly $\mathbb{P}_J(C_S \geq R \mid j, (m_q)_{q \in \n})$. From (\ref{eqn:proofappendixresnetapproximationbranchinglemmastep2limofdpL}) we know that
\begin{equation}
    \lim_{L \to \infty} |d_{\mathbf{p}_\pi, L}|^2 = \frac{\alpha_J^2 J!}{k! \lambda_g^J} \prod_{q \in \n}(q! \beta_q^2)^{m_q}, \quad J = j + \sum_{q \in \n}m_q.
\end{equation}
Hence, averaging over $\pi$ for a fixed configuration $j, (m_q)_{q \in \n}$ with $N=k$ yields
\begin{align}
 \frac{j! \prod_{q \in \n}(q!)^{m_q} m_q!} {k!}\sum_{\substack{\mathbf{p} \in \mathcal{I}_k,
 \nu_0(\mathbf{p}) = j \\
\#\{1 \leq t \leq s:\;  \nu_t(\mathbf{p}) =q\} = m_q \; \forall q \in \n \\
 C_{S_k}(\mathbf{p}) \geq R}} \lim_{L \to \infty} |d_{\mathbf{p}, L}|^2 = \frac{\alpha_J^2 J!}{k! \lambda_g^J} \prod_{q \in \n}(q! \beta_q^2)^{m_q} \mathbb{P}_J\left(C_S \geq R \mid j, (m_q)_{q \in \n}\right)
\end{align}
or in other words that
\begin{align}
    \sum_{\substack{\mathbf{p} \in \mathcal{I}_k,
 \nu_0(\mathbf{p}) = j \\
\#\{1 \leq t \leq s:\;  \nu_t(\mathbf{p}) =q\} = m_q \; \forall q \in \n \\
 C_{S_k}(\mathbf{p}) \geq R}} \lim_{L \to \infty} |d_{\mathbf{p}, L}|^2 &= \frac{\alpha_J^2 J!}{j! \prod_{q \in \n}m_q!} \left(\frac{1}{\lambda_g}\right)^j \prod_{q \in \n} \left(\frac{\beta_q^2}{\lambda_g}\right)^{m_q} \\&\quad \cdot \mathbb{P}_J\left(C_S \geq R \mid j, (m_q)_{q \in \n}\right) \\
 &= \alpha_J^2 \mathbb{P}_J\left(j, (m_q)_{q \in \n}\right)\cdot \mathbb{P}_J\left(C_S \geq R \mid j, (m_q)_{q \in \n}\right),
\end{align}
where the second line applied the definition of $\mathbb{P}_J$ in (\ref{eqn:proofappendixapproximationloweroundhermitebranchingpdfofJ}). 
We exchange the limit and the finite sum and then sum over all configurations for which $N=k$ (which includes summing over $J$). However, since $J \leq N$ we need only consider $J \leq k$, which yields
\begin{equation}
\label{eqn:proofappendixappromxiationResnetfrobenisunormstep2}
    \lim_{L \to \infty}\sum_{\substack{\mathbf{p} \in \mathcal{I}_k \\ C_{S_k}(\mathbf{p}) \geq R}} |d_{\mathbf{p}, L}|^2 = \sum_{J =1}^k\alpha_J^2 \mathbb{P}_J(N=k,\; C_S \geq R).
\end{equation}
\textbf{(Step 3).} Under $\mathbb{P}_J$, let $B := \sum_{q \geq 2}m_q$ be the number of positive labels occurring at least twice (i.e. the number of nonlinear factors). Because the probability mass function of $\mathbb{P}_J$ from (\ref{eqn:proofappendixapproximationloweroundhermitebranchingpdfofJ}) is a multinomial law, we find that $B \sim \operatorname{Binomial}(J, 8\mathfrak{v}_g)$ (since $\sum_{q \geq 2}\beta_q^2 / \lambda_g = 8 \mathfrak{v}_g$). Moreover, since $g$ is Lipschitz we know that
\begin{equation}
    \sum_{q \in \n} q\beta_q^2 = \mathbb{E}_{G \sim \gamma_1}|g'(G)|^2 \leq \operatorname{Lip}(g)^2,
\end{equation}
which implies that
\begin{equation}
    \mathbb{E}_J[N] = \mathbb{E}_J \left[j + \sum_{q \in \n}qm_q\right] = \frac{J}{\lambda_g} \left(1 + \sum_{q \in \n}q \beta_q^2\right) \leq \frac{JB_g}{8}. 
\end{equation}
By Markov's inequality, we find that
\begin{equation}
    \mathbb{P}_J(N > B_g J) \leq \frac{1}{8}.
\end{equation}
Next, by the binomial Chernoff bound we know
\begin{equation}
    \mathbb{P}_J(B < 4 \mathfrak{v}_gJ) \leq e^{-\mathfrak{v}_gJ},
\end{equation}
and so a union bound gives that for $J \geq K_g$ we have
\begin{equation}
\label{eqn:proofappendixhermitebranchingchernoffconsequence}
    \mathbb{P}_J(N \leq B_g J, \; B \geq 4 \mathfrak{v}_gJ) \geq 1 - \frac{1}{8} - e^{-\mathfrak{v}_g J} \geq \frac{13}{16},
\end{equation}
where we denote $K_g := \lceil 4/\mathfrak{v}_g\rceil$. Now, condition on a configuration $j, (m_q)_{q \in \n}$ with $B \geq 1$. For each positive label occurring at least twice, select two positions carrying that label. A uniformly balanced random cut separates these positions with probability $\geq \frac{1}{2}$, which implies
\begin{equation}
    \mathbb{E}\left[C_S  \mid j, (m_q)_{q \in \n}\right] \geq \frac{B}{2}.
\end{equation}
Next, we know that
\begin{equation}
   C_S = C_S \1_{C_S < B/4} + C_S \1_{C_S \geq B/4} \leq \frac{B}{4} + \frac{3B}{4} \1_{C_S \geq B/4},
\end{equation}
where we use that $0 \leq C_S \leq B$ (since any pair contributing to $C_S$ must also contribute to $B$). Taking conditional expectations,
\begin{equation}
    \frac{B}{2} \leq \mathbb{E}\left[C_S  \mid j, (m_q)_{q \in \n}\right] \leq \frac{B}{4} + \frac{3B}{4}\mathbb{P}_J\left(C_S \geq B/4 \mid  j, (m_q)_{q \in \n}\right).
\end{equation}
Since $B > 0$, we rearrange to find that
\begin{equation}
    \mathbb{P}_J\left(C_S \geq B/4 \mid  j, (m_q)_{q \in \n}\right) \geq \frac{1}{3}.
\end{equation}
For a configuration with $B=0$ this holds trivially, and therefore it always holds. Combining with (\ref{eqn:proofappendixhermitebranchingchernoffconsequence}) gives
\begin{equation}
\label{eqn:proofappendixappromxiationResnetfrobenisunormstep3}
    \mathbb{P}_J(N \leq B_gJ, \; C_S \geq \mathfrak{v}_g J) \geq \frac{13}{48}.
\end{equation}

We now finish the proof of the main claim (\ref{eqn:proofappendixthehermitebranchinglemmamainestimate}). Combining the estimates (\ref{eqn:proofappendixappromxiationResnetfrobenisunormstep1NEW}) and (\ref{eqn:proofappendixappromxiationResnetfrobenisunormstep2}) from steps 1 and 2, respectively, gives
\begin{align}
     \liminf_{L \to \infty} \sum_{j > L^{R-1}} \sigma_j (M_k(F_L))^2 &\geq  \sum_{J =1}^k\alpha_J^2 \mathbb{P}_J(N=k,\; C_S \geq R).
\end{align}
For $K_g \leq K_1 \leq K_2$, summing over $K_1 \leq k \leq \lceil B_gK_2\rceil$ and taking $R = \lfloor \mathfrak{v}_g K_1\rfloor$ gives
\begin{align}
    \liminf_{L \to \infty}\sum_{k=K_1}^{\lceil B_gK_2\rceil} \sum_{j > L^{\lfloor \mathfrak{v}_g K_1\rfloor-1}} \sigma_j (M_k(F_L))^2 &\geq \sum_{k=K_1}^{\lceil B_gK_2\rceil} \sum_{J =1}^k\alpha_J^2 \mathbb{P}_J(N=k,\; C_S \geq \lfloor \mathfrak{v}_g K_1\rfloor)
    \\ &= \sum_{J=1}^{\lceil B_gK_2\rceil} \alpha_J^2\sum_{k = \max\{K_1, J\}}^{\lceil B_gK_2\rceil} \mathbb{P}_J(N=k,\; C_S \geq \lfloor \mathfrak{v}_g K_1\rfloor)
    \\ &= \sum_{J=1}^{\lceil B_gK_2\rceil} \alpha_J^2 \mathbb{P}_J(K_1 \leq N\leq \lceil B_gK_2\rceil, \; C_S \geq \lfloor \mathfrak{v}_g K_1\rfloor) \\
    &\geq\sum_{J=K_1}^{K_2} \alpha_J^2 \mathbb{P}_J(K_1 \leq N\leq \lceil B_gK_2\rceil, \; C_S \geq \lfloor \mathfrak{v}_g K_1\rfloor) \\
    &= \sum_{J=K_1}^{K_2} \alpha_J^2 \mathbb{P}_J(N\leq \lceil B_gK_2\rceil, \; C_S \geq \lfloor \mathfrak{v}_g K_1\rfloor)\\
&\geq \sum_{J=K_1}^{K_2} \alpha_J^2 \mathbb{P}_J(N\leq B_gJ, \; C_S \geq \mathfrak{v}_g J)\\
    &\geq \frac{13}{48} \sum_{J=K_1}^{K_2}\alpha_J^2,
\end{align}
where the third and fifth lines use that $N \geq J$ almost surely, the fourth uses that $V_g = \operatorname{Var}(g) \leq \operatorname{Lip}(g)^2$ and so $B_g \geq 1$, the sixth uses that $\lceil B_g K_2\rceil \geq B_g J$ and $\lfloor \mathfrak{v}_g K_1\rfloor \leq \mathfrak{v}_g J$, and the last uses (\ref{eqn:proofappendixappromxiationResnetfrobenisunormstep3}) from step 3. This is exactly the claim (\ref{eqn:proofappendixthehermitebranchinglemmamainestimate}). 
\end{proof}

\end{document}